\documentclass[12pt,oneside]{book}
\usepackage[letterpaper, left=1in, right=1in, top=1in, bottom=1in]{geometry}
\usepackage{fancyhdr}
\usepackage{caption}
\usepackage{subcaption}
\usepackage{titlesec}
\usepackage{titletoc}
\usepackage{lipsum}
\usepackage{enumitem}
\usepackage{makecell}
\usepackage{float}
\usepackage{amsmath,amssymb,amsthm,amsfonts}
\usepackage{algorithmic}
\usepackage{afterpage}
\usepackage{graphicx}
\usepackage{booktabs, multirow} 
\usepackage{tocloft}
\usepackage{tocbibind}
\usepackage[table]{xcolor} 
\usepackage{pdfpages}
\usepackage{soul}
\usepackage{setspace}
\usepackage{hyperref}
\hypersetup{
bookmarks=true,
linktoc=all,
hidelinks=true,
breaklinks=true
}
\usepackage[
backend=bibtex,
style=numeric,
sorting=none,
maxbibnames=99
]{biblatex}
\usepackage[T1]{fontenc}
\usepackage[utf8]{inputenc}
\usepackage{appendix}
\usepackage[ruled,vlined,linesnumbered]{algorithm2e}
\usepackage{listings}
\usepackage{bm}
\usepackage{csquotes}
\bibliography{Reference}
\usepackage{titlesec}
\usepackage[acronym]{glossaries}
\usepackage{nomencl}
\usepackage{etoolbox}
\usepackage[export]{adjustbox}
\usepackage{threeparttable}

\renewcommand{\chaptermark}[1]{\markboth{\MakeUppercase{#1}}{}}

\titleformat{ \chapter}[display]{\normalfont\huge\bfseries}{\chaptertitlename\thechapter}{30pt}{\Huge} 
\titlespacing*{\chapter}{0pt}{-30pt}{30pt}

\usepackage{pifont}
\newcommand{\cmark}{\ding{51}}%
\newcommand{\xmark}{\ding{55}}

\newcommand{\mb}[1]{\boldsymbol{#1}}
\newcommand{\mset}[1]{\mathcal{#1}}
\newcommand{\mseq}[1]{\mb{#1}}
\newcommand{\mseqdef}[1]{\mb{[}#1\mb{]}}
\newcommand{\mvecdef}[1]{\mb{[}#1\mb{]}}
\newcommand{\msetdef}[1]{\mb{\{}#1\mb{\}}}
\newcommand{\ctcblank}[0]{\varnothing}

\newcommand{\mt}[2]{^{\mb{#1}}_{#2}}
\newcommand{\mtseq}[2]{\mt{#1}{#2}}
\newcommand{\mcat}[2]{\left[#1;#2\right]}
\newcommand{\mcats}[1]{\mathrm{concat}(#1)}
\newcommand{\mconv}[2]{#1\odot#2}

\DeclareMathOperator*{\argmax}{argmax}
\DeclareMathOperator*{\concatenate}{concatenate}
\DeclareMathAlphabet{\mathcal}{OMS}{cmsy}{m}{n}

\makeglossaries
\newacronym{ann}{ANN}{Artificial Neural Network}
\newacronym{blstm}{BLSTM}{Bidirectional Long-Short Term Memory}
\newacronym{cnn}{CNN}{Convolutional Neural Network}
\newacronym{ctc}{CTC}{Connectionist Temporal Classification}
\newacronym{cv}{CV}{Computer Vision}
\newacronym{ce}{CE}{Cross Entropy}
\newacronym{cer}{CER}{Character Error Rate}
\newacronym{dan}{DAN}{Document Attention Network}
\newacronym{dl}{DL}{Deep Learning}
\newacronym{dla}{DLA}{Document Layout Analysis}
\newacronym{dnn}{DNN}{Deep Neural Network}
\newacronym{dpi}{DPI}{Dots Per Inch}
\newacronym{dsc}{DSC}{Depthwise Separable Convolution}
\newacronym{fcn}{FCN}{Fully Convolutional Network}
\newacronym{ffn}{FFN}{Feed-Forward Network}
\newacronym{gpu}{GPU}{Graphic Processing Unit}
\newacronym{gcrl}{GCRL}{Gated Convolutional Recurrent Network}
\newacronym{gcnn}{GCNN}{Gated Convolutional Neural Network}
\newacronym{ged}{GED}{Graph Edit Distance}
\newacronym{gfcn}{GFCN}{Gated Fully Convolutional Network}
\newacronym{gru}{GRU}{Gated Recurrent Unit}
\newacronym{hdr}{HDR}{Handwritten Document Recognition}
\newacronym{hmm}{HMM}{Hidden Markov Model}
\newacronym{htr}{HTR}{Handwritten Text Recognition}
\newacronym{iou}{IoU}{Intersection over Union}
\newacronym{lm}{LM}{Language Model}
\newacronym{loer}{LOER}{Layout Ordering Error Rate}
\newacronym{lstm}{LSTM}{Long-Short Term Memory}
\newacronym{map}{mAP}{mean Average Precision}
\newacronym{mdlstm}{MDLSTM}{Multi-Dimensional Long-Short Term Memory}
\newacronym{mdrnn}{MDRNN}{Multi-Dimensional Recurrent Neural Network}
\newacronym{ml}{ML}{Machine Learning}
\newacronym{mlp}{MLP}{Multi-Layer Perceptron}
\newacronym{ner}{NER}{Named Entity Recognition}
\newacronym{nlp}{NLP}{Natural Language Processing}
\newacronym{nmt}{NMT}{Neural Machine Translation}
\newacronym{ocr}{OCR}{Optical Character Recognition}
\newacronym{pper}{PPER}{Post Processing Edition Rate}
\newacronym{rnn}{RNN}{Recurrent Neural Network}
\newacronym{rpn}{RPN}{Region Proposal Network}
\newacronym{seq2seq}{seq2seq}{Sequence-to-sequence}
\newacronym{sgd}{SGD}{Stochastic Gradient Descent}
\newacronym{snn}{SNN}{Self-Normalizing Network}
\newacronym{span}{SPAN}{Simple Predict \& Align Network}
\newacronym{van}{VAN}{Vertical Attention Network}
\newacronym{vit}{ViT}{Vision Transformer}
\newacronym{wer}{WER}{Word Error Rate}
\newacronym{xml}{XML}{eXtensible Markup Language}

\renewcommand\nomgroup[1]{%
  \item[\bfseries
    \ifstrequal{#1}{E}{Other symbols}{%
        \ifstrequal{#1}{D}{Set}{%
            \ifstrequal{#1}{C}{Matrix / Tensor}{
                \ifstrequal{#1}{B}{Vector}{}
            }
        }
    }%
]}
\makenomenclature
\nomenclature{$a$ or $A$}{A variable or function.}

\nomenclature[B, 00]{$\mb{a}$}{A vector or a sequence.}
\nomenclature[B, 01]{$\mb{a}_i$}{The element of $\mb{a}$ at position $i$.}
\nomenclature[B, 02]{$\mb{a}\mt{t}{}$}{$\mb{a}$ at time step $t$.}
\nomenclature[B, 03]{$\mb{a}\mt{t}{i}$}{The element of $\mb{a}$ at position $i$ and time step $t$.}
\nomenclature[B, 04]{$\mb{a}^T$}{The transpose operator over the vector $\mb{a}$.}
\nomenclature[B, 05]{$\mb{0}$}{A vector of zeros.}
\nomenclature[B, 06]{$\mseqdef{a,b}$}{A vector made up of $a$ and $b$.}
\nomenclature[B, 07]{$\mcat{\mb{a}}{\mb{b}}$}{The concatenation of vectors $\mb{a}$ and $\mb{b}$.}
\nomenclature[B, 08]{$\lVert\mb{a}\rVert_i$ }{The norm $i$ of vector $\mb{a}$.}

\nomenclature[C, 00]{$\mb{A}$}{A matrix or tensor.}
\nomenclature[C, 01]{$\mb{A}_{i,j}$}{The element of $\mb{A}$ at position $(i,j)$.}
\nomenclature[C, 02]{$\mb{A}\mt{t}{}$}{$\mb{A}$ at time step $t$.}
\nomenclature[C, 03]{$\mb{A}\mt{t}{i,j}$}{The element of $\mb{A}$ at position $(i,j)$ and time step $t$.}
\nomenclature[C, 04]{$\mb{A}^T$}{The transpose operator over the matrix $\mb{A}$.}
\nomenclature[C, 06]{$\mb{A} \circ \mb{B}$}{The Hadamart (element-wise) product between $\mb{A}$  and $\mb{B}$.}
\nomenclature[C, 07]{$\mconv{\mb{A}}{\mb{K}}$}{Convolution operation between the input matrix $\mb{A}$ and the kernels $\mb{K}$.}
\nomenclature[C, 08]{$\lVert\mb{A}\rVert_i$}{The norm $i$ of matrix $\mb{A}$.}

\nomenclature[D, 00]{$\mset{A}$}{A set.}
\nomenclature[D, 01]{$\msetdef{a,b}$}{The set made up of $a$ and $b$.}
\nomenclature[D, 02]{$\mathrm{card}(\mset{A})$}{The cardinality of set $\mset{A}$.}
\nomenclature[D, 03]{$\mset{A}^n$}{The n-ary Cartesian power of the set $\mset{A}$.}

\nomenclature[E, 00]{$\mathbb{R}$}{The set of real numbers.}
\nomenclature[E, 01]{$\mset{L}$}{The loss function.}
\nomenclature[E, 02]{$\mset{C}$}{The cost function.}
\nomenclature[E, 03]{$\mset{N}(a,b)$}{The normal distribution with mean $a$ and variance $b$.}
\nomenclature[E, 04]{$\mset{U}(a,b)$}{The uniform distribution with lower bound $a$ and upper bound $b$.}
\nomenclature[E, 05]{$\left[a,b\right]$}{The closed interval between $a$ and $b$.}
\nomenclature[E, 06]{$(a,b)$}{Couple made up of $a$ and $b$.}

\begin{document}



\includepdf{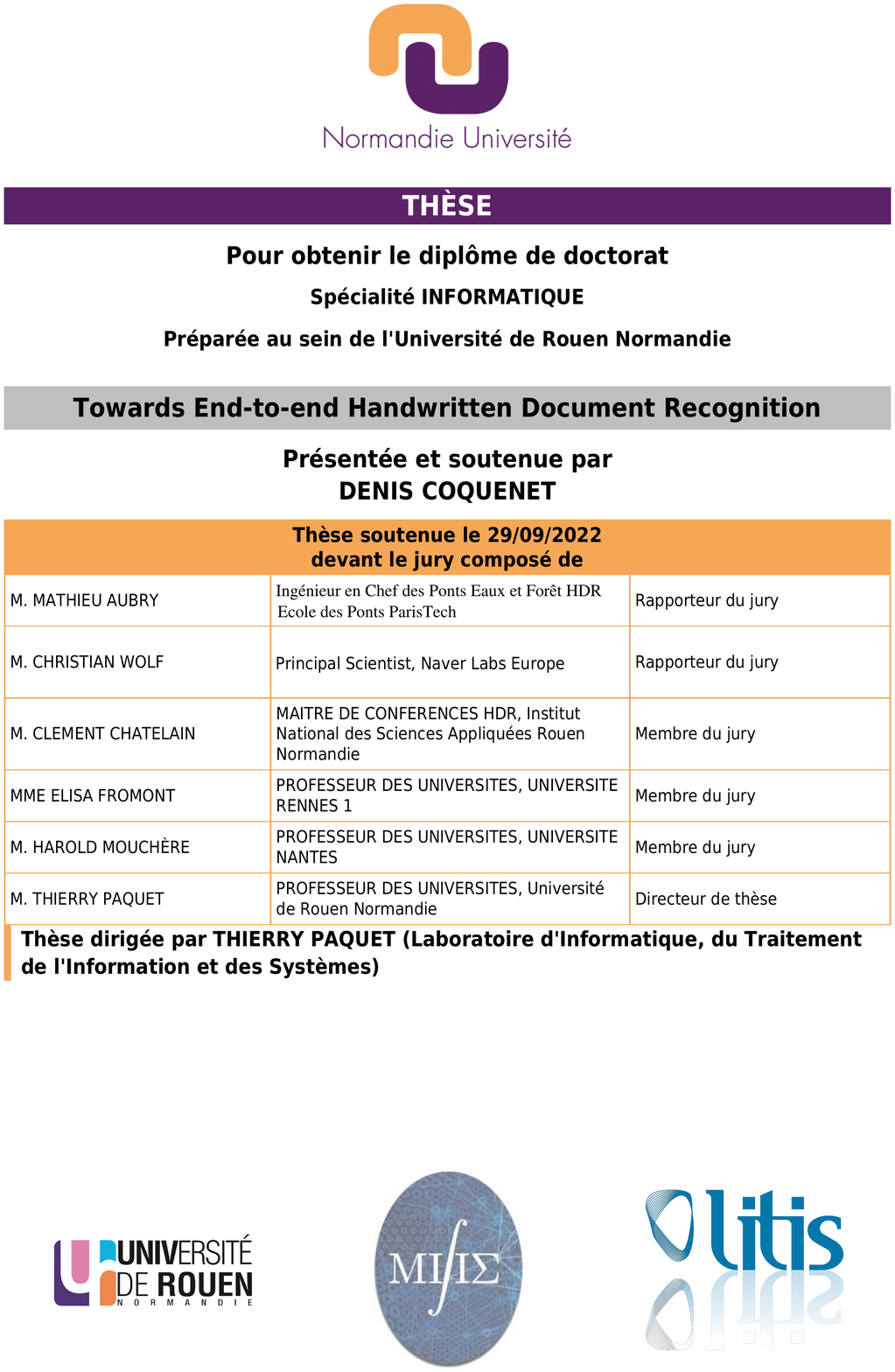}
\restoregeometry
\afterpage{
    \thispagestyle{empty}
    \clearpage
    \newpage \ \newpage
}

\pagenumbering{roman}
\setcounter{page}{2}
\clearpage

\chapter*{Acknowledgments}
\chaptermark{Acknowledgments}
\label{chap:Acknowledgments}
\addcontentsline{toc}{chapter}{Acknowledgments}

First of all, I would like to thank my thesis supervisor Thierry Paquet and my advisor Clément Chatelain for their confidence, their support and all the advice and knowledge they have shared with me. I am grateful to Nicolas Malandain for this conversation which led me to consider undertaking a thesis. Thanks also to Yann Soullard who supervised and advised me during my first research works.

I would like to thank Christian Wolf and Mathieu Aubry who kindly accepted to review my thesis, and more generally to the whole defense committee: Harold Mouchere and Elisa Fromont. I am very honored and humbled by their review of my work.

I would like to thank the whole LITIS, for its welcome and its kindness, and more specifically the Apprentissage team.
I would like to particularly thank Mélodie Boillet, with whom I went through the different stages of this thesis, for the laughs, the stimulating discussions and the "adventure moments" we shared together. I would like to thank Guillaume Renton, Thomas Constum and Theo Larcher without whom lunch breaks would not have been as pleasant. I would also like to thank all the administrative staff, who have accompanied me and whose daily work is remarkable: Fabienne Bocquet, Mathieu Baum and Brigitte Diarra, just to name a few.

I would also like to thank all the people who have been by my side during these three years of thesis. I would like to thank my family who has always supported me and believed in me. I thank my friends for the great moments shared with them, far from the worries of the thesis. I would like to take this opportunity to thank my love who has given me invaluable support.

I would like to thank the staff of CRIANN (Regional HPC Center, Normandy, France), notably Benoist Gaston and Beatrice Charton, and of GENCI-IDRIS (Grant 2020-AD011012155). 
This thesis was financially supported by the French Defense Innovation Agency and by the Normandy region, which I also thank.

\clearpage

\chapter*{Résumé}
\chaptermark{Résumé}
\label{chap:Resume}
\addcontentsline{toc}{chapter}{Résumé}

La reconnaissance de textes manuscrits a été largement étudiée au cours des dernières décennies pour ses nombreuses applications. Aujourd'hui, l'approche à l'état de l'art repose sur un processus en trois étapes. Le document est segmenté en lignes de texte, qui sont ensuite ordonnées et reconnues. Cependant, cette approche en trois étapes présente de nombreux inconvénients. Les trois étapes sont traitées indépendamment alors qu'elles sont étroitement liées. Les erreurs s'accumulent d'une étape à l'autre. L'étape d'ordonnancement est basée sur des règles heuristiques qui empêchent son utilisation pour des documents à la mise en page complexe ou pour des documents hétérogènes. L'étape de segmentation nécessite ses propres annotations supplémentaires. 

Dans cette thèse, nous proposons de résoudre ces problèmes en effectuant la reconnaissance du texte du document en une seule étape, de bout en bout. Pour ce faire, nous augmentons progressivement la difficulté de la tâche de reconnaissance, en passant de lignes isolées à des paragraphes, puis à des documents entiers. Nous avons proposé une approche au niveau ligne, basée sur un réseau entièrement convolutif, afin de concevoir un premier module générique d'extraction de caractéristiques pour la tâche de reconnaissance d'écritures manuscrites. 

Sur la base de ce travail préliminaire, nous avons étudié deux approches différentes pour reconnaître des paragraphes manuscrits. D'une part, nous avons conçu une approche non récurrente qui vise à aligner les prédictions du paragraphe entier dans un espace 2D afin de préserver la nature de l'image d'entrée. D'autre part, nous avons conçu une approche récurrente dans laquelle les lignes de texte sont traitées de manière itérative. Nous avons obtenu des résultats à l'état de l'art au niveau paragraphe sur les jeux de données RIMES 2011, IAM et READ 2016 et nous avons également dépassé l'état de l'art au niveau ligne sur ces jeux de données.

Nous avons enfin proposé la première approche de bout en bout dédiée à la reconnaissance à la fois du texte et de la mise en page, au niveau document. Les caractères et les symboles représentant la mise en page sont prédits séquentiellement en suivant un ordre de lecture appris. Nous avons proposé deux nouvelles métriques que nous avons utilisées pour évaluer cette tâche sur les jeux de données RIMES 2009 et READ 2016, au niveau page et double page.

Dans une optique de reproductibilité, de transparence et de partage scientifique auquel nous sommes attachés, tous les codes et poids des modèles sont publiquement disponibles en ligne.

\textbf{\textit{Mots-clés -}} Reconnaissance de textes manuscrits, mécanismes d'attention, réseaux de neurones entièrement convolutifs, Transformer.

\clearpage

\chapter*{Abstract}
\chaptermark{Abstract}
\label{chap:Abstract}
\addcontentsline{toc}{chapter}{Abstract}

Handwritten text recognition has been widely studied in the last decades for its numerous applications. Nowadays, the state-of-the-art approach consists in a three-step process. The document is segmented into text lines, which are then ordered and recognized. However, this three-step approach has many drawbacks. The three steps are treated independently whereas they are closely related. Errors accumulate from one step to the other. The ordering step is based on heuristic rules which prevent its use for documents with a complex layouts or for heterogeneous documents. The need for additional physical segmentation annotations for training the segmentation stage is inherent to this approach. 

In this thesis, we propose to tackle these issues by performing the handwritten text recognition of whole document in an end-to-end way. To this aim, we gradually increase the difficulty of the recognition task, moving from isolated lines to paragraphs, and then to whole documents. We proposed an approach at the line level, based on a fully convolutional network, in order to design a first generic feature extraction step for the handwriting recognition task. 

Based on this preliminary work, we studied two different approaches to recognize handwritten paragraphs. First, we designed a one-shot approach which aims at aligning the predictions of the whole paragraph in a two-dimensional space to preserve the nature of the the input image. Second, we designed a recurrent approach in which text lines are processed iteratively. We reached state-of-the-art results at paragraph level on the RIMES 2011, IAM and READ 2016 datasets and outperformed the line-level state of the art on these datasets.

We finally proposed the first end-to-end approach dedicated to the recognition of both text and layout, at document level. Characters and layout tokens are sequentially predicted following a learned reading order. We proposed two new metrics we used to evaluate this task on the RIMES 2009 and READ 2016 dataset, at page level and double-page level.

For reproducibility, transparency and scientific sharing purposes to which we are attached, all codes and weights of the models are publicly available online.

\textbf{\textit{Keywords -}} Handwritten Text Recognition, Attention mechanism, Image-to-sequence, Fully Convolutional Network, Transformer.

\clearpage




\renewcommand\contentsname{Table of Contents}
\tableofcontents{}
\clearpage

\listoftables
\clearpage

\listoffigures
\clearpage

\addcontentsline{toc}{chapter}{List of Algorithms}
\listofalgorithms
\clearpage

\addcontentsline{toc}{chapter}{Acronyms}
\printglossary[type=\acronymtype,nonumberlist]
\clearpage



\pagenumbering{arabic}
\setcounter{page}{1} 

\glsresetall
\chapter{Introduction}
\label{chap:Introduction}

\section{Context and motivation}
One of the inherent characteristics of human beings is their ability to communicate, to exchange information with each other. Since the birth of writing, a few thousand years ago, human beings used various media to store, save and distribute information: from gravings and papyri for the oldest to whiteboards, road signs, and mostly paper sheets nowadays. 
We are now living in a digital era. However, most of the information is only stored on physical media. Indeed, handwriting has long been ubiquitous in our daily lives no matter what form it takes: shopping lists, student notes, letters or even books. 

Extracting the textual content from these media can be particularly useful for the automation of many industrial tasks such as bank check, invoice and form analysis and processing for example. It is also useful for historians needs, to avoid experts to transcribe the texts themselves, which is very time-consuming. \gls{htr} can be used as a first step among more complex systems to translate documents or spot areas of interest through keywords. It also has interesting military applications: real-time document translation for spying for instance. This has induced a great interest in the automation of the handwriting recognition task. 

\gls{htr} consists in recognizing the handwritten text from a digitized document, by producing a sequence of characters in ASCII format. More specifically, this thesis is dedicated to offline \gls{htr}, \textit{i. e.} that the input is an image, a digital representation obtained after a scanning process. 
\gls{htr} is a difficult task which remains challenging even for modern computer vision systems. As a matter of fact, there is a large diversity of writing styles from one person to another: block letters, cursive,  more or less inclined handwritings. In addition, the size of the characters, the thickness of the strokes and even the inter-character or inter-word spacing can vary significantly. The text can also be colored, with various backgrounds. All this variability makes this task more difficult than the recognition of printed text. 

Nowadays, state-of-the-art approaches for the recognition of handwritten documents in the literature rely on deep neural networks. This task is mainly handled through a complex pipeline implying three main steps that are sequentially applied: the segmentation of the document into text regions (and sometimes other regions of interest such are drawings or tables), the ordering of these regions, and the recognition of the text regions. However, this standard approach has several drawbacks. By definition, \gls{htr} only corresponds to the recognition of the text. The use of this three-step paradigm (segmentation, ordering, recognition) involves the training of a segmentation model which requires its own segmentation annotations, which are costly to produce. In addition, the ordering step is usually based on heuristic rules. Designing a rule-based ordering step is tedious for complex documents and is impossible without prior knowledge about the input documents, whereas the reading order is crucial to process whole documents. Moreover, the recognition errors accumulate with the errors made by the two previous stages. 

Breaking down the \gls{htr} task into three steps prevents a global understanding of the document. Indeed, recognizing a whole document implies to read the different textual regions in a specific order to preserve the global coherence of the content; this is only feasible with an understanding of the layout. The layout may be very different from one document to another: one-column for this thesis, two-column for some scientific papers, multi-column for journals or more complex for maps and schemes. Many non-textual elements can also be found in the image, such as illustrations or mathematical expressions. \gls{htr} is a more complicated task than it seems: it implies the localization and the recognition of the different characters in the correct order, ignoring the non-textual parts, and based on the understanding of the document layout.

In addition, managing the three steps altogether is difficult: the different stages being interdependent, the modification of one of the steps may imply the adaptation of the whole pipeline. This way, the work proposed in the literature mainly focused on only one of these steps.
The recognition of isolated character, word and line images, and their corresponding segmentation stage, have been widely studied by the community, leading to state-of-the-art performances. 
To alleviate the need for segmentation labels, very few works have been devoted to the end-to-end recognition of paragraphs. 
Although essential, the reading order aspect is almost non-existent in the literature. 

This way, processing the \gls{htr} task in an end-to-end way, \textit{i. e.} designing a model which takes advantage of the layout analysis and which integrates the notion of reading order to recognize the text, seems to be to consider to solve these different issues and limitations. 

\section{Contributions}
This thesis aims at advancing the field of handwriting recognition through four contributions which are articulated around a common goal: to move towards the end-to-end recognition of whole documents. To this end, we started from the recognition of isolated text lines to go towards the recognition of paragraphs, then of whole documents.

\paragraph*{Contribution for line-level \gls{htr}.}
We started with the idea of designing a generic feature extractor module (encoder) for the \gls{htr} task, which could be used for text line recognition, but also for paragraph and document recognition. To this end, we proposed a \gls{fcn} architecture we applied on the \gls{htr} task at line level. We achieved competitive results on two public datasets: RIMES 2011 \cite{RIMES_paragraph} and IAM \cite{IAM}. 


\paragraph*{Contributions for paragraph-level \gls{htr}.}
Based on the previous contribution, we then increased the level of reading order difficulty by dealing with paragraph images: this represents an additional vertical axis to take into account. We proposed two approaches for the \gls{htr} task at paragraph level:
\begin{itemize}
    \item A one-shot approach \textit{i. e.} the prediction of the whole paragraph is carried out in a single iteration, without any recurrence. The idea is to preserve the two-dimensional nature of the task for the prediction. A reshaping operation is then used to switch from a two-dimensional prediction to a one-dimensional character sequence. The order of the characters is preserved through a vertical alignment during the prediction. We proposed the \gls{span} following this approach. This recurrence-free model reached competitive results on three public datasets: RIMES 2011, IAM and READ 2016 \cite{READ2016}.
    \item A line-level recurrent approach \textit{i. e.} the paragraph image is processed iteratively, each iteration being dedicated to the recognition of a specific text line. We proposed the \gls{van} following this approach through the use of a line-level hybrid attention module. The \gls{van} also learns to detect the end of the paragraph so as to stop the recurrent process. It outperformed state-of-the-art results on the three public datasets: RIMES 2011, IAM and READ 2016.
\end{itemize}


\paragraph*{Contribution for \gls{hdr}.}
The last step was to handle whole documents. It means handling images containing multiple paragraphs, sometimes spread over several columns.
We proposed the first approach able to deal with whole document images in an end-to-end fashion. In addition to the text recognition, the proposed approach aims at recognizing the layout by labeling text parts using begin and end tags in an XML-like fashion. We proposed the \gls{dan} to illustrates this approach. It consists in a segmentation-free transformer-based model which involves a character-level attention. This way, the \gls{dan} is trained without using any physical segmentation annotation. We proposed two new metrics for this task and we evaluated the model at single-page and double-page levels on the RIMES 2009 \cite{RIMES_page} and READ 2016 datasets.

\section{Overview of the thesis}
This manuscript is divided into 6 chapters. Chapter \ref{chap:background} formally presents the \gls{htr} task as an image-to-sequence problem. The general deep learning background is provided, with a particular focus on the areas of \gls{cv} and \gls{seq2seq} transformation problems. The three following chapters are guided by the evolution of the approaches to deal with the \gls{htr} task and our contributions. Chapter \ref{chap:line} is dedicated to line-level \gls{htr} approaches. The paragraph-level \gls{htr} approaches are studied in Chapter \ref{chap:paragraph}. We present the document-level approach we proposed in Chapter \ref{chap:document}. Conclusion and future directions of work are given in Chapter \ref{chap:conclusion}.

\glsresetall
\chapter{Handwritten text recognition: overview of the problem}
\label{chap:background}

The task of \gls{htr} can be formulated as follows. The aim is to recognize all the text from an input image. This image is noted $\mb{X} \in \mathbb{R}^{H \times W \times C}$, where $H$ is its height, $W$ is its width and $C$ is the number of channels (1 for gray-scaled images and 3 for RGB images). The image contains a sequence $\mseq{y}$ of $L$ characters $\mseq{y}\mtseq{i}{}$ from an alphabet $\mset{A}$. This sequence of characters is structured in multiple text regions (or paragraphs) according to a specific layout. The goal is to develop a system that takes as input the image $\mb{X}$ and outputs all the characters, in a human reading order, as the prediction $\hat{\mseq{y}}$. For an English paragraph for example, it means from top to bottom and from left to right. The reading order can be more complicated for whole documents, from one paragraph to another for instance.

\begin{figure}[h]
    \centering
    \includegraphics[width=0.9\textwidth]{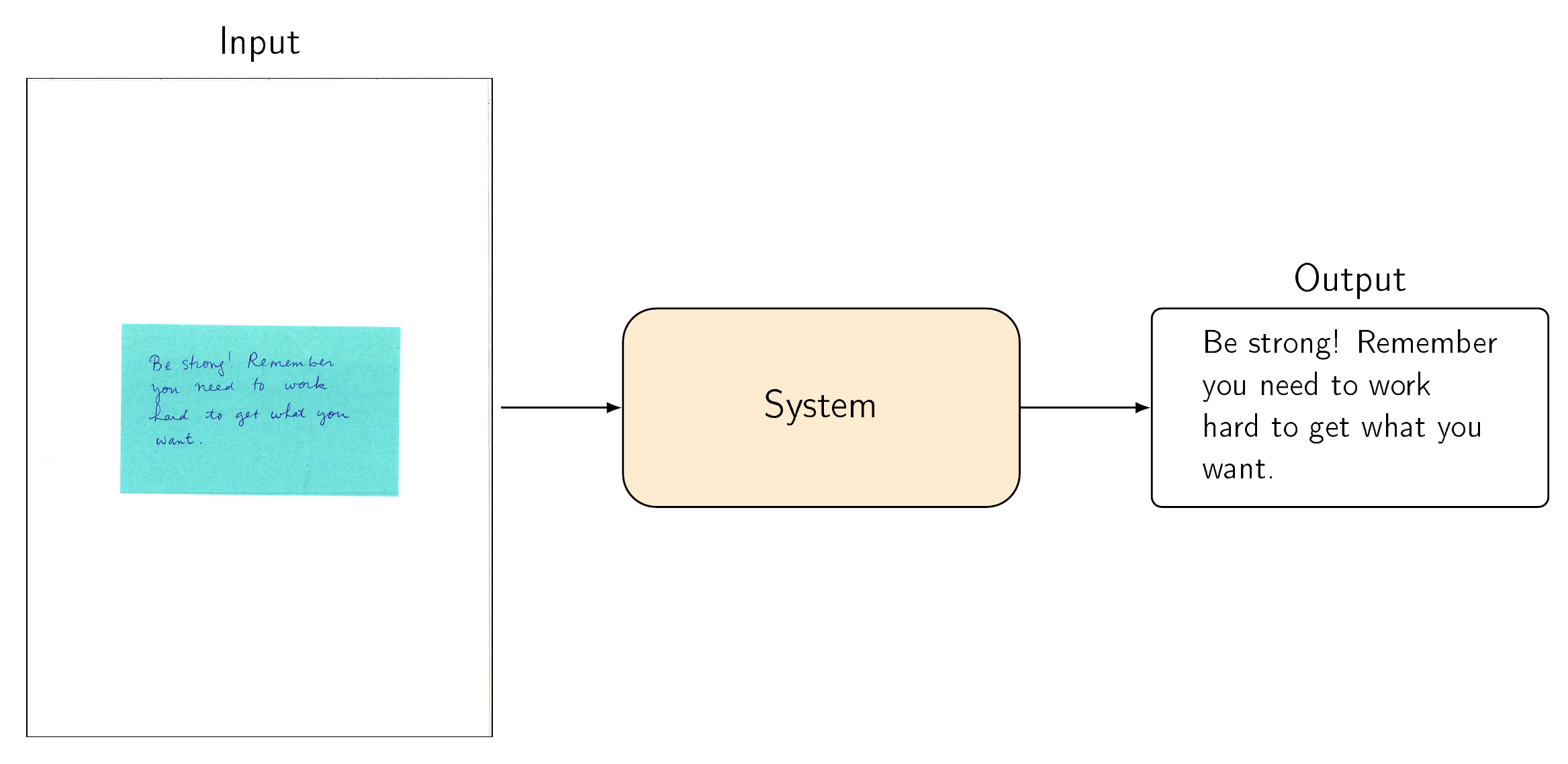}
    \caption{Handwritten text recognition task. The aim is to extract the text from the input image, in a correct order.}
    \label{fig:htr-overview}
\end{figure}

In other words, during training, the aim is to maximize $p(\hat{\mseq{y}}=\mseq{y}|\mb{X})$, the probability to predict the ground truth sequence $\mseq{y}$, given the input image $\mb{X}$. 
One can implement the maximization of this quantity using different statistical models; in this thesis, we focus on optimization approaches associated with \gls{dnn} architectures. The task to be solved is depicted in Figure \ref{fig:htr-overview}. Here, the line breaks are considered as any other character, but it could also be considered as a space character. 

One should note several points to understand the complexity of the task:
\begin{itemize}
    \item We do not make any assumptions about the digitized handwritten documents in terms of writing styles, layout, resolution, background or color/format encoding.
    \item We do not make any assumptions about the size of the digitized document, so the input size can vary a lot from one image to another. Moreover, the image size is not related to the number of characters in the image \textit{i. e.} the length of the output sequence. This length is not known beforehand.
    \item We do not make any assumptions about the reading order of the documents. For example, a single-column document follows a fixed reading order: from top to bottom and from left to right whereas a two-column document implies another reading-order level: column by column.
    \item For complex documents such as schemes or maps, many reading orders would be humanly acceptable: there is not a unique valid output sequence. This should be taken into account when training the system.
\end{itemize}

Nowadays, state-of-the-art results for such a complex task are obtained with deep neural networks. Instinctively, one could classify \acrshort{htr} as a \gls{cv} task since it involves an input image. But it could also be seen as a sequence-to-sequence (\acrshort{seq2seq}) problem due to the expected output character sequence. In fact, we will see that it would be better to qualify \acrshort{htr} as an image-to-sequence problem.

In the following sections, we explain how a \acrshort{dnn} system works and present the main challenges when dealing with such systems. We go over the main discoveries and improvements in the fields of computer vision and sequence-to-sequence problems, which will give us the fundamental elements to handle image-to-sequence tasks. The remainder of this chapter is dedicated to the presentation of the public datasets and metrics we used to evaluate the performance of the trained networks.

\section{Deep Learning background}
\gls{dl} is part of \gls{ml}: it aims at estimating a function $F: \mb{x} \rightarrow \hat{\mb{y}}$ in order to solve a given task, be it regression or classification, so as to optimize the objective $p(\hat{\mb{y}}=\mb{y}|\mb{x})$. While older \gls{ml} approaches rely on a handcrafted feature extraction process, deep learning techniques are based on learning this feature extraction process, through the use of \glspl{ann}, applied on raw data. The \glspl{ann} used in \gls{dl} differ from the other \gls{ml} approaches by the way they combine many elementary functions to estimate a complex function $F$. Indeed, \acrshort{ann}s are inspired by the biological neural networks found in living organisms: they are made of a large quantity of neurons, connected to each other. 

\subsubsection*{Perceptron}
\begin{figure}[h]
    \centering
    \includegraphics[width=0.5\linewidth]{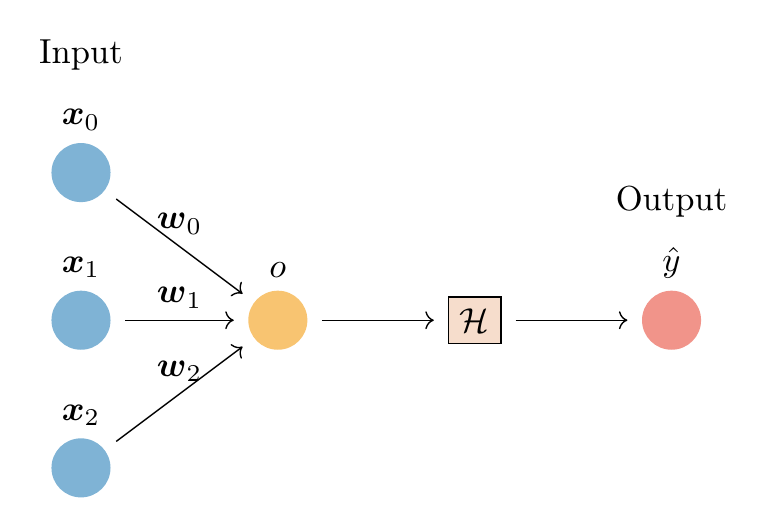}
    \caption{Example of Perceptron for a 2-feature input. The output $\hat{y}$ is computed as a weighted sum between the input values $\mb{x}_i$ and the trainable weights $\mb{w}_i$, to which is applied the heaviside function $\mset{H}$.}
    \label{fig:perceptron}
\end{figure}

In 1943, the authors of \cite{McCulloch43} described the first artificial neuron. Few years later, in 1958, Frank Rosenblatt proposed an algorithm to adjust the weights, also adding few changes to this first neuron, leading to the Perceptron. The Perceptron is a binary classifier based on a unique formal neuron followed by a heaviside function $\mset{H}$. It is depicted in Figure \ref{fig:perceptron} for a 2-feature input.

The formal neuron is fully connected to the inputs $\mb{x} = \mvecdef{\mb{x}_1,...,\mb{x}_n}$ \textit{i. e.} its output $o$ is a weighted sum of its inputs with its trainable weights $\mb{w}_i$:

\begin{equation}
   o = \displaystyle \sum_{i=1}^n \mb{w}_i \mb{x}_i + b,
\end{equation}
where $b$ is known as bias. It is also a trainable parameter. By setting $x_0=1$ and $w_0=b$, one can re-write the equation:
\begin{equation}
    o = \displaystyle \sum_{i=0}^n \mb{w}_i \mb{x}_i.
\end{equation}

The output of the Perceptron $\hat{y}$ is then computed by applying the heaviside function $\mset{H}$ on $o$:
\begin{equation}
\hat{y} = \mathcal{H}(o) =
\begin{cases}
    0 & \text{if } o < 0,\\
    1 & \mathrm{otherwise}.
\end{cases}
\end{equation}
One can note that the decision function of the Perceptron is linear, which is an important limitation for real-world classification tasks.

\subsubsection*{Multi-Layer Perceptron}
\begin{figure}[ht!]
    \centering
    \includegraphics[width=0.5\linewidth]{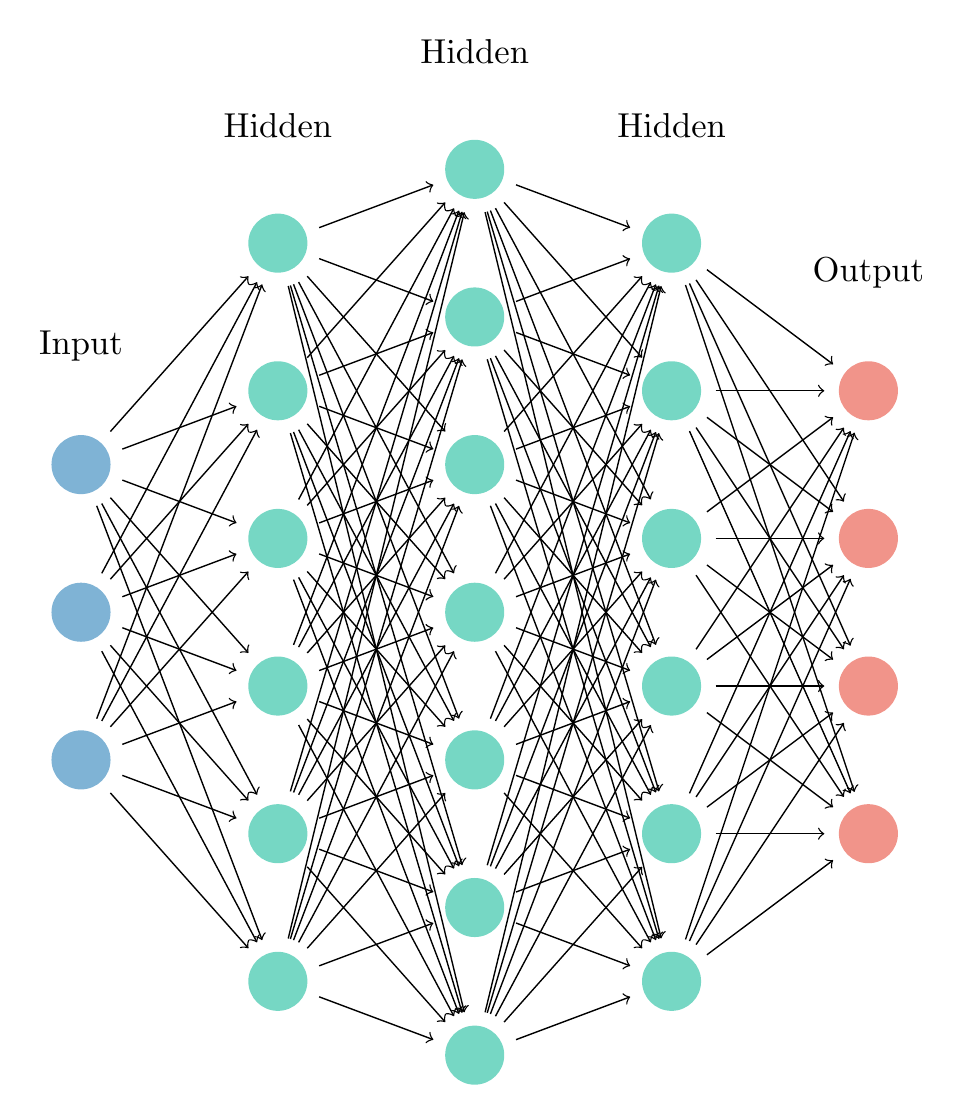}
    \caption{Example of Multi-Layer Perceptron with 3 hidden layers for a 2-feature input and a 4-feature output.}
    \label{fig:mlp}
\end{figure}

The \gls{mlp} was designed to tackle the linear limitation of the Perceptron. It is made up of a succession of neuronal layers, which are arranged one after the other; connections are only made between adjacent layers, and always from the input to the output, thus the name \gls{ffn}. More precisely, in an \acrshort{mlp}, the layers are fully-connected; it means that each neuron of a layer is connected to all the neurons of the previous layer. An example of a \acrshort{mlp} with 3 hidden layers is represented in Figure \ref{fig:mlp}. Activation functions are used at the output of each layer. The introduction of non-linear activation functions enables to overcome the major drawback of the original Perceptron. \acrshort{mlp} was the most popular and simple neural network architecture before the advent of deep learning. It was common to only use a single hidden layer. Nowadays, deep learning mainly refers to the use of neural networks made up of several neuronal layers. The number of neuronal layers corresponds to the depth of the network.

\begin{figure}[ht!]
    \centering
    \includegraphics[width=0.9\linewidth]{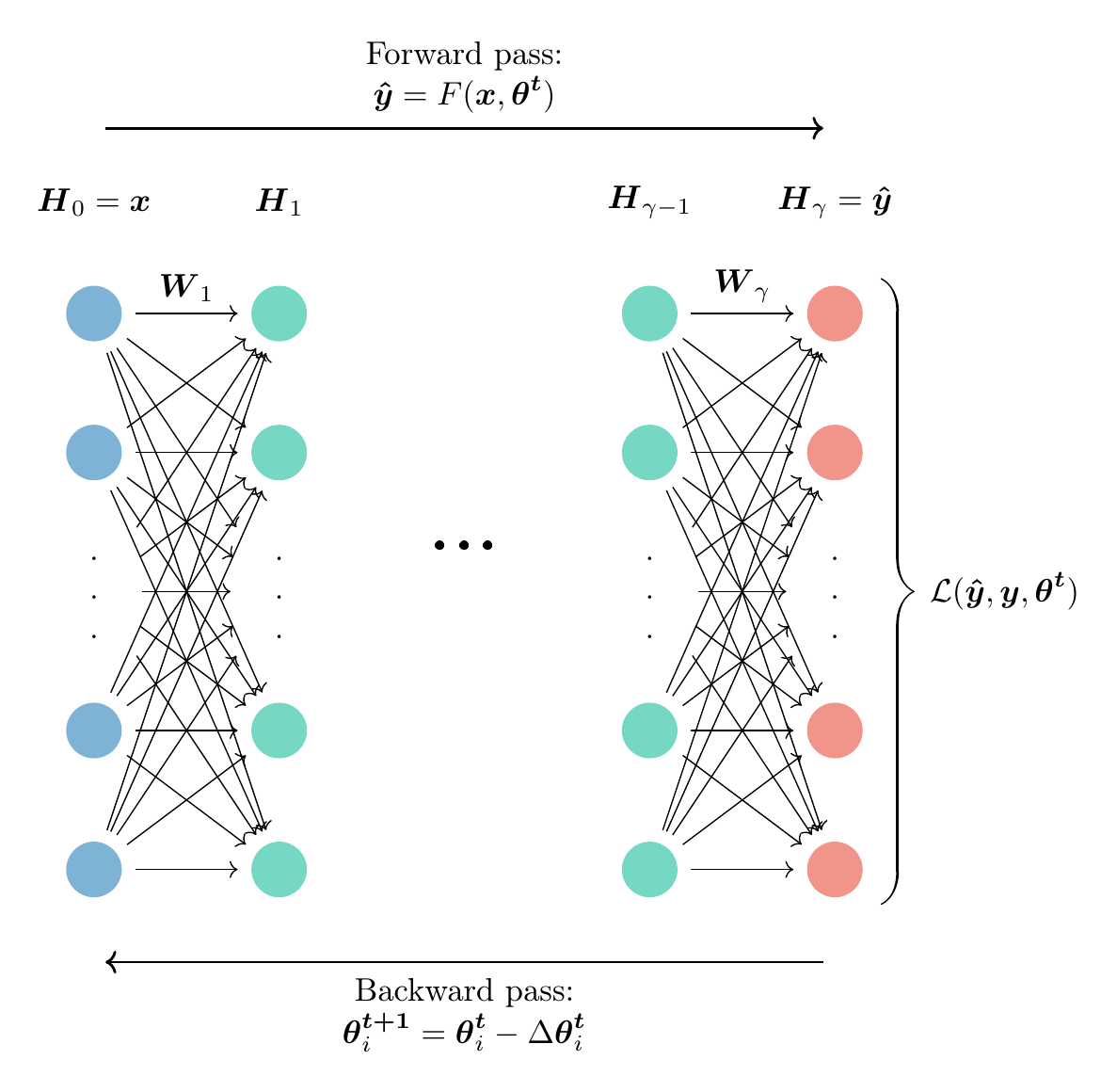}
    \caption{Training of the Multi-Layer Perceptron (generalization for $\gamma$ layers).}
    \label{fig:nn}
\end{figure}

More generally, an \acrshort{ffn} can be represented as a stack of $\gamma$ neuronal layers.  Such a system is depicted in Figure \ref{fig:nn}. 
Each neuronal layer $\mb{H}_i \in \mathbb{R}^{n_i}$ is the result of an operation performed on the neurons of the previous layer $\mb{H}_{i-1}$. This operation can imply some variables, known as trainable weights (or trainable parameters). These weights can be represented as a matrix $\mb{W}_i \in \mathbb{R}^{n_i \times n_{i-1}}$. But it can also be a parameter-free operation, such as an average computation. 
The aggregation of all these elementary functions enables to model a complex function. To make the notations simpler, we denote $\mb{\theta}$ the vector composed of all the weights from all the layers ($\{\mb{W}_1, ..., \mb{W}_\gamma \}$) . 

\subsubsection*{Training}

Training a neural network consists in finding the optimal weights $\mb{\theta}^*$ that minimize the errors made by the model. Deep learning approaches rely on gradient-based algorithms, as many other machine learning approaches. Indeed, \gls{dl} approaches use gradient backpropagation algorithms \cite{BackProp}, known as optimizers, in order to iteratively improve the weights through examples \textit{i. e.} to modify the values of each weight in such a way that the system improves its predictions. This falls into the supervised learning paradigm. 

It implies the use of a dataset $\mset{D}_\mathrm{train}$ composed of couples $(\mb{x},\mb{y})$, where $\mb{x}$ are the inputs and $\mb{y}$ are their respective expected output (known as the ground truth).
One must then define a loss function $\mset{L}$. It must be a differentiable function with respect to the parameters $\mb{\theta}$, measuring the prediction error for a given sample \textit{i. e.} the distance between the prediction of the model $\mb{\hat{y}}$ and the expected output $\mb{y}$. Mean Square Error (MSE) and \gls{ce} are losses that are commonly used for regression and classification tasks, respectively. The average of the loss functions over the whole training dataset is usually taken as the objective function to minimize. We refer to it as the cost function $\mset{C}$.
To sum up, the goal is to find:
\begin{equation}
    \mb{\theta}^* = \arg \min_{\mb{\theta}} \mset{C}(\mset{D}_\mathrm{train},\mb{\theta}) = \arg \min_{\mb{\theta}} \displaystyle \sum_{(\mb{x},\mb{y}) \in \mset{D}_\mathrm{train}}\mset{L}(\mb{x},\mb{y},\mb{\theta}),
\end{equation}
which is solved by gradient descent. 

The gradient descent algorithm consists in iteratively updating the weights $\mb{\theta}_i$ so as to decrease the loss function. $\mb{\theta}\mt{t}{i}$ denotes the weight $i$ at iteration $t$. By iteration, we mean the process of updating the weights. The weights update consists in increasing the weights $\mb{\theta}\mt{t}{i}$ by a fraction of the opposite of the gradient vector of the loss function $\mset{L}$ with respect to $\mb{\theta}\mt{t}{i}$, so as to get closer to the minimum of $\mset{L}$. This gradient vector can be computed on a single example or averaged over many examples, known as a batch:
\begin{equation}
\mb{\theta}\mt{t+1}{i} = \mb{\theta}\mt{t}{i} - \Delta \mb{\theta}\mt{t}{i},
\end{equation}
with
\begin{equation}
\label{eq:weight_diff}
\Delta \mb{\theta}\mt{t}{i} = \eta \cdot \frac{1}{\mathrm{card}(\mset{X}_\mathrm{batch})} \cdot \displaystyle \sum_{(\mb{x},\mb{y})\in (\mset{X}_\mathrm{batch},\mset{Y}_\mathrm{batch})} \frac{\delta \mset{L}(\mb{x}, \mb{y},\mb{\theta}\mt{t}{})}{\delta \mb{\theta}\mt{t}{i}},
\end{equation}
where $\eta$ is a fixed learning rate. This is known as the backward pass.

To compute this gradient vector, one must first compute the prediction $\mb{\hat{y}}$ based on the input $\mb{x}$ and on the current weights $\mb{\theta}\mt{t}{}$, this is the forward pass:
    \begin{equation}
    \hat{y} = F(\mb{x}, \mb{\theta}\mt{t}{}).
    \end{equation}

In addition, this weight update must be computed for each single weight of the neural network, no matter the neural layer. This is made possible by computing the gradient layer by layer, using the chain rule between two successive layers:
\begin{equation}
\label{eq:chain_w}
    \frac{\delta \mset{L}}{\delta \mb{\theta}_i} = \frac{\delta \mset{L}}{\delta \mb{H}_i} \cdot \frac{\delta \mb{H}_i}{\delta \mb{\theta_i}},
\end{equation}
with
\begin{equation}
\label{eq:chain_h}
    \frac{\delta \mset{L}}{\delta \mb{H}_{i}} = \frac{\delta \mset{L}}{\delta \mb{H}_{i+1}} \cdot \frac{\delta \mb{H}_{i+1}}{\delta \mb{H}_{i}},
\end{equation}
thus the name error gradient backpropagation, or backward pass.

Algorithm \ref{alg:optimizer} sums up these different steps by presenting the standard optimizer to solve this problem. Weights are randomly initialized. The algorithm consists in going through the whole training dataset, again and again, successively updating the weights. These loops are known as epochs. The process stops when a stopping condition is met. Generally, it is a threshold on the variation of the value of a loss or metric through the last epochs; but it can also be a fixed number of epochs or training time. Between each epoch, training samples are shuffled to avoid bias during training. The update of the weights is carried out after processing $N$ examples; $N$ is called the mini-batch size. The choice of $N$ is discussed in section \ref{section:optimizer}. Processing a labeled example $(\mb{x},\mb{y})$ consists in two steps: the forward pass and the backward pass, as defined previously.

\begin{algorithm}[ht!]
\caption{Standard error gradient backpropagation optimizer.}
\label{alg:optimizer}
\SetKwInOut{Input}{input}
\Input{$F$, initialized with random weights $\mb{\theta}\mt{0}{i}$,\\
Learning rate $\eta$,\\
Dataset $\mset{D_\mathrm{train}} = (\mset{X}_\mathrm{train},\mset{Y}_\mathrm{train})$}
$t \gets 0$ \\
\While{stopping criterion not reached}{
    $(\mset{X}_\mathrm{shuffled},\mset{Y}_\mathrm{shuffled}) \gets \mathrm{shuffle}(\mset{X}_\mathrm{train},\mset{Y}_\mathrm{train})$\\
    \For{$(\mset{X}_\mathrm{batch},\mset{Y}_\mathrm{batch}) \in \mathrm{split}(\mset{X}_\mathrm{shuffled},\mset{Y}_\mathrm{shuffled})$}{
        $\Delta \mb{\theta}\mt{t}{i} \gets 0 \;\forall i$ \\
        \For{$(\mb{x},\mb{y}) \in (\mset{X}_\mathrm{batch},\mset{Y}_\mathrm{batch})$}{
            $\mb{\hat{y}} \gets F(\mb{x}, \mb{\theta}\mt{t}{})$ \\
            $\Delta \mb{\theta}\mt{t}{i} \gets \Delta \mb{\theta}\mt{t}{i} + \frac{\delta \mset{L}(\mb{x}, \mb{y},\mb{\theta}\mt{t}{})}{\delta \mb{\theta}\mt{t}{i}} \;\forall i$ \\
        }
        $\Delta \mb{\theta}\mt{t}{i} \gets \eta \frac{\Delta \mb{\theta}\mt{t}{i}}{\mathrm{card}(\mset{X}_\mathrm{batch})} \;\forall i$\\
        $t \gets t +1$ \\
        $\mb{\theta}\mt{t}{i} \gets \mb{\theta}\mt{t-1}{i} - \Delta \mb{\theta}\mt{t-1}{i} \;\forall i$ \\
    }
}
\end{algorithm}

This iterative optimization algorithm seems relatively simple. However, training a deep neural network faces three main difficulties:
\begin{itemize}
    \item Ensuring generalization. The network is trained to minimize the cost function on a training dataset. The model must learn enough from these annotated examples to perform well on unseen data. However, it should not learn too much in order to avoid learning the training data exactly, without capacity to generalize on unseen data.
    \item Converging to the global minimum of $\mset{C}$ is the main objective, \textit{i. e.} reaching the optimal weights $\mb{\theta}^*$. Real-world tasks involve non-convex optimizations, and reaching the global minimum is not guaranteed. This way, one tries to converge as efficiently as possible. It means converging towards a local minimum near the global minimum, as fast as possible. 
    \item Dealing with vanishing and exploding gradients. Neural networks are getting deeper and deeper in order to approximate more complex functions. This leads to vanishing or exploding gradients for the lowest layers of the network, implying a stagnation of the weights or overflows, respectively.
\end{itemize}

In the following, we discuss the main techniques that have helped to overcome or mitigate these various challenges, leading to more stable, robust and efficient deep networks.

\subsection{Generalization}
\begin{figure}[h]
    \centering
    \begin{subfigure}{0.3\textwidth}
    \includegraphics[width=\textwidth]{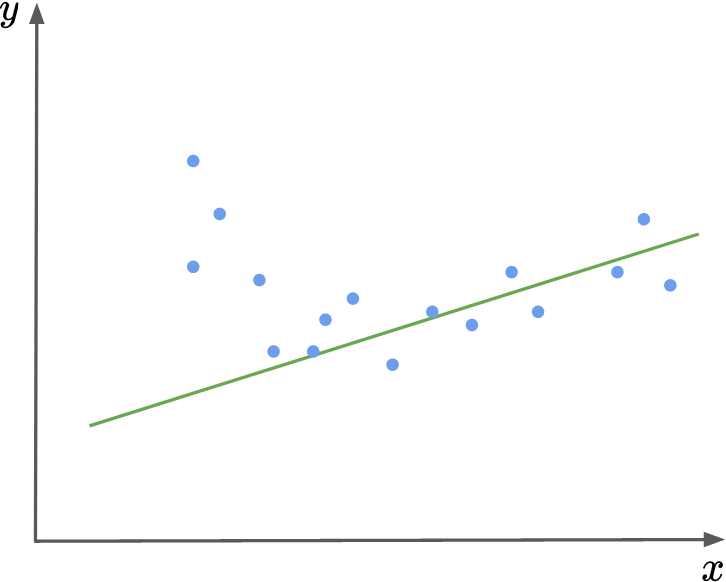}
    \caption{Underfitting.}
    \label{fig:underfit}
    \end{subfigure}
    \hfill
    \begin{subfigure}{0.3\textwidth}
    \includegraphics[width=\textwidth]{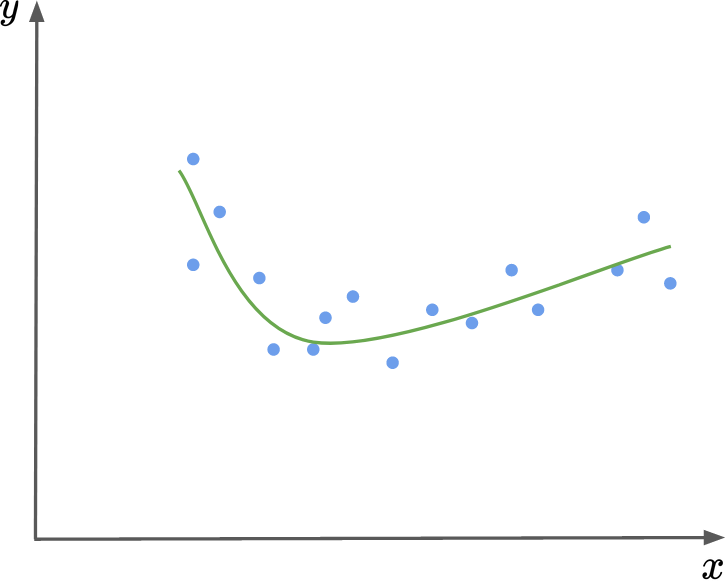}
    \caption{Generalization.}
    \end{subfigure}
    \hfill
    \begin{subfigure}{0.3\textwidth}
    \includegraphics[width=\textwidth]{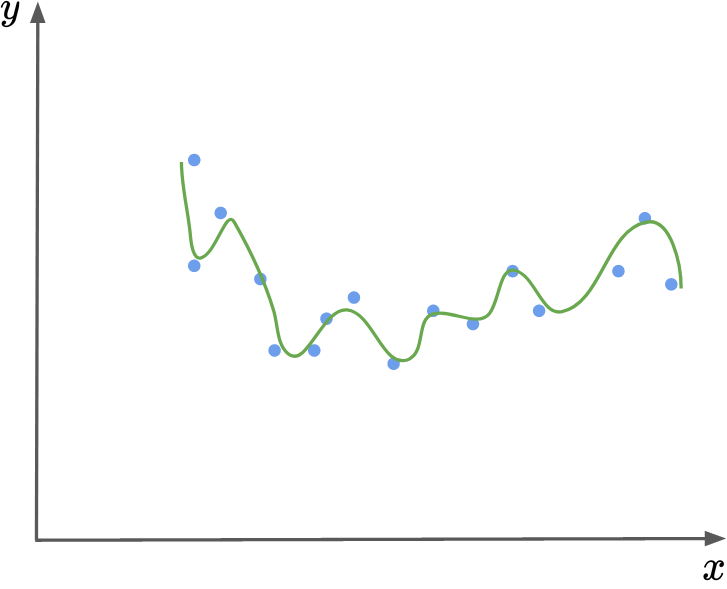}
    \caption{Overfitting.}
    \label{fig:overfit}
    \end{subfigure}
    \caption{Function approximation on the training dataset. The green curves represent the approximated functions, and the blue dots correspond to the annotated training data.}
    \label{fig:fit}
\end{figure}

So far, we have seen the global approach to minimize the cost function over the training dataset, through gradient descent. However, the real purpose of the network is to generalize on data never seen during training. Here comes the bias-variance dilemma \cite{Geman1992}. 
It is defined as the conflict induced when trying to minimize these two sources of errors:
\begin{itemize}
    \item The bias error. It corresponds to the error made from wrong assumptions of the model. A high bias is characterized by a poor modeling of the relationship between inputs and outputs. It is called underfitting.
    \item The variance error. It refers to errors induced by a sensitivity to small fluctuations in the training dataset. A high variance results in modeling noise from the training dataset, which is not relevant. This phenomenon is called overfitting.
\end{itemize}
These two cases are depicted in Figure \ref{fig:fit}, with a simple case of monovariate linear regression task. As one can note, the aim is to have a good generalization property \textit{i. e.} a trade-off between bias and variance to adapt as best as possible to unseen data.

Choosing the network complexity \textit{i. e.} the right number of parameters can be seen as the main issue. Indeed, the underfitting phenomenon depicted in Figure \ref{fig:underfit} can be explained by a too low number of trainable parameters (only two for a straight line). In contrast, too many parameters result in learning the training data too much, which leads to overfitting (Figure \ref{fig:overfit}). Generalization is obtained when choosing a consistent number of parameters with respect to the complexity of the task. However, this value cannot be known in advance and one generally chooses a number of parameters high enough to avoid underfitting. The overfitting issue is commonly handled by using regularization techniques and a specific criterion to stop the learning process.

Another key point regarding the generalization issues is the amount of available labeled data: the more training data, the less overfitting. This is all the more true as the number of parameters increases. 
However, the annotation of the data is mainly a hand-made process, which is costly. This way, it is crucial to deal with as few manually labeled data as possible. 

In the following, we present the main techniques we mentioned to deal with the generalization issues.

\subsubsection{Regularization}
Neural networks are prone to overfitting \textit{i. e.} the model learns features from the training set that are not relevant for the task, leading to poor generalization to unseen data. Regularization techniques aims at reducing this phenomenon. 

A first regularization technique consists in adding a regularization term on the weights to the loss function $\mset{L}$. The idea is that one does not know in advance the complexity of the model one wants to approximate trough the neural network. By default, one chooses a model with a high complexity \textit{i. e.} with a high number of parameters. However, the use of too many parameters amplifies the overfitting phenomenon. Intuitively, one would like to add a penalization on the number of parameters used in the model; but the 0-norm is not differentiable and cannot be used. A way to counteract this issue is to use the L1 and L2 regularizations \cite{Andrew2004}, in order to introduce sparsity, with some weights tending to 0. This way, it gets a similar impact as controlling the number of parameters of the network, while keeping a differentiable cost function.

L1 regularization applies 1-norm to the weights, it is defined as follows:
\begin{equation}
    \mset{L}_\mathrm{L1}(\mb{x}, \mb{y}, \mb{\theta}) = \mset{L}(\mb{x}, \mb{y}, \mb{\theta}) + \lambda \displaystyle \sum_j |\mb{\theta}_j|,
\end{equation}
where $\lambda$ is the regularization parameter.
In the same way, L2 regularization applies 2-norm to the weights:
\begin{equation}
    \mset{L}_\mathrm{L2}(\mb{x}, \mb{y}, \mb{\theta}) = \mset{L}(\mb{x}, \mb{y}, \mb{\theta}) + \lambda \displaystyle \sum_j \mb{\theta}_j^2.
\end{equation}

The introduction of sparsity aims at mimicking a network with fewer parameters. It leads to less complex models, avoiding them to overfit too much. However, L1 and L2 regularizations do not prevent overfitting totally. Moreover, they do not solve the co-adaptation issue.

Co-adaptation is the phenomenon whereby all neuronal connections do not carry the same amount of information. This leads to some connections being very prevalent and others insignificant in the decision. Dropout \cite{Dropout} was proposed in 2014 as a new regularization approach to tackle this issue. Dropout is a technique which is used at training time only. It consists in randomly zeroing some elements of a hidden neural layer, given a probability $\tau$, using a Bernoulli distribution. To compensate the introduction of zeros, the others values are scaled by $\displaystyle \frac{1}{1-\tau}$ to preserve the behavior between training time and evaluation time. This way, Dropout pushes all neurons to contribute to the prediction since a part of them is randomly ignored at each training step, making the network more robust.
Spatial Dropout (or 2D Dropout) \cite{SpatialDropout} is a similar approach proposed for the case of \acrshort{cnn} where standard Dropout is less efficient since convolutional kernels are applied on neighbors, which may be strongly correlated. With Spatial Dropout, entire feature maps are ignored to prevent the network from using these correlations. 

Even using regularization techniques, the model still tends to overfit through the epochs, notably due an excessive over-parameterization. One way to tackle this issue is to stop the training process before learning too much the training samples.

\subsubsection{Stopping criterion}
\begin{figure}[h]
    \centering
    \includegraphics[width=0.4\linewidth]{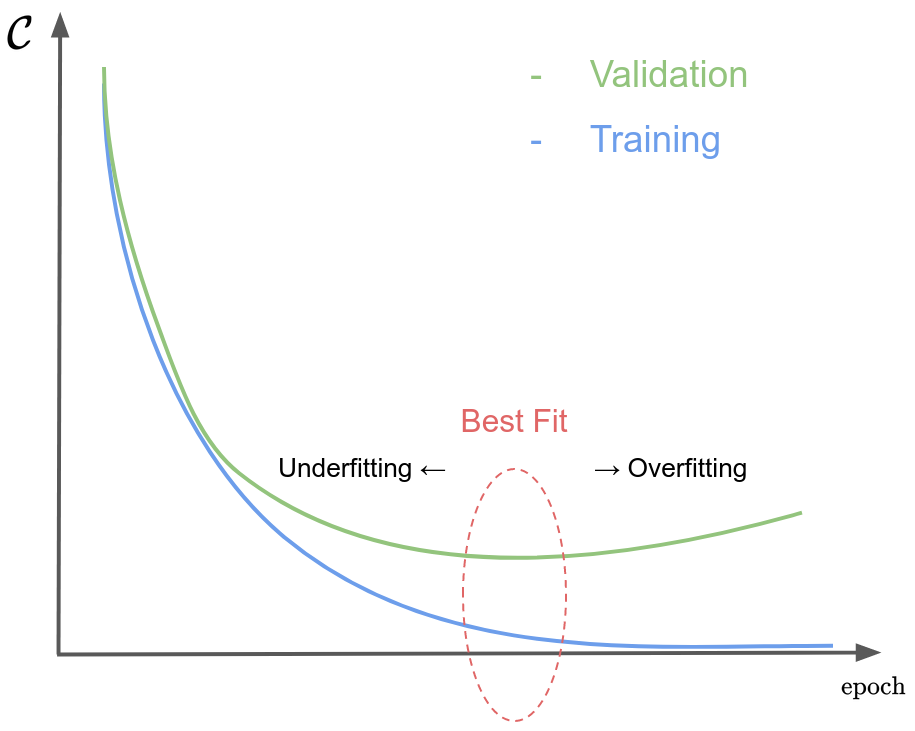}
    \caption{Stopping criterion. The chosen weights are those of the epoch that precedes the rise of the validation curve.}
    \label{fig:select_weight}
\end{figure}

The network is trained on a training dataset $\mset{D}_\mathrm{train}$. But the aim is to have the best generalization over unseen data. To this end, the original dataset $\mset{D}$ is generally  split into three sub-datasets: $\mset{D}_\mathrm{train}$, $\mset{D}_\mathrm{validation}$ and $\mset{D}_\mathrm{test}$. The model weights are updated to minimize the prediction errors with respect to $\mset{D}_\mathrm{train}$.  During training, one regularly estimates the error on $\mset{D}_\mathrm{validation}$, which contains samples unseen from the model during training. This enables to select the weights that best generalize on these unseen examples. As shown on Figure \ref{fig:select_weight}, one should select weights just before the validation curve goes back up. The validation curve to look at can also be a metric curve. Indeed, metrics are similar to losses in that they evaluate a distance between prediction and ground truth. But metrics do not need to be differentiable since they are not used in the backpropagation process. This way, they can be designed with more flexibility and give a better idea of the performance of the model. 

However, this validation error estimation is not sufficient since it is biased. Indeed, we selected the weights for which the error was the lowest on these examples. In addition, hyperparameters are tuned to reach the lowest error on this sub-dataset. That is why a better estimation of the performance of the model on never-before-seen data is computed on $\mset{D}_\mathrm{test}$.

Combining regularization techniques with this stopping criterion enables to alleviate the overfitting phenomenon. But it can be not sufficient if one does not have enough labeled data to train the model.

\subsubsection{Dealing with few training data}
A key point to reach a good convergence, avoiding overfitting and reaching great results, is the amount of available labeled data. There are several ways to deal with few labeled data.

\subsubsection*{Data augmentation}
Data augmentation is a simple technique that enables to augment the number of training data artificially.
It consists in applying transformations on the input data while preserving the relevant information \textit{i. e.} the associated ground truth remains the same. The idea is to generate new training samples without additional cost. Moreover, it enables the model to determine which features are important and which are not since, for a given sample, the different data augmentation techniques generate different inputs while the expected outputs are the same. These data augmentation techniques are as varied as there are tasks to solve. For instance, color distortion, Gaussian noise addition, or rotation are common data augmentation techniques used for computer vision tasks. The data augmentation techniques we used for \gls{htr} are described in Appendix \ref{section-data-augmentation}.

Using synthetic data is another way to artificially augment the number of training data. Synthetic data are computer-generated examples based on heuristics or generative models. The aim is to produce correct couples $(\mb{x},\mb{y})$ as realistic as possible for a given task without any human cost for annotation. 

\subsubsection*{Transfer learning}
\begin{figure}[h]
    \centering
    \includegraphics[width=0.8\linewidth]{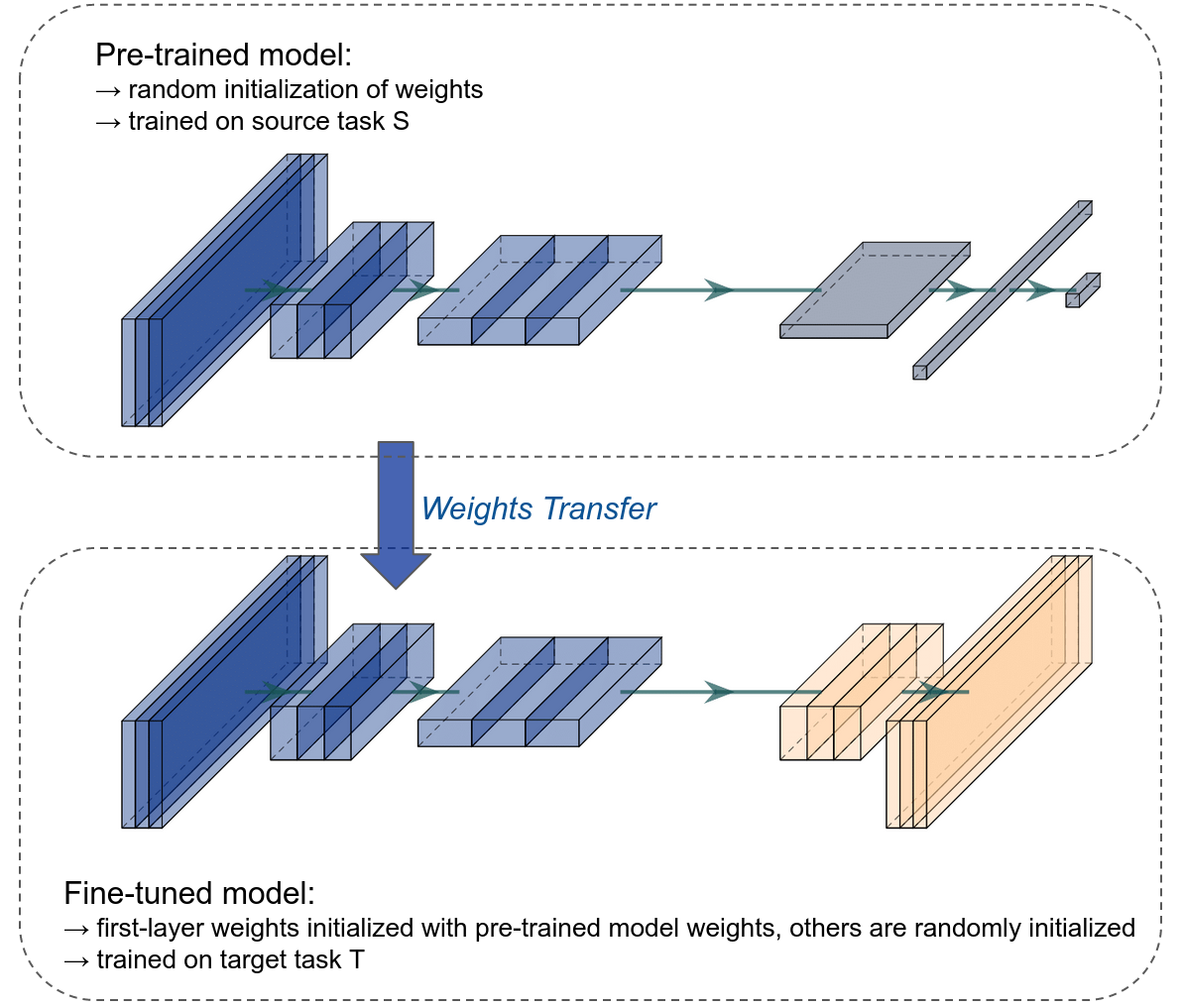}
    \caption{Transfer learning concept. Only the layers in blue are pre-trained for the fine-tuned model.}
    \label{fig:transfer_learning}
\end{figure}

Transfer learning is another way to avoid overfitting. Indeed, in many cases, one can face a lack of data for a target task T. Transfer learning is a way to reduce this issue by transferring knowledge from a network trained on a source task S correlated with T. This notion is illustrated in Figure \ref{fig:transfer_learning}. 

It consists in two steps:
\begin{itemize}
    \item Pre-training: training a model on a source task S, whose weights are randomly initialized.
    \item Fine-tuning: training another model on a target task T, with a part of the weights initialized with the weight values of the pre-trained model.
\end{itemize}
It implies that both networks share, at least in part, the same architecture. It usually corresponds to the lowest layers of the network. The closer the tasks, the more effective this technique. As a matter of fact, transfer learning is inspired by human learning behavior. Someone who knows how to play a musical instrument will learn another instrument more easily than someone who has never played any instrument. On the other hand, the same observation can be made for learning a new score, for the same instrument. This way, transfer learning can also be applied for a same task, if we have a source dataset, different from the target one.

\subsubsection*{Multi-tasking}
Multi-tasking is close to transfer learning. Indeed, the aim is to benefit from learning of one or many other tasks. This induces the use of several models that share some of their weights. The main difference with standard transfer learning strategies is that the different models are trained jointly and not one after the other. The models can either be learned by changing the task between each weight update, or be learned together via a combination of the different losses if the input is the same for the different tasks. It also enables to reduce the total amount of trainable weights. This have been applied in many fields such as \gls{nlp} \cite{MultiTaskingNLP}, combining six tasks, namely \gls{ner}, part-of-speech tagging, chunking, semantic role labeling, language modeling and semantically related words or \acrshort{cv} with an object detection approach \cite{MultitaskingRCNN}.

\subsubsection*{Semi-supervised learning}
Semi-supervised learning \cite{SemiSupervised} is another way to deal with small datasets. It is a special case where one has many unlabeled data (without human annotation), in addition to some annotated data. The aim is to take benefits from the use of the unlabeled data. One way to do this is to first train a model with the annotated data until reaching rather good results. Then, the model generates predictions for part of the unseen unlabeled data. If the predictions are judged correct enough according to a given criterion, they are used as pseudo labels, leading to new training samples. This process is repeated while the performance is increasing.

\subsubsection*{Self-supervised learning}
Self-supervised learning is a specific kind of pre-training strategy which consists in training part of the model from unlabeled data, still using a supervised learning strategy. The idea is to produce artificial labels (known as pseudo labels) from the input itself automatically \textit{i. e.} without any human effort to generate the annotations. Those labels are then used for pre-training, this is also known as training on a pretext task. Auto-encoders \cite{AutoEncoder} are an example of self-supervised learning. It consists in reconstructing the input signal. It means that the expected output (label) is the input. Another example is the completion of sequences of tokens whose some parts are masked, as in \cite{BERT}. 

A recent trend among the self-supervised learning strategies is contrastive learning \cite{ContrastiveSurvey}. It is based on the idea of learning similarities and dissimilarities between examples. It generally consists in applying different data augmentation strategies on a same input. This training aims at learning to bring closer representations that come from the same input and move away those of examples coming from different inputs, or containing different information given a desired task. 

\subsection{Converging efficiently}
We have studied the general gradient descent algorithm used for training and we have seen how to handle the issues related to underfitting and overfitting. We now discuss the convergence issues. Indeed, the aim is to converge \textit{i. e.} to reach the global minimum of the cost function over the training dataset by updating the values of the weights. The main issue is to find how to modify the values of the weights in the right proportion, \textit{i. e.} to go quickly towards the local minimum. 

The Newton's method was proposed to tackle this problem. It consists in iteratively minimizing the derivative of the loss through its second-order Taylor approximation:
\begin{equation}
    \mset{L}(\mb{\theta}\mt{t}{} + \epsilon) \approx \mset{L}(\mb{\theta}\mt{t}{}) + \nabla \mset{L}(\mb{\theta}\mt{t}{}) \epsilon + 
    \nabla^2 \mset{L}(\mb{\theta}\mt{t}{}) \epsilon^2
\end{equation}
with $\mb{\theta}\mt{t+1}{} = \mb{\theta}\mt{t}{} + \epsilon$.

To simplify the notations in the following, we consider a batch size of 1:
\begin{equation}
\mb{g}\mt{t}{i} = \frac{1}{\mathrm{card}(\mset{X}_\mathrm{batch})} \cdot \displaystyle \sum_{(\mb{x},\mb{y})\in (\mset{X}_\mathrm{batch},\mset{Y}_\mathrm{batch})} \frac{\delta \mset{L}(\mb{x}, \mb{y},\mb{\theta}\mt{t}{})}{\delta \mb{\theta}\mt{t}{i}} = \frac{\delta \mset{L}}{\delta \mb{\theta}\mt{t}{i}}.
\end{equation}

Based on Equation \ref{eq:weight_diff}, this parameter-free approach (\textit{i. e.} which does not involve the learning rate parameter), leads to the following formula for the weight update:
\begin{equation}
    \Delta \mb{\theta}\mt{t}{i} = \displaystyle  \frac
    {\mb{g}\mt{t}{i}}
    {\displaystyle \frac{\delta^2 \mset{L}}{\delta {\mb{\theta}\mt{t}{i}}^2}}.
\end{equation}
The Newton's method provides a better approximation of the cost function than the first-order approximation used so far in the standard gradient descent algorithm. Moreover, there is not any learning rate to tune in the equation; the step is given by the inverse of the Hessian. However, this approach requires a cost function twice differentiable and the involved computations are not tractable for training on real tasks. As a consequence, one must use an optimizer which includes a learning rate, which involves many issues.

\begin{figure}[h]
    \centering
    \begin{subfigure}{0.3\textwidth}
    \includegraphics[width=\textwidth]{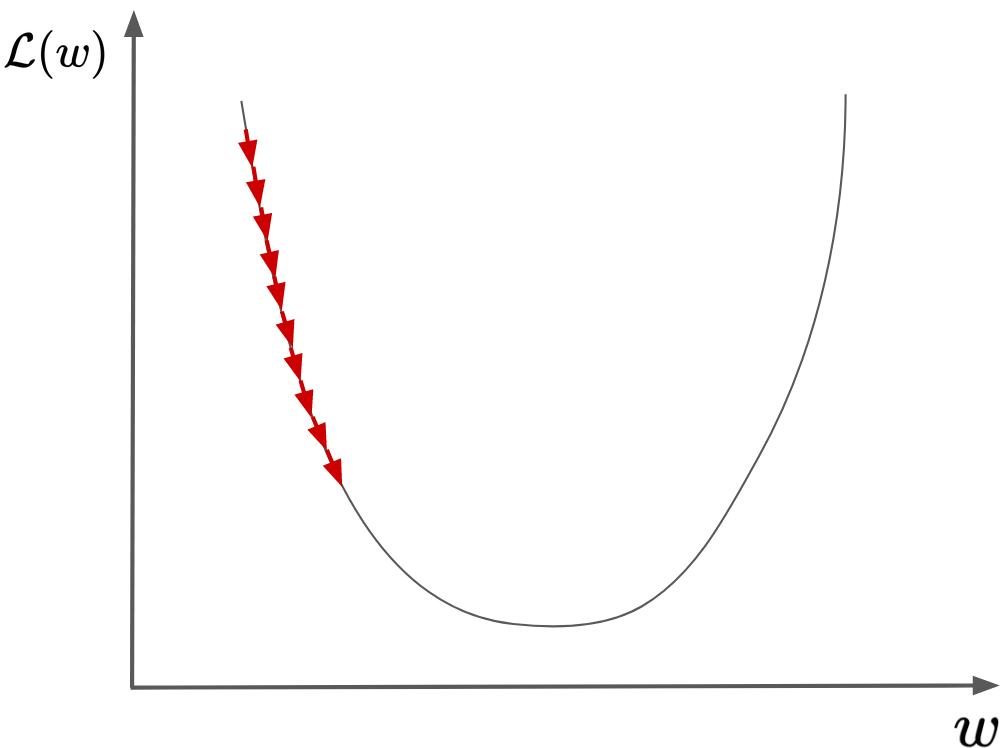}
    \caption{Learning rate too small.}
    \end{subfigure}
    \hfill
    \begin{subfigure}{0.3\textwidth}
    \includegraphics[width=\textwidth]{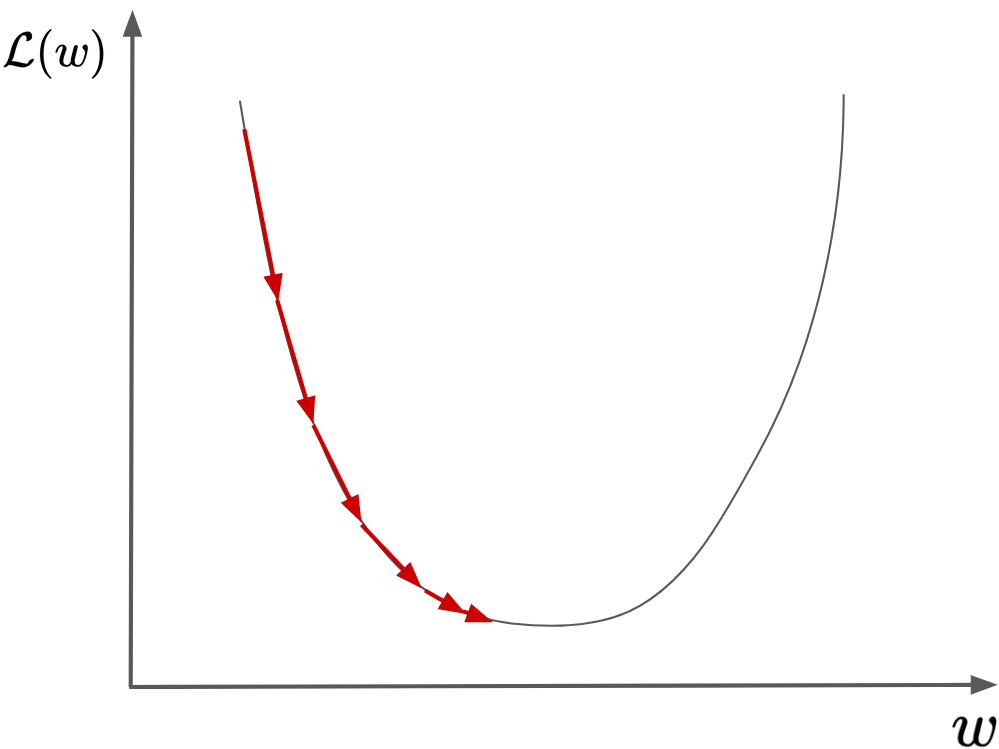}
    \caption{Adequate learning rate.}
    \end{subfigure}
    \hfill
    \begin{subfigure}{0.3\textwidth}
    \includegraphics[width=\textwidth]{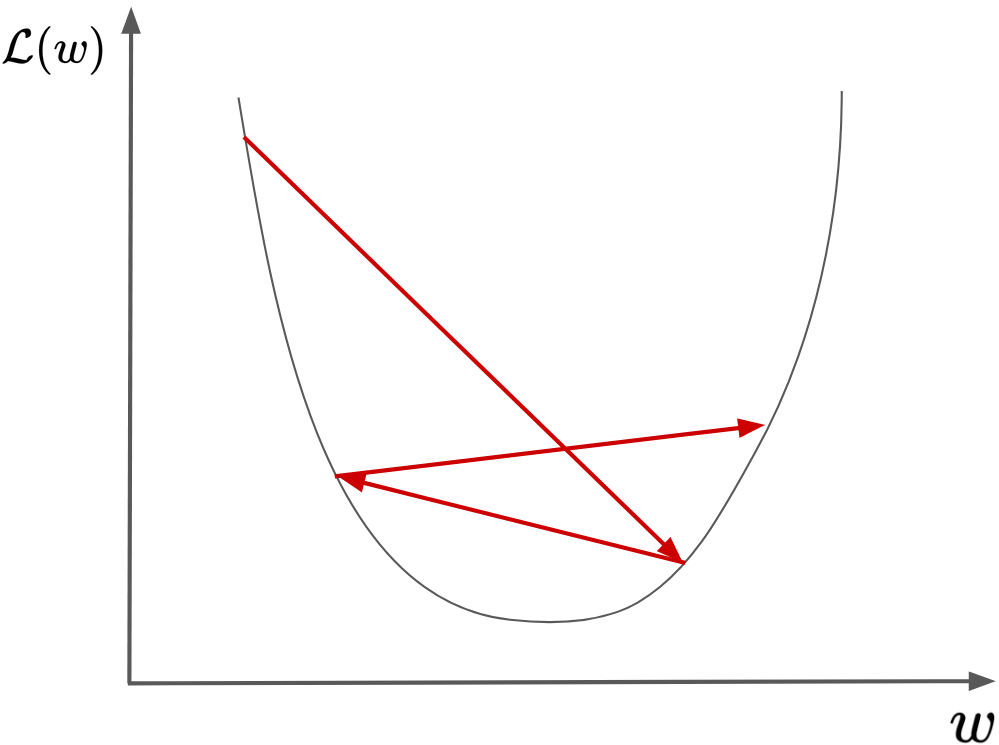}
    \caption{Learning rate too big.}
    \end{subfigure}
    \caption{Learning rate importance. The red arrows indicate the evolution of the weights through the different updates.}
    \label{fig:lr}
\end{figure}

Let's take an easy example with a convex function and only one weight $w$, as depicted in Figure \ref{fig:lr} for different training scenarii. If the learning rate is too small, the convergence is slow. If the learning rate is too big, it prevents the convergence. The adequate learning rate should lead to convergence in few iterations.

As one can see, convergence is inherently dependent on finding the appropriate learning rate. This is why one can improve the convergence by choosing an appropriate optimizer. This optimizer can be combined with a learning rate scheduler. Other techniques, such as curriculum strategies and teacher forcing can be used at training time with the same goal.

\subsubsection{Optimizers}
\label{section:optimizer}
Optimizers are algorithms that update the model's weights in order to minimize the cost function. The first choice is the weight update frequency \textit{i. e.} the number of samples to process between each weight update. As mentioned previously, samples are grouped into mini-batches characterized by their size. The vanilla approach is Batch Gradient Descent (BGD). It consists in updating the weights after each epoch (once all the samples of the training dataset have been processed). It can lead to slow convergence since there is only one update per epoch. A variant is \gls{sgd}, where the update is performed after each sample $(\mb{x},\mb{y})$. The updates are more frequent but it is computationally expensive. Mini-batch Gradient Descent (MGD) is the balance between the BGD and \acrshort{sgd}: the dataset is randomly split into sub-datasets after each epoch, known as mini-batches. The update of the weights occurs after processing each mini-batch. MGD is the most used gradient descent technique.

Another important key point in optimizers is about the learning rate, or more generally, how much to change the current weights with respect to the gradient. Indeed, a too small learning rate will lead to slow convergence and a too high learning rate will prevent reaching the minimum. 

The naive approach is to use a fixed learning rate through the epochs, leading to poor performance. Intuitively, one would need a fairly high learning rate at the beginning to quickly approach the global minimum, and then a lower learning rate as it gets closer to it, to avoid diverging. 

A first proposition was adding momentum to \acrshort{sgd} \cite{SGD_momentum}. The idea is to retain the direction of the previous update, combining it to the new computed gradient. This way, it speeds up convergence and reduces oscillations. Equation \ref{eq:weight_diff} becomes:
\begin{equation} 
    \Delta \mb{\theta}\mt{t}{i} = \displaystyle  \eta \mb{g}\mt{t}{i} + \nu \Delta \mb{\theta}\mt{t-1}{i},
\end{equation}
where $\nu$ is the momentum, a fixed hyperparameter.

In 2011, the Adagrad (Adaptive gradient) algorithm \cite{Adagrad}  was proposed with the idea of a custom learning rate per weight:
\begin{equation} 
    \Delta \mb{\theta}\mt{t}{i} = \displaystyle  \frac{\eta}{\sqrt{\mb{G}\mt{t}{i}+\epsilon}} \cdot \mb{g}\mt{t}{i},
\end{equation}
where $\mb{G}\mt{t}{i}$ is the sum of the square of the gradients up to $t$ for the weight $\mb{\theta}_i$. This way, it performs larger updates for parameters whose past gradients were small. The drawback of this method is that the denominator keeps increasing during training, leading to infinitesimally small weight updates.

To solve this problem, G. Hinton proposed in his lecture notes RMSProp (Root Mean Squared Propagation). It consists in replacing $\mb{G}\mt{t}{}$ by $\mb{D}\mt{t}{}$, the exponentially decaying average of previous squared gradients:
\begin{equation}
    \mb{D}\mt{t}{} = \nu_D\mb{D}\mt{t-1}{} + (1-\nu_D){\mb{g}\mt{t}{}}^2.
\end{equation}

The authors of \cite{Adadelta} noted that there is a unit mismatching issue in the update rule of the previously mentioned algorithms (Adagrad, SGD with momentum, RMSProp). They proposed Adadelta to tackle this problem. They added an exponentially decaying average of squared parameter update term:
\begin{equation}
    \mb{E}\mt{t}{} = \mu_E \mb{E}\mt{t-1}{} + (1-\mu_E)\Delta {\mb{\theta}\mt{t}{}}^2.
\end{equation}

Equation \ref{eq:weight_diff} becomes:
\begin{equation}
    \Delta \mb{\theta}\mt{t}{i} = \displaystyle  \frac{\sqrt{\mb{E}\mt{t-1}{i} + \epsilon}}{\sqrt{\mb{D}\mt{t}{i}+\epsilon}} \cdot \mb{g}\mt{t}{i}.
\end{equation}
One can note that Adadelta does not require any learning rate hyperparameter.
In 2015, Adam \cite{Adam} (Adaptive moment estimation) was proposed as an evolution of the previous algorithms. It is also based on exponentially decaying averages, as for RMSProp and Adadelta. But this time, it is computed for both past squared gradients ($\mb{v}\mt{t}{}$) and past gradients ($\mb{m}\mt{t}{}$):

\begin{equation}
    \mb{m}\mt{t}{} = \beta_1 \mb{m}\mt{t-1}{} + (1- \beta_1)\mb{g}\mt{t}{},
\end{equation}
\begin{equation}
    \mb{v}\mt{t}{} = \beta_2 \mb{v}\mt{t-1}{} + (1- \beta_2){\mb{g}\mt{t}{}}^2.
\end{equation}
The authors noted that $\mb{m}\mt{t}{}$ and $\mb{v}\mt{t}{}$ are biased toward zero due to the initial values $\mb{m}\mt{0}{}=\mb{v}\mt{0}{}=\mb{0}$. They computed $\mb{\hat{m}}\mt{t}{}$ and $\mb{\hat{v}}\mt{t}{}$ to counteract these biases:
\begin{equation}
    \mb{\hat{m}}\mt{t}{} = \frac{\mb{m}\mt{t}{}}{1-\beta_1^t},
\end{equation}
\begin{equation}
    \mb{\hat{v}}\mt{t}{} = \frac{\mb{v}\mt{t}{}}{1-\beta_2^t}.
\end{equation}
It leads to the following equation:
\begin{equation}
        \Delta \mb{\theta}\mt{t}{i} = \displaystyle  \frac{\eta}{\sqrt{\mb{\hat{v}}\mt{t}{i}+\epsilon}} \cdot \mb{\hat{m}}\mt{t}{i}.
\end{equation}

Adam is one of the most used optimizers nowadays. Since then, many alternatives based on this algorithm have been proposed such as AdamW \cite{AdamW}, Nadam \cite{Nadam} or AdaMax \cite{AdaMax}.

On top of these optimizers, one can also add an additional algorithm to control the evolution of the initial learning rate $\eta$, which becomes $\eta\mt{t}{}$. These algorithms are known as learning rate schedulers.

\subsubsection{Learning rate schedulers}
\begin{figure}[h!]
    \centering
    \includegraphics[width=\linewidth]{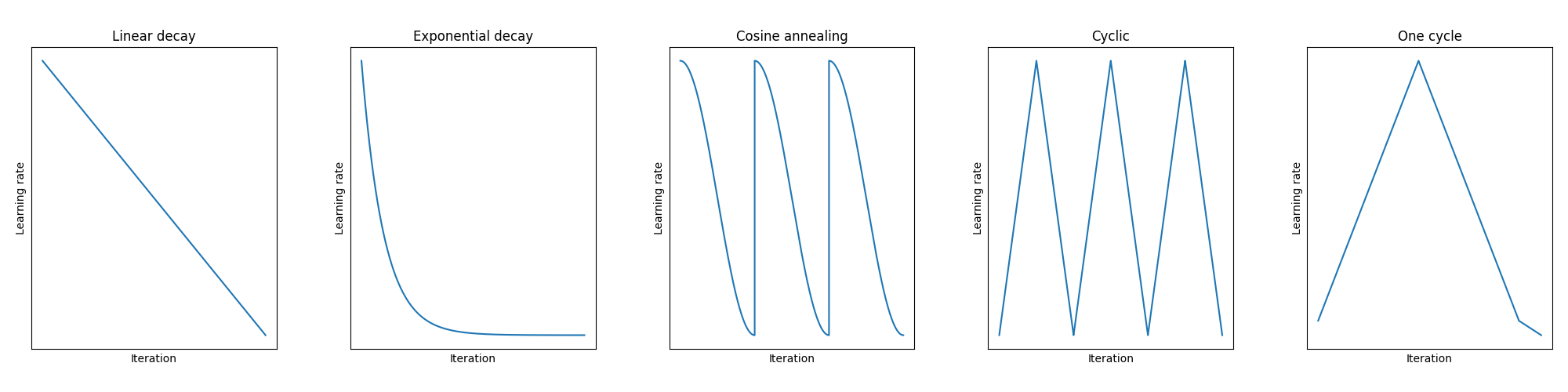}
    \caption{Examples of learning rate scheduler.}
    \label{fig:lrs}
\end{figure}

Learning rate schedulers can be used with any optimizer, even adaptive ones like Adam, as soon as they integrate an initial learning rate. The aim is to have a dynamic learning rate between the successive updates of the weights. For instance, it can enable to have a globally decreasing learning rate, even if there is one learning rate per parameter in the case of adaptive optimizers. Learning rate schedulers can vary the learning rate in many ways. In the following, we describe the most popular learning rate schedulers; they are depicted in Figure \ref{fig:lrs}. The simplest is the linear decay:

\begin{equation}
    \eta\mt{t}{} = \eta\mt{t-1}{} - \gamma ,
\end{equation}
where $\gamma$ is a constant step to decrease the learning rate. $t$ can count the number of weight updates or the number of epochs, depending on what one wants to achieve.
One can linearly vary the $\gamma$ term, so that the learning rate decreases slower and slower.

Another learning rate scheduler consists in exponentially decaying the learning rate:
\begin{equation}
    \eta\mt{t}{} = \gamma \cdot \eta\mt{t-1}{}.
\end{equation}

These two learning rate schedulers are strictly decreasing, mainly to avoid diverging and to reach a minimum. One can define a minimum value for the learning rate or a maximum number of learning rate updates to prevent the learning rate to become too small.
In 2017, the authors of \cite{WarmRestart} proposed the idea of warm restarts in order to avoid getting stuck in local minima or saddle points. They introduce the cosine annealing scheduler, with warm restarts \textit{i. e.} the learning rate decreases following a cosine curve. It gets back to its initial value after $T$ iterations, $T$ being a hyperparameter.

The same year, the authors of \cite{cylicLR} proposed a cyclic learning rate scheduler with the same warm restart idea: the learning rate varies linearly between a minimum value and maximum value, in a cyclic way. The next year, a one-cycle policy was proposed \cite{onecycleLR}, improving by far the convergence. It follows the same concept of linear variation between lower and upper bound, but this is done only once. Then, when the cycle is ended, the learning rate is annealed during the last iterations, with an important decrease. These two last approaches are based on a range test approach to determine the bounded values to choose: a first training is carried out by increasing the learning rate after each weight update, leading to diverging and thus an increasing loss. The maximum learning rate is generally the learning rate before the loss increase. The minimum learning rate is taken as one tenth of the maximum.

Learning rate schedulers can also be chained together, to have a dynamic strategy during training, or be adapted to the evolution of the metrics for instance. Learning rate schedulers can enable to reach better results or simply to converge more efficiently. However, one can note that in most papers, the authors use the Adam optimizer without any learning rate scheduler. Indeed, these schedulers often need to tune some hyperparameters, sometimes very sensitive to the model or the data used. To summarize, adaptive optimizers, like Adam, allow to guarantee a certain convergence without needing to set hyperparameters in a precise way, it is very useful in the research phase. However, for an industrial application, it may be relevant to take the time to find the right hyperparameters for the use of a learning rate scheduler to reach even better performances.

\subsubsection{Curriculum strategies}
Curriculum strategies are inspired by the way humans learn. Curriculum learning is the most known strategy. It consists in learning a task with an increasing difficulty. For instance, one first learns to write isolated letters, words, then sentences before writing whole dissertations. 

In \cite{CurriculumLearning}, the authors show improving results when gradually increasing the difficulty on two different tasks: language modeling, with an increasing vocabulary, and shape recognition, with an increasing shape variability. This idea of curriculum learning is very related to the task and must be adapted on a case-by-case basis.

Curriculum paradigm can also be applied to Dropout for instance: in \cite{CurriculumDropout}, the authors proposed to slightly increase the probability of dropout $\tau$ during training:
\begin{equation}
    \tau\mt{t}{} = (1 - \bar{\tau}) \exp(- \gamma t) + \bar{\tau}, \gamma > 0,
\end{equation}
where $\bar{\tau}$ is the final dropout rate, $t$ is the number of iterations and $\gamma = \frac{1}{T}$ ($T$ being the total estimated number of weight updates during training).
This dropout scheduler showed promising results on various image classification datasets. 

\subsubsection{Teacher forcing}
Teacher forcing is another strategy, applied at training time, to improve the convergence speed of neural networks. This is used for tasks in which the expected output is a sequence of tokens $\mseq{y}$ which are recurrently predicted \textit{i. e.} that the prediction of $\hat{\mseq{y}}\mtseq{t}{}$ depends on the previous predicted tokens $\hat{\mseq{y}}\mtseq{t-i}{},\ i \in \left[0, t-1\right]$. The idea is to provide the ground truth tokens instead of the previous predicted tokens to predict the next token. It has two main advantages. First, the computations can be parallelized during training to predict the whole expected sequence at once, thus reducing the training time per epoch. Second, if one considers training from scratch, the predictions are very poor at the beginning. Giving the ground truth tokens instead of the poor predictions enables the model to make predictions based on realistic previous tokens, reducing its prediction errors and improving the convergence.

However, teacher forcing should not be used all the time. Indeed, if one always provides the ground truth as previous predicted tokens, the model only sees perfect context to make the next prediction and is not robust: it does not generalize at prediction time, when some errors may occur. One way to alleviate this issue is to introduce some errors in the provided ground truth in order to train the model to predict correct tokens even when its previous predictions are wrong. One can also use teacher forcing for the first epochs only, and then train the model with its own predictions in a second step.

Now that we have seen how to handle the generalization and convergence problems, we can now focus on the vanishing and exploding gradient issues.

\subsection{Dealing with vanishing and exploding gradients}
Vanishing and exploding gradients \cite{Pascanu2013} are phenomena appearing in the lowest layers of deep networks. It is induced by the chained rule used during backward propagation (Equation \ref{eq:chain_w} and \ref{eq:chain_h}). Indeed, based on those equations, the formula for the gradient computation for the weights of the lowest layer $\mb{W}_1$ is:
\begin{equation}
\label{eq:chain_vanish}
    \frac{\delta \mset{L}}{\delta \mb{W}\mt{t}{1}} = \frac{\delta \mset{L}}{\delta \mb{H}_1} \cdot \frac{\delta \mb{H}_1}{\delta \mb{W}\mt{t}{1}} = \frac{\delta \mset{L}}{\delta \mb{H}_\gamma} \cdot \left(\displaystyle \prod_{i=2}^\gamma \frac{\delta \mb{H}_i}{\delta \mb{H}_{i-1}}\right) \cdot \frac{\delta \mb{H}_1}{\delta \mb{W}\mt{t}{1}}.
\end{equation}
It includes a multiplication in chain implying more and more terms as one approaches the lowest layers of the network. It is at the origin of the well-known vanishing and exploding gradient issues:
\begin{itemize}
    \item if $ \displaystyle 0 \leq |\frac{\delta \mb{H}_i}{\delta \mb{H}_{i-1}}| < 1, \forall i$, the gradient will tend to zero: this is the vanishing gradient. It means that the weights are not updated anymore for the lowest layers.
    \item  if $ \displaystyle |\frac{\delta \mb{H}_i}{\delta \mb{H}_{i-1}}| > 1, \forall i$, the gradient will tend to infinity: this is the exploding gradient. It will lead to numerical instabilities and overflows, due to hardware limitations.
\end{itemize}

$\displaystyle \frac{\delta \mb{H}_i}{\delta \mb{H}_{i-1}}$ corresponds to the gradient of the output of layer i with respect to its input \textit{i. e.} the gradient of the elementary functions. The main used functions are activation functions and linear layers.

For an activation function $f$, there is no trainable weights involved:
\begin{equation}
    \frac{\delta \mb{H}_i}{\delta \mb{H}_{i-1}} = f'(\mb{H}_{i-1}).
\end{equation}

Linear layers are layers in which each neuron is computed as a weighted sum of the previous neurons: $\mb{H}_i = \mb{W}_i\mb{H}_{i-1}$. Densely-connected layers, fully-connected layers and formal neuron are different names used for linear layers. Its associated gradient is:
\begin{equation}
    \frac{\delta \mb{H}_i}{\delta \mb{H}_{i-1}} = \mb{W}_i.
\end{equation}

To sum up, one must pay attention to the values of the activation functions, and more precisely to their derivatives, as well as those of the trainable weights. In the following, we demonstrate how the choice of the activation functions and of the weight initialization method is important. We also present some other techniques to avoid vanishing and exploding gradient issues: normalization and residual connections.

\subsubsection{Activation functions}
In neural networks, each neuron layer is followed by an activation function (no activation can be seen as identity function). Since real-world tasks are complex, linear activation functions are not sufficient to model adequate functions. This way, non-linear activation functions are mainly used such as tanh, sigmoid or ReLU. 

\begin{figure}[h]
    \centering
    \includegraphics[width=\linewidth]{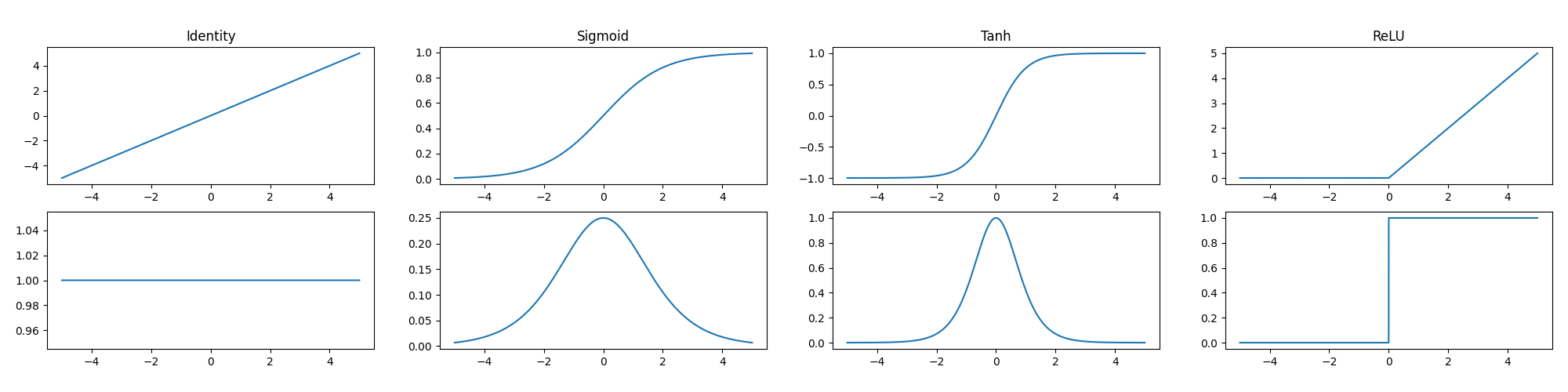}
    \caption{Activation functions (top) and their associated derivatives (bottom).}
    \label{fig:activations}
\end{figure}

Figure \ref{fig:activations} depicts the main activation functions and their corresponding derivatives. As one can note, tanh and sigmoid derivatives lead to bounded values between 0 and 1, and 0 and 0.25, respectively. As we have seen previously, this will lead to vanishing gradient issues. The impact grows with the number of layers in the network. ReLU activation was introduced to tackle this problem, its derivative being either 0 or 1. 

Another important activation function is softmax, which enables to get a probability distribution over an input vector $\mb{x} = \mvecdef{\mb{x}_1, ..., \mb{x}_n}$:
\begin{equation}
    \mathrm{softmax}(\mb{x}_i) = \displaystyle \frac{\exp(\mb{x}_i)}{\displaystyle \sum_{j=1}^n \exp(\mb{x}_j)}.
\end{equation}
It is notably used as last activation function, as a decision layer, in the case of classification problems.

\subsubsection{Weights Initialization}
The initialization of the weights is a crucial element regarding convergence. Let's consider an \acrshort{ffn}. A naive approach would be to initialize all weights to zero, or with a constant value. If one follows this approach, it would lead to each neuron of a same layer computing the exact same value, since their associated weights and their inputs are identical. The associated gradients would also be identical leading to all neurons of a same layer learning the same features. Training would not be efficient at all.

To tackle this problem, one can use a normal $\mset{N}(0, \sigma^2)$ or uniform $\mset{U}(-u,u)$ distribution to bring randomness, and make the different neurons behave differently, thus learning different features. But one should keep in mind that too small weights will lead to vanishing gradient issues, and too high weights to exploding gradient issues, as explained before. A solution is then to choose a weight initialization that enables to preserve the variance from one layer to another, but this  depends on the activation functions used.

In 2010, Xavier Glorot \textit{et. al.} proposed in \cite{XavierGlorot} some hyperparameters for uniform and normal distributions to preserve the variance from one layer to another when using tanh or sigmoid activation functions:
\begin{equation}
   u = \sqrt{\frac{6}{n_\mathrm{in} + n_\mathrm{out}}},
\end{equation}
and
\begin{equation}
   \sigma = \sqrt{\frac{2}{n_\mathrm{in} + n_\mathrm{out}}},
\end{equation}
where $n_\mathrm{in}$ and $n_\mathrm{out}$ are the numbers of neurons in the previous layer and in the next layer, respectively.

In 2015, Kaiming He \textit{et al.} proposed in \cite{KaimingHe} another hyperparameter for the use of the ReLU activation, in the same spirit of preserving the variance from one layer to another:
\begin{equation}
   \sigma = \sqrt{\frac{2}{n_\mathrm{in}}}.
\end{equation}

\subsubsection{Normalization}
Normalization aims at fixing the distribution of the output of a layer by normalizing the values \textit{i. e.} to have 0 mean and unit variance. It is a good way to reduce vanishing gradient when using sigmoid or tanh activation functions for example. As a matter of fact, based on Figure \ref{fig:activations}, 0 mean and unit variance force the derivative of sigmoid and tanh to be far from their 0-boundary: the associated gradient values are bigger, reducing the vanishing gradient issue induced by the chain rule (Equation \ref{eq:chain_vanish}). The authors of \cite{BatchNorm} proposed Batch Normalization to also tackle the internal covariate shift issue. It is defined in \cite{BatchNorm} as the modification in the distribution of network activations due to the update of the network weights during training. 

Given a set of values $\mset{X} = \msetdef{x_1, ...,x_N}$, normalization consists in computing $\mset{\hat{X}} = \msetdef{\hat{x}_1, ...,\hat{x}_N}$ with $\mset{\hat{X}} \sim \mset{N}(\mu_{\hat{X}},\sigma_{\hat{X}}^2)$ so that $\mu_{\hat{X}}=0$ and $\sigma_{\hat{X}}^2=1$. The normalization is carried out in three steps: 

\begin{equation}
	\mu_X = \frac{1}{N} \sum_{i=1}^N {x}_i	,
\end{equation}

\begin{equation}
	\sigma^2_X = \frac{1}{N} \sum_{i=1}^N ({x}_i - \mu_X)^2,
\end{equation}

\begin{equation}
	{\hat{x}}_i = \frac{{x}_i-\mu_{X}}{\sqrt{\sigma^2_X + \epsilon}},
\end{equation}
where $\epsilon$ is an arbitrary positive hyperparameter close to zero, to avoid a division by zero.
An additional step is proposed in this paper to scale and shift the obtained normalized values  respectively by $\gamma$ and $\beta$, two trainable weights:

\begin{equation}
    {y}_i = \gamma {\hat{x}}_i + \beta
\end{equation}

Given an input of shape $(N, L, C)$ where $N$ is the mini-batch size, $L$ is the sequence length ($L=H \times W$ for an image) and $C$ is the number of channels, Batch Normalization consists in applying normalization through the samples of a same mini-batch, but considering the channels independently. Batch Normalization is widely used nowadays; it showed interesting properties such as quicker convergence and regularization, especially for large mini-batches. It also avoids values to grow too much through the layers in the case of ReLU activation for instance. However, it comes at the cost of additional computations, impacting prediction time for example. At training time, the loss due to the additional computations may be compensated by the better convergence.

Running mean and variance is usually computed during training and used at inference time to preserve an adequation between training and prediction. Otherwise, distribution differences between these two configurations could lead to poorer results.

In 2018, the authors of \cite{ICS} showed that the efficiency of the batch normalization is not mainly due to its impact on the internal covariate shift. They suggest that it would be due to its reparametrization of the optimization problem making it more stable and smoother in the sense of loss Lipschitzness and loss $\beta$-smoothness respectively.

\begin{figure}[h!]
	\includegraphics[width=\textwidth]{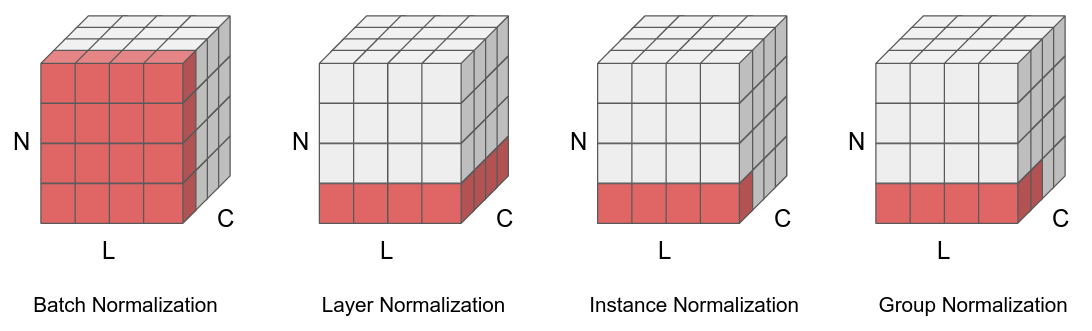}
	\caption{Visualization of the different normalization techniques. The red area indicates the axes on which the normalization is performed (N: the samples, C: the feature maps, L: the frames or pixels). }
	\label{fig:norm}
\end{figure}

In 2016, other normalization techniques have been proposed: Instance Normalization \cite{InstanceNorm} and Layer Normalization \cite{LayerNorm}. They follow the same idea; the key difference is the dimensions on which normalization is performed. Instance Normalization consists in normalizing over the L axis, considering the different channels and the different samples independently. Layer Normalization applies normalization over the channels and the L axis, still considering the samples independently. More recently, Group Normalization \cite{GroupNorm} has been proposed as an in-between, grouping some channels together for the normalization. Normalization techniques are visually compared in Figure \ref{fig:norm}. 

\subsubsection{Residual connections}
\begin{figure}[h]
    \centering
    \includegraphics[width=0.5\linewidth]{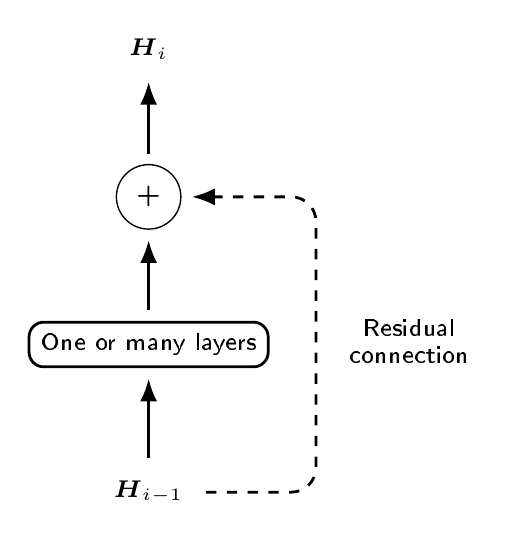}
    \caption{Residual connection between hidden state $\mb{H}_{i-1}$ and $\mb{H}_{i}$.}
    \label{fig:residual}
\end{figure}

Residual connections (or skip connections) have been introduced in \cite{ResNet}, in which the authors proposed the ResNet. A residual connection is a way to introduce features from other part of the network than just the previous layer. Figure \ref{fig:residual} provides an illustration of the concept of residual connection. As one can note, $\mb{H}_{i-1}$ is followed by two sub-pipelines before being merged through an operation. This operation is usually an element-wise addition but it can also be a concatenation or an element-wise multiplication for instance.

The use of residual connections bring two main advantages. First, it aggregates information from different layers, leading to a multi-scale feature extraction, which can lead to better results. Second, it helps reducing the vanishing gradient. Let us take the example of the element-wise addition as merging operation between layer $\mb{H}_{i-1}$ and $\mb{H}_i$, and identity function as second branch. Equation \ref{eq:chain_h} becomes:
\begin{equation}
    \frac{\delta \mset{L}}{\delta \mb{H}_{i-1}} = \frac{\delta \mset{L}}{\delta \mb{H}_{i}} \cdot \left(\frac{\delta \mb{H}_i}{\delta \mb{H}_{i-1}} + 1\right).
\end{equation}
As one can note, it avoids the gradient from decreasing from one layer to another during backpropagation.

\subsubsection{Hardware limitations}
\acrlong{dnn}s are trained with computers, leading to some limitations. Computation time for forward and backward passes are dependent on the hardware used. Modern frameworks (Tensorflow, Pytorch) provide highly optimized algorithms for commonly used operations, notably on \gls{gpu}. Nonetheless, as networks get deeper and deeper and datasets get larger and larger, it can be complicated to reach convergence in a reasonable training time or to use large mini-batches. One way to reduce this issue is to use the concept of automatic mixed precision \textit{i. e.} to perform the operations with 16-bit floats instead of 32-bit floats, when the operation is sufficiently numerically stable. Indeed, it enables to reduce both computation time and \acrshort{gpu} memory consumption in many cases. 

As mentioned previously, exploding gradient induces overflows, due to the limitation of 32-bit floats encoding. One way to counteract this issue is to use gradient clipping. It consists in setting a maximum value that the gradient values cannot exceed.

We showed that neural networks can estimate complex functions, through gradient descent algorithms. We presented the main components and techniques to deal with the generalization and convergence issues as well as the vanishing and exploding gradients. All these techniques enable to have very powerful, stable and deep neural networks, explaining their impressive performance in many tasks. We now study the main components and architectures which enables to tackle computer vision and sequence-to-sequence problems, which will enable us to understand how to handle image-to-sequence tasks such as \gls{htr}.

\section{HTR: a computer vision problem}
\acrfull{cv} refers to the various techniques that enable computers to see and understand the content of images or videos. It includes numerous applications such as image captioning, object detection, facial recognition or image classification. Therefore, \gls{htr} naturally falls into the computer vision field. Indeed, in \gls{htr}, we aim at recognizing the text from an input image $\mb{X}$ of shape $(H, W, C)$. The model must learn to extract relevant features from the input images in order to detect the different characters. We limit ourselves here to study the evolution of the deep learning practices and architectures which made it possible to improve the performances for these computer vision tasks.

\subsection{Multi-Layer Perceptron for computer vision}
\gls{mlp} is one of the first neural network proposed to tackle computer vision problems. However, \acrshort{mlp} presents a main drawback. The input needs to be of fixed size since each neuron of a given layer is connected to all neurons of the previous layer (the input layer included). This way, the number of trainable weights is dependent on the input size. It implies another drawback: the number of weights exponentially increases with the input size. Considering our input image $\mb{X}$, it would require $H \times W \times C \times n$ weights for a first neuron layer containing $n$ neurons. This is shown in Table \ref{tab:mlp_weights} for different input sizes, resolutions (number of pixels per inch) and color encodings, and for a first layer containing 1028 neurons (taken arbitrarily). As one can note, the number of weights is really high (26.8G) for a A4 RGB input with a resolution of 300 \acrshort{dpi}, which represents the majority of paper documents used with standard scanning quality. Decreasing the input size to A6 format (4 times smaller than A4), and the resolution to 100 \acrshort{dpi}, leads to 38 times less parameters required. However, it still represents a high number of trainable parameters, just to consider the first layer. 

\begin{table}[h]
    \centering
    \begin{tabular}{c c c c c c c c}
         \hline
         Format & Resolution & Color & H & W & C & n & \# weights \\
         \hline
         \multirow{4}{*}{A4 (210x297 mm)} & \multirow{2}{*}{300 \acrshort{dpi}} & RGB & 3,508 & 2,480 & 3 & 1028 & 26,8G\\
         & & Gray-scaled  & 3,508 & 2,480 & 1 & 1028 & 8.9G\\
         & \multirow{2}{*}{100 \acrshort{dpi}} & RGB & 1,170 & 827 & 3 & 1028 & 3.0G\\
         && Gray-scaled & 1,170 & 827 & 1 & 1028 &1.0G\\
         \hline
         \multirow{2}{*}{A6 (105x148 mm)} & 300 \acrshort{dpi} & RGB & 1,754 & 1,241 & 3 & 1028 & 6.7G\\
         & 100 \acrshort{dpi} & RGB & 585 & 414 & 3& 1028 & 0.7G\\
         \hline
         MNIST & \xmark & B\&W & 28 & 28& 1& 1028 & 0.8M\\
         \hline
    \end{tabular}
    \caption{Number of weights implied by the first layer of an MLP with respect to the input.}
    \label{tab:mlp_weights}
\end{table}

\acrshort{mlp} have been successfully applied for classification tasks on toy datasets such as MNIST \cite{MLP_MNIST} which consists in black-and-white handwritten digit images of 28 pixels by 28 pixels. Nonetheless, this relation between weights and input size prevents the usage of \acrshort{mlp} for the majority of computer vision tasks, when the input images are large and/or cannot be compressed due to the importance of low-level information. 

\subsection{Convolutional Neural Network}
\begin{figure}[h]
    \centering
    \includegraphics[width=\linewidth]{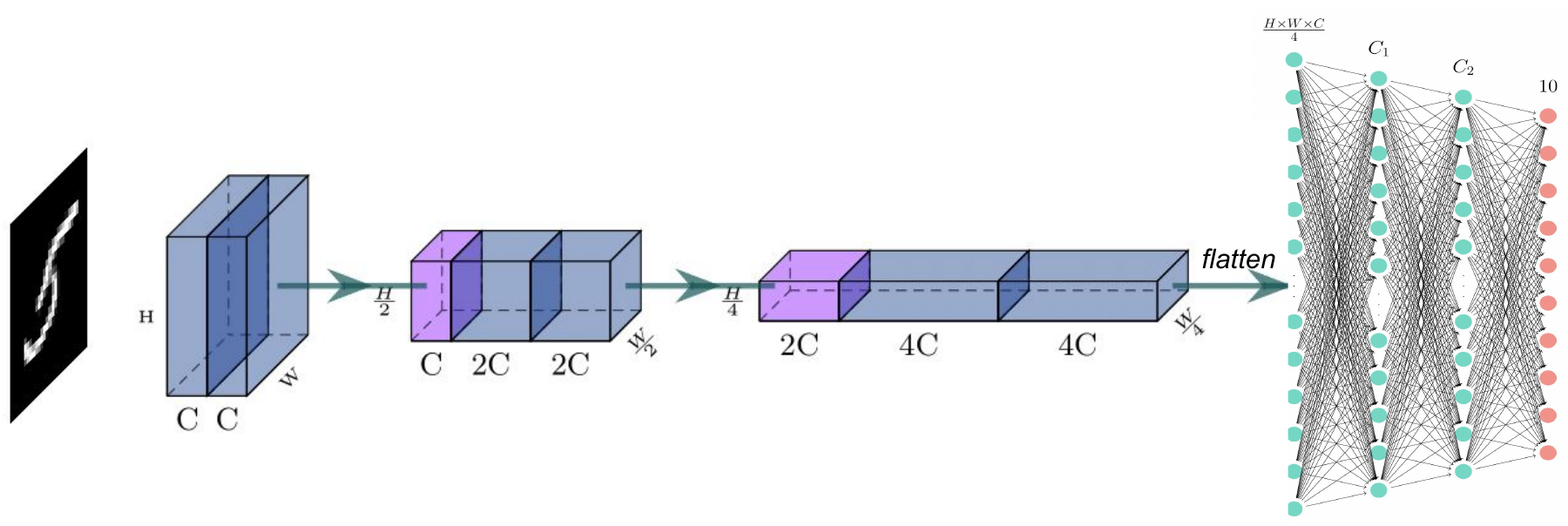}
    \caption{Example of Convolutional Neural Network applied to digit recognition. Blue boxes correspond to convolutional layers, and pink boxes to pooling layers. $C$, $C_1$ and $C_2$ are arbitrarily chosen numbers of feature maps/neurons.}
    \label{fig:cnn}
\end{figure}

In 1989, LeCun \textit{et al.} \cite{LeCun1989} proposed the first \gls{cnn}, LeNet-5, based on the Neocognitron \cite{Fukushima1982} of Fukushima  \textit{et al.}. It was inspired by discoveries related to visual mechanisms in living organisms. \acrshort{cnn} corresponds to \acrshort{mlp} in which some convolutional and pooling layers are added, usually at the beginning of the network. Convolutions are now the key element of almost all neural networks applied to \gls{cv} tasks. A \acrshort{cnn}, applied to the task of digit recognition, is depicted in Figure \ref{fig:cnn}. The flattening operation denotes the concatenation of the feature maps, thus leading from a 3D tensor (height, width, channels) to a vector.

\subsubsection{Convolutional layers}
\label{section-convolution}
\begin{figure}[h]
    \centering
    \includegraphics[width=\linewidth]{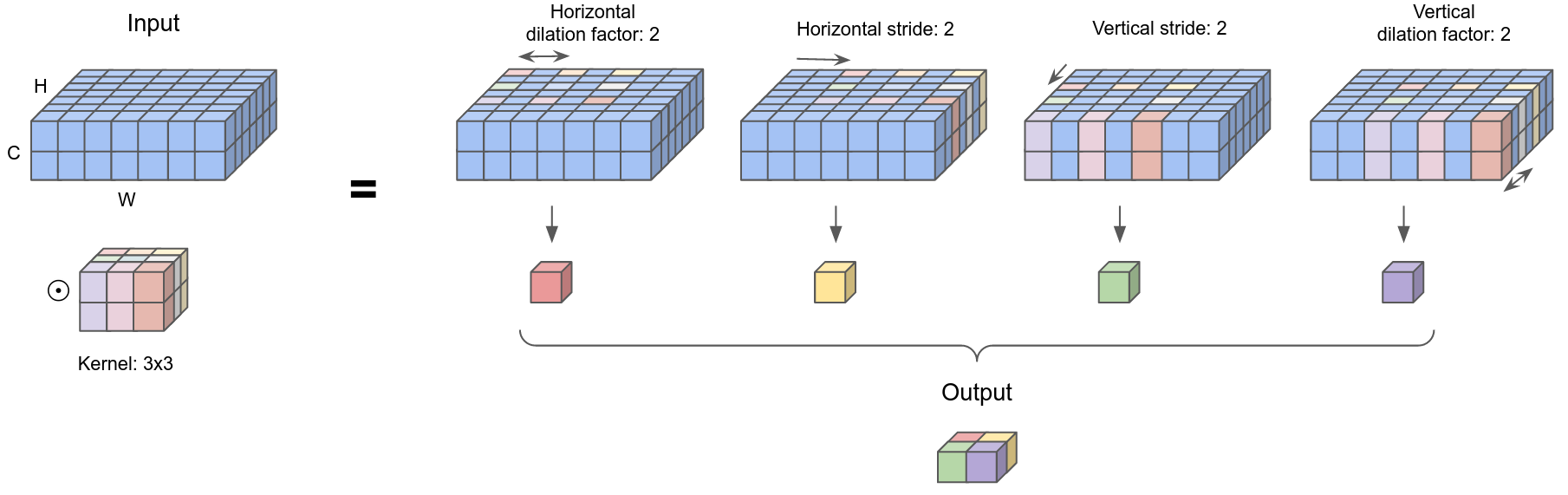}
    \caption{Convolution applied on a 2-channel input with a single $3 \times 3$ kernel, $2 \times 2$ stride and $2 \times 2$ dilation factor.}
    \label{fig:conv_sliding}
\end{figure}

A convolutional layer consists in applying $n_\mathrm{k}$ weighted sums between the elements of $n_\mathrm{k}$ kernels and the input, through a sliding window over the input. A convolution is defined by the size of the kernels $k_\mathrm{h} \times k_\mathrm{w}$, their stride $s_\mathrm{h} \times s_\mathrm{w}$, which corresponds to the displacement step of a sliding window on the two dimensions and their dilation factor $d_\mathrm{h} \times d_\mathrm{w}$, which corresponds to the spacing between two consecutive values to be considered in the calculation. These different parameters are represented through an example in Figure \ref{fig:conv_sliding}. The convolution operation is represented by the operator $\odot$. 

Mathematically, for a single kernel $\mb{k}$:
\begin{equation}
    \mb{Y} = \mb{X} \odot \mb{k},
\end{equation}
with 
\begin{equation}
    \mb{Y}_{i,j} = \displaystyle \sum^{k_\mathrm{h}}_{h=1} \sum^{k_\mathrm{w}}_{w=1} \sum^{C}_{c=1} \mb{X}_{s_\mathrm{h} * (j-1) + d_\mathrm{h} * (h-1) + 1, s_\mathrm{w} * (i-1) + d_\mathrm{w} * (w-1) + 1, c} \mb{k}_{h, w, c}
\end{equation}

\begin{figure}[h]
    \centering
    \includegraphics[width=0.3\linewidth]{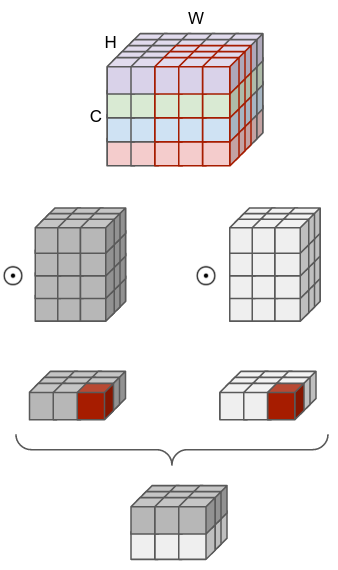}
    \caption{Convolution applied to a 4-channel input with 2 kernels of size $3 \times 3$.}
    \label{fig:convolution}
\end{figure}

The standard configuration is as follows: $3 \times 3$ kernel, $1 \times 1$ dilation factor, $1 \times 1$ stride. An extension to 2 kernels with this standard configuration is depicted in Figure \ref{fig:convolution}. Increasing the stride mainly aims at reducing the size of the tensors and thus the memory consumption. The dilation factor can be increased to capture relationships over a longer distance or to model higher level elements using fewer convolutional layers. Padding can also be added to preserve the original size of the input image. 

As one can note, convolutions imply local connectivity: the output at a specific position only depends on neighboring inputs. In addition, the same kernel is applied at different locations of the input: this is a property of convolutional layers called weight sharing. It assures shift-equivariance \textit{i. e.} a same pattern will be recognized in the same way no matter where it is in the image. It also enables to dramatically reduce the number of trainable parameters when compared to a dense layer. Indeed, contrary to fully-connected layers, the number of weights induced by a convolutional layer does not depend on the width and the height of the input. Considering the input image $\mb{X}$, a convolution applying $n_\mathrm{k}$ kernels of size $k_\mathrm{H} \times k_\mathrm{W}$ implies $C \times k_\mathrm{H} \times k_\mathrm{W} \times n_\mathrm{k}$ weights. Well-known \gls{cnn} architectures such as ResNet \cite{ResNet} are made up of convolutions using up to 512 channels, sometimes more. A convolution applied on a 1028-channel input with 1028 kernels of size $3 \times 3$ would then lead to 9.5M of weights, no matter the input image dimensions. This is far from the billions of parameters required for one fully-connected layer applied on an A4 input image (Table \ref{tab:mlp_weights}). But contrary to the convolutional layer, each neuron of a fully-connected layer take advantage of the global context of the whole input.

\begin{figure}[h]
    \centering
    \includegraphics[width=0.8\linewidth]{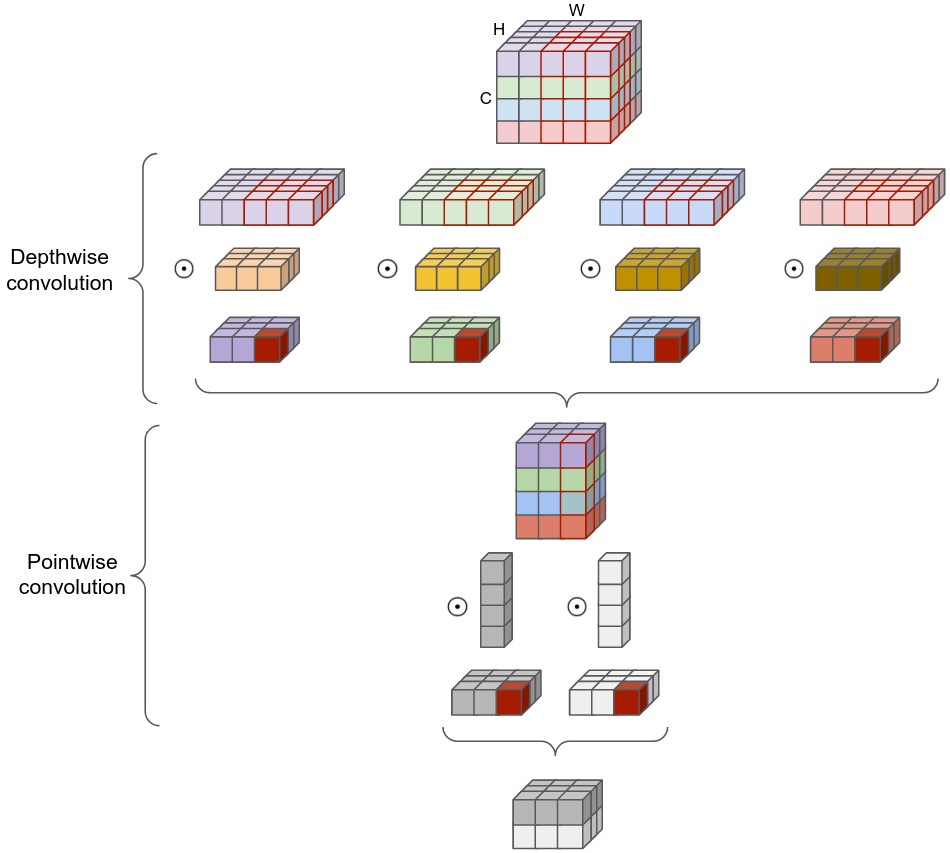}
    \caption{Depthwise Separable Convolution applied to a 4-channel input with 2 kernels of size $3 \times 3$.}
    \label{fig:dsc}
\end{figure}

Stacking several convolutional layers while maintaining a large number of channels still leads to a huge number of trainable weights. One way to alleviate this issue is to use \glspl{dsc} \cite{DSC} instead of the standard convolutions. This kind of convolutions is depicted in Figure \ref{fig:dsc}.

\acrshort{dsc} is a composition of two convolutions:
\begin{itemize}
    \item Depthwise convolution: each of the $C$ channels of the input is treated independently by a specific kernel of shape $k_\mathrm{H} \times k_\mathrm{W} \times 1$, leading to $C$ features maps which are merged together. This first step is not sufficient since it would be to consider that the channels are independent, which is false when we speak of color channels red, green and blue of an image for example.
    \item Pointwise convolution: $n_\mathrm{k}$ kernels of shape $1 \times 1 \times C$ are applied on these intermediate feature maps, leading to the final results. This second operation aims at modeling the relations between the different channels. 
\end{itemize}

Then, the associated number of weights is $k_\mathrm{H} \times k_\mathrm{W} \times C$ for the depthwise convolution and $C \times n_\mathrm{k}$ for the pointwise convolution. It results in $C \times (k_\mathrm{H} \times k_\mathrm{W} + n_\mathrm{k})$ parameters for the global operation. We can now compare the difference with standard convolution for different parameters through Table \ref{tab:conv_weights}.

\begin{table}[h]
    \centering
    \begin{tabular}{c c c c c c }
         \hline
         \multirow{3}{*}{C} & \multirow{3}{*}{$n_\mathrm{k}$} & \multirow{3}{*}{$k_\mathrm{H}$} & \multirow{3}{*}{$k_\mathrm{W}$} & \# weights & \# weights\\
         & & & & convolution & \acrshort{dsc} \\
         & & & & $C \times k_\mathrm{H} \times k_\mathrm{W} \times n_\mathrm{k}$ & $C \times (k_\mathrm{H} \times k_\mathrm{W} + n_\mathrm{k})$ \\
         \hline
         512 & 512 & 3 & 3 & 2.4M & 0.3M\\
         1028 & 1028 & 3 & 3 & 9.5M & 1.1M \\
         1028 & 1028& 5& 5& 26.4M & 1.1M \\
         \hline
    \end{tabular}
    \caption{Comparison of the number of weights implied by a standard convolution and by a Depthwise Separable Convolution (DSC).}
    \label{tab:conv_weights}
\end{table}

As one can note, using \acrshort{dsc} enables to decrease the number of parameters by a factor of around 9 for the standard kernel size $3 \times 3$. This factor is all the more important as the size of the kernels is large.

\subsubsection{Pooling layers}
Pooling layers are the other component of \acrshort{cnn}. They consist in downsampling the input so as to decrease the memory consumption, leading to shorter calculation times. To this end, a kernel is applied with stride, similarly to convolutional layers, but there is no trainable parameters involved: the returned value depends on a predefined function such as max or mean. The idea is to compress the input signal by only keeping the relevant information, discarding the non-relevant one. This way, during back-propagation, only gradients of selected values are propagated.

\begin{figure}[h]
    \centering
    \includegraphics[width=\linewidth]{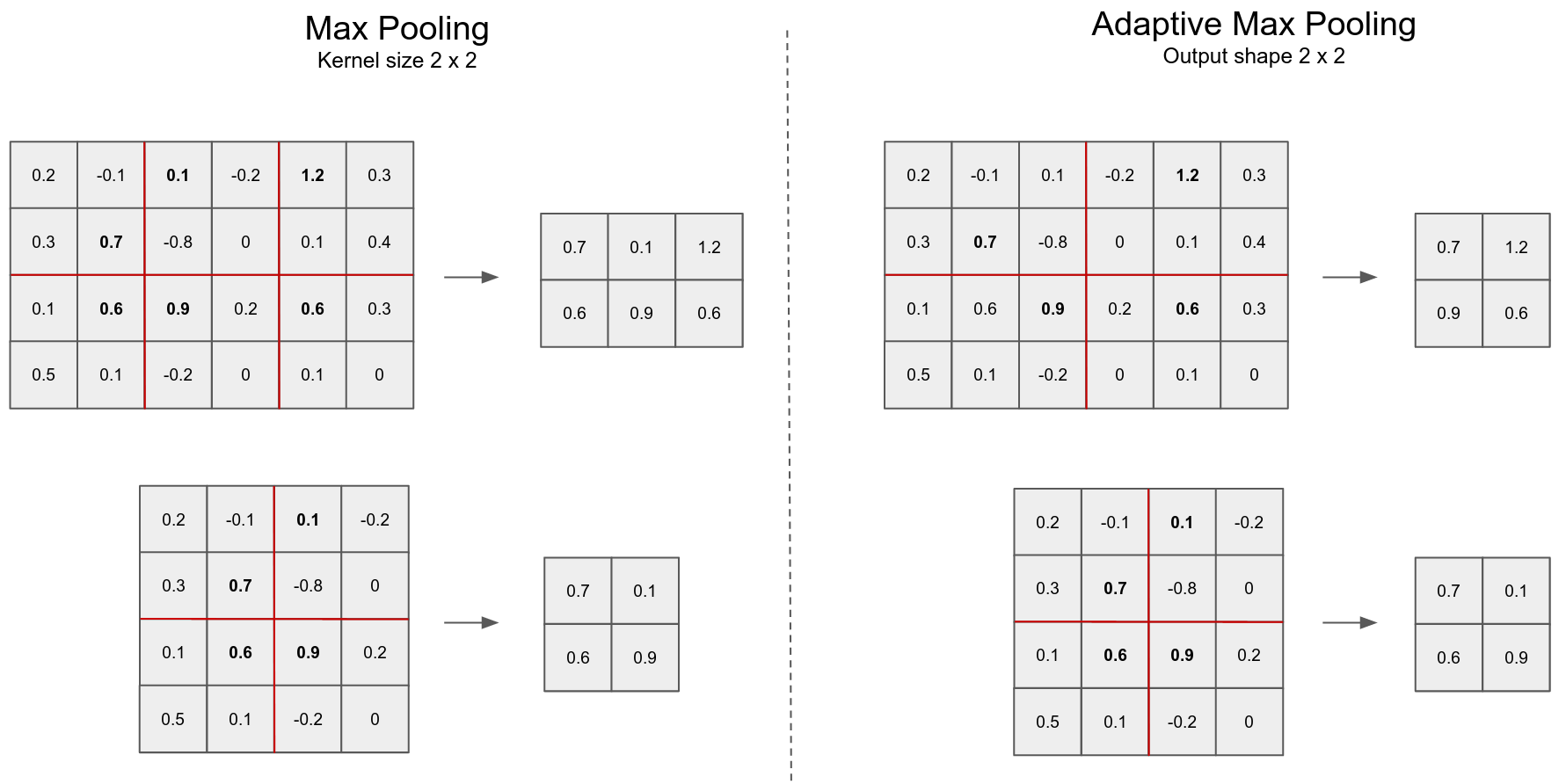}
    \caption{Max Pooling compared to Adaptive Max Pooling.}
    \label{fig:pooling}
\end{figure}

\acrshort{cnn} are traditionally made up of a stack of blocks of convolutional and pooling layers, ended by one or multiple fully-connected layers as shown in Figure \ref{fig:cnn}. \acrshort{cnn} still suffer from the fixed-input-shape requirement induced by those fully-connected layers. Adaptive pooling is a way to alleviate this issue. It consists in a pooling layer with a dynamic kernel in order to have a fixed output shape. It is compared with standard pooling in Figure \ref{fig:pooling}, for the max function. 

Given the input and output shapes, the kernel size is automatically adapted to cover all the input signal, giving the correct number of values expected as output. This way, adaptive pooling enables to take inputs of variable size, still using fully-connected layers, by fixing intermediate tensor sizes in the network. It enables to preserve the original resolution and ratio of the image for the first layers for tasks implying outputs of fixed size. 

\subsubsection{Receptive field}
\begin{figure}[ht]
    \centering
    \includegraphics[width=0.7\linewidth]{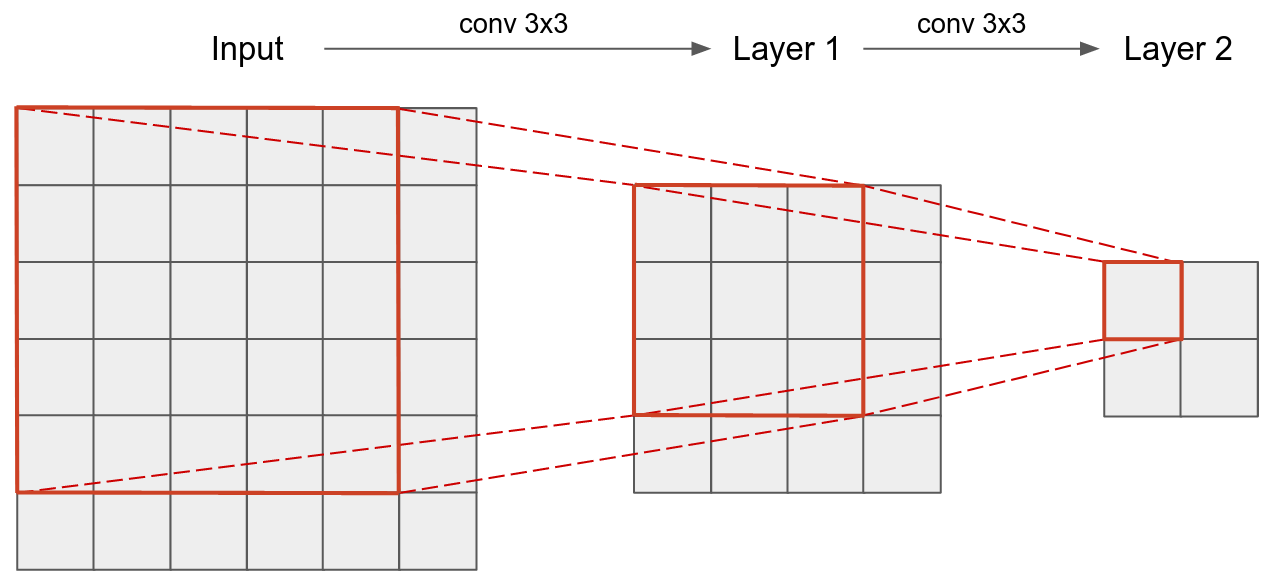}
    \caption{Receptive field visualization for 2 convolutions of kernel $3 \times 3$, stride $1 \times 1$ and dilation $1 \times 1$.}
    \label{fig:receptive_field}
\end{figure}

As shown previously, convolution and pooling operations compute outputs based only on neighboring inputs \textit{i. e.} each output has a partial and unique view of the input image. The receptive field corresponds, for a given layer, to the size of the region in the input from which each value of this layer has been calculated. In other words, it quantifies what is seen by a given neuron. It is an important parameter to understand the highest level of representation that a neural network is able to model. This notion is represented in Figure \ref{fig:receptive_field}. In this example, the receptive field for layer 2 is (5, 5) for height and width, respectively.

The receptive field is fixed for strictly sequential feed-forward neural networks. However, when one uses residual connections, multiple receptive fields are combined for a same layer, leading to multi-scale representations.

\acrshort{cnn} have gradually become the default architecture for computer vision tasks bringing state-of-the-art results on several tasks. For example, well-known architectures such as AlexNet \cite{AlexNet} and VGGNet \cite{VGGNet} have been proposed for image classification. However, we have highlighted the limitations of such architectures: even using adaptive pooling techniques, they can only handle tasks for which the size of the output is fixed. There are some tasks for which preserving the original size, or at least the dimension ratio, is important. For instance, most segmentation tasks imply pixel-to-pixel classification, which requires to preserve the image size. As a consequence, \acrshort{cnn} cannot be used to solve these tasks. 

\subsection{Fully Convolutional Network}
\glspl{fcn} have been proposed in response to the need for outputs of variable size. It corresponds to a \acrshort{cnn} without any fully-connected layer: it is only made up of convolutional components: convolutions and poolings. This way, the output of the network is the output of the last convolutional layer: the input shape can be preserved from the beginning to the end of the network.

\begin{figure}[ht]
    \centering
    \includegraphics[width=\textwidth]{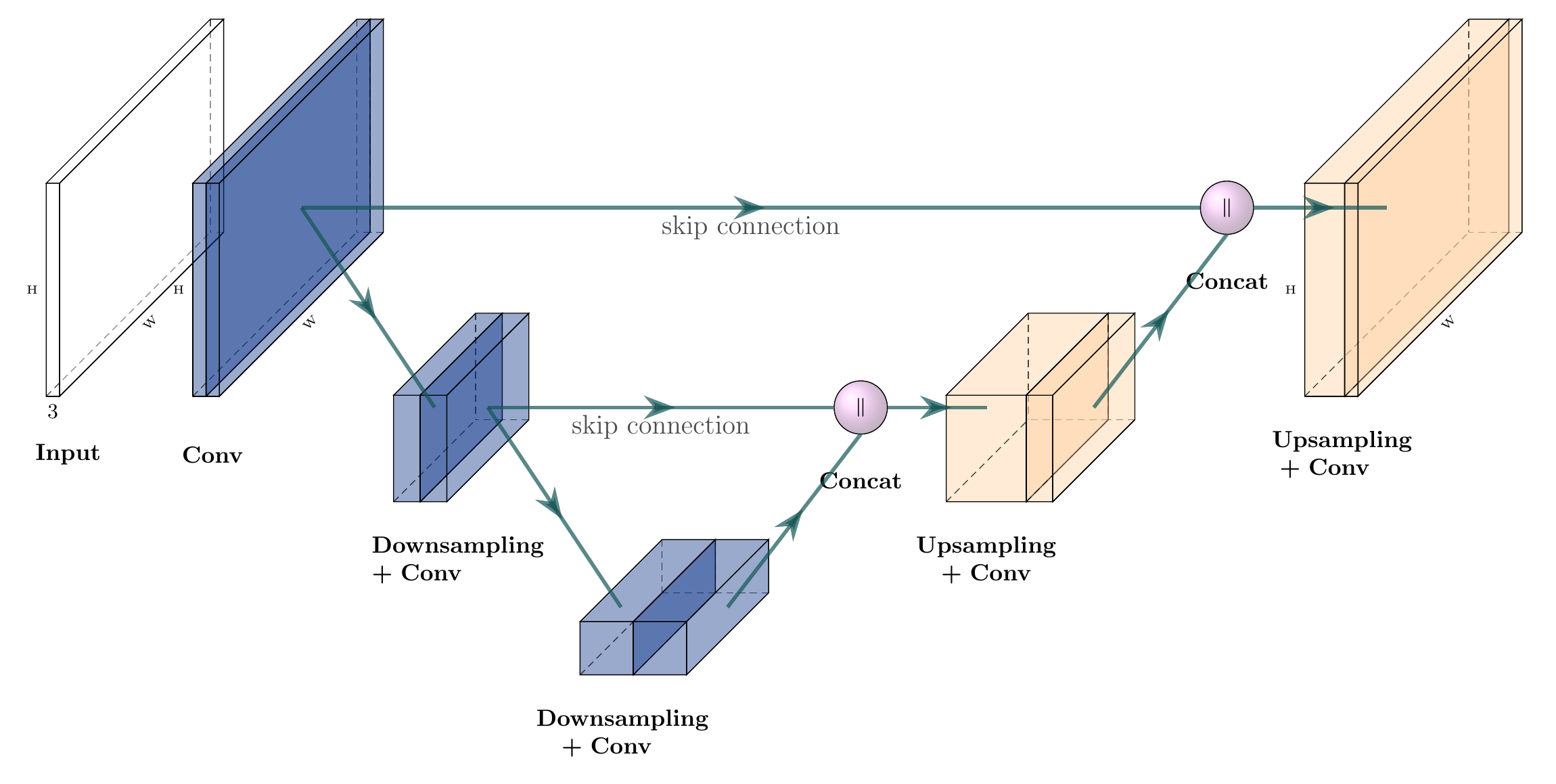}
    \caption{Overview of the U-Net architecture. The U-net integrates skip connections to preserve low-level features, extracted from the encoder (blue part), through concatenations in the decoder part (beige part).}
    \label{fig:u-net}
\end{figure}

U-Net is a well-known \gls{fcn} architecture; it is depicted in Figure \ref{fig:u-net}. In this architecture, hidden representations are compressed through convolution or pooling operations (blue part) with stride higher than $1 \times 1$ to save memory consumption. Then, to get back to the original size of the input image, it relies on different upsampling techniques (beige part).

The naive way to upsample the hidden representation is to locally duplicate the information. One can also use an interpolation based on the values of the neighboring cells (a bi-linear interpolation for example). Another common technique is the inverse operation of max pooling. To this end, the indices of the max values selected during max pooling are kept in memory for the max unpooling operation. They are used to place the values of the current hidden representation; other locations are set to zero.

Transposed convolution consists in applying $n_k$ kernels of size $k_\mathrm{H} \times k_\mathrm{W} \times C$ to each location of an input $\mb{X} \in \mathbb{R}^{H \times W \times C}$. This way, a $k_\mathrm{H} \times k_\mathrm{W} \times n_k$ representation is computed for each location of the input, before being merged together. In the standard configuration, it leads to an output $\mb{Y} \in \mathbb{R}^{(H+k_\mathrm{H}-1) \times (W+k_\mathrm{W}-1) \times n_k}$. However, it can also be used with a stride and a dilation factor different from $1 \times 1$ to perfectly fit the dimensions of the previous neural layers, no matter the configuration used for the convolutions.

The ability of \gls{fcn} to process inputs of variable sizes as well as their low number of parameters make them one the most popular model for many computer vision tasks, including segmentation tasks such as semantic segmentation of biomedical images \cite{UNet} and instance segmentation \cite{MaskRCNN}.

As we have seen so far, the computer vision tasks are mainly handled by \gls{cnn}, notably to reduce the number of parameters of the models when compared to \acrshort{mlp}. However, \gls{cnn} also faces fixed-size input/output issues and their modeling ability is dependent on their receptive field. \acrshort{fcn} architectures enable to deal with inputs of variable sizes, but it remains two main issues. 
To model dependencies across the whole input signal with an \acrshort{fcn}, it would require to have a receptive field as big as the input size, which could require many and many layers (and thus many trainable parameters). Moreover, \acrshort{fcn} cannot deal with outputs whose size is independent to that of the inputs as is. It makes the computer vision models presented in this section unsuitable for the \gls{htr} task which implies output sequences of variable lengths. To solve this issue, we will now study the \gls{htr} under the sequence-to-sequence point of view.

\section{HTR: a sequence-to-sequence problem}
There is no consensus about the definition of a \gls{seq2seq} problem. In this thesis, we refer to sequence-to-sequence problems when inputs and outputs are sequences of variable and independent sizes (no assumption is made about the length of the output sequence with respect to the length of the input sequence). 
\acrlong{htr} must handle images as inputs and sequences of tokens as outputs. It can be considered as a sequence-to-sequence problem.
Indeed, contrary to the classification task, where there is globally only one entity present in the image, a handwritten document image contains several characters, ordered horizontally and vertically. The order can be even more complex depending on the layout of the document. Moreover, the image can be seen as a 2D sequence of pixels or a sequence of columns or rows of pixels $\mb{x}$ of length $L_\mathrm{x}$. This way, the input can be considered as a sequence, and the output is a sequence $\mb{y}$ of $L_\mathrm{y}$ characters: this is a sequence-to-sequence problem.

Sequence-to-sequence tasks imply three main issues:
\begin{itemize}
    \item Extracting features from the whole input sequence. It requires to understand the relations between the different items of the sequence \textit{i. e.} to model the input dependencies.
    \item Handling input and output of variable sizes. The architecture must be able to predict sequences of any length, independently of the input length.
    \item Extracting features from the previous predictions. Modeling the output dependencies enables to take into account the output context for the prediction of the next item.
\end{itemize}

In the following, we will go over the main components that enable to deal with these issues.

\subsection{Modeling dependencies}
Modeling dependencies aims at extracting features from a whole sequence, by modeling the relations between the different items of this sequence, rather than considering them independently of each other.
\glspl{cnn} enable to model local dependencies from the input, but it does not model the output dependencies.
Until the end of the 2010's, state-of-the-art approaches for output sequence modeling were mainly based on \gls{hmm}. 

\subsubsection{Hidden Markov Model}
\begin{figure}[h]
    \centering
    \includegraphics[width=0.8\linewidth]{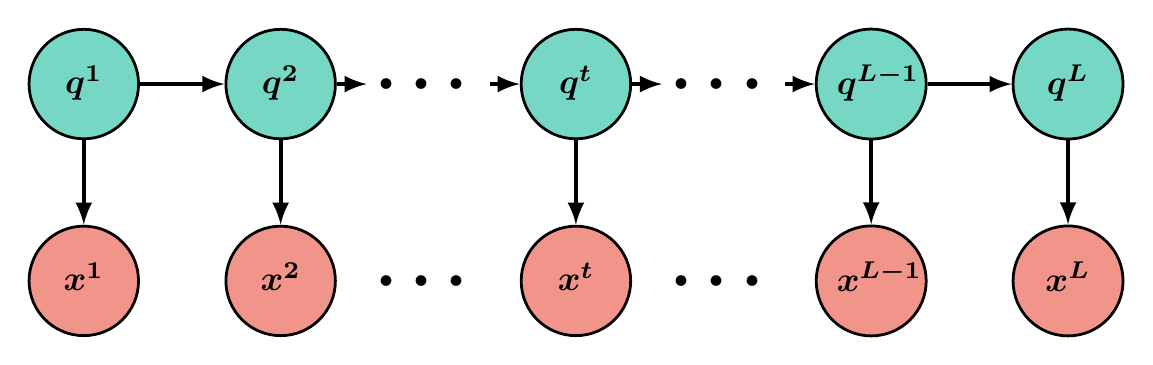}
    \caption{Representation of a Hidden Markov Model for a sequence of $L$ observations $\mb{x}$, given the states $\mb{q}$.}
    \label{fig:hmm}
\end{figure}

\acrshort{hmm} are statistical models which aims at capturing hidden information from sequential observations. \acrshort{hmm} can be represented as a finite state machine  as in Figure \ref{fig:hmm}. It is defined by:
\begin{itemize}
    \item A set of states $S = \msetdef{s_1,...,s_N}$, and a sequence of $L$ states $\mb{q} = \mseqdef{\mb{q}\mtseq{1}{},...,\mb{q}\mtseq{L}{}}$ with $\mb{q}\mtseq{t}{} \in S$.
    \item A sequence of observations $\mb{x} = \mseqdef{\mb{x}\mtseq{1}{},...,\mb{x}\mtseq{L}{}}$, which can be real valued observation vectors.
    \item An emission model $B$, which defines the probabilities of an observation $\mb{x}\mtseq{t}{}$ given a state $i$: $p(\mb{x}\mtseq{t}{}|\mb{q}\mtseq{t}{}=s_i)$. $B$ can either be composed of discrete probabilities or continuous gaussian mixtures.
    \item A transition model $A$, which defines the probabilities $\mb{a}_{i,j}$ of moving from state i to state $j$: $p(\mb{q}\mtseq{t}{}|\mb{q}\mtseq{t-1}{})$
    \item An initial probability distribution model $\Pi$, which defines the probabilities $\mb{\pi}_i$ of starting in state $i$: $p(\mb{q}\mtseq{1}{})$.
\end{itemize}

\acrshort{hmm} follow the Markov assumption implying that the state $\mb{q}\mtseq{t}{}$ at timestep $t$ only depends on the previous state $\mb{q}\mtseq{t-1}{}$:
\begin{equation}
    p(\mb{q}\mtseq{t}{}|\mb{q}\mtseq{1}{},...,\mb{q}\mtseq{t-1}{})=p(\mb{q}\mtseq{t}{}|\mb{q}\mtseq{t-1}{}).
\end{equation}
In a first step, $B$, $A$ and $\Pi$ are learned through the forward-backward algorithm (also known as Baum-Welch algorithm) \cite{BaumWelch}, which is a special case of the Expectation-Maximization algorithm \cite{ExpectationMaximization}. Then, the most likely state sequence $\mb{q}^*$ given an observed sequence $\mseq{x}$ is obtained using the Viterbi algorithm \cite{Viterbi}: 
\begin{equation}
    \mb{q}^* = \arg\max_{\mb{q} \in S^L} p(\mb{q}|\mseq{x})
    = \arg\max_{\mb{q} \in S^L} \frac{p(\mseq{x}|\mb{q})p(\mb{q})}{p(\mb{x})}
    =\arg\max_{\mb{q} \in S^L} p(\mseq{x}\mtseq{1}{}|\mb{q}\mtseq{1}{})p(\mb{q}\mtseq{1}{}) \prod_{t=2}^{L} p(\mseq{x}\mtseq{t}{}|\mb{q}\mtseq{t}{})p(\mb{q}\mtseq{t}{}|\mb{q}\mtseq{t-1}{}).
\end{equation}

\gls{hmm} are unable to deal with long term dependencies in sequences. Deep learning approaches have made it possible to overcome this problem through the use of \glspl{rnn}, and more particularly those including a memory process. 

\subsubsection{Recurrent Neural Network}
\glspl{rnn} are specific neural networks which aim at dealing with input signals of variable lengths. They are based on the works of David Rumelhart from 1986 \cite{RNN}. Let's consider an input signal $\mb{x}=\mvecdef{\mb{x}\mtseq{1}{},...,\mb{x}\mtseq{L}{}}$, which is made up of $L$ frames $\mb{x}\mtseq{t}{}$, with $C$ features per frame ($\mb{x}\mtseq{t}{} \in \mathbb{R}^C$). The idea is to process the input sequentially, frame per frame, applying the same weights. The key principle is that the output does not only depend on the input, but also on the previous outputs. It means that the decision process is not local but keeps tracks of what have been output so far ($\mb{y}\mtseq{1}{},...,\mb{y}\mtseq{t-1}{}$) and takes it into account for the current output $\mb{y}\mtseq{t}{}$. It introduces a time dependency between successive states, as for \gls{hmm}. 

\begin{figure}[h]
    \centering
    \includegraphics[width=\linewidth]{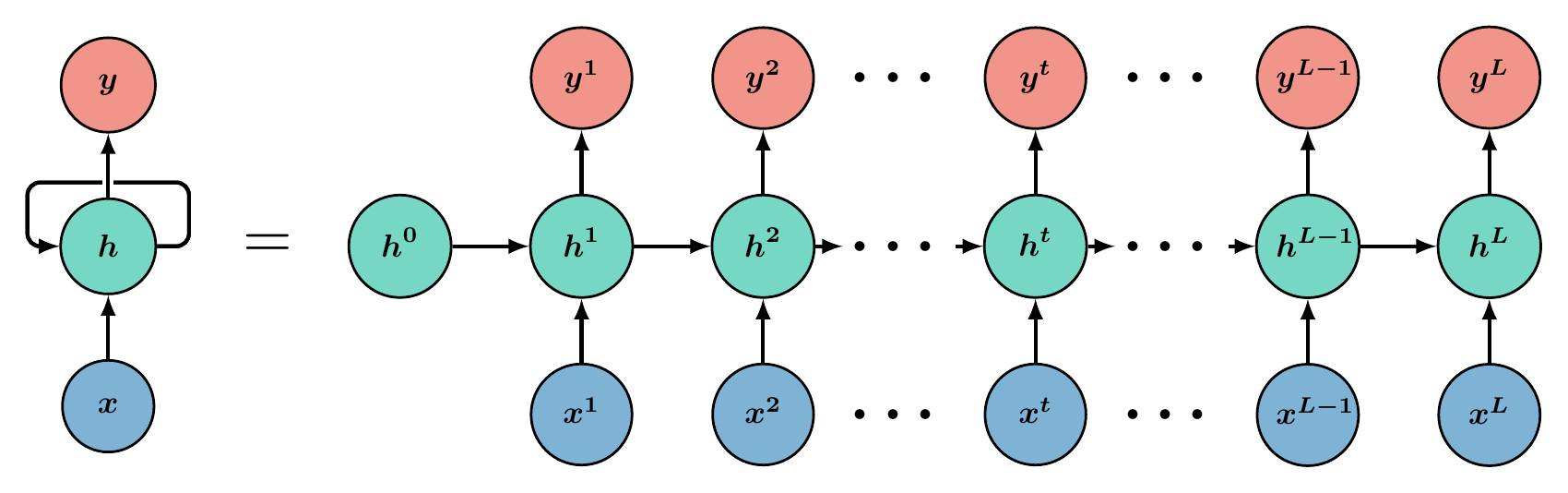}
    \caption{Recurrent Neural Network: compressed (left) and unfolded (right) representations.}
    \label{fig:rnn}
\end{figure}

Figure \ref{fig:rnn} presents a visualization of such neural networks. As one can note, $\mb{y}\mtseq{t}{}$ is computed based on $\mb{h}\mtseq{t}{}$, given that $\mb{h}\mtseq{t}{}$ is itself computed based on $\mb{x}\mtseq{t}{}$ and $\mb{h}\mtseq{t-1}{}$. This way, $\mb{y}\mtseq{t}{}$ is computed based on all $\mb{x}\mtseq{i}{}$ with $i\leq t$. $\mb{h}\mtseq{t}{}$ is called the hidden state at time step $t$; it acts as a memory of past predictions.

There are many ways of computing $\mb{y}\mtseq{t}{}$ from $\mb{x}\mtseq{t}{}$ and $\mb{h}\mtseq{t}{}$. Jordan \cite{RNNJordan} and Elman \cite{RNNElman} networks are examples of "Simple Recurrent Networks". 

In 1986, M. Jordan proposed the Jordan Network. It follows these equations:
\begin{equation}
    \begin{split}
        &\mb{h}\mtseq{t}{} = \phi_h(\mb{W}_x\mb{x}\mtseq{t}{} + \mb{U}_h\mb{y}\mtseq{t-1}{} + \mb{b}_h),\\
        &\mb{y}\mtseq{t}{} = \phi_y(\mb{W}_y\mb{h}\mtseq{t}{} + \mb{b}_y),
    \end{split}
\end{equation}
where $\phi_h$ and $\phi_y$ are activation functions, $\mb{W}_x$, $\mb{W}_y$ and $\mb{U}_h$ are trainable weights of fully-connected layers and $\mb{b}_h$ and $\mb{b}_y$ are trainable biases.

The Elman network was proposed in 1990 and is defined as follows:
\begin{equation}
\begin{split}
        &\mb{h}\mtseq{t}{} = \phi_h(\mb{W}_x\mb{x}\mtseq{t}{} + \mb{U}_h\mb{h}\mtseq{t-1}{} + \mb{b}_h),\\
        &\mb{y}\mtseq{t}{} = \phi_y(\mb{W}_y\mb{h}\mtseq{t}{} + \mb{b}_y).
\end{split}
\end{equation}
The difference is about the $\mb{h}\mtseq{t}{}$ calculation where $\mb{h}\mtseq{t-1}{}$ is superseded by $\mb{y}\mtseq{t-1}{}$.

In 1997, the authors of \cite{BRNN} presented the Bidirectional Recurrent Neural Network (BRNN) with the following idea. Some tasks can process whole (offline) data sequences, as opposed to online approaches where the signal is made available over time. It means that one can access to the whole sequence $\mb{x}$, and instead of just modeling forward unidirectional representations from $\mb{x}\mtseq{1}{}$ to $\mb{x}\mtseq{L}{}$, the authors suggested combining forward and backward representations simultaneously. This way, the model computes two independent hidden states: $\overrightarrow{\mb{h}\mtseq{t}{}}$ for forward pass and $\overleftarrow{\mb{h}\mtseq{t}{}}$ for backward pass.  $\overrightarrow{\mb{h}\mtseq{t}{}}$ and $\overleftarrow{\mb{h}\mtseq{t}{}}$ are then concatenated, leading to $\mb{h}\mtseq{t}{}$, aggregating knowledge from past and future. This bidirectional strategy can be used whichever the kind of \acrshort{rnn} architecture used.

The \acrshort{rnn} we have studied so far are limited to one-dimensional sequences. They can be used for tasks from \acrshort{nlp} field such as Speech Recognition. \acrshort{rnn} have been generalized to multi-dimensional inputs with the \gls{mdrnn} \cite{MDRNN}. \acrshort{mdrnn} can be used to handle 2D inputs such as images or 3D inputs like videos. Considering image inputs, $\mb{h}\mtseq{t}{}$ would contain context from both top and left pixels for instance.
In the same way, the bidirectional concept has been generalized to Multi-Directional \acrshort{rnn} for \acrshort{mdrnn}. This time, $\mb{h}\mtseq{t}{}$ would be the concatenation of 4 passes: from top left, top right, bottom left and bottom right.
Increasing the number of dimensions and directions leads to longer and longer computation time, accompanied by a growing number of trainable weights.

\acrshort{rnn} are trained using the Backpropagation Through Time algorithm. It is an adapted version of the standard backpropagation we have seen before in which the recurrent layers are unfolded over time, as shown in Figure \ref{fig:rnn}. It shows that \gls{rnn} can be considered as stacks of feed-forward layers. The number of layers in these stacks is dependent to the length of the input, making \acrshort{rnn} prone to vanishing and exploding gradient issues. In fact, it induces vanishing gradients on the lowest layers \textit{i. e.} for the first inputs. It means that there is no learning for these first steps, leading to a short memory.

\subsubsection{Long-Short Term Memory layers}
\begin{figure}[h]
    \centering
    \includegraphics[width=0.7\linewidth]{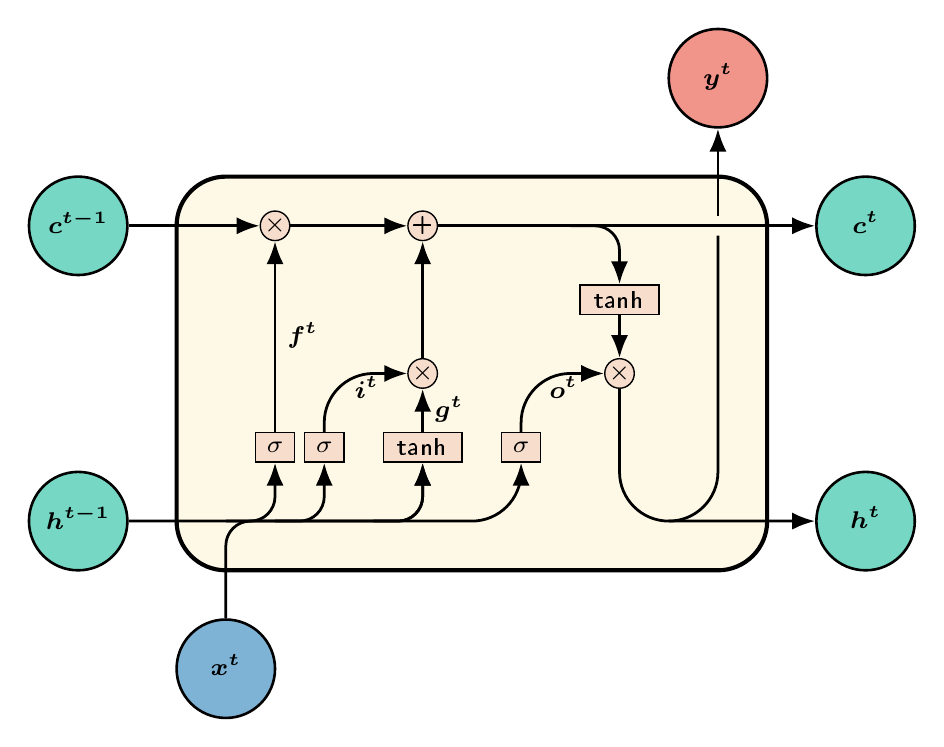}
    \caption{Long-Short Term Memory cell.}
    \label{fig:lstm}
\end{figure}

\gls{lstm} \cite{LSTM} layers have been proposed in 1997 to tackle the issue of short memory, or in other words, the gradient vanishing problem over time. \acrshort{lstm} integrate internal mechanisms, known as gates, whose aim is to regulate the flow of information. An \acrshort{lstm} cell is depicted in Figure \ref{fig:lstm}. It is made up of 4 gates $\mb{f}\mtseq{t}{}$, $\mb{i}\mtseq{t}{}$, $\mb{o}\mtseq{t}{}$, $\mb{g}\mtseq{t}{}$ for forget, input, output and cell gate, respectively. It also introduces a cell state $\mb{c}\mtseq{t}{}$, which acts as an additional memory unit. 
Each gate is computed from the previous hidden state  $\mb{h}\mtseq{t-1}{}$ as well as the current input $\mb{x}\mtseq{t}{}$, using the sigmoid (or tanh for the cell gate) activation function to output values between 0 and 1, in order to select the relevant information. First, $\mb{g}\mtseq{t}{}$ is computed as the new information to integrate to the cell state $\mb{c}\mtseq{t}{}$. The forget gate $\mb{f}\mtseq{t}{}$ and the input gate $\mb{i}\mtseq{t}{}$ quantify the amount of information to keep from the previous cell state $\mb{c}\mtseq{t-1}{}$ and to integrate from the new input $\mb{x}\mtseq{t}{}$, respectively. Finally, the output gate is applied on the new cell state $\mb{c}\mtseq{t}{}$ to compute $\mb{h}\mtseq{t}{}$ which is also the output $\mb{y}\mtseq{t}{}$.

Mathematically, the \acrshort{lstm} cell is defined by:
\begin{equation}
    \begin{split}
        & \mb{i}\mtseq{t}{} = \sigma(\mb{W}_i\mb{x}\mtseq{t}{} + \mb{U}_i\mb{h}\mtseq{t-1}{} + \mb{b}_i),\\
        & \mb{f}\mtseq{t}{} = \sigma(\mb{W}_f\mb{x}\mtseq{t}{} + \mb{U}_f\mb{h}\mtseq{t-1}{} + \mb{b}_f),\\
        & \mb{o}\mtseq{t}{} = \sigma(\mb{W}_o\mb{x}\mtseq{t}{} + \mb{U}_o\mb{h}\mtseq{t-1}{} + \mb{b}_o),\\
        & \mb{g}\mtseq{t}{} = \tanh(\mb{W}_g\mb{x}\mtseq{t}{} + \mb{U}_g\mb{h}\mtseq{t-1}{} + \mb{b}_g),\\
        & \mb{c}\mtseq{t}{} = \mb{f}\mtseq{t}{} \circ \mb{c}\mtseq{t-1}{} + \mb{i}\mtseq{t}{} \circ \mb{g}\mtseq{t}{},\\
        & \mb{h}\mtseq{t}{} = \mb{o}\mtseq{t}{} \circ \tanh(\mb{c}\mtseq{t}{}),\\
        & \mb{y}\mtseq{t}{} = \mb{h}\mtseq{t}{}.
    \end{split}
\end{equation}
$\mb{W}_i$, $\mb{W}_f$, $\mb{W}_o$, $\mb{W}_g$, $\mb{U}_i$, $\mb{U}_f$, $\mb{U}_o$, $\mb{U}_g$ are trainable weights, $\mb{b}_i$, $\mb{b}_f$, $\mb{b}_o$, $\mb{b}_g$ are trainable biases, $\sigma$ is the sigmoid activation function and $\circ$ denotes the Hadamard product \textit{i.e .} the element-wise product.

By controlling the information flow, the \acrshort{lstm} solved the long-term memory issue, but it comes at the cost of many additional trainable weights.

\subsubsection{The Gated Recurrent Unit}
\begin{figure}[h]
    \centering
    \includegraphics[width=0.7\linewidth]{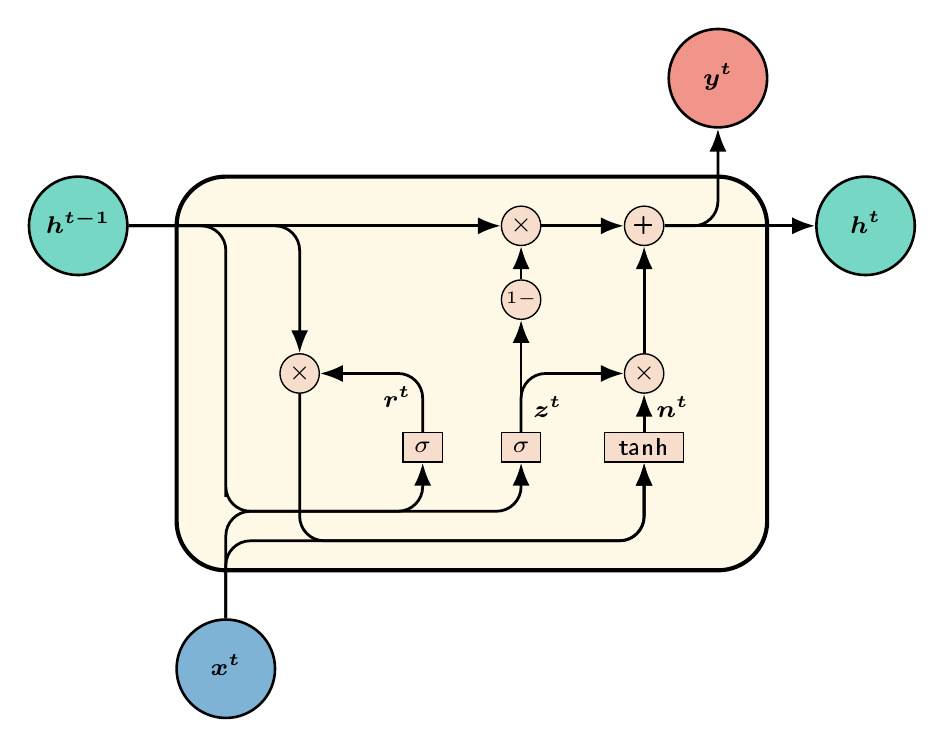}
    \caption{Gated Recurrent Unit cell.}
    \label{fig:gru}
\end{figure}

In 2014, the authors of \cite{GRU} proposed the \gls{gru} as an alternative to \acrshort{lstm}, in order to reduce the number of trainable weights, while keeping the long-term memory property. A \acrshort{gru} cell is presented in Figure \ref{fig:gru}. It relies on two gates: an update gate $\mb{z}\mtseq{t}{}$ and a reset gate $\mb{r}\mtseq{t}{}$.

Contrary to \acrshort{lstm}, \acrshort{gru} do not rely on a cell state. The gates are also computed on the previous hidden state $\mb{h}\mtseq{t-1}{}$ and the current input $\mb{x}\mtseq{t}{}$. In \acrshort{gru},  the update gate $\mb{z}\mtseq{t}{}$ is computed to specify how much to keep from the previous step through $\mb{h}\mtseq{t-1}{}$ and how much to integrates from the new input through $\mb{n}\mtseq{t}{}$. The reset gate $\mb{r}\mtseq{t}{}$ also impacts how much information to forget from the past. 

\acrshort{gru} cell can be mathematically formulated this way:
\begin{equation}
    \begin{split}
        & \mb{r}\mtseq{t}{} = \sigma(\mb{W}_r\mb{x}\mtseq{t}{} + \mb{U}_r\mb{h}\mtseq{t-1}{} + \mb{b}_r),\\
        & \mb{z}\mtseq{t}{} = \sigma(\mb{W}_z\mb{x}\mtseq{t}{} + \mb{U}_z\mb{h}\mtseq{t-1}{} + \mb{b}_z),\\
        & \mb{n}\mtseq{t}{} = \tanh(\mb{W}_n\mb{x}\mtseq{t}{} + \mb{r}\mtseq{t}{} \circ \mb{U}_n\mb{h}\mtseq{t-1}{} + \mb{b}_n),\\
        & \mb{h}\mtseq{t}{} = (1-\mb{z}\mtseq{t}{}) \circ \mb{h}\mtseq{t-1}{} + \mb{z}\mtseq{t}{} \circ \mb{n}\mtseq{t}{},\\
        & \mb{y}\mtseq{t}{} = \mb{h}\mtseq{t}{}.
    \end{split}
\end{equation}
$\mb{W}_r$, $\mb{W}_z$, $\mb{W}_n$, $\mb{U}_r$, $\mb{U}_z$, $\mb{U}_n$ are trainable weights, $\mb{b}_r$, $\mb{b}_z$, $\mb{b}_n$ are trainable biases, $\sigma$ is the sigmoid activation function and $\circ$ denotes the element-wise product.

There is no consensus on the superiority of one of these two cells (\acrshort{gru} and \acrshort{lstm}). It seems that each one enables to reach better results in particular conditions.

We have seen that \gls{rnn} could be used on multi-dimensional inputs of variable sizes such as images and could combine passes over multiple directions to gather more context. In addition, \gls{lstm} and \gls{gru} enables to deal with long-term dependencies. However, whether it is \gls{hmm} or \gls{rnn}, both approaches lead to output sequences of the same length than that of the input sequences, which made them unsuitable for the \gls{htr} task. In the following, we will see how to deal with this issue.

\subsection{Dealing with inputs and outputs of variable lengths}
When dealing with sequence-to-sequence tasks, one faces the problem of the difference between the input sequence length and the target sequence length, which varies from one sample to another. It means that the expected target can be shorter, longer or of the same length than the input sequence. But the point is that one does not know in advance what is the length of the expected target. There are two main approaches to tackle this problem. The first is to predict as many tokens as there are inputs. The challenge is then to find a transformation that enables to adapt the length of the predicted sequence via a post-processing step. The \gls{ctc} follows this approach. The second is to use a recurrent process which outputs the tokens sequentially, step by step, based on the whole input. This recurrent process ends when a specific <end-of-prediction> token is predicted, which enables to have outputs of variable length. This is the idea behind the sequence-to-sequence models.

\subsubsection{Connectionist Temporal Classification}
The \gls{ctc} \cite{CTC} was proposed by A. Graves \textit{et al.} in 2006 as an output layer for temporal classification with \acrshort{rnn}s. It was first introduced for the tasks of speech and handwriting recognition. The aim was then to solve the alignment problem between the inputs (image or voice recording) and their expected sequence of characters. Indeed, for an image, a character is written on more than a single column of pixels, and for an audio file, the sound corresponding to a phoneme is spread over multiple audio frames.

Let $\mseq{x}$ and $\mseq{y}$ be a couple of input and target output sequences, of lengths $L_\mathrm{x}$ and $L_\mathrm{y}$, respectively, with $L_\mathrm{x} \geq L_\mathrm{y}$. $\mseq{y}$ is a sequence of tokens $\mseq{y}\mtseq{t}{} \in \mset{A}$, where $\mset{A}$ is a given alphabet. The \acrshort{ctc} introduces an additional specific null symbol $\ctcblank$, also known as blank token, to the original alphabet $\mset{A}$: $\mset{A}'=\mset{A} \cup \msetdef{\ctcblank}$. The aim is to determine $p(\mseq{y}|\mseq{x})$. 

The \acrshort{ctc} consists in a softmax output layer. Let $\mseq{p}$ be the probability lattice ($\mseq{p} \in \mathbb{R}^{L_\mathrm{x} \times \mathrm{card}(\mset{A}')}$) corresponding to the input $\mb{x}$, giving probabilities at each frame for each token of the alphabet and for the \acrshort{ctc} null symbol. 

\begin{figure}[h!]
    \centering
    \includegraphics[width=\textwidth]{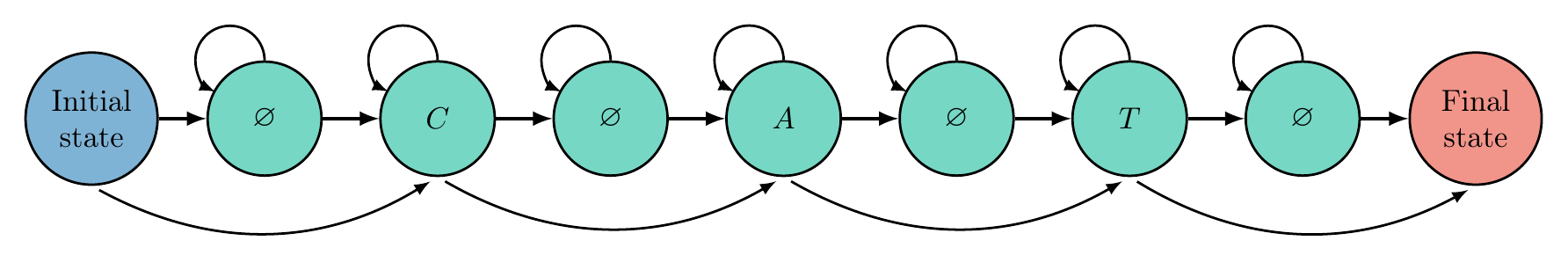}
    \caption{CTC automaton for the word "CAT".}
    \label{fig:ctc-automaton}
\end{figure}

The \acrshort{ctc} involves the use of a many-to-one map $\beta$, which enables to map the input sequence with the output sequence: $\beta : \mset{A}'^L \mapsto \mset{A}^{\leq L}$. By many-to-one, we mean that different input sequences can lead to a same output sequence. $\beta$ consists in a two-step processing: first, the successive identical tokens are removed from the input; then, the null symbols are removed too. This way, the null symbol has two functions: it enables to "predict nothing" and to separate two successive identical tokens in the ground truth. $\beta$ can be represented as an automaton defined by a binary transition matrix. Figure \ref{fig:ctc-automaton} shows the automaton for the target sequence "CAT". Here, $\beta(\mathrm{CAAAT}) = \beta(\mathrm{CAT}) = \beta(\mathrm{C}\ctcblank\mathrm{AAT}) = \mathrm{CAT}$, but $\beta(\mathrm{CCA}\ctcblank\mathrm{AT}) = \mathrm{CAAT}$.

$p(\mseq{y}|\mseq{x})$ can then be computed as the sum of all the valid paths $\mseq{\pi} \in \mset{A}'^{L_\mathrm{x}}$ leading to $\mseq{y}$ through $\beta$:
\begin{equation}
p(\mseq{y}|\mseq{x}) = \sum_{\mseq{\pi} \in \mathcal{B}^{-1}(\mseq{y})}p(\mseq{\pi}|\mseq{x}),
\end{equation}
with 
\begin{equation}
    p(\mseq{\pi}|\mseq{x}) = \prod_{t=1}^{L_\mathrm{x}}\mseq{p}\mtseq{t}{\mseq{\pi}\mtseq{t}{}}, \forall \mseq{\pi} \in \mset{A}'^{L_x},
\end{equation}
where $\mseq{p}\mtseq{t}{\mseq{\pi}\mtseq{t}{}}$ is the probability of observing label $\mseq{\pi}\mtseq{t}{}$ at position $t$ in the input sequence $\mseq{x}$.

The \acrshort{ctc} is used as a loss, dynamically computed through the forward-backward algorithm, similarly to \gls{hmm}, but by using a binary transition matrix that corresponds to the CTC automaton:
\begin{equation}
\mset{L}_{\mathrm{CTC}}(\mseq{x},\mseq{y}) = -\ln{p(\mseq{y}|\mseq{x})}.
\end{equation}

The \acrshort{ctc} approach enables to deal with sequence-to-sequence tasks in a straightforward way: all the input tokens are provided at once and $\beta$ maps it to the final prediction $\hat{\mseq{y}}$. However, there is some important limitations:
\begin{itemize}
    \item The \acrshort{ctc} is only defined for 1D sequences. 
    \item The input length must be greater than or equal to the output length, due to how $\beta$ is defined.
    \item The CTC only allows monotonic alignments. This is not a problem for the recognition of text lines but it prevents its use for \gls{nmt} for instance.
\end{itemize}

We now present the sequence-to-sequence paradigm, which solves the second and third \gls{ctc} limitations.

\subsubsection{Sequence-to-sequence paradigm}
\begin{figure}[h]
    \centering
    \includegraphics[width=\linewidth]{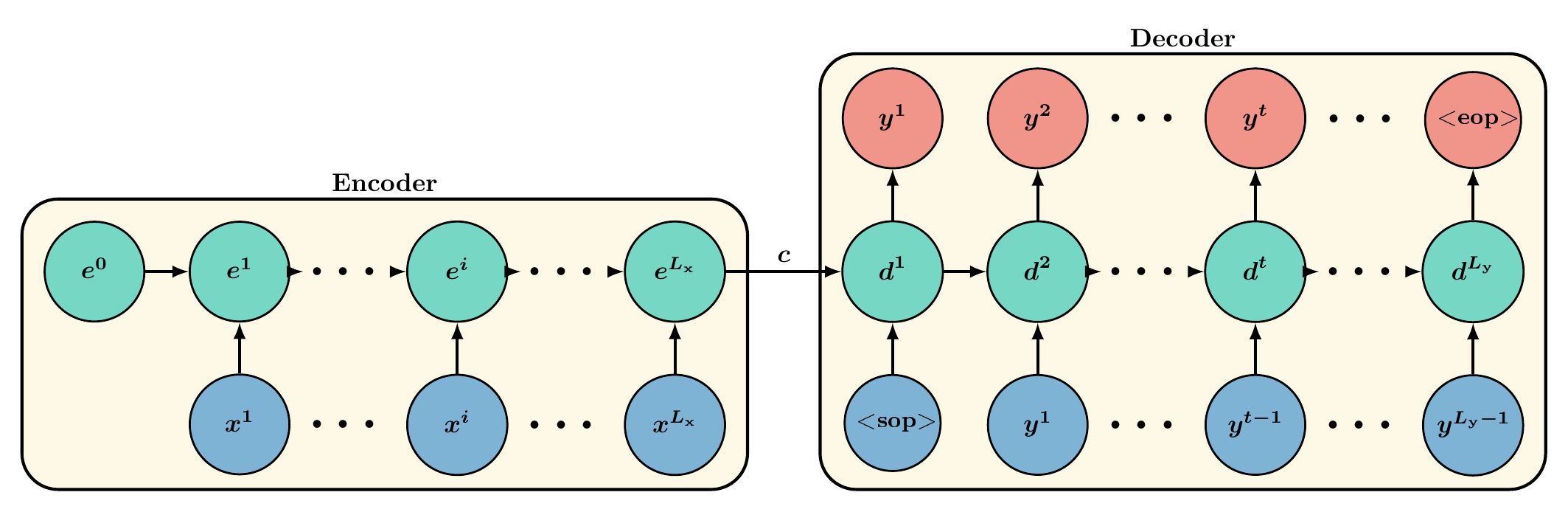}
    \caption{Sequence-to-sequence architecture. The encoder computes a context vector $\mb{c}$ from an input sequence $\mb{x}$, which is used to initialize the decoder. The decoding process is recurrent, and stops when a special end-of-prediction token (<eop>) is predicted.}
    \label{fig:seq2seq}
\end{figure}

The \gls{seq2seq} paradigm was proposed in \cite{Seq2Seq} for the task of \acrshort{nmt}. It consists in using a neural network which is made up of two modules: an encoder and a decoder. As represented in Figure \ref{fig:seq2seq}, both encoder and decoder are based on recurrent neural layers. The encoder extracts a representation of the whole input sequence $\mseq{x}$ through the last hidden state $\mb{e}_{L_\mathrm{x}}$. This representation, known as context vector $\mb{c}$, has a fixed size, no matter the input length $L_\mathrm{x}$. It enables to handle inputs of variable length. This representation is used to initialize the decoder. The first input of the decoder is a special start-of-prediction token (<sop>). Then, it uses the previous prediction $\mseq{y}\mtseq{t-1}{}$ as input. The prediction is stopped when a special end-of-prediction token (<eop>) is output. This way, the length of the output $L_\mathrm{y}$ is defined by the prediction of the <eop> token: this is how this paradigm handles the prediction of outputs of variable length.

More generally, the encoder extracts some features from the input sequence $\mseq{x}$ in the form of a context vector $\mb{c}$. And the decoder recurrently predicts some tokens based on $\mb{c}$ and on its own predictions.

The main drawback of this encoder-decoder architecture is the fixed-size representation $\mb{c}$. No matter the size of the input sequence, $\mb{c}$ must represent the whole sequence. It leads to poorer results for long sequences. Indeed, the signal cannot be compressed without loss, especially for very long sequences. Intuitively, we can assume that the first elements of the input sequence are under-represented due to the high number of successive gates. The notion of attention mechanism was introduced to tackle this issue. 

\subsubsection*{Attention mechanisms}
\begin{figure}[ht]
    \centering
    \includegraphics[width=0.6\linewidth]{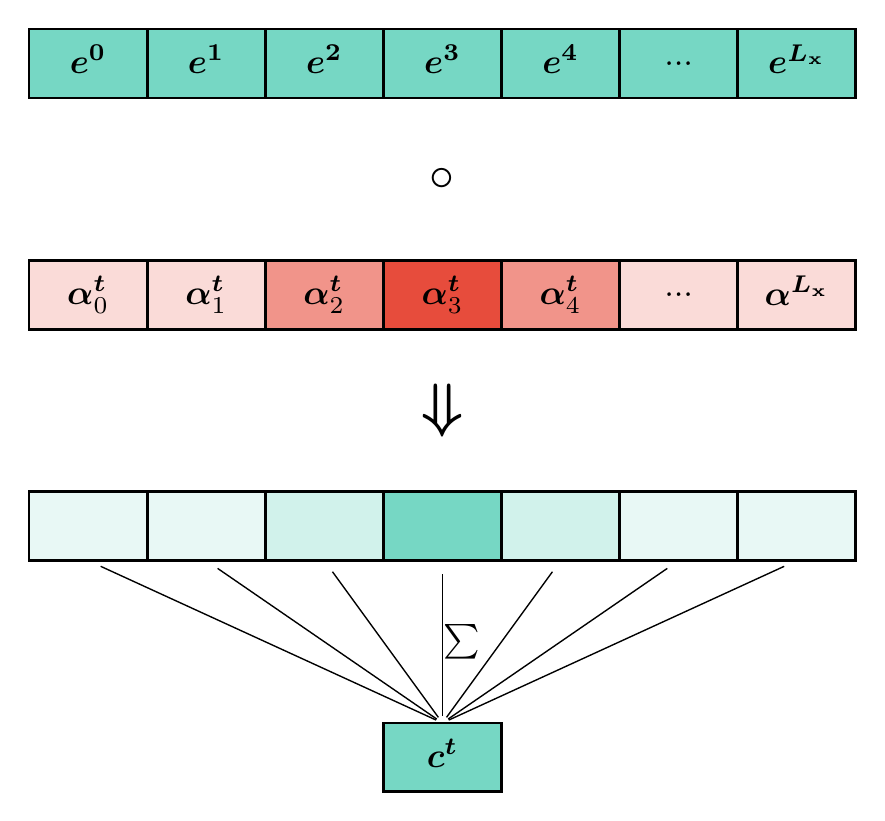}
    \caption{Attention mechanism. The context vector $\mb{c}\mt{t}{}$ is computed for each decoding step $t$ through a weighted sum between the attention weights $\mseq{\alpha}\mt{t}{}$ and the features extracted from each input frame $\mb{e}\mtseq{}{}$.}
    \label{fig:attention}
\end{figure}

The idea behind the attention mechanisms is to keep the extracted features of each input frame ($\mb{e}\mtseq{1}{}, ..., \mb{e}\mtseq{L_\mathrm{x}}{}$) and to design a way to select the most relevant input frames for the current prediction. Indeed, at each time step $t$, an attention weight $\mseq{\alpha}\mt{t}{i}$ is computed for each feature frame $\mb{e}\mtseq{i}{}$.
Attention weights can be seen as a weighted mask, following a probability distribution: $\sum_{i=1}^{L_\mathrm{x}}\mseq{\alpha}\mt{t}{i} = 1, \, \forall t$. It is used to compute a weighted sum of the feature frames, leading to a dynamic context vector $\mb{c}\mt{t}{}= \sum_{i=1}^{L_\mathrm{x}} \mseq{\alpha}\mt{t}{i}\mb{e}\mtseq{i}{}$. As shown in Figure \ref{fig:attention}, this attention mechanism enables to get a fixed-size representation at time step $t$ from an input of arbitrary length. $\mb{c}\mt{t}{}$ can then be used as input for the decoder module.
As one can note, the use of an attention mechanism enables to preserve the representation of the whole input sequence. 

Attention mechanisms can be classified according to various aspects. First aspect is about the feature frames: attention can be soft, hard or local. Soft attention, as in \cite{Bahdanau}, corresponds to probabilistic attention weights over all the feature frames. The operation is differentiable but it can be expensive since it depends on the input size. Hard attention is the opposite, it consists in selecting only one frame (one attention weight to 1 and the others to 0). This way, there is less cost at inference time but the model is non-differentiable. Local attention \cite{Luong} is an in-between with probabilistic weights over a sub-part of the feature frames. However, it implies some heuristics to choose these sub-parts.

Monotonic attention \cite{Monotonic} refers to hard attention in which the feature frames are read only in a given order, from left to right for instance. This way, the frames are read from left to right and processed only if a given score is over a specified threshold. This can only be used for specific tasks implying strict sequentiality. A variant is monotonic chunkwise attention \cite{MonotonicChunkwise}, which consists in a monotonic local attention.

The most used attention is the soft attention. It was first proposed in \cite{Bahdanau} for the task of \acrshort{nmt}. This attention mechanism computes an attention score for each feature frame $\mseq{e}\mtseq{i}{}$ at time step $t$:
\begin{equation}
\label{eq:badhanau}
    \mseq{s}\mt{t}{i} = \mb{v}^T\tanh(\mb{W}\mseq{d}\mtseq{t-1}{} + \mb{V}\mseq{e}\mtseq{i}{}),
\end{equation}
where $\mb{v}$, $\mb{W}$ and $\mb{V}$ are weight tensors of fully connected layers. These scores reflect the importance of a given feature frame $\mseq{e}\mtseq{i}{}$ with respect to the previous hidden state $\mseq{d}\mtseq{t-1}{}$ in deciding the next hidden state $\mseq{d}\mtseq{t}{}$ and in generating the associated output $\mseq{y}\mtseq{t}{}$.

Scores are associated with probabilities through a softmax activation, leading to the attention weights:
\begin{equation}
    \mseq{\alpha}\mt{t}{i} = \displaystyle \frac{\exp(\mseq{s}\mt{t}{i})}{\displaystyle \sum_{k=1}^{L_\mathrm{x}}\exp(\mseq{s}\mt{t}{k})}.
\end{equation}

This attention is known as content-based since the scores are based on the input information $\mseq{e}\mtseq{i}{}$. In addition, it is qualified as additive attention due to the addition between the two sources of information $\mseq{d}\mtseq{t-1}{}$ and $\mseq{e}\mtseq{i}{}$ in Equation \ref{eq:badhanau}.

The next year, Luong \textit{et al.} proposed other ways to compute content-based attention \cite{Luong}:
\begin{itemize}
    \item Dot: $\mseq{s}\mt{t}{i} = {\mseq{d}\mtseq{t}{}}^T\mseq{e}\mtseq{i}{}$.
    \item General: $\mseq{s}\mt{t}{i} = {\mseq{d}\mtseq{t}{}}^T\mb{W}\mseq{e}\mtseq{i}{}$.
    \item Concat: $\mseq{s}\mt{t}{i} = {\mb{v}}^T\tanh(\mb{W}\mcat{\mseq{d}\mtseq{t}{}}{\mseq{e}\mtseq{i}{}})$.
\end{itemize}

They also proposed a location-based attention \textit{i. e.} the attention weights are only computed from the decoder hidden state:
\begin{equation}
    \mseq{\alpha}\mt{t}{} = \mathrm{softmax}(\mb{W}\mseq{d}\mtseq{t}{}).
\end{equation}

The authors of \cite{Chorowski} proposed to extend the content-based attention from \cite{Bahdanau} by including some location information in the computation of the attention scores:
\begin{equation}
     \mseq{s}\mt{t}{i} = \mb{v}^T\tanh(\mb{W}\mseq{d}\mtseq{t-1}{} + \mb{V}\mseq{e}\mtseq{i}{} + \mb{U}\mseq{f}\mtseq{t}{i}),
\end{equation}
with
\begin{equation}
    \mseq{f}\mtseq{t}{} = \mconv{\mb{F}}{\mseq{\alpha}\mt{t-1}{}},
\end{equation}
where $\mb{F}$ is a weight matrix used for a convolution operation. 
In the following, we will refer to it as hybrid attention since it is both content-aware and location-aware. The authors showed that combining these two information enables a better choice of the feature maps leading to better results in the context of speech recognition.

Whether it is location-based, content-based or hybrid, the attention mechanisms presented so far involve many computations due to the recurrent layers used: each of these attentions is based on $\mseq{d}\mtseq{t}{}$. It results in long training and prediction times.

\subsubsection*{Gating mechanism}
In 2017, the authors of \cite{Dauphin2017} proposed a gating mechanism for \gls{cnn} architectures for the language modeling task. It is similar to the gating mechanisms of \gls{lstm} or \gls{gru} cells. However, instead of controlling the information flow through time, the gating mechanism proposed in \cite{Dauphin2017} focuses on the depth of the network \textit{i. e.} which information must be preserved from one layer to another.
This gating mechanism is defined as follows. Given a 2D input representation $\mb{X} \in \mathbb{R}^{H \times W \times C}$, two projections are computed the same way:
\begin{equation}
    \begin{split}
        & \mb{A} = \mb{X}\mb{W} + \mb{b},\\
        & \mb{B} = \mb{X}\mb{V} + \mb{c},\\
    \end{split}
\end{equation}
with $\mb{W}$ and $\mb{V}$ matrices of weights and $\mb{b}$ and $\mb{c}$ biases.
The output $\mb{Y}$ is defined as the element-wise product between $\mb{A}$ and the results of the sigmoid function applied on $\mb{B}$. This second operand takes its values between 0 and 1, leading to a filtering effect:
\begin{equation}
    \mb{Y} = \mb{A} \circ \sigma(\mb{B}).
\end{equation}

This gating mechanism can be seen as a form of self attention because it operates a selection over the input $\mb{X}$, which is based on $\mb{X}$ itself. It has also been successfully applied to \gls{nmt} in \cite{Gehring2017}.
Nowadays, transformers constitute the state-of-the-art model for a majority of tasks that process sequences.

\subsubsection*{The Transformer architecture}
\begin{figure}[h]
    \centering
    \includegraphics[width=0.8\linewidth]{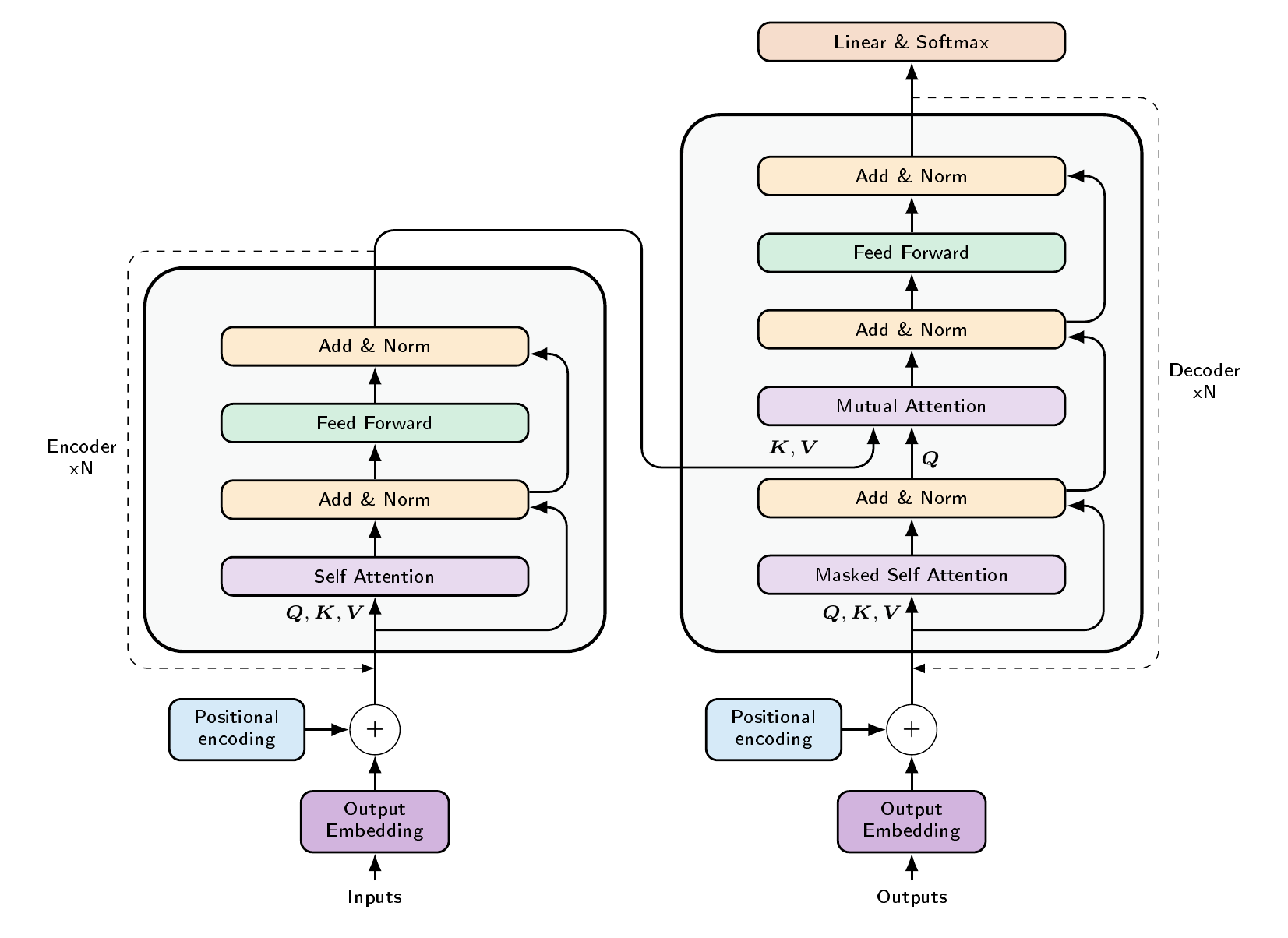}
    \caption{Transformer architecture. }
    \label{fig:transformer}
\end{figure}

In 2017, Vaswani \textit{et al.} proposed the Transformer architecture \cite{Vaswani2017}, depicted in Figure \ref{fig:transformer}, for the \acrshort{nmt} task. This architecture is made up of $N$ stacked encoder layers followed by $N$ stacked decoder layers. It is based on multi-head, scaled dot-product attention used for self and mutual attention. It also includes embedding, positional encoding and residual components. Contrary to the previous attention-based approaches, the Transformer architecture, although following a recurrent process, does not involve any recurrent layer. It means that the prediction $\hat{\mseq{y}}\mtseq{t}{}$ at timestep $t$ is based on the previous prediction $\hat{\mseq{y}}\mtseq{t-1}{}$. The main difference is that here, $\hat{\mseq{y}}\mtseq{t}{}$ is a raw token, while in the previous presented approaches, the model relied on the decoder hidden state at the previous time step. This difference enables to use teacher forcing during the Transformer training.

\begin{figure}[h!]
    \centering
    \includegraphics[width=\textwidth]{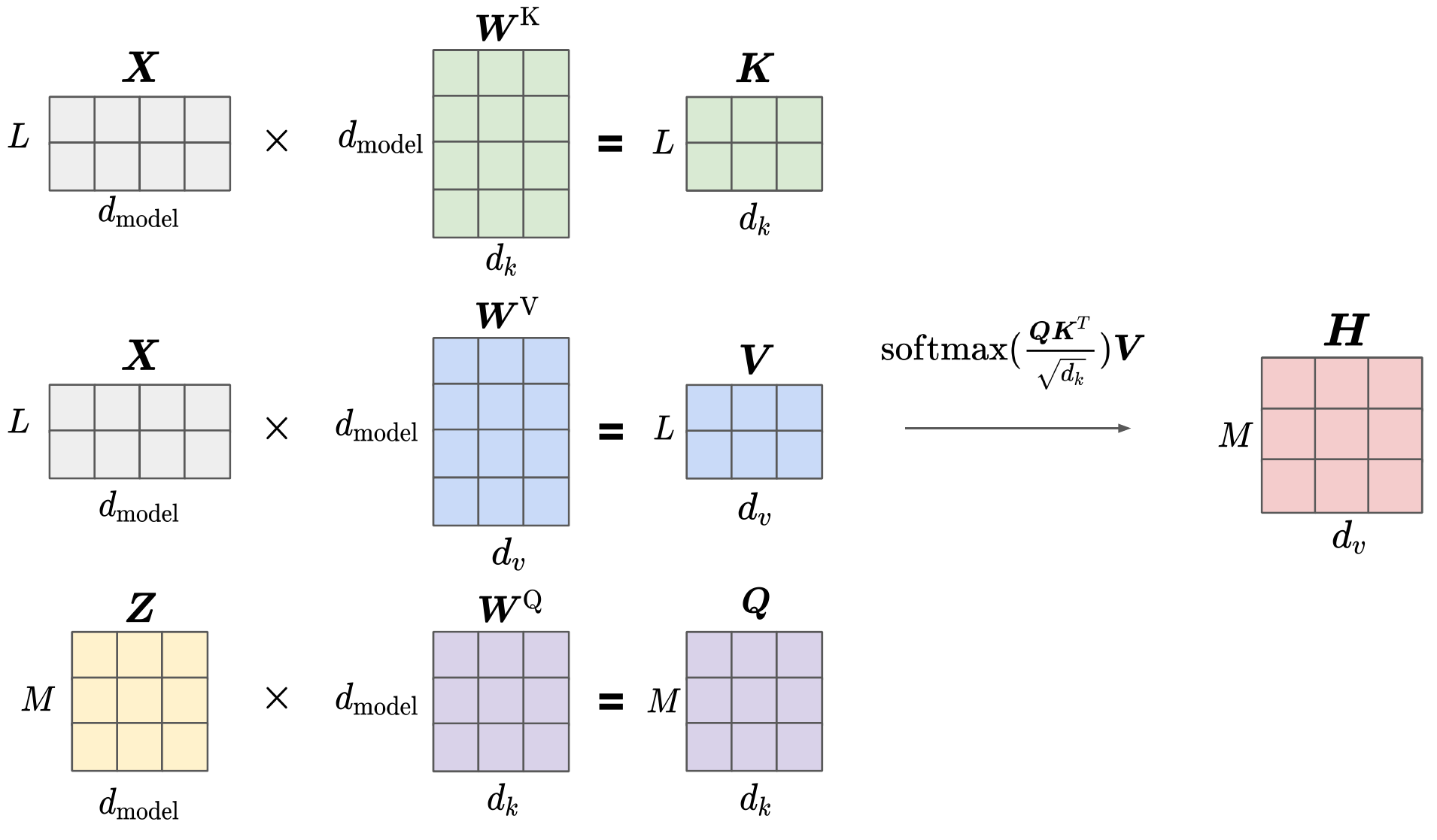}
    \caption{Scaled-dot product attention used in the Transformer architecture.}
    \label{fig:scaled-dot-product}
\end{figure}

Scaled dot-product attention follows the query-key-value paradigm. It is depicted in Figure \ref{fig:scaled-dot-product}. Let's take a set of items $\mset{X}=\msetdef{\mb{x}_1,...,\mb{x}_L}$, with $\mb{x}_i \in \mathbb{R}^{d_\mathrm{model}}$. They can be grouped in a matrix $\mb{X} \in \mathbb{R}^{L \times d_\mathrm{model}}$ with $\mb{X}_{i} = \mb{x_i}$. The aim is to retrieve the relevant information among these $\mb{x_i}$ with respect to a given query representation $\mb{z} \in \mathbb{R}^{d_\mathrm{model}}$. To this aim, we associate to each item $\mb{x}_i$ a key $\mb{k}_i \in \mathbb{R}^{d_k}$ and a value $\mb{v}_i \in \mathbb{R}^{d_v}$ through linear projection. In the same way, we associate a query vector $\mb{q} \in \mathbb{R}^{d_k}$ to its representation $\mb{z}$ trough linear projection to fit the dimension of the keys. This can be generalized for a set of $M$ query representations ($\mb{Z} \in \mathbb{R}^{M \times d_\mathrm{model}}$) with matrix computations:
\begin{equation}
    \begin{split}
        &\mb{K} = \mb{X}\mb{W}^K, \\
        &\mb{V} = \mb{X}\mb{W}^V, \\
        &\mb{Q} = \mb{Z}\mb{W}^Q, \\
    \end{split}
\end{equation}
with $\mb{W}^K \in \mathbb{R}^{d_\mathrm{model} \times d_k}$, $\mb{W}^V \in \mathbb{R}^{d_\mathrm{model} \times d_v}$ and $\mb{W}^Q \in \mathbb{R}^{d_\mathrm{model} \times d_k}$ weight matrices, and $\mb{K} \in \mathbb{R}^{L \times d_k}$, $\mb{V} \in \mathbb{R}^{L \times d_v}$ and $\mb{Q} \in \mathbb{R}^{M \times d_k}$.

The key $\mb{k}_i$ is used to compute a matching score with respect to a query $\mb{q}_j$ through dot product, scaled by a factor $\frac{1}{\sqrt{d_k}}$. The value $\mb{v}_i$ consists in the relevant information of the item $\mb{x}_i$ with respect to a given task. This way, given a query $\mb{q}_j$, a score is computed for each item $\mb{x}_i$ through their key $\mb{k}_i$. Probabilities are computed from these scores through softmax and associated to the corresponding values $\mb{v}_i$ through weighted sum:
\begin{equation}
   \mb{H} =\mathrm{Attention}(\mb{Q}, \mb{K}, \mb{V}) = \mathrm{softmax}\left(\frac{\mb{Q}\mb{K}^T}{\sqrt{d_k}}\right)\mb{V}.
\end{equation}

\begin{figure}[h!]
    \centering
    \includegraphics[width=\textwidth]{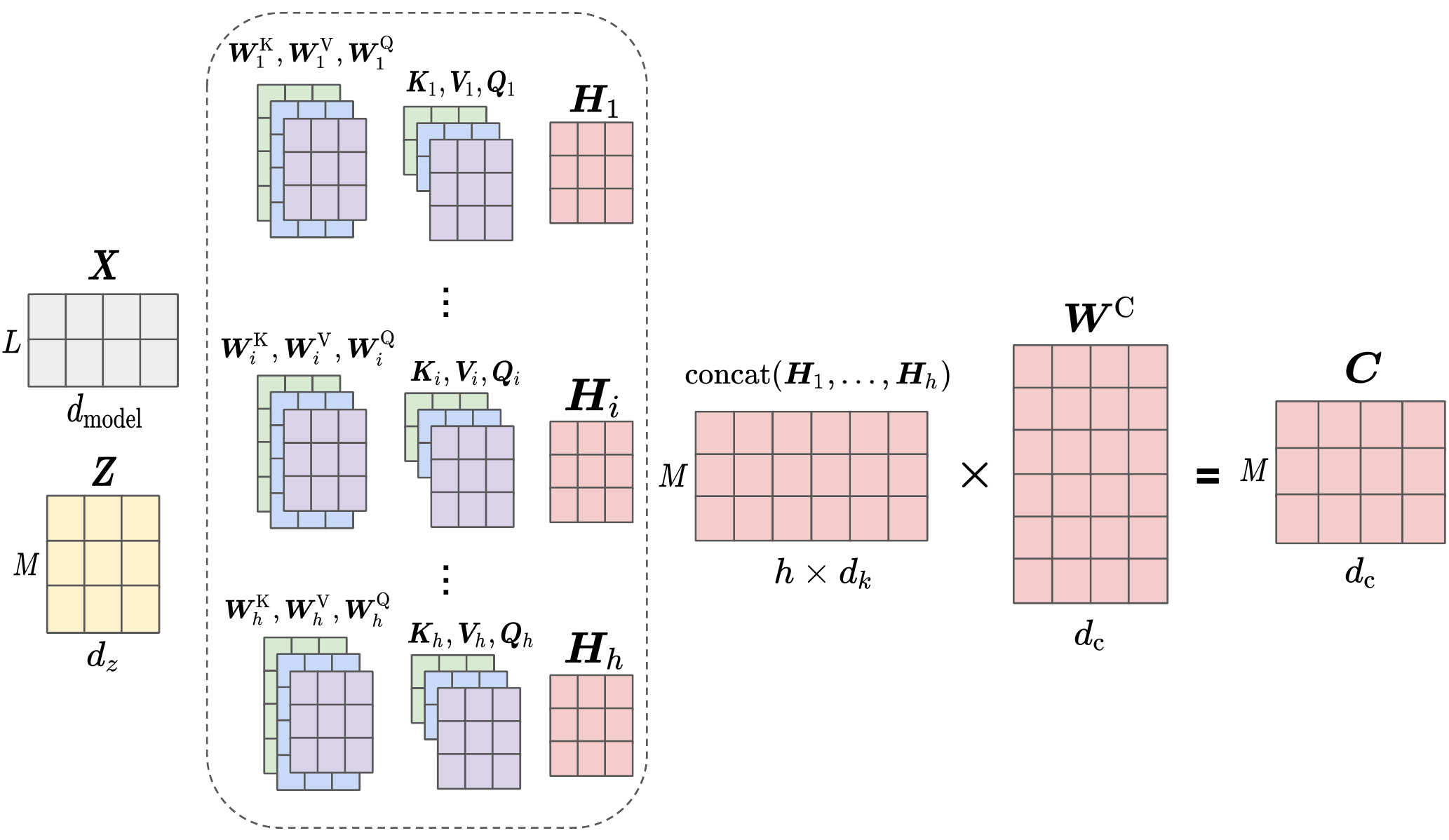}
    \caption{Multi-head scaled-dot product attention used in the Transformer architecture.}
    \label{fig:multi-head-attention}
\end{figure}

The authors also proposed the concept of multi-head attention, depicted in Figure \ref{fig:multi-head-attention}. It consists in computing the attention weights $h$ times (once per head $\mb{H}_i$) in parallel using different projections for the same queries, keys and values. The final context vectors (one per query) $\mb{C}$ is the projection of the concatenation of these intermediate results:
\begin{equation}
    \mb{C} = \mathrm{MultiHead}(\mb{Q}, \mb{K}, \mb{V}) = \mcats{\mb{H}_1, ..., \mb{H}_h}\mb{W}^C,
\end{equation}
with 
\begin{equation}
    \mb{H}_i = \mathrm{Attention}(\mb{Z}\mb{W}_i^Q, \mb{X}\mb{W}_i^K,\mb{X}\mb{W}_i^V).
\end{equation}
$\mb{W}^K_i \in \mathbb{R}^{d_\mathrm{model} \times d_k}$, $\mb{W}^V_i \in \mathbb{R}^{d_\mathrm{model} \times d_v}$, $\mb{W}^Q_i \in \mathbb{R}^{d_\mathrm{model} \times d_k}$ and $\mb{W}^C \in \mathbb{R}^{h \cdot d_v \times d_c}$ are matrices of weights.
The authors set $d_v=d_k=d_\mathrm{model}/h$ and $d_c=d_\mathrm{model}$.
Multi-head attention aims at combining information from different positions and in different sub-spaces to improve the expressiveness of the context vector. 

Based on these definitions, self attention refers to multi-head scaled dot-product applied with queries, keys and values from a same source sequence. In the encoder, the source sequence corresponds to the input. In the decoder, the source sequence is the previous predicted tokens.
Mutual attention is also a multi-head scaled dot-product attention, but the queries come from a target sequence, different from the source sequence, from which the keys and the values are computed. The target sequence is the already predicted tokens and the source sequence is the output of the encoder \textit{i. e.} a representation of the input. 

As mentioned previously, the Transformer architecture does not rely on any recurrent layer. This way, the queries cannot be based on a hidden state of a recurrent layer; it is directly based on raw predicted tokens. One can then use teacher forcing to parallelize the computations at training time. This way, a mask is used to prevent using future predictions to compute the current one. 
Since inputs and outputs are raw tokens, some embedding layers are used to increase the dimension of the model. 

In addition to removing the recurrent layers, the Transformer architecture does not rely on any convolutional component either. Dependencies are only modelized through the attention layers, by weighted sums of the sequence frames. However, there is no notion of sequence: frames are considered as a set, not as a sequence. To tackle this problem, the authors proposed to add positional encoding to the embedding in order to inject information about this sequentiality. The positional encoding used in this work corresponds to sine and cosine functions of different frequencies:
\begin{equation}
\begin{split}
    & \mathrm{PE}(j, 2k) = \sin(w_k \cdot j),\\
    & \mathrm{PE}(j, 2k+1) = \cos(w_k \cdot j),\\
    & \forall k \in [0, d_\mathrm{model}/2],\\
\end{split}
\end{equation}
with 
\begin{equation}
w_k = 1/10000^{2k/d_\mathrm{model}},
\end{equation}
where $j$ is the position and $k$ is the dimension.
Computing the positional encoding in this way bring many advantageous properties:
\begin{itemize}
    \item It is fixed and deterministic: it does not imply trainable weights.
    \item Positional encoding does not depend of the total length of the sequence.
    \item Encoding values are bounded and can generalize to unseen lengths thanks to the sine and cosine functions. 
    \item The encoding is unique for each position thanks to the combination of frequencies.
\end{itemize}

The Transformer architecture has exceeded the state of the art in many areas. However, there is still room for improvement. Indeed, the self-attention layer is a bottleneck when dealing with very long sequences: it requires quadratic computation time to produce the attention scores. In addition, the recurrent process combined with the stacked layers leads to inference time that could not satisfy most of industry applications. Many works attempted to improve the Transformer architecture to alleviate these issues: \cite{Transformer-XL,TurboTransformer,LongFormer,Reformer}.

The Transformer architecture, and more generally the attention-based and sequence-to-sequence architectures, are designed to deal with one-dimensional input sequences. As a result, standard sequence-to-sequence approaches cannot be used as is for \acrshort{htr}. On the other hand, we have seen that powerful computer vision architectures (\gls{cnn}, \gls{fcn}) were proposed to handle input images.
This way, it would be better to consider \acrshort{htr} as an in-between: an image-to-sequence (or document-to-sequence) problem, which is closer to reality. 

\section{HTR: an image-to-sequence problem}
On the one hand, the emergence of \acrshort{cnn} enabled to process images more efficiently, modeling spatially dependent information, also bringing translation equivariance and weight sharing. One can also reduce the number of trainable weights by using \acrlong{dsc}. In addition, one can deal with inputs of variable sizes using \acrshort{fcn}. On the other hand, recurrent layers and sequence-to-sequence architectures enabled to deal with one-dimensional input sequences and one-dimensional output sequences of variable lengths. 

\acrshort{htr}, as well as image captioning or scene text recognition, are part of the tasks that lie between computer vision and sequence-to-sequence problems. The evolution of the architectures dedicated to these tasks has followed those of these two fields. 

For instance, the image captioning task was first handled by a hybrid architecture mixing \acrshort{cnn} and \acrshort{lstm} as in \cite{ImageCaptioningCNNLSTM}. A \acrshort{cnn} encoder is used to extract features from the input image, generating a vector of fixed size. The \acrshort{lstm} layer is then used to predict the caption, word by word, based on this feature vector.

The authors of \cite{ImageCaptioningCNNLSTMAttention} proposed to introduce an attention mechanism in this kind of models. By doing this, one can use an \acrshort{fcn} in order to avoid collapsing the horizontal and vertical axes of the image for the extracted features. A flatten operation enables to reshape the features into a one-dimensional sequence. A self content-based attention mechanism, combined with an \acrshort{lstm} layer, is used to select the feature frames to focus on, while recurrently predicting the caption.

\acrshort{cnn} was then combined with a transformer decoder in \cite{Singh} for \acrshort{htr}. 1D positional encoding is superseded by 2D positional encoding for the flattened features to preserve the 2D nature of the image. It means that the position is still encoded using sine and cosine functions, but half of the dimensions are dedicated to the horizontal position and the other half to the vertical position.

Whether it is \acrshort{cnn}+\acrshort{lstm}, \acrshort{cnn}+\acrshort{lstm} with soft attention or \acrshort{cnn}+transformer, these architectures only represent a concatenation of an encoder sub-network coming from computer vision for feature extraction, and a decoder sub-network from the sequence-to-sequence paradigm for sequence prediction. The connection of these two sub-networks is made possible by some tricks: the flatten operation, the 2D positional encoding. 

More recently, the authors of \cite{VisionTransformer} proposed the \gls{vit} for the task of image captioning. It consists in applying the transformer encoder on the raw image pixels, without using any convolutional component. To this end, they apply a pre-processing stage to the input image: the image is split into fixed-size patches and flattened, before being linearly projected. An \acrshort{mlp} is used after the transformer encoder to predict the class probabilities. This concept has also been used in \cite{Chen2021} for image classification and adapted to image-to-sequence tasks by also using the transformer decoder part, such as in \cite{VITIC} for image captioning or in \cite{VITSTR} for scene text recognition. 
 
\acrlong{vit} provides similar or even better results compared to \acrshort{cnn}-based approaches. It could be due to the difficulty of \acrshort{cnn} to model very long dependencies which depends on the size of the receptive field. On the contrary, transformer-based approaches rely on self attention which consists in a soft attention over the whole input sequence: dependencies between first and final items can theoretically be modeled as of the first encoder layer. 

We can therefore ask ourselves if the \acrshort{cnn} are still relevant for image-to-sequence problems. From these first works, it seems that \acrshort{vit} are prevalent only for very large datasets. Indeed, \acrshort{cnn} would perform better with few data compared to \acrshort{vit} because it encodes prior knowledge about the image domain like translation equivariance and spatial dependencies; \acrshort{vit} must learn it.

We have formulated the \gls{htr} task as an image-to-sequence problem and we have seen the major issues related to the use of deep neural networks to handle such a complex task. We will now focus on the evaluation of the performance of such systems.

 \section{Evaluation environment for HTR}
To compare objectively the performance of different neural networks, one must evaluate the different approaches in the same conditions. This way, we evaluated our proposed models on public datasets and using the same metrics.

\subsection{Datasets}
\label{section:dataset}
There are very few available public datasets for the \gls{htr} task. To evaluate the approaches we proposed, we used the three following public datasets: RIMES, IAM and READ. Although these datasets provide images of whole pages, they are generally annotated for line-level approaches, as we will see in the next chapter.

\subsubsection{RIMES}
RIMES (Recognition and Indexing of handwritten documents and faxes) is a project funded by the French ministries of defense and research. Its goal was to create a new dataset of mixed (handwritten and typed) documents, dedicated to varied tasks representative of industrial applications. The idea was to cover the following tasks: layout analysis, handwriting recognition, information extraction, writer identification and logo identification.

\begin{figure}[ht!]
    \centering
    \begin{subfigure}[b]{0.3\textwidth}
    \includegraphics[width=\textwidth,frame]{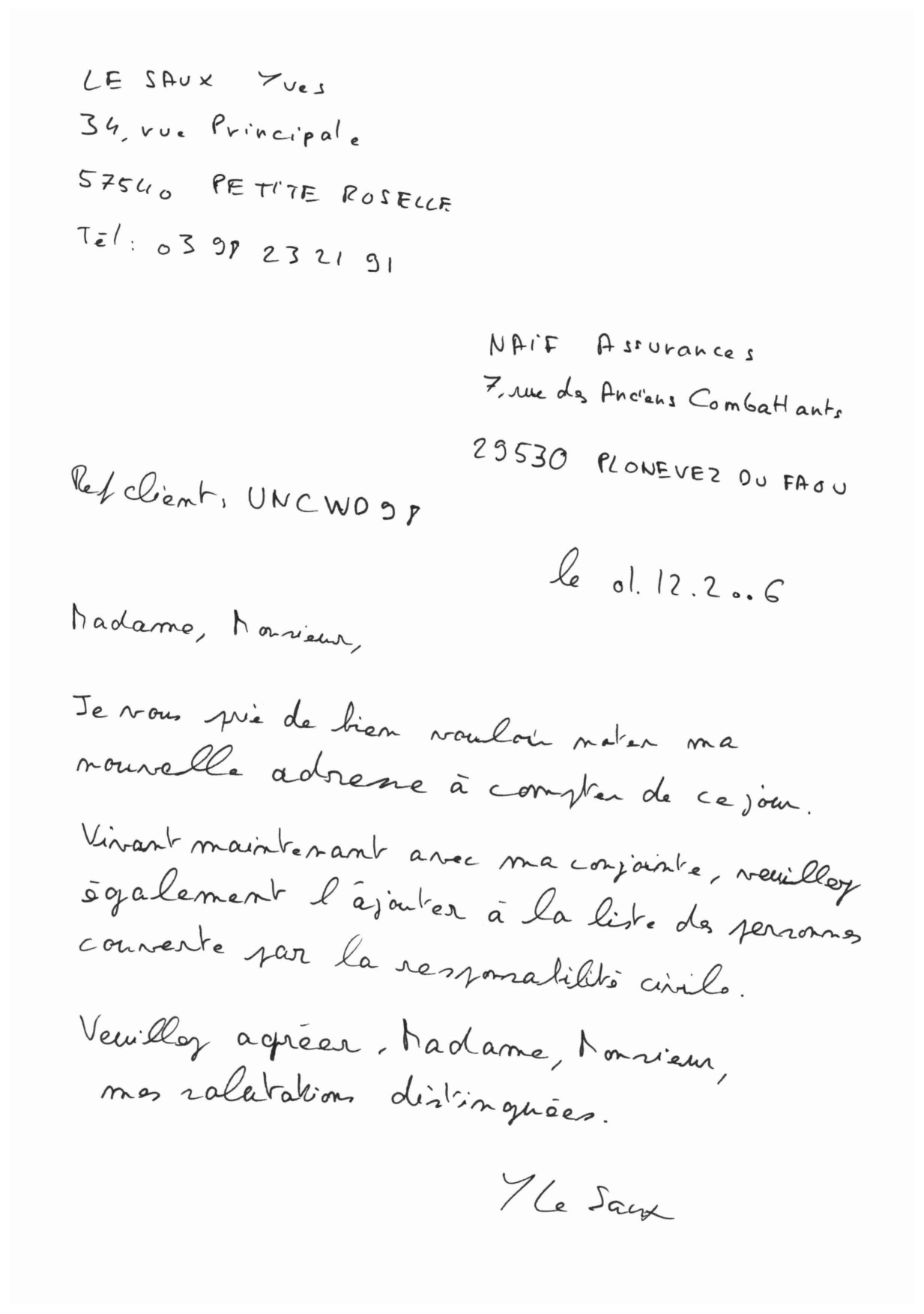}
    \caption{RIMES 2009 - page.}
    \end{subfigure}
    \hfill
    \begin{minipage}[b]{0.3\textwidth}
    \begin{subfigure}[b]{\textwidth}
    \centering
    \includegraphics[width=0.5\textwidth,frame]{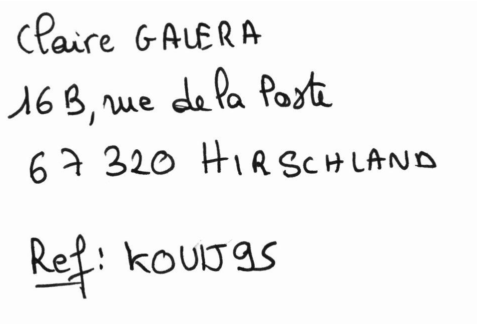}
    \par\bigskip
    \caption{RIMES 2009 - paragraph.}
    \end{subfigure}
    \par\bigskip
    \begin{subfigure}[b]{\textwidth}
    \centering
    \includegraphics[width=\textwidth,frame]{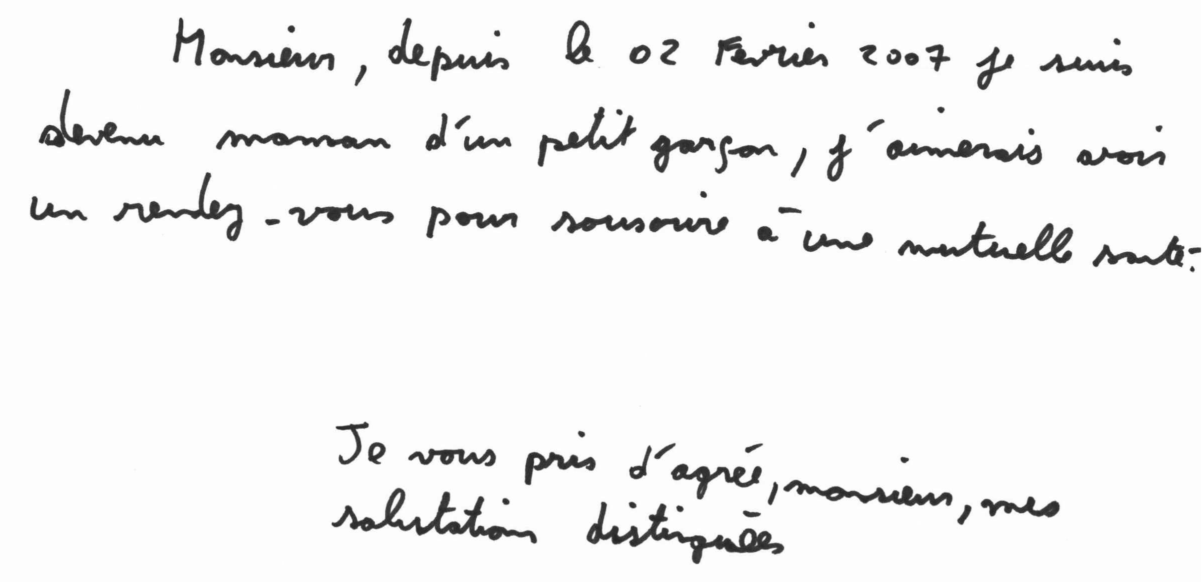}
    \par\bigskip
    \caption{RIMES 2011 - paragraph.}
    \end{subfigure}
    \end{minipage}
    \hfill
    \begin{subfigure}[b]{0.3\textwidth}
    \centering
    \includegraphics[width=0.7\textwidth, frame]{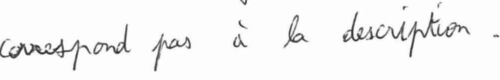}
    \par\bigskip
    \par\bigskip
    \includegraphics[width=0.5\textwidth, frame]{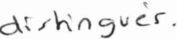}
    \par\bigskip
    \par\bigskip
    \includegraphics[width=\textwidth, frame]{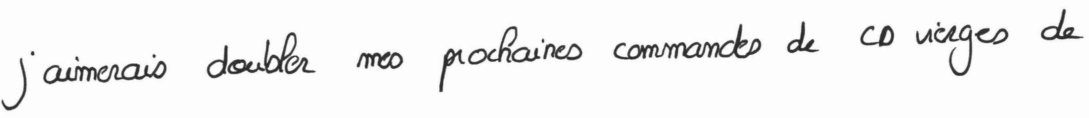}
    \par\bigskip\par\bigskip\par\bigskip\par\bigskip
    \caption{RIMES 2011 - line.}
    \end{subfigure}
    \caption{RIMES dataset examples.}
    \label{fig:rimes_dataset}
\end{figure}

The RIMES dataset corresponds to 12,723 pages collected from volunteers in the form of 5,605 handwritten mails of two to three pages, whose content corresponds to a fictional correspondence between an individual and a company. Each mail is composed of a handwritten letter page, a form page and an optional fax cover sheet page. The volunteers could write up to 5 mails, each one following one of the 9 proposed scenarii, such as change of personal information or modification of contract. These mails have been scanned at a resolution of 300 \acrshort{dpi}, with a gray-scale encoding. 

It led to two main evaluation campaigns and a competition, providing two sub-datasets from this RIMES dataset in 2009 and 2011, respectively. We named them RIMES 2009 and RIMES 2011.

\subsubsection*{RIMES 2009}
RIMES 2009 corresponds to the dataset used in the second evaluation campaign \cite{RIMES_page}. It is made up of 1,050 pages for training, 100 pages for validation and 100 pages for test. All the pages are the first of a subset of the mails \textit{i. e.} single-page handwritten letters. RIMES 2009 was proposed for letter layout analysis, handwriting recognition and writer identification. Text regions are localized and classified into 8 categories: sender, recipient, date \& location, subject, opening, body, signature and PS \& attachment. This way, for a given document image, there are many sequences of characters which represent the different text regions, each one associated to a given class. Bounding boxes are also provided for each text region; this enables to also process the dataset for paragraph-level \gls{htr}. 

It has to be noted that this version of the dataset was only made available to participants of the evaluation campaign. At that time, the recognition tasks only consisted in recognizing isolated characters or words. To our knowledge, we are the first to use it at paragraph and page levels.

\subsubsection*{RIMES 2011}
RIMES 2011 corresponds to the dataset provided in the handwriting competition at ICDAR 2011 \cite{RIMES_paragraph}. Images are bodies of a subset of the handwritten letters. RIMES 2011 dataset also provides annotations of bounding boxes of the text lines, leading to a line-level version of RIMES 2011. This way, the images of bodies are split into text line images and associated to the corresponding ground truth transcription.

Samples of the different versions of the RIMES dataset are shown in Figure \ref{fig:rimes_dataset}.
Table \ref{tab:rimes_stats} provides statistical information about the different RIMES datasets. As one can note, samples from RIMES 2009 contain up to 2,719 characters at page level, with a mean of 578 characters. One can also notice that the paragraphs from RIMES 2009 are more diverse in terms of image size and text length. This is due to the fact that paragraphs from RIMES 2011 only correspond to bodies of letters whereas paragraphs from RIMES 2009 can be of any class (body, recipient or sender coordinates, or subject for instance).
\begin{table}[ht!]
    \caption{RIMES datasets statistics.}
    \label{tab:rimes_stats}
    \centering
    \begin{tabular}{l c c c c c c}
    \hline
         & Min. & Q1 & Median & Q3 & Max. & Mean \\
        \hline
        \hline
        \textbf{RIMES 2009 - page}\\
        Width (px) & 2,446 & 2,466 & 2,470 & 2,472 & 2,480 & 2,469 \\ 
        Height (px) & 3,464 & 3,500 & 3,502 & 3,504 & 3,512 & 3,502 \\ 
        Lines & 5 & 16 & 18 & 20 & 43 & 18 \\ 
        Words & 36 & 93 & 113 & 138 & 558 & 120 \\ 
        Chars & 157 & 454 & 550 & 660 & 2,719 & 578 \\ 
        \textbf{RIMES 2009 - paragraph}\\
        Width (px) & 112 & 738 & 928 & 1,297 & 2,464 & 1,123 \\ 
        Height (px) & 64 & 150 & 298 & 514 & 2,276 & 431 \\ 
        Lines & 1 & 1 & 3 & 4 & 24 & 3 \\ 
        Words & 1 & 5 & 9 & 20 & 414 & 21 \\ 
        Chars & 3 & 21 & 47 & 79 & 2,043 & 103 \\ 
        \textbf{RIMES 2011 - paragraph}\\
        Width (px) & 1,464 & 2,108 & 2,228 & 2,336 & 2,468 & 2,204 \\ 
        Height (px) & 346 & 846 & 1,061 & 1,282 & 2,132 & 1,081 \\ 
        Lines & 2 & 6 & 7 & 9 & 18 & 8 \\ 
        Words & 12 & 48 & 65 & 85 & 245 & 70 \\ 
        Chars & 71 & 244 & 322 & 422 & 1,182 & 348 \\ 
        \textbf{RIMES 2011 - line} \\
        Width (px) & 96 & 1,346 & 1,828 & 2,044 & 2,462 & 1,636 \\ 
        Height (px) & 34 & 104 & 126 & 152 & 364 & 129 \\ 
        Words & 1 & 7 & 9 & 12 & 24 & 9 \\ 
        Chars & 2 & 34 & 47 & 57 & 110 & 45 \\ 
         \hline
    \end{tabular}
\end{table}

In Table \ref{tab:rimes_split}, we specified the split used in terms of training, validation and test, as well as the charset size and the total number of characters in the dataset. For RIMES 2011 at line level, there are two splits used in the literature: line-1, which is the line version of the RIMES 2011 dataset at paragraph level; and line-2 which is the mainly used split by the community. For line-2, text lines from a same paragraph can be found in both training and validation sets for example. This way, line-1 is to be preferred when pre-training on line images before training on paragraph images. The difference of characters from page to paragraph or from paragraph to lines can be explained by the addition of some line breaks. It can also be due to annotation errors.

\begin{table}[h!]
    \caption{RIMES datasets split.}
    \label{tab:rimes_split}
    \centering
    \begin{tabular}{l c c c c c}
    \hline
     & Training & Validation & Test & charset size & \# chars\\
    \hline
    \hline
    RIMES 2009 - page & 1,050 & 100 & 100  & 108 & 723k\\
    RIMES 2009 - paragraph & 5,875 & 540 &  559 & 108 & 717k\\
    RIMES 2011 - paragraph & 1,400 & 100 & 100  & 98 & 556k\\
    RIMES 2011 - line-1 & 10,530 & 801 & 778  & 97 & 546k\\
    RIMES 2011 - line-2 & 9,947 & 1,333 & 778  & 97 & 543k\\
    \hline
    \end{tabular}
\end{table}

\subsubsection{IAM}
The IAM dataset was first published in 1999 in the International Conference on Document Analysis and Recognition (ICDAR) \cite{IAM99}, and updated in \cite{IAM}. We used the third version of this dataset. It consists in 1,539 form pages of handwritten English text scanned at a resolution of 300 \acrshort{dpi}. Scans are saved as PNG images with 256 gray levels.  The writers, 657 in number, were asked to copy the printed text at the top of a form. The printed texts correspond to fragments of text from the Lancaster-Oslo/Bergen (LOB) corpus \cite{LOB}, a collection of 500 English texts.

\begin{figure}[h!]
    \centering
    \hfill
    \begin{subfigure}[b]{0.3\textwidth}
    \centering
    \includegraphics[width=\textwidth,frame]{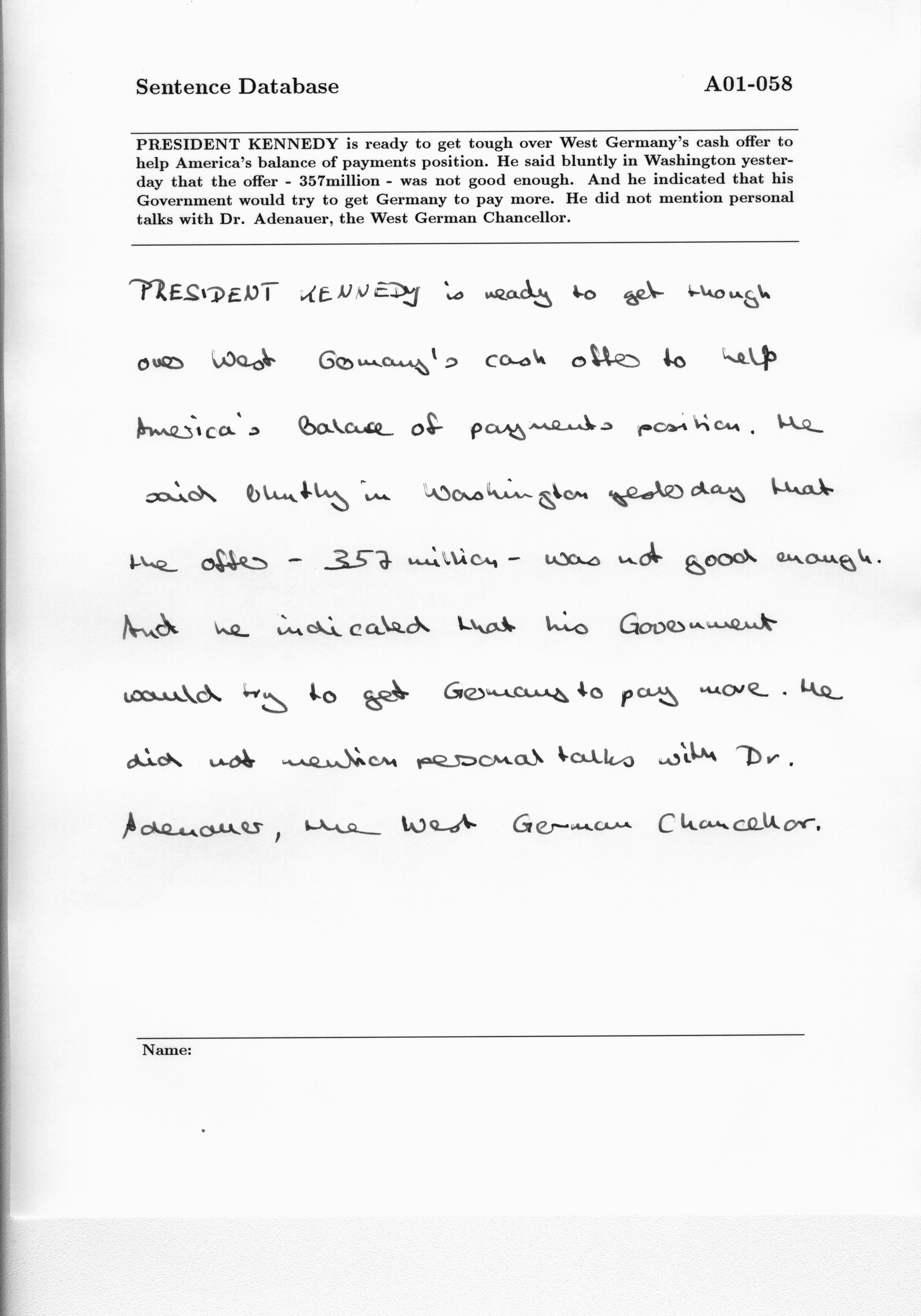}
    \caption{IAM - Original form.}
    \end{subfigure}
    \hfill
    \begin{subfigure}[b]{0.3\textwidth}
    \centering
    \includegraphics[width=0.9\textwidth,frame]{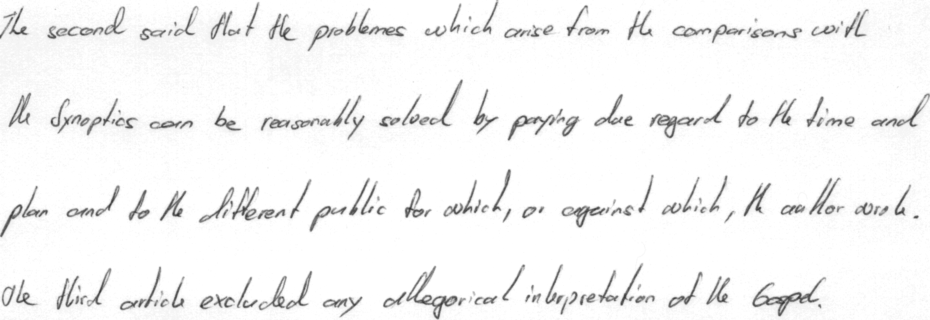}
    \par\bigskip
    \includegraphics[width=\textwidth,frame]{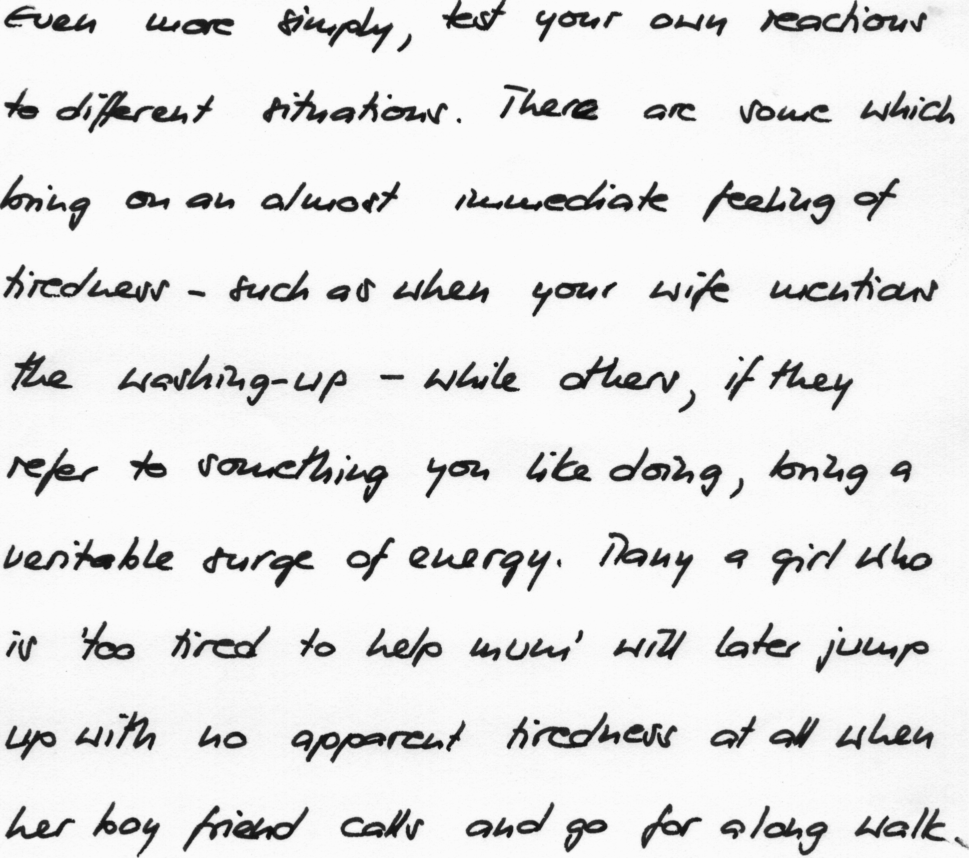}
    \caption{IAM - paragraph.}
    \end{subfigure}
    \hfill
    \begin{subfigure}[b]{0.3\textwidth}
    \centering
    \includegraphics[width=\textwidth, frame]{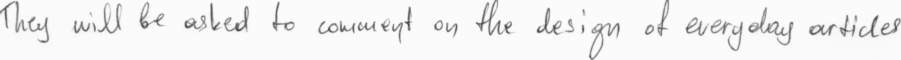}
    \par\bigskip\par\bigskip\par\bigskip
    \includegraphics[width=0.2\textwidth, frame]{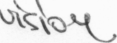}
    \par\bigskip\par\bigskip\par\bigskip
    \includegraphics[width=0.9\textwidth, frame]{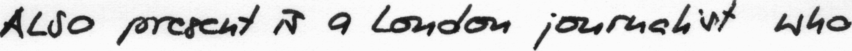}
    \par\bigskip\par\bigskip\par\bigskip
    \caption{IAM - line.}
    \end{subfigure}
    \hfill
    \caption{IAM dataset examples.}
    \label{fig:iam_dataset}
\end{figure}

The IAM dataset includes ground truth transcription as well as segmentation annotation at several levels: paragraph (body), line and word. We only used the paragraph and line levels.

Samples of the IAM dataset are shown in Figure \ref{fig:iam_dataset}.
We did not use the original forms from the IAM dataset because the printed text part consists in the exact same text than the handwriting part. Indeed, this would have biased the training. As one can note, the writing style can differ a lot from one writer to another, comparing the two paragraph images. 

\begin{table}[h!]
    \caption{IAM datasets statistics.}
    \label{tab:iam_stats}
    \centering
    \begin{tabular}{l c c c c c c}
    \hline
         & Min. & Q1 & Median & Q3 & Max. & Mean \\
        \hline
        \hline

        \textbf{Paragraph}\\
        Width (px) & 1,574 & 1,816 & 1,874 & 1,950 & 2,266 & 1,891 \\ 
        Height (px) & 356 & 1,253 & 1,522 & 1,730 & 2,102 & 1,488 \\ 
        Lines & 2 & 7 & 9 & 10 & 13 & 9 \\ 
        Words & 13 & 70 & 78 & 87 & 150 & 79 \\ 
        Chars & 52 & 340 & 379 & 420 & 634 & 382 \\ 
        \textbf{Line} \\
        Width (px) & 100 & 1,672 & 1,758 & 1,834 & 2,260 & 1,699 \\ 
        Height (px) & 38 & 100 & 120 & 144 & 342 & 124 \\ 
        Words & 1 & 7 & 9 & 11 & 53 & 9 \\ 
        Chars & 1 & 38 & 43 & 49 & 93 & 43 \\ 
         \hline
    \end{tabular}
\end{table}

Statistics about IAM datasets are given in Table \ref{tab:iam_stats}. Looking at both paragraph and line levels, one can notice that the IAM dataset follows the same trend than the RIMES 2011 dataset, with an increasing size variability from paragraphs to lines. IAM paragraphs are made up of 2 to 13 lines, with a mean of 9 lines. It is a bit less than for paragraphs of RIMES 2011. The number of characters per line is, on average, nearly identical compared to the RIMES 2011 dataset with 43 characters compared to 45.

We used the Aachen split \cite{Aachen} for both paragraph and line datasets, as detailed in Table \ref{tab:iam_split}.
\begin{table}[h!]
    \caption{IAM datasets split.}
    \label{tab:iam_split}
    \centering
    \begin{tabular}{l c c c c c}
    \hline
         & Training & Validation & Test & charset size & \# chars \\
        \hline
        \hline
        Paragraph & 747 & 116 & 336  & 80 & 458k\\
        Line & 6,482 & 976 & 2,915  & 79 & 449k\\
         \hline
    \end{tabular}
\end{table}

\subsubsection{READ}
READ (Recognition and Enrichment of Archival Documents) is a European project whose mission was to improve the access to archival documents through the use of \gls{htr} and keyword spotting technologies. One of the READ datasets was proposed in a competition of the International Conference on Frontiers in Handwriting Recognition (ICFHR) in 2016 \cite{READ2016}. We named it READ 2016. 

\begin{figure}[ht!]
    \centering
    \begin{subfigure}[b]{0.64\textwidth}
    \centering
    \includegraphics[width=0.7\textwidth,frame]{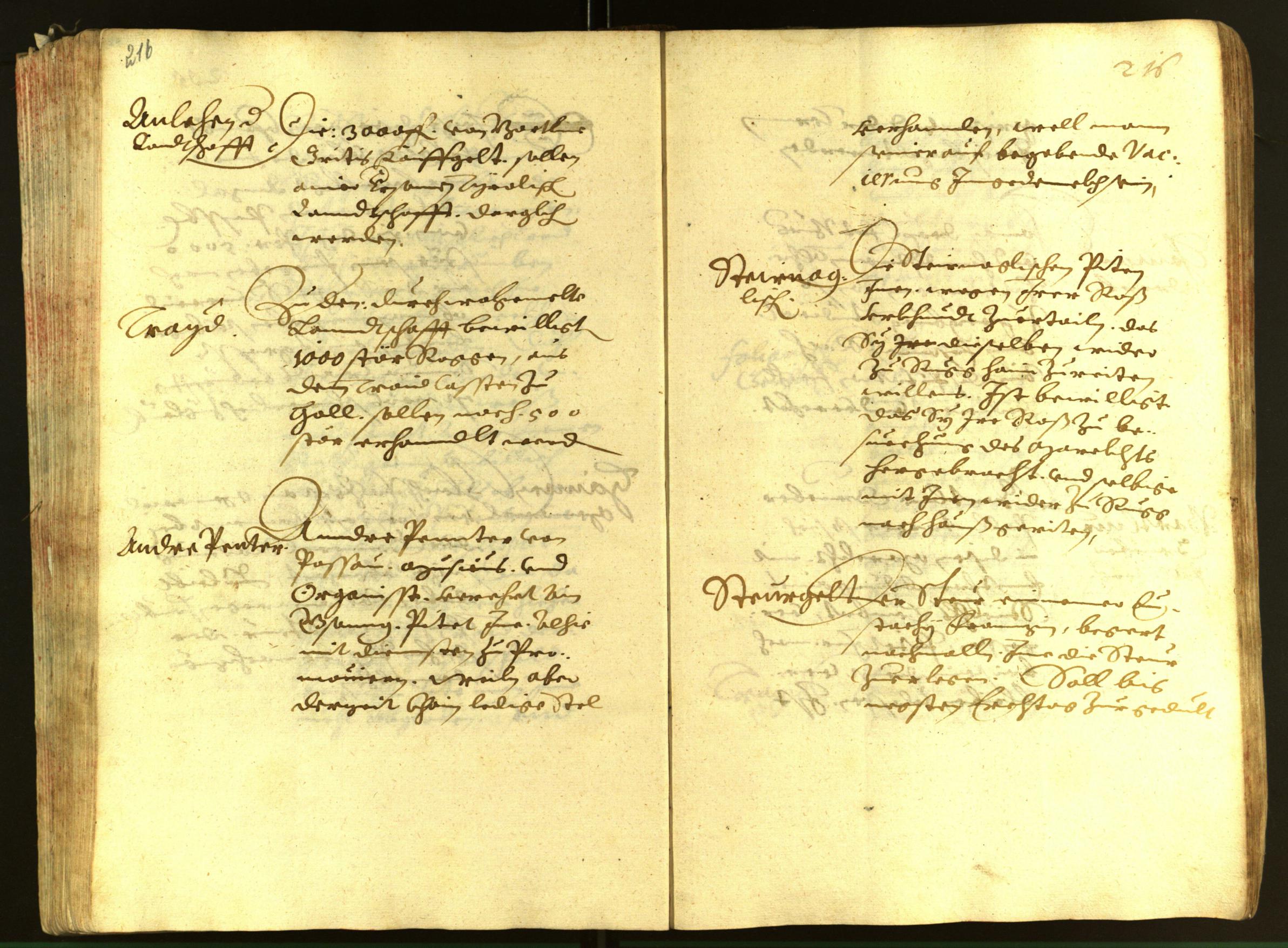}
    \caption{READ-double-page.}
    \end{subfigure}
    \hfill
    \begin{subfigure}[b]{0.32\textwidth}
    \centering
    \includegraphics[width=0.7\textwidth,frame]{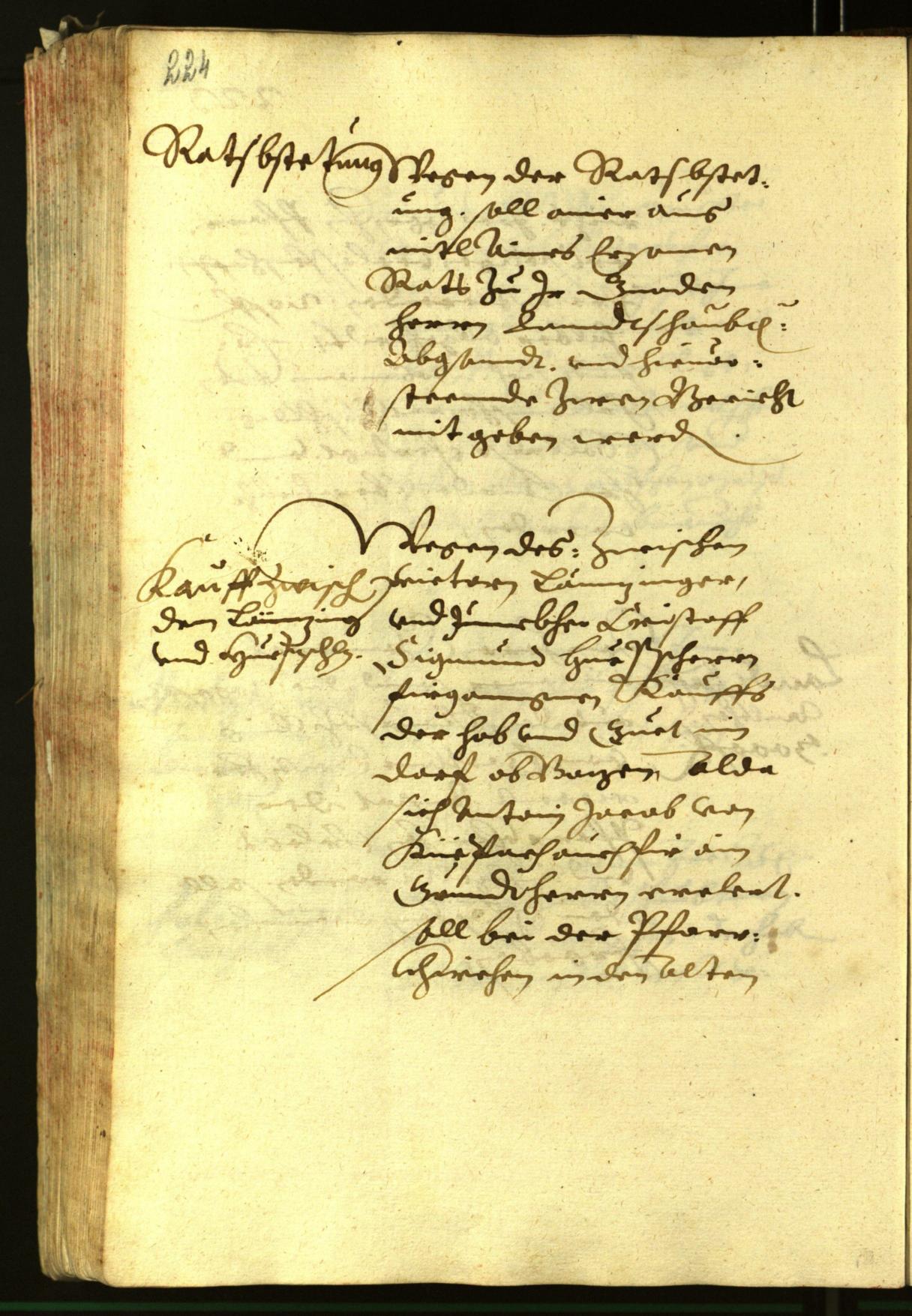}
    \caption{READ-page.}
    \end{subfigure}
    \hfill
    \par\bigskip
    \begin{subfigure}[b]{0.45\textwidth}
    \centering
    \includegraphics[width=0.4\textwidth,frame,valign=c]{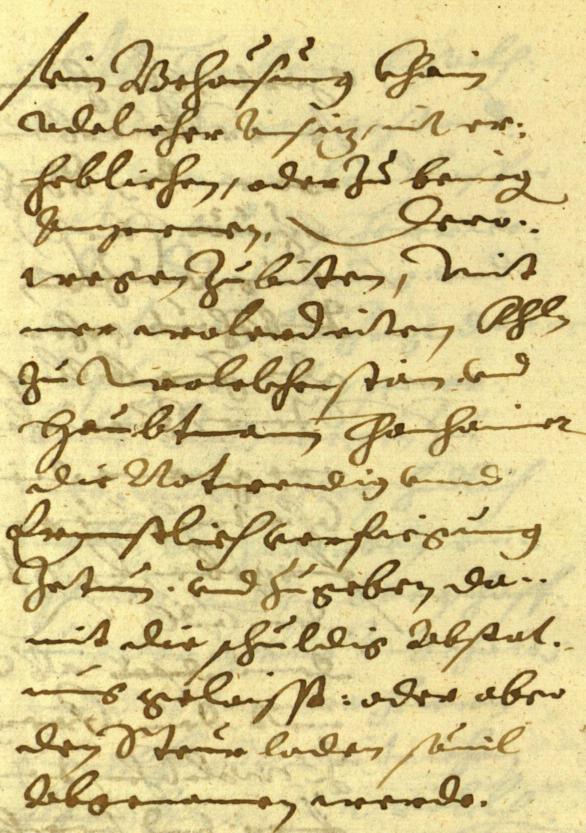}
    \centering
    \includegraphics[width=0.2\textwidth,frame,valign=c]{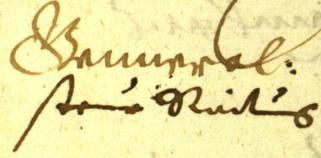}
    \caption{READ-paragraph.}
    \end{subfigure}
    \hfill
    \begin{subfigure}[b]{0.45\textwidth}
    \centering
    \includegraphics[width=0.4\textwidth, frame]{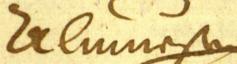}
    \par\bigskip
    \includegraphics[width=0.5\textwidth, frame]{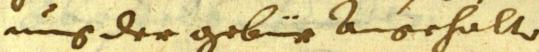}
    \par\bigskip
    \includegraphics[width=0.05\textwidth, frame]{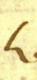}
    \caption{READ-line.}
    \end{subfigure}
    \hfill
    \caption{READ 2016 dataset examples.}
    \label{fig:read_dataset}
\end{figure}

The READ 2016 dataset is made up of 450 pages of Early Modern German. It consists of a subset of documents from the Ratsprotokolle collection, which is composed of about 30,000 pages of minutes of the council meetings, dated from 1470 to 1805. The number of writers is unknown. Contrary to RIMES or IAM, it is encoded as RGB images.

This dataset is originally at single-page level. We generated a double-page version of this dataset by merging successive pages. Only very few pages were not paired: we removed them from the dataset. This dataset also provides, in addition to the text annotation, segmentation labels at paragraph and line levels. We used them for the \gls{htr} task at these levels.

Samples of the READ 2016 datasets are shown in Figure \ref{fig:read_dataset}. As one can notice, handwriting from READ 2016 are far more difficult to read than those of RIMES and IAM. This is mainly due to the background, noised by the bleed-through effect, combined with the historical way of writing and the language.

Statistics about this dataset are provided in Table \ref{tab:read_stats}.
As for RIMES 2009 at page level, samples from READ 2016 at single-page and double-page levels are close in terms of size. 
One should note that the paragraphs also integrate images of page number leading to very small images of a single character. It results in a huge variety in the text length going from 1 to 758 characters.
\begin{table}[h!]
    \caption{READ 2016 datasets statistics.}
    \label{tab:read_stats}
    \centering
    \begin{tabular}{l c c c c c c}
    \hline
         & Min. & Q1 & Median & Q3 & Max. & Mean \\
        \hline
        \hline
        \textbf{Double-page}\\
        Width (px) & 4,714 & 4,770 & 4,786 & 4,802 & 4,874 & 4,793 \\ 
        Height (px) & 3,494 & 3,510 & 3,510 & 3,534 & 3,630 & 3,523 \\ 
        Lines & 18 & 45 & 48 & 50 & 59 & 47 \\ 
        Words & 48 & 177 & 196 & 218 & 276 & 194 \\ 
        Chars & 296 & 963 & 1,072 & 1,204 & 1,455 & 1,061 \\ 
        \textbf{Single-page}\\
        Width (px) & 2,168 & 2,346 & 2,394 & 2,450 & 2,562 & 2,398 \\ 
        Height (px) & 3,494 & 3,510 & 3,510 & 3,534 & 3,630 & 3,523 \\ 
        Lines & 6 & 22 & 24 & 25 & 31 & 23 \\ 
        Words & 18 & 86 & 98 & 111 & 153 & 97 \\ 
        Chars & 95 & 466 & 536 & 604 & 807 & 528 \\ 
        \textbf{Paragraph}\\
        Width (px) & 64 & 430 & 716 & 1,340 & 1,992 & 863 \\ 
        Height (px) & 34 & 176 & 352 & 910 & 3,000 & 636 \\ 
        Lines & 1 & 1 & 3 & 8 & 26 & 5 \\ 
        Words & 1 & 2 & 8 & 33 & 143 & 22 \\ 
        Chars & 1 & 7 & 39 & 186 & 758 & 119 \\ 
        \textbf{Line} \\
        Width (px) & 42 & 930 & 1,056 & 1,164 & 1,626 & 962 \\ 
        Height (px) & 22 & 100 & 114 & 132 & 514 & 121 \\ 
        Words & 1 & 3 & 4 & 5 & 13 & 4 \\ 
        Chars & 1 & 19 & 23 & 27 & 43 & 22 \\ 
         \hline
    \end{tabular}
\end{table}

The different splits used in this thesis are detailed in Table \ref{tab:read_split}. One can note that there are fewer examples at single-page and double-page levels than in RIMES 2009 at page level. However, for the line-level version, it is comparable to RIMES and IAM.

\begin{table}[h!]
    \caption{READ 2016 datasets split.}
    \label{tab:read_split}
    \centering
    \begin{tabular}{l c c c c c}
    \hline
         & Training & Validation & Test & charset size & \# chars \\
        \hline
        \hline
        Double-page & 169 & 24 & 24  & 89 & 230k\\
        Single-page & 350 & 50 & 50  & 89 & 238k\\
        Paragraph & 1,602 & 182 & 199  & 89 & 236k\\
        Line & 8,367 & 1,043 & 1,140  & 88 & 227k\\
         \hline
    \end{tabular}
\end{table}

\subsection{Metrics}
\label{section:metrics}

The \gls{htr} task is generally handled in three steps, including a segmentation stage and a recognition stage. The segmentation part aims at detecting the text regions in the input images, and the recognition part aims at recognizing the sequence of characters inside these different text regions.
We now go over the main metrics used in both segmentation and recognition stages.

\subsubsection{Segmentation metrics}
\label{section-metric-segmentation}
Segmentation predictions and annotations are generally bounding boxes representing text regions. The main difficulty to compute the metrics is that the number of predictions may not match the actual number of text regions in the ground truths: the model can predict 3 text lines whereas there are 4 text lines in the image, for instance. 
Two metrics are generally used to evaluate the performance of the segmentation stage: the \gls{iou} and the \gls{map}. 

\subsubsection*{IoU}
The \acrlong{iou} is defined as the area of the intersection between the ground truth $\mb{Y}$ and the prediction $\hat{\mb{Y}}$, divided by the area of the union of these two areas:
\begin{equation}
\mathrm{IoU} = \frac{\mb{Y} \cap \hat{\mb{Y}}}{\mb{Y} \cup \hat{\mb{Y}}}.
\end{equation}
Calculated in this way, the \gls{iou} enables to measure the quality of the segmentation process. On the one hand, it ensures that the prediction encompasses the annotation as much as possible, via the intersection. And on the other hand, it avoids the over-segmentation thanks to the division by the union, preventing to predict the whole document as a text region.

In the case of multiple text regions in a same input image, the \gls{iou} can be averaged between the different predictions. Predictions are ordered by their confidence score and associated to the ground truth which leads to the highest \gls{iou}. 

\subsubsection*{mAP}
An alternative metric to evaluate the segmentation stage is the \acrlong{map} \cite{mAP_COCO,mAP_PascalVOC}. It corresponds to a mean of the Average Precision (AP) of the different classes to be recognized. AP is computed as the area under the precision-recall curve. 
Precision and recall are metrics based on ratios of True Positive (TP), False Positive (FP) and False Negative (FN).
The precision corresponds to the ratio of correct predictions over all the predictions:
\begin{equation}
    \mathrm{Precision} = \frac{\mathrm{TP}}{\mathrm{TP}+\mathrm{FP}},
\end{equation}
while the recall measures the ratio of predictions that match the annotation:
\begin{equation}
    \mathrm{Recall} = \frac{\mathrm{TP}}{\mathrm{TP}+\mathrm{FN}}.
\end{equation}

\begin{figure}[h!]
    \begin{subfigure}{0.4\textwidth}
        \centering
        \resizebox{\textwidth}{!}{
        \begin{tabular}{c c c c c }
        \hline
         Rank & TP/FP & Precision & Recall & $p_\mathrm{inter}$ \\
        \hline
        \hline
        1 & TP & 1/1 & 1/4 & 1 \\
        2 & FP & 1/2 & 1/4 & 1 \\
        3 & TP & 2/3 & 2/4 & 3/4 \\
        4 & TP & 3/4 & 3/4 & 3/4 \\
        5 & FP & 3/5 & 3/4 & 3/4 \\
        6 & TP & 4/6 & 4/4 & 4/6 \\
        7 & FP & 4/7 & 4/4 & 4/6 \\
         \hline
        \end{tabular}
        }
    \end{subfigure}
    \hfill
    \begin{subfigure}{0.55\textwidth}
    \centering
    \includegraphics[width=\textwidth]{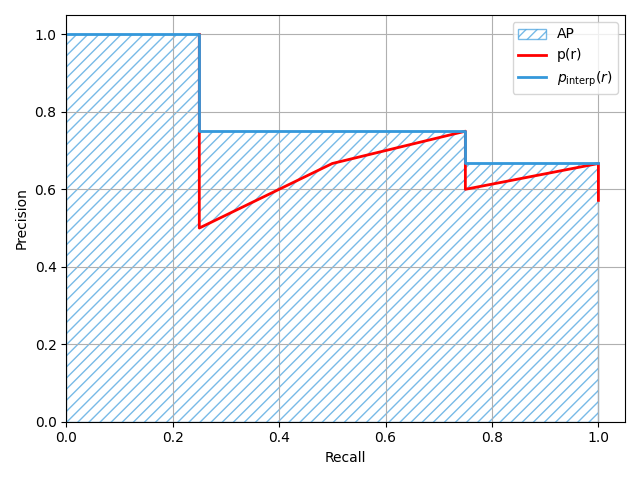}
    \end{subfigure}
    \caption{AP computation example. Left: ordered predictions. Right: associated Precision/Recall curve.}
    \label{fig:map-computation}
\end{figure}

A prediction is considered as true positive with respect to a bounding box annotation if the \gls{iou} between them is higher than a given threshold.
This way, one can compute the AP for a given class and a specific \gls{iou} threshold as follows:
\begin{itemize}
    \item The predictions are ordered by their confidence score.
    \item Precision $p(r_n)$ and recall $r_n$ are computed following this order, for each prediction.
    \item For $k$ predictions, the Average Precision (AP) is computed as the interpolation of the area under the precision/recall curve, considering $r_0=0$:
    \begin{equation}
        \mathrm{AP} = \sum_{n=0}^k (r_{n+1} - r_n) \cdot p_\mathrm{interp}(r_{n+1}),
    \end{equation}
    with 
    \begin{equation}
        p_\mathrm{interp}(r_{n+1}) = \max_{\tilde{r}>r_{n+1}} p(\tilde{r}).
    \end{equation}
\end{itemize}

Figure \ref{fig:map-computation} provides an example of AP computation for 7 predictions and 4 ground truth annotations. It leads to:
\begin{equation}
    \mathrm{AP} = \displaystyle 
    (\frac{1}{4}-0) \times 1 + 
    (\frac{2}{4} - \frac{1}{4}) \times \frac{3}{4} + 
    (\frac{3}{4} - \frac{2}{4}) \times \frac{3}{4} + 
    (\frac{4}{4} - \frac{3}{4}) \times \frac{4}{6}
    \approx 79 \%
\end{equation}

The \gls{map} can then be computed as the mean of the AP over a set of classes $\mset{C}$:
\begin{equation}
    \mathrm{mAP} = \frac{1}{\mathrm{card}(\mset{C})} \displaystyle \sum_{c \in \mset{C}} \mathrm{AP}_c.
\end{equation}
There are various ways of computing the \gls{map}. A common way is to weight the AP values by the number of items per class or by the number of pixels involved in each class. 
In addition, the \gls{map} is generally computed for several \gls{iou} thresholds so as to to take their average. Standard \gls{iou} thresholds are from 50\% to 95\% with a step of 5\%.

Both \gls{iou} and \gls{map} are debatable. More generally, designing a metric to evaluate a segmentation task is very complex in nature. Indeed, the annotation of the different elements to be segmented is subject to variations due to the subjectivity of the annotators. Thus, the main metrics to rely on are the ones that evaluate the target task \textit{i. e.} the recognition metrics.

\subsubsection{Recognition metrics}
To evaluate the accuracy of the text recognition, two main metrics are computed: the \gls{cer} and the \gls{wer}. Both metrics are based on the Levenshtein distance. It is defined between two sequences of symbols $\mseq{s}_\mathrm{A}$ and $\mseq{s}_\mathrm{B}$ as the minimal cost to transform $\mseq{s}_\mathrm{A}$ into $\mseq{s}_\mathrm{B}$ while only using the following elementary operations: insertion, deletion and substitution. Edit cost is set to one for each operation. The Levenshtein distance is computed through dynamic programming as follows:

\begin{equation}
\mathrm{lev}(\mseq{s}_\mathrm{A}, \mseq{s}_\mathrm{B})=
    \begin{cases}
        \max(|\mseq{s}_\mathrm{A}|, |\mseq{s}_\mathrm{B}|) & 
            \text{if } \min(|\mseq{s}_\mathrm{A}|, |\mseq{s}_\mathrm{B}|) = 0,\\
        \mathrm{lev}(\mseq{s}_{\mathrm{A}_{[1:]}}, \mseq{s}_{\mathrm{B}_{[1:]}}) &
            \text{if } \mseq{s}_{\mathrm{A}_{0}} = \mseq{s}_{\mathrm{B}_{0}}\\
         1 + \min \begin{cases}
           \mathrm{lev}(\mseq{s}_{\mathrm{A}_{[1:]}}, \mseq{s}_{\mathrm{B}})\\
           \mathrm{lev}(\mseq{s}_{\mathrm{A}}, \mseq{s}_{\mathrm{B}_{[1:]}})\\
           \mathrm{lev}(\mseq{s}_{\mathrm{A}_{[1:]}}, \mseq{s}_{\mathrm{B}_{[1:]}})\\
        \end{cases} & \text{otherwise}.
    \end{cases}
\end{equation}
In this equation, $\mseq{s}_{\mathrm{B}_{[1:]}}$ denotes the sequence $\mseq{s}_{\mathrm{B}}$ in which the first item is removed. $|\mseq{s}_\mathrm{A}|$ is the length (cardinality) of the sequence $\mseq{s}_\mathrm{A}$.

\acrshort{cer} is defined as the levenshtein distance between two strings: the ground truth text $\mseq{y}$ and the predicted text $\mseq{\hat{y}}$. They are processed as sequences of characters, normalized by the length of the ground truth. The global \acrshort{cer} for a set of $K$ strings is computed as follows:

\begin{equation}
    \mathrm{CER} = \frac{\displaystyle \sum_{i=1}^K \mathrm{lev}(\hat{\mseq{y}}_i, \mseq{y}_i)}{\displaystyle\sum_{i=1}^K{|\mseq{y}_i}|},
\end{equation}

\acrshort{wer} follows the exact same formula but strings are processed as sequences of words. If not stated otherwise, punctuation characters are considered as words, as in \cite{READ2016}.

\section{Conclusion}
In this chapter, we formalized the \gls{htr} task as an image-to-sequence problem. We provided fundamental background on deep learning approaches and presented the main components and techniques proposed in the literature to deal with common issues: training, generalization, convergence and vanishing and exploding gradients. We particularly focused on computer vision and sequence-to-sequence fields of research which are the two pillars of image-to-sequence problems. This way, we covered the main neuronal components and how to train them. We also presented the evaluation part, including datasets and metrics. 


At the beginning of this work, line-level approaches dominated the state of the art (this is still the mainly used approach currently). In fact, to our knowledge, the only works proposing to recognize multiple lines of text \textit{i. e.} whole paragraphs, in an end-to-end way, without any explicit line segmentation stage, were only two \cite{Bluche2016,Bluche2017b}. Our ultimate goal was, from the early beginning, to be able to recognize the text of whole documents, without any explicit segmentation step, in the hope of further improving the performance. To this end, we decided to go step by step, from single text line recognition to paragraph recognition, and then to the recognition of whole documents. In addition to propose end-to-end architectures for \gls{htr} at line, paragraph and document levels, the following constraints were also of primary consideration with the aim of designing modules as generic as possible:
\begin{itemize}
    \item The architecture must be able to deal with input images of variable sizes to preserve the height/width ratio so as to avoid deformations and information losses. In addition, it enables transfer learning between line-level architectures and paragraph-level or document-level architectures.
    \item The model must reach at least competitive results. To this aim, and to be comparable with state-of-the-art approaches, the model must be trained without using any pre-trained model on external data, and without being trained itself on external data. In addition, to really compare the performances of raw architectures, they must be evaluated without using an external language model.
    \item The number of trainable parameters, the training time as well as the prediction time must be reasonable, to respect hardware limitations.
    \item The model must be trained with few training data. This is a limitation due to the public datasets we used. In addition, since the annotation cost is essential, we aim at using as few annotations as possible.
\end{itemize}

Based on the image-to-sequence study we provided in this chapter, and on these objectives, we decided to design a generic \gls{fcn} encoder for the feature extraction part from the input image. Indeed, \gls{fcn} involves a reasonable number of parameters, operations are parallelizable, inputs can be of variable sizes, and it requires fewer labeled data compared to transformer-based encoder for instance, thanks to its translation equivariance and spatial dependencies properties. This way, an \gls{fcn} encoder could be used by any system dedicated to the \gls{htr} task, it is especially useful for transfer learning purposes. On the other hand, we focused on the attention mechanisms, in particular the transformer architecture, for the decoder part. As a matter of fact, the soft attention enables to preserve the whole encoded representation, without compression through decoding. In addition, it also enables to model dependencies across the whole signal and to focus on specific part of the input signal: this seemed to be very relevant when dealing with reading orders. However, as we have seen, designing and training deep and complex architectures raises several issues (overfitting, convergence) that we must handle.

We will now focus on the contributions of this thesis in details. The three following chapters are dedicated to line-level, paragraph-level and document-level recognition, respectively. For each one, we introduce the context and the challenges and we present the related works and the contributions.
\glsresetall
\chapter{Handwritten text line recognition}
\label{chap:line}

Historically, \gls{htr} was performed at character level by splitting the task into four main steps \cite{Arica2001}: pre-processing, segmentation, representation and \gls{ocr} \cite{Blumenstein2003,Rani2011,Choudhary2013,Choudhary2013-2}.
Many pre-processings were used to make the segmentation and the recognition easier. For example, binarization, noise reduction, skew correction or slant removal are common pre-processing that were used. 
Segmentation was performed to extract isolated images of characters. This character segmentation was mainly based on heuristics, such as inter-line and inter-word spacing, to first segment the document into lines, then words, and finally characters. 
The representation step consisted in detecting features \cite{DueTrier1996} through transformation or statistical approaches.
The character recognition part was mainly carried out by template matching approaches, statistical methods such as \gls{hmm} or clustering analysis, and more recently neural networks.

These systems have a poor generalization capacity due to the use of heuristics. They imply many handcrafted features, requiring a priori knowledge from experts. It is costly to generate such systems and they hardly can adapt to data variability, from one language to another for example. 
Nowadays, the majority of \acrshort{htr} approaches still lies on a prior segmentation step, at word or line level, implying the recognition of whole words or lines.

\section{Problem statement}
\begin{figure}[h]
    \centering
    \includegraphics[width=\linewidth]{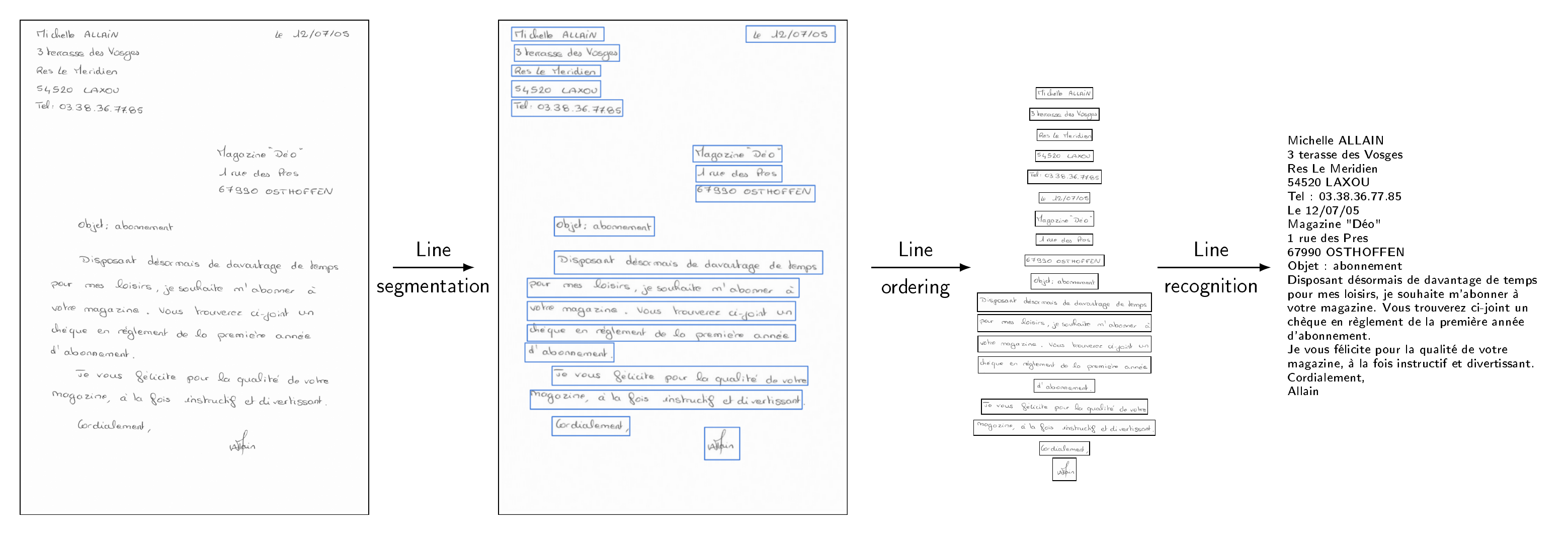}
    \caption{Three-step line-level approach for Handwritten Text Recognition.}
    \label{fig:two-step-overview}
\end{figure}

Recognizing the text of a whole document is a difficult task, it involves many challenges. Not only does the system need to identify where the text is located in the image but it also needs to recognize it. In addition, the different identified text regions must be ordered in order to preserve the meaning of the text. This process is represented in Figure \ref{fig:two-step-overview}.

The most common way to tackle this task is to perform these three sub-tasks as three independent steps.

\subsection*{The segmentation stage}
\label{section-segmentation-line}
The text line segmentation task consists in localizing all the text lines from a given input image of a document. It involves many challenges. One does not know in advance the number of text lines to be recognized in the document. On top of that, they can start and end wherever in the document. As shown in Figure \ref{fig:two-step-overview}, the size of the text lines can vary a lot. 

\begin{figure}[h]
    \centering
    \begin{subfigure}[b]{0.4\textwidth}
    \includegraphics[width=\textwidth]{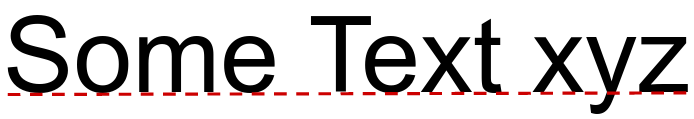}
    \caption{Baseline.}
    \end{subfigure}
    ~
    \begin{subfigure}[b]{0.4\textwidth}
    \includegraphics[width=\textwidth]{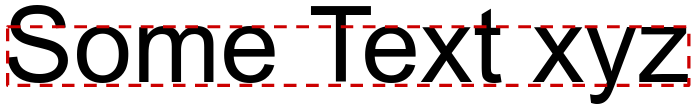}
    \caption{X-height.}
    \end{subfigure}   
    \par\medskip
    \begin{subfigure}[b]{0.4\textwidth}
    \includegraphics[width=\textwidth]{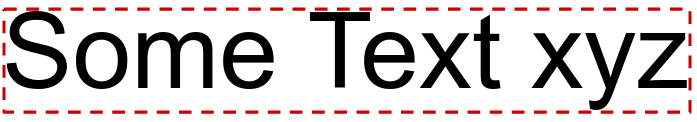}
    \caption{Bounding box.}
    \end{subfigure}
    ~
    \begin{subfigure}[b]{0.4\textwidth}
    \includegraphics[width=\textwidth]{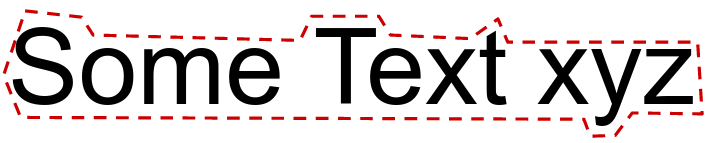}
    \caption{Polygon.}
    \end{subfigure}  
    \caption{Examples of line segmentation annotations.}
    \label{fig:seg_labels}
\end{figure}

Using an explicit segmentation stage also raises the question of the definition of a line. Baseline, X-height, bounding box and polygon are examples of target labels for segmentation that have been frequently used in the literature, all with their pros and cons \cite{Renton2018}. These different kinds of annotation are represented in Figure \ref{fig:seg_labels}. In addition, due to the visual nature of this task, segmentation labels are prone to the subjectivity of the annotators, leading to heterogeneity in the labeling process.

\begin{figure}[h]
    \centering
    \begin{subfigure}[c]{0.3\textwidth}
    \includegraphics[width=\textwidth]{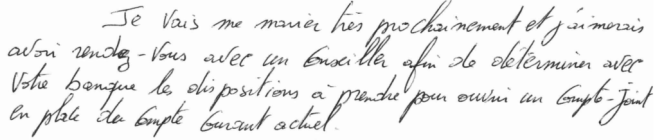}
    \caption{Overlapping text lines.}
    \end{subfigure}
    \begin{subfigure}[c]{0.3\textwidth}
    \includegraphics[width=\textwidth]{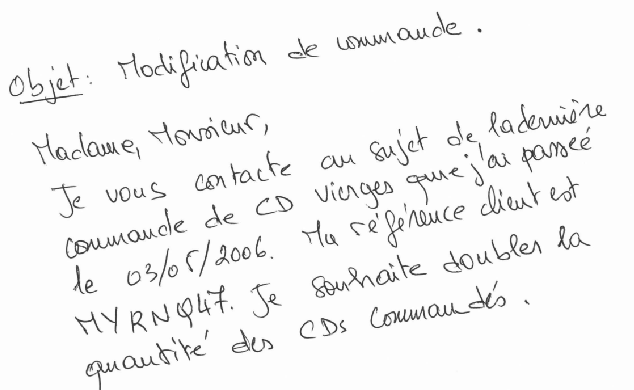}
    \caption{Slanted text lines.}
    \end{subfigure}    
    \begin{subfigure}[c]{0.35\textwidth}
    \includegraphics[width=\textwidth]{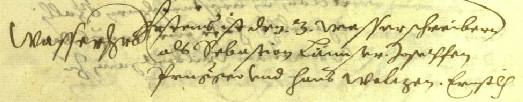}
    \caption{Blurred limit with annotation.}
    \end{subfigure}
    \caption{Line segmentation issues.}
    \label{fig:seg_issues}
\end{figure}

There are several cases in which the line segmentation can be difficult, as shown in Figure \ref{fig:seg_issues}. Ascenders and descenders can lead to overlapping text lines; it results in noise coming from strokes of adjacent text lines. Text lines can be more or less slanted, preventing the use of strictly horizontal bounding boxes for instance. The last example shows how it can be complicated to distinguish the end of a text line and the beginning of another one in the context of multiple columns of text.

\subsection*{The ordering stage}
\begin{figure}[h]
    \centering
    \begin{subfigure}[c]{0.4\textwidth}
    \includegraphics[width=\textwidth]{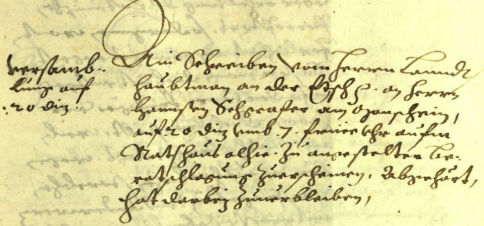}
    \caption{Expected reading order by column.}
    \end{subfigure}
    \begin{subfigure}[c]{0.4\textwidth}
    \includegraphics[width=\textwidth]{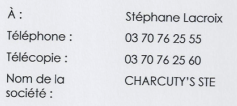}
    \caption{Expected reading order by row.}
    \end{subfigure}    
    \caption{The line reading ordering depends on the global analysis of the document.}
    \label{fig:order_issues}
\end{figure}

The ordering step can be formalized as follows: the input is a set of text lines, whatever the line definition chosen, and the output is an ordered sequence of text lines. One can note two main challenges for this task. First, there is not a single possible reading order for a given document. Indeed, for documents with a complex layout, such as tables or drawings, there are multiple ways of reading the textual content, while following a humanly logical order. The second challenge is about the need for the global context. Figure \ref{fig:order_issues} illustrates this issue. For both images, there is only one correct reading order to preserve the global meaning of these documents. They both follow a similar double-column layout. However, the document on the left must be read column by column and the document on the right must be read row by row. Using only the segmented text line coordinates, it seems impossible to choose between those two orders. For such complex cases, it is required to have a global understanding of the document layout to understand the notion of margin or fields of forms. The textual content itself can also be helpful to choose how to order the text lines.

\subsection*{The recognition stage}
The recognition step consists in recognizing the sequence of characters from a text line image. 
The main difficulty of this recognition stage is about the writing style variability: each person has his own way of writing with specific character shapes, slant, alignment and character spacing. The background and the quality of the input images can also dramatically affect the performance of the model. This difficulty is intrinsic to the task, it is not related to the technique used: it is the same for humans.

In addition, the recognition of text lines involves two main technical issues:
\begin{itemize}
    \item A one-dimensional sequence of characters is expected as output whereas the input is a two-dimensional image.
    \item One does not know in advance the number of characters to be recognized, and it cannot be deduced from the width of the image neither.
\end{itemize}

Although handwritten text line recognition is \textit{a priori} limited to visual character recognition, a \gls{lm} is often associated, as post-processing, to improve the performance. A language model is a probability distribution over sequences of characters or words which represents a given language. The idea is to detect and correct sequences of characters or words that have low probability \textit{i. e.} which are statistically incorrect. It is generally handled by a statistical approach \cite{Rosenfeld2000,Swaileh2017}. It consists in N-gram character or word language models which estimate probabilities over sequences of N successive characters or words. In this thesis, we focus on raw architecture performances, we do not study the use of such external language model.

\section{Related works}
We now present the related works for each of these three sub-tasks.

\subsection{Text line segmentation}
Early text line segmentation techniques \cite{Sulem2007} can be classified into three categories as suggested in \cite{Papavassiliou2011}. First, the projection-based methods, which consider the boundaries between lines as valleys of vertical projection profile \cite{Weliwitage2005}. Second, the grouping methods; it consists in grouping rows of connected components according to heuristic rules \cite{Feldbach2001}. The last category is the smearing methods which use blurring filters combined with binarization or active contours for example \cite{Swaileh2015}.

Nowadays, it is generally handled by deep neural networks. One can note two main approaches: pixel-level classification and object detection.

\begin{figure}[h!]
    \centering
    \includegraphics[width=0.9\textwidth]{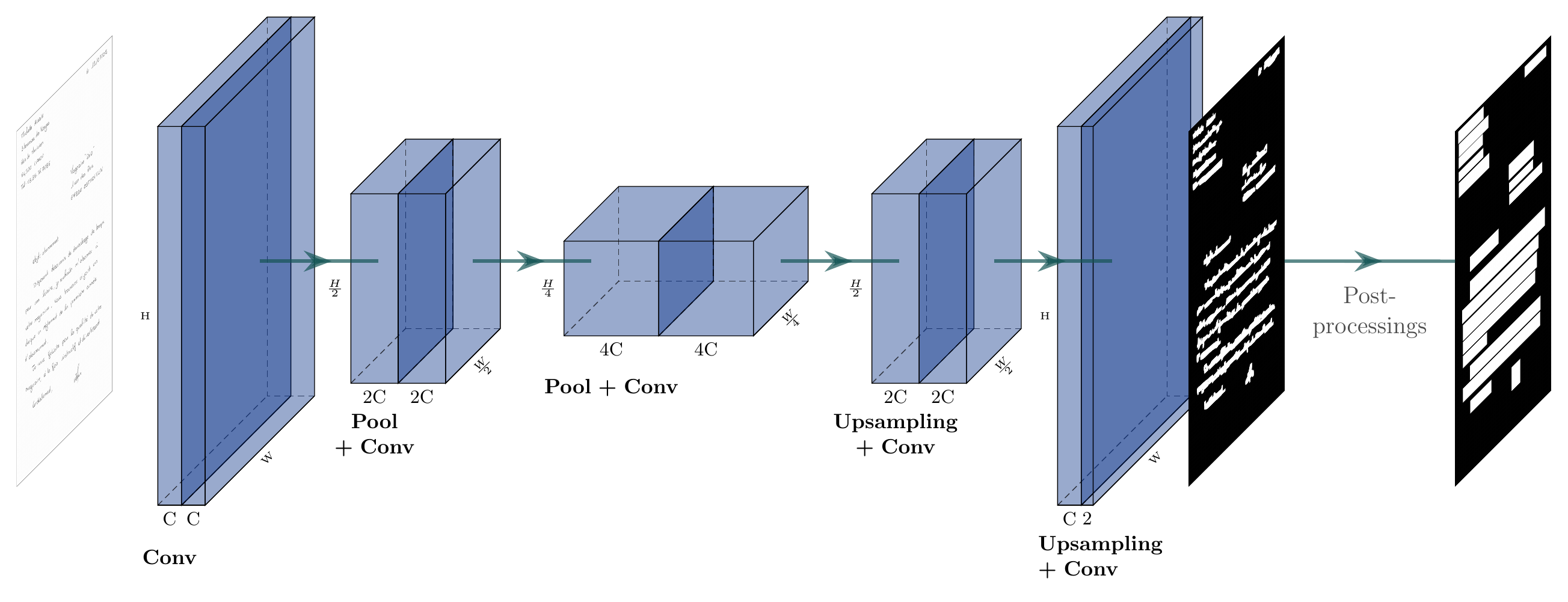}
    \caption{Line-level segmentation by pixel-level classification. An FCN is used to classify each pixel of the input image as text or background. Then, the text line bounding boxes are obtained through some rule-based post-processings.}
    \label{fig:line-segmentation-pixel}
\end{figure}

Pixel-level classification, in the context of text line segmentation, consists in classifying each pixel of the input image among two classes, namely background and text.  It is generally handled by a \gls{fcn} as in \cite{Renton2018, DHSegment, ARUNet, Boillet2020, Boillet2022}, as shown in Figure \ref{fig:line-segmentation-pixel}. 

Since the prediction is carried out at pixel level, the prediction may present some scattered anomalies, even if the overall prediction seems correct. Some post-processings can be used to deal with this issue, through the use of heuristics based on neighboring predictions. The adjacent pixels labeled as text are grouped into blocks of pixels: each block corresponds to a text line. However, this is where the major drawback of this approach lies. Indeed, all the text line predictions are predicted in the same 2D representation. In the case of closely spaced text lines, the pixel-level predictions of successive text lines could touch and merge: the segmentation fails in such cases.

\begin{figure}[h!]
    \centering
    \includegraphics[width=0.9\textwidth]{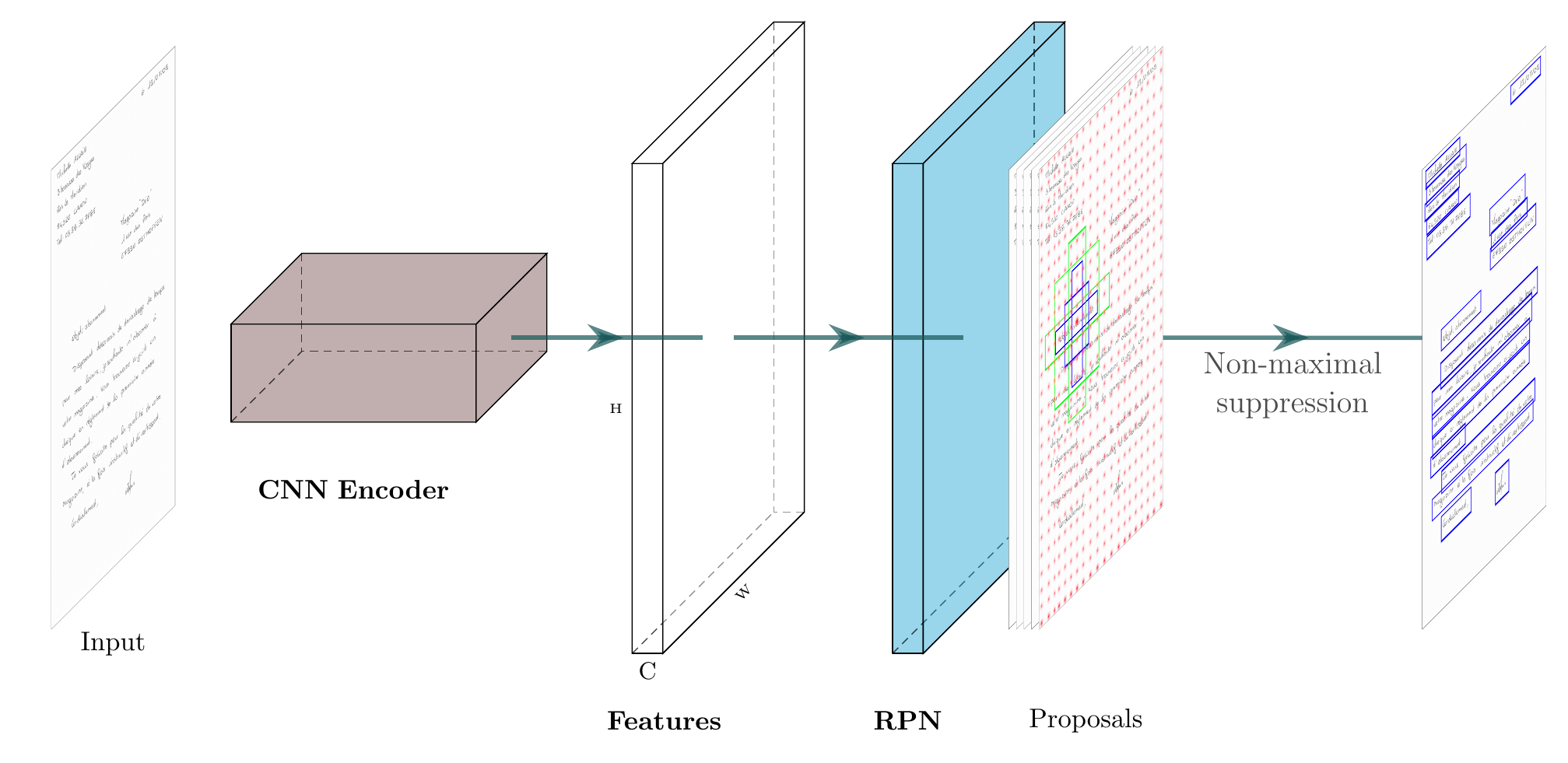}
    \caption{Line-level segmentation with an object-detection approach. A CNN encoder is used to extract 2D features from the input image. The RPN provides a set of proposals (green and blue rectangles) for each anchor (red dots). The best proposals are selected through the non-maximal suppression algorithm.}
    \label{fig:line-segmentation-rpn}
\end{figure}

The object-detection approach, through the prediction of bounding box coordinates, solve this issue. This approach is depicted in Figure \ref{fig:line-segmentation-rpn}. The models proposed in \cite{Carbonell2019,Carbonell2020,Chung2020} follow this approach at word level. A \acrshort{cnn} is used to extract features from the input image. It is followed by a \gls{rpn}. The idea is to assign an anchor point to certain coordinates in the image, following a sliding window approach, to cover the whole image. It leads to a grid of anchor points. These anchor points correspond to the center of potential regions containing a word. For each anchor, rectangles of varied shapes and sizes are predefined, leading to a high number of potential bounding boxes. For each one, offsets are predicted to give more flexibility in the prediction of the bounding box coordinates. This region proposal network is  combined with a non-maximal suppression process in order to filter these proposals. It is an iterative process in which proposals with highest confidence scores are kept and overlapping one are discarded. The remaining proposals are the final predicted word bounding boxes. 

In \cite{Carbonell2019,Carbonell2020}, the authors propose such a model following a multi-task end-to-end architecture. The segmentation part is jointly trained with the word recognition part. In \cite{Chung2020}, the word bounding boxes are merged into line bounding boxes using some rules based on heuristics: the recognition is carried out at line level.
In \cite{Moysset2016}, the authors proposed to separately detect bottom-left and top-right corners of text line bounding boxes, using an additional matching step.

The idea of region proposal is also used in \cite{Moysset2017,Wigington2018,Wigington2019}. However, these approaches do not attempt to predict text line bounding boxes but only the coordinates of the start position of the text lines, as well as their height. In \cite{Moysset2017}, a \acrshort{cnn}+\acrshort{mdlstm} is used to predict these start-of-line references. The line width is considered as the image width. A specific end-of-line token is added for the recognition part to handle multi-column texts.
In \cite{Wigington2018,Wigington2019}, a \acrshort{cnn} is used as start-of-line predictor. Then, a recurrent process predicts the next position based on the current one until the end of the line, generating a normalized line. The idea is to deal with curved lines. The approach proposed in \cite{Wigington2019} is similar to the one of \cite{Wigington2018}, but it can handle transcriptions without line breaks.

\subsection{Text line ordering}
Very few works proposed whole line-level \gls{htr} pipeline: text line segmentation and text line recognition are mostly processed independently. This way, the text line ordering aspect has not been much studied or detailed in the literature. It seems that the majority of the works focused on a rule-based ordering approach. The reading order is computed based on the coordinates of the text regions and some heuristics. Generally, for Latin languages, the reading order is fixed from top to bottom and from left to right. It is sufficient for datasets in which the layout is regular. It becomes tedious for heterogeneous documents. Very few works have been working on learning-based approaches for text line ordering: \cite{Quiros2020,Quiros2022}. In these works, the authors considered the problem as a binary relation ordering issue between each pair of text lines.

\subsection{Text line recognition}
\label{section:related-work-htr-line}
Regarding handwritten text line recognition, it was first solved using handcrafted features and \gls{hmm} \cite{Ploetz2009,Yacoubi1999}. However, those models lacked discriminative power and were limited when dealing with long term dependencies in sequences. For a long time, hybrid systems combining \acrshort{hmm} with neural networks were proposed, leading to better results over standard \acrshort{hmm}, thanks to the discriminative power of neural networks: \acrshort{hmm}+\acrshort{mlp} \cite{Bengio1995,Gilloux1995,Salicetti1996,Knerr1998}, \acrshort{hmm}+\acrshort{cnn} \cite{Bluche2013} or \acrshort{hmm}+\acrshort{rnn} \cite{Senior1995,Frinken2009}. Currently, the state of the art is reached with deep neural networks, without the use of HMM anymore.

The recognition of whole words or lines by deep neural networks was mainly made possible by the \gls{ctc} to handle the alignment issue between the one-dimensional predicted sequence and the one-dimensional ground truth sequence, which do not have the same length (issue previously handled by HMM). However, it remains a main challenge: the input is a two-dimensional image whereas the expected output is a one-dimensional sequence of characters. This way, \gls{htr} architectures are usually made up of three components: an encoder, a transformation mechanism and a decoder. The encoder aims at extracting features from the input image. The transformation mechanism is used to go from a two-dimensional feature space to a one-dimensional feature space. The decoder enables to associate to each feature frame a probability for each character of a given alphabet, as well as a probability for the \gls{ctc} null symbol (blank).

\begin{figure}[h!]
    \centering
    \includegraphics[width=\textwidth]{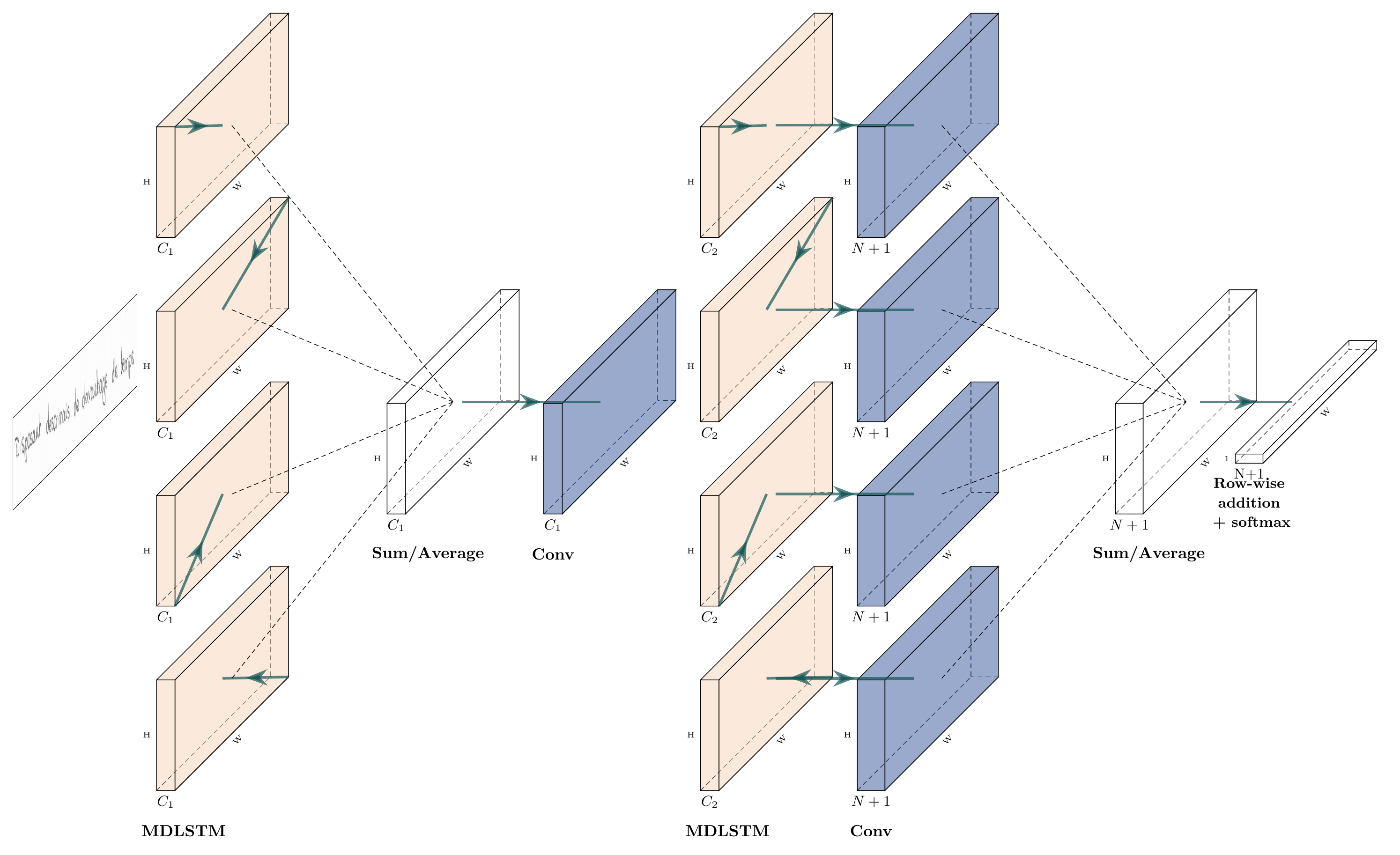}
    \caption{CNN+MDLSTM architecture for text line recognition. The model aggregates context from 4 directions, before collapsing the vertical axis through row addition, and computing the probabilities of the characters, for an alphabet of size $N$.}
    \label{fig:line-htr-cnn-mdlstm}
\end{figure}

In 2008, A. Graves \textit{et al.} proposed to use a \gls{mdlstm} as encoder for text line recognition \cite{Graves2008} so as to model the dependencies of the whole input image. The transformation mechanism consists in summing all rows of the features maps element-wise, to collapse the vertical dimension, leading to a one-dimensional sequence. The decoder is only a softmax activation. With the emergence of \gls{cnn} and their efficiency for computer vision tasks, the authors of \cite{Pham2014} proposed to alternate between convolutional layers and \gls{mdlstm} layers (\acrshort{cnn}+\acrshort{mdlstm}). The same idea is studied in \cite{Voigtlaender2016}. An example of \acrshort{cnn}+\acrshort{mdlstm} architecture is depicted in Figure \ref{fig:line-htr-cnn-mdlstm}.

One should note that recurrent layers present a main drawback: the computations are sequential, leading to longer training and prediction times. On the other hand, the computations of the convolutional layers are parallelizable and the algorithms have been highly optimized to this end. In addition, for the specific case of \gls{lstm} cells, it can lead to an important number of trainable parameters. 

\begin{figure}[h!]
    \centering
    \includegraphics[width=\textwidth]{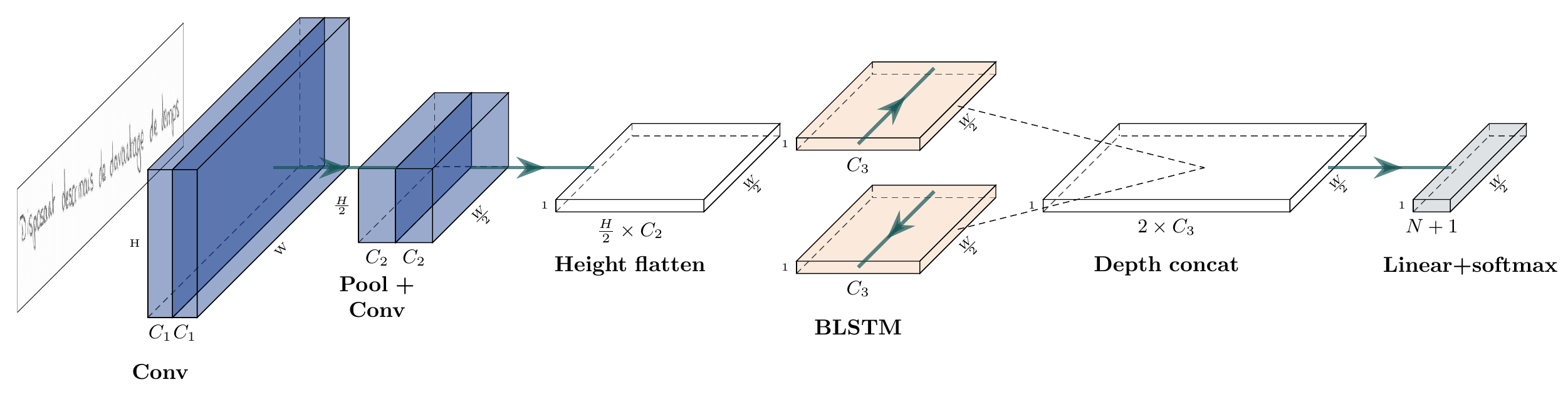}
    \caption{CNN+BLSTM architecture for text line recognition. The model extracts 2D features, before flattening the representation over the vertical axis. It then aggregates context from left and right. $C_1$, $C_2$ and $C_3$ correspond to arbitrary numbers of feature maps, and $N$ is the size of the alphabet.}
    \label{fig:line-htr-cnn-blstm}
\end{figure}

It leads to a trend toward the reduction of the use of recurrent layers. In 2017, the authors of \cite{Wigington2017,Puigcerver2017} proposed to use a \acrshort{cnn} as encoder and a \acrshort{blstm} as decoder (\acrshort{cnn}+\acrshort{blstm}), as shown in Figure \ref{fig:line-htr-cnn-blstm}. In \cite{Puigcerver2017}, the transformation mechanism is a flatten operation over the vertical dimension. The authors of \cite{Moysset2019} argued that \gls{mdlstm} would provide better results for complex handwriting compared to \acrshort{cnn}+\acrshort{blstm} while the performance would be equivalent for simpler ones.

\begin{figure}[h!]
    \centering
    \includegraphics[width=\textwidth]{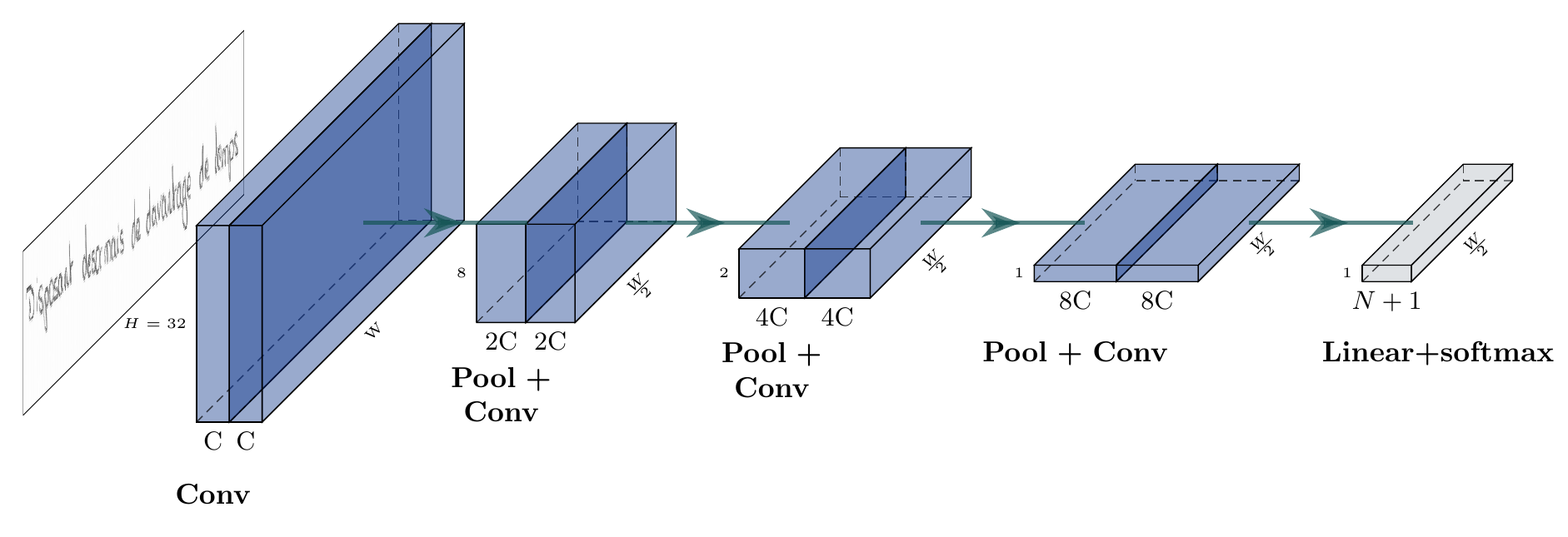}
    \caption{CNN architecture for text line recognition. The height of the input image is fixed to 32px. It is gradually decreased through pooling until it reaches one, while the number of feature maps is gradually increased.}
    \label{fig:line-htr-cnn}
\end{figure}

Based on the successful application of gates in \cite{Dauphin2017} for language modeling and in \cite{Gehring2017} for \gls{nmt} in 2017, the authors of \cite{Bluche2017a} and \cite{Ingle2019} proposed \gls{gcrl}: \acrshort{cnn}+\acrshort{blstm} architectures with gates in the convolutional part. The idea is to counteract the decrease in the number of \acrshort{lstm} layers, which integrate gates to select information along the time axis, by integrating this information flow control over the depth axis.
Following this idea, we proposed a \gls{gcnn} in 2019, removing every recurrent layers. The same year, \cite{Ptucha2019} achieved state-of-the-art results on isolated words on the IAM dataset, using a \gls{cnn} without any gate component. Figure \ref{fig:line-htr-cnn} shows an example of \gls{cnn} architecture for line-level \gls{htr}.

The authors of \cite{Yousef_line} proposed a \gls{gfcn} \textit{i. e.} a \acrshort{gcnn} without any dense layer, reaching state-of-the-art results on many datasets, including IAM at line level. Their architecture consists in a large stack of convolutional layers, leading to 26 million of trainable parameters.

All the works we have just mentioned use the \gls{ctc} loss to train their models. However, based on \cite{Bahdanau} and the emergence of attention models, \cite{Sueiras2018} and \cite{Michael2019} successfully applied attention-based \acrshort{cnn}+\acrshort{blstm} to the task of \gls{htr} at line level. The models follow a recurrent process, generating the output character by character. They use specific <start-of-line> and <end-of-line> tokens to initialize and stop the recurrent process, respectively. The main drawback of this approach lies in the sequentiality of the process, inherent to the recurrence, which leads to prediction time dependent on the sequence length. The authors of \cite{Poulos2021} also proposed an attention-based approach. However, while in \cite{Sueiras2018} and \cite{Michael2019} the features are flattened before the attention mechanism, in  \cite{Poulos2021}, the features are still in two dimension and the attention weights are computed over the horizontal axis only: the same weight is applied for each vertical position of a same column.

Inspired by these previous works, we proposed a \gls{gfcn}, in 2020, to take advantage of the parallelization of the computations, with the aim of having fast training and prediction times, with few trainable parameters. As a matter of fact, despite the use of convolutional components only in \cite{Yousef_line}, the high number of parameters reduces the advantage of using them by increasing the computation time. We now present the \gls{gfcn} model we proposed. 

\section{Gated Fully Convolutional Network for Handwritten Text Line Recognition}

We proposed a \acrlong{gfcn} for the task of \gls{htr} at line level in \cite{Coquenet2020}. Trained with the \gls{ctc} loss, the model reached competitive results on the RIMES and IAM datasets at line level.

In this section, we introduce the architecture of the proposed \gls{gfcn} and we provide an experimental study to evaluate the performance of this model in the context of text line recognition. We provide all source code and pre-trained model weights at \url{https://github.com/FactoDeepLearning/LinePytorchOCR}.

\subsection{Architecture}

The aim of the proposed network is to combine the following expected properties: a large receptive field, to efficiently model the dependencies in the input space, a small number of parameters, to generalize well and for computational efficiency, and the only use of convolutional components for computation parallelization. These properties are obtained by a deep \gls{fcn} architecture enhanced by a strong gating mechanism and by the use of \gls{dsc}.

The proposed architecture is a \gls{gfcn} inspired from our preliminary work on \gls{gcnn} applied to handwritten text recognition \cite{Coquenet2019}. In other words, it is only made up of convolutional components: convolutional and pooling layers, enabling it to deal with input images of various sizes. No recurrent nor fully connected layers are used.

We opted for a \gls{gfcn} because convolutional layers are light components: operations can be parallelized on \acrshort{gpu} and they do not require a lot of parameters. Therefore, we can stack many of these layers without major impact on memory usage or training and prediction times. The proposed model contains 22 convolutional layers, enabling to reach a receptive field of size (v=196, h=240) where v and h stand for the vertical and horizontal dimensions, respectively. We mostly use \gls{dsc} to reduce the number of parameters. As defined in \cite{DSC} (Section \ref{section-convolution}), \gls{dsc} consists in performing a depthwise spatial convolution followed by a pointwise convolution. The advantage is that these operations need less trainable parameters than standard convolutions, while providing comparable results. However, we keep standard convolutional layers at the beginning and at the end of our model because these layers are more crucial to extract features and predict probabilities. Through our experiments \gls{dsc} turned out to be less efficient when introduced on these layers.

We used another component which is the Max Pooling. It enables to increase the size of the receptive fields while reducing the tensors size and thus the memory usage, preserving the most relevant information. We used it until reducing the vertical dimension to unity while only dividing by four the horizontal dimension so as to keep enough frames to align the horizontal representation with the ground truth using the \acrshort{ctc} loss.

\begin{figure}[htbp!]
    \centering
    \begin{subfigure}[b]{\linewidth}
    \centering
        \includegraphics[width=\linewidth]{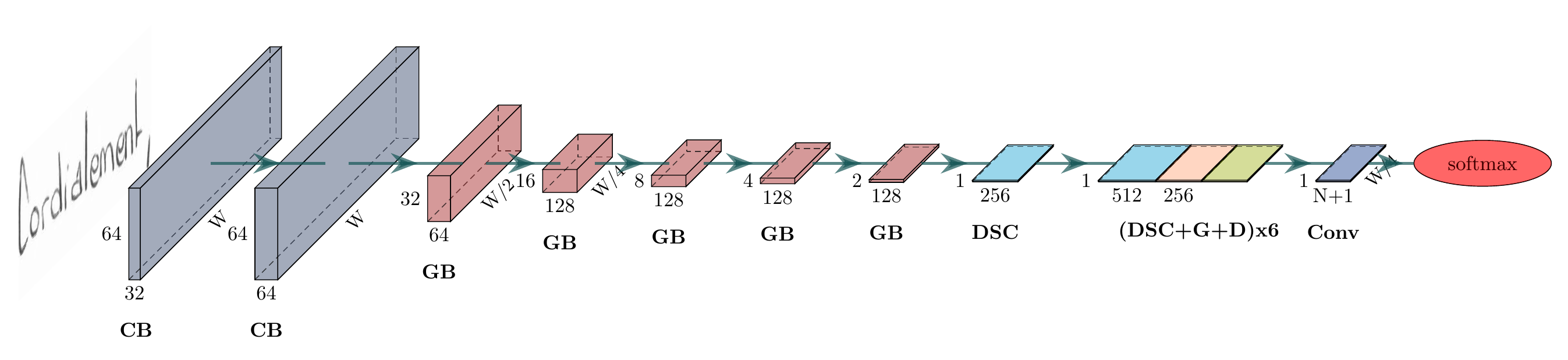}
        \caption{Overview of the GFCN.}
        \textit{CB: ConvBlock, GB: GateBlock, G: Gate, D: Dropout.}
        \label{fig:gfcn-archi_global}
    \end{subfigure}
    
    \par\bigskip
    \par\bigskip
    
    \begin{subfigure}{\linewidth}
        \centering
        \includegraphics[width=0.6\linewidth]{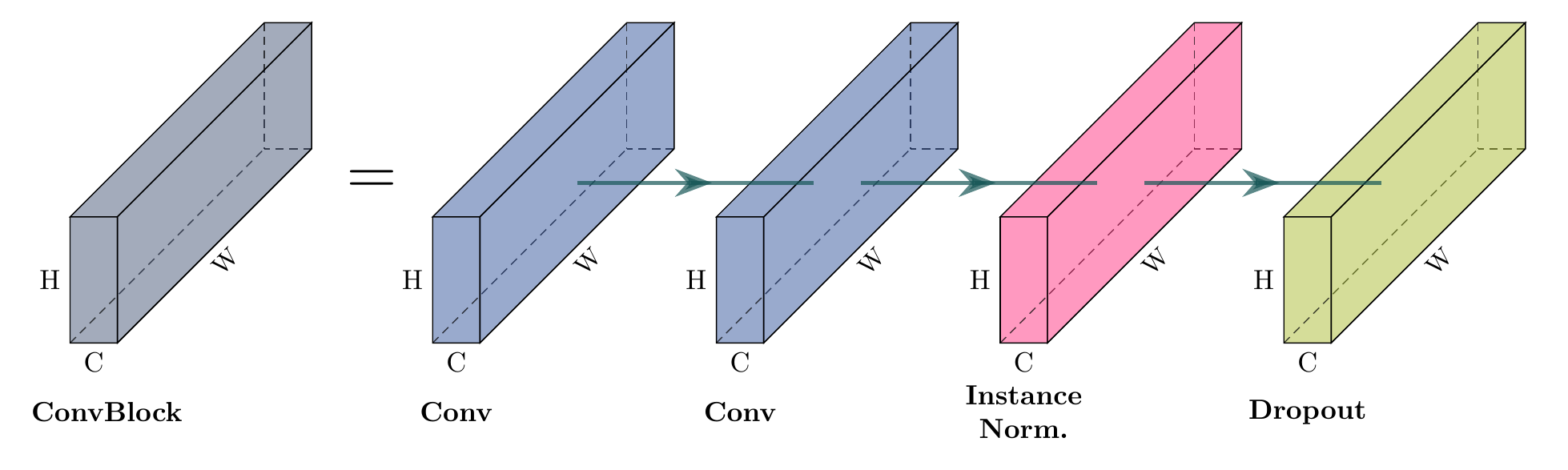}
        \caption{ConvBlock (CB).}
        \label{fig:gfcn-archi_convblock}
    \end{subfigure}
    
    \par\bigskip
    \par\bigskip
    
    \begin{subfigure}{\linewidth}
        \centering
        \includegraphics[width=0.75\linewidth]{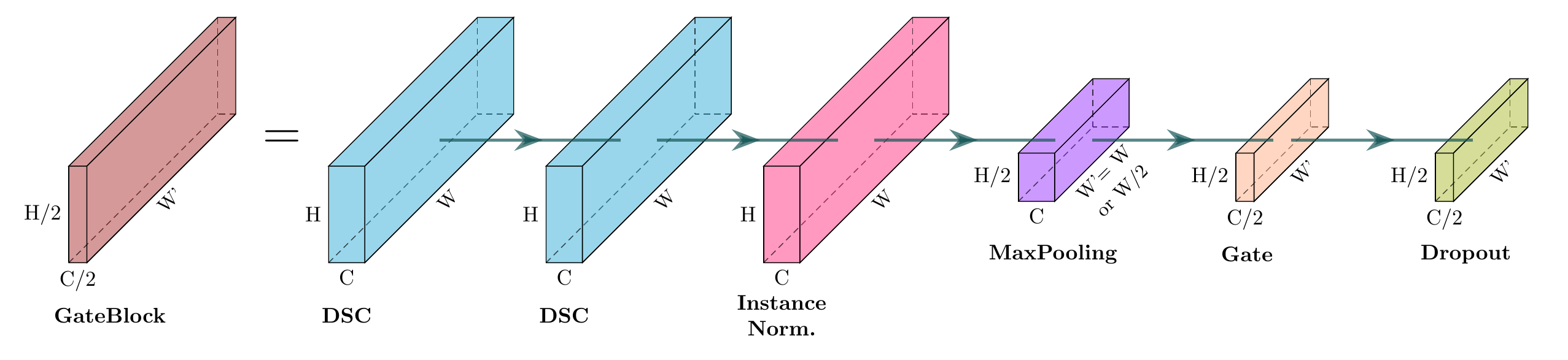}
        \caption{GateBlock (GB).}
        \label{fig:gfcn-archi_gateblock}
    \end{subfigure}
    
    \par\bigskip
    \par\bigskip
    
    \begin{subfigure}{\linewidth}
        \centering
        \includegraphics[width=0.7\linewidth]{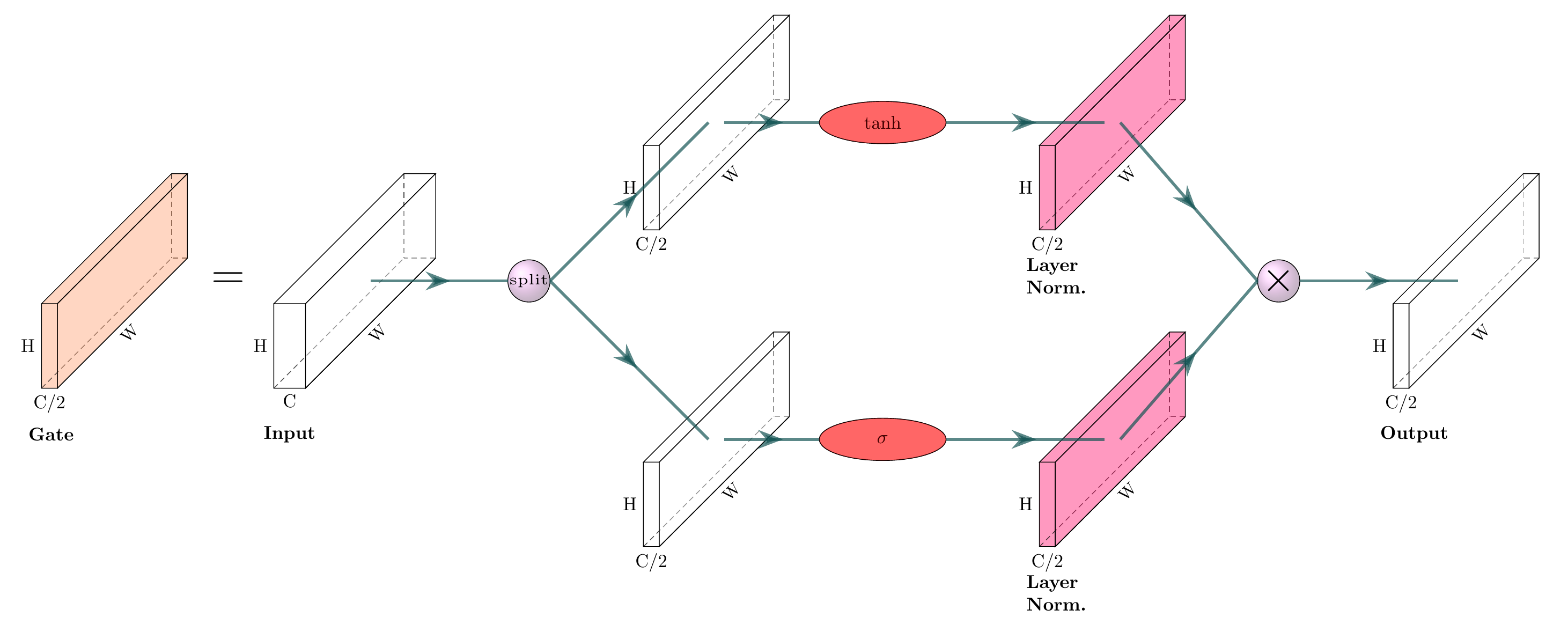}
        \caption{Gating mechanism (G). White elements are just representations of the tensors at a given step (no specific operation is performed).}
        \label{fig:gfcn-archi_gate}
    \end{subfigure}
    
    \caption{Proposed GFCN architecture. The model is made up of some ConvBlocks and GateBlocks, and relies on a gating mechanism.}

    \label{fig:gfcn-archi}
\end{figure}

The full model is depicted on Figure \ref{fig:gfcn-archi}. In order to provide a better view of the proposed  model, we have gathered some layers into GateBlocks and ConvBlocks. The \gls{gfcn} takes as input images of 64 pixels in height with variable width. The model starts with 2 ConvBlocks of respectively 32 and 64 filters. It preserves the original image size to extract low level details. It is followed by 5 GateBlocks, which progressively decrease the height and the width. A \acrshort{dsc} with $2 \times 1$ kernels is then used to collapse the vertical dimension, leading to a one-dimensional sequence of feature frames. A succession of 6 \acrshort{dsc}+Gate+Dropout is applied with $1 \times 8$ kernels  to enlarge the receptive field along the horizontal axis. A last convolution with $1 \times 1$ kernels, combined with a softmax activation, enables to provide the probabilities of each character ($N$ being the charset size + 1 for the \acrshort{ctc} blank) for each output frame. The model is trained in an end-to-end fashion with the \acrshort{ctc} loss.

\subsubsection*{ConvBlocks}
A ConvBlock consists in 2 standard convolutions with $C$ filters, followed by instance normalization and dropout with a probability of $0.4$. Each convolution has the following parameters: $3 \times 3$ kernel, $1 \times 1$ stride, $1 \times 1$ padding and ReLU activation. A visualization of a ConvBlock is provided on Figure \ref{fig:gfcn-archi_convblock}.

\subsubsection*{GateBlocks}
A GateBlock is defined as a succession of two \acrshort{dsc} (C filters, $3 \times 3$ kernel, $1 \times 1$ stride,  $1 \times 1$ padding, each followed by a ReLU activation) followed by instance normalization, Max Pooling, Gate and dropout layers. For the two first GateBlocks, Max Pooling is carried out with a kernel size of $2 \times 2$,  the three following blocks have a kernel size of $2 \times 1$. It means that the height is divided by 32 and the width by only 4. Dropout is applied with a probability of $0.4$. Figure \ref{fig:gfcn-archi_gateblock} shows a GateBlock in a more visual way.

\subsubsection*{The gating mechanism}
The model integrates the concept of Gate which enables to select the relevant features throughout the layers. The gating mechanism is inspired from \cite{Yousef_line} and is defined as follows: the input tensor is split over the channel axis into two sub-tensors of same dimensions. A $\tanh$ activation is applied to one sub-tensor and a sigmoid to the other. Both are then layer normalized separately before being multiplied element-wise. Thanks to the sigmoid function, the gate can be seen as a selection operator which is used over the layers. A visualization of this mechanism is given on Figure \ref{fig:gfcn-archi_gate}.

\subsubsection*{Regularization techniques}
Regularization techniques are used to improve the training stability and to reduce overfitting. In this way, the architecture starts with a gaussian noise layer. Instance normalization is added after the convolutional layers in the ConvBlocks and the GateBlocks, and dropout is also applied multiple times. We choose to use the instance normalization instead of batch normalization which is widely used in the literature because it is independent of the mini-batch size and it showed great results in other tasks such as stylized image generation \cite{InstanceNorm}. As a matter of fact, since we used small mini-batches, the batch normalization is not that stable in our case. This choice is analyzed in Section \ref{section:gfcn-norm-choice}.

\subsection{Experimental study}

The experimental environment is as follows:
\begin{itemize}
    \item Datasets: to evaluate the proposed model, we used two public datasets at line level: RIMES 2011 (line-2) and IAM, as defined in Section \ref{section:dataset}.
    \item Pre-processings: images are resized to obtain a height of 64 pixels conserving their initial width. Mean and variance are computed over the training data set to normalize the grey-level pixel intensity.
    \item Training details: the experiments are carried out with Pytorch, using a single \acrshort{gpu} V100 (16 Go). We used the \gls{ctc} loss to train the model.
We used Adam optimizer with an initial learning rate of $10^{-4}$.
Mini-batches of size 2 are used during training and evaluation. Trainings are stopped if no improvement is observed during 50 consecutive epochs.
We do not use any data augmentation technique in the experiments. We do not use any external data nor external \gls{lm}. 
    \item Metrics: we used \gls{cer} and \gls{wer} as metrics to evaluate the performance of the proposed approach, as detailed in Section \ref{section:metrics}. For the \acrshort{wer}, punctuation marks are not counted as words: they are considered as belonging to the previous word.
\end{itemize}

We evaluate the proposed \gls{gfcn} on both RIMES and IAM datasets and compare it to the state-of-the-art approaches. Then, we discuss the normalization technique used and the impact of the receptive field through experiments.

\subsubsection*{Comparison with state-of-the art handwritten text line recognition systems}

The following comparisons are made with approaches under similar
conditions, \textit{i. e.} at line level, without the use of external data, data augmentation strategy, lexicon constraints nor external \gls{lm}.

Table \ref{table:gfcn-rimes} shows the state-of-the-art results on the RIMES datasets. As we can see, the proposed model reaches better results than some recurrent architectures even if it is not at the leading position. The proposed \gls{gfcn} reaches a \acrshort{cer} of 4.35\% on the test set, which is not far from the Puigcerver \acrshort{cnn} + \acrshort{lstm} model that reaches the best \acrshort{cer} with 3,3\%. One can notice that, any of the architectures (\acrshort{cnn}+\acrshort{blstm}, \acrshort{cnn}+\acrshort{mdlstm} and \gls{gfcn}) reach similar \acrshort{cer}.

\begin{table}[!ht]    
\caption{Comparison of the GFCN results with the state of the art on the RIMES dataset without LM, lexicon, nor data augmentation.}
    \centering
    \resizebox{0.75\linewidth}{!}{
    \begin{tabular}{ c c c c c}
    \hline
    \multirow{2}{*}{Architecture} & \gls{cer} (\%) & \gls{wer} (\%) & \gls{cer} (\%) & \gls{wer} (\%)\\ 
    & validation & validation & test & test\\
    \hline
    \hline
    \acrshort{cnn}+\acrshort{mdlstm} \cite{Moysset2019} & 3.32 & 13.24 & 4.94 & 16.03\\
    \acrshort{cnn}+\acrshort{mdlstm}-X2 \cite{Moysset2019}& 3.14 & 12.48 & 4.80 & 16.42 \\
    \acrshort{cnn}+\acrshort{blstm} \cite{Moysset2019} & \textbf{2.9} & 11.68 & 4.39 & 14.05\\
    \acrshort{cnn}+\acrshort{blstm} \cite{Puigcerver2017} & 3.0 & & \textbf{3.3}\\
    Ours (\acrshort{gfcn}) \cite{Coquenet2020} & 3.82 & 15.60 & 4.35  & 18.01\\
    \hline
    \end{tabular}
    }

    \label{table:gfcn-rimes}
\end{table}
\vspace{-0.3cm}

The results obtained on the IAM dataset are presented in Table \ref{table:gfcn-iam}.

The proposed model reaches 7.99\% of \gls{cer} which is very close to the best recurrent model that achieves 7.73\%.
\begin{table}[!ht]
    \caption{Comparison of the GFCN results with the state of the art on the IAM dataset without LM, lexicon nor data augmentation.}
    \centering
    \resizebox{0.75\linewidth}{!}{
    \begin{tabular}{c c c c c}
    \hline
    \multirow{2}{*}{Architecture} & \gls{cer} (\%) & \gls{wer} (\%) & \gls{cer} (\%) & \gls{wer} (\%)\\ 
    & validation & validation & test & test \\
    \hline
    \hline
    \acrshort{cnn}+\acrshort{mdlstm} \cite{Moysset2019} & 5.41 & 20.15 & 8.88 & 29.15 \\
    \acrshort{cnn}+\acrshort{mdlstm}-X2 \cite{Moysset2019}& 5.40 & 20.40 & 8.86 & 29.31 \\
    \acrshort{cnn}+\acrshort{blstm} \cite{Puigcerver2017} & 5.1 & & 8.2 & \\
    \acrshort{cnn}+\acrshort{blstm} \cite{Moysset2019} & \textbf{4.62} & 17.31 & \textbf{7.73} & 25.22 \\
    Ours (\acrshort{gfcn}) \cite{Coquenet2020} & 5.23 & 21.12 & 7.99 & 28.61 \\
    \hline
    \end{tabular}
    }

    \label{table:gfcn-iam}
\end{table}

\subsubsection*{Comparison of the models}

We now compare the previously mentioned models by considering additional features, and not only the performance (\gls{cer}): we also consider the training time, the prediction time and the number of parameters at stake. We have reproduced the best model seen previously to get those results, for the others (\acrshort{cnn}+\acrshort{mdlstm} and \acrshort{cnn}+\acrshort{mdlstm}-X2) we only give the number of parameters. This experiment was conducted on the IAM dataset with images resized to 128 px height and preserving the original width. We added one GateBlock (and thus one Max Pooling layer) to our model in order to be compatible with images of that height. The results are presented in Table \ref{table:gfcn-comp}. Training time corresponds to the time spent to train over a full epoch (train set) and prediction time is the mean time to predict a sample of the IAM test set. Both are computed with a mini-batch size of 2 on a \acrshort{gpu} V100 32Go.

\begin{table}[!ht]
    \caption{Comparison of the GFCN with the state-of-the-art architectures for the IAM dataset, for an input image height of 128px, preserving the original width.}
    \centering
    \resizebox{0.75\linewidth}{!}{
    \begin{tabular}{c c c c}
    \hline
    \multirow{2}{*}{Architecture} & Training time & Prediction time & \multirow{2}{*}{Parameters}\\ 
    & (min/epoch) & (ms/sample)& \\
    \hline
    \hline
    \acrshort{cnn}+\acrshort{mdlstm} \cite{Moysset2019} & & & 0.8 M\\
    \acrshort{cnn}+\acrshort{mdlstm}-X2 \cite{Moysset2019} & & & 3.3 M\\
    CNN + \acrshort{cnn}+\acrshort{blstm} \cite{Puigcerver2017, Moysset2019} & 11.25 & 57 & 9.6 M\\
    Ours (\acrshort{gfcn}) \cite{Coquenet2020} & 13.75 & 74 & 1.4 M \\
    \hline
    \end{tabular}
    }
    \label{table:gfcn-comp}
\end{table}

As we can see among the best models, ours is the one with the lowest number of parameters. Training time and prediction time of the \acrshort{cnn}+\acrshort{blstm} \cite{Puigcerver2017, Moysset2019} have the same order of magnitude than that of the proposed \gls{gfcn}.
This can be explained by the high number of normalization layers used in our model and its depth that counterbalance with the sequential computations of the \gls{lstm} layers of the other models. 
It has to be noted that \acrshort{cnn}+\acrshort{mdlstm} and \acrshort{cnn}+\acrshort{mdlstm}-X2 architectures performed well too and they do not imply so many parameters. That is especially true for the \acrshort{cnn}+\acrshort{mdlstm} which uses the smallest number of parameters of all of the reported models. 

\subsubsection*{Comparison of the normalization layers}
\label{section:gfcn-norm-choice}
In this experiment, we are looking for the best normalization technique for the proposed model. In this respect, we trained the model on the RIMES dataset with different normalization layers in place of the instance normalization layers shown in Figure \ref{fig:gfcn-archi}. We tested the 4 most common normalization techniques namely batch normalization \cite{BatchNorm}, layer normalization  \cite{LayerNorm}, instance normalization  \cite{InstanceNorm} and group normalization  \cite{GroupNorm} for a group of size 32. Results are shown in Table \ref{table:gfcn-norm}. Reported \gls{cer} are the best \gls{cer} on the validation set over the first 50, 100, 150 or 200 epochs. The training time for one epoch is also given since it can be an important criterion. 

\begin{table}[!ht]
    \centering
    \caption{Comparison of the different kinds of normalization for the proposed GFCN with the RIMES dataset. CER is computed on the valid set.}
    \resizebox{0.75\linewidth}{!}{
    \begin{tabular}{c c c c c c}
    \hline
    \multirow{2}{*}{Normalization} & \gls{cer} (\%) & \gls{cer} (\%) & \gls{cer} (\%) & \gls{cer} (\%) & Time\\ 
    & 50 epochs & 100 epochs & 150 epochs & 200 epochs & (/epoch)\\
    \hline
    \hline
    Instance & 6.87 & \textbf{5.03} & \textbf{4.47} & \textbf{4.28} & \textbf{8.5 min} \\
    Layer & \textbf{6.75} & 5.04 & \textbf{4.47} & \textbf{4.28} & 15 min\\
    Group (32) & 7.10 & 5.30 & 4.86 & 4.32 & 8.75 min \\
    Batch & 9.6 & 5.7 & 5.4 & 4.8 & \textbf{8.5 min}\\
    \hline
    \end{tabular}
    }
    \label{table:gfcn-norm}
\end{table}

As one can see, from 100 epochs, the best \acrshort{cer} is obtained with instance normalization. Layer normalization is acting similarly to instance normalization but it implies longer training time which makes it less relevant.
Group normalization performs well too with tiny differences. 
Batch normalization is the one with the worst \acrshort{cer}. This can be explained by the small mini-batches used that lead to a slightly less stable batch normalization. As a matter of fact, it is the only normalization that is mini-batch size dependent. The others use the samples in an independent way to compute means and variances. 
Given these results we can conclude that, apart from batch normalization, any of these normalization techniques can be considered. Batch normalization  should be preferred with larger mini-batches only.
Instance normalization turned out to be an option to be considered when dealing with images and small mini-batches; that's why we kept this one.

\subsubsection*{Impact of the receptive field}
This last experiment aims at determining the impact of the receptive field in a \gls{gfcn} architecture. In this respect, we vary the number of (\gls{dsc} + Gate + Dropout) used in our model which is fixed to 6 in our baseline. We made this number vary from 1 to 6; for each one we kept the best \gls{cer} on the valid set over the first 100 and 200 epochs and we report the respective number of parameters and receptive field.  The results are presented in Table \ref{table:gfcn-rf}. 

\begin{table}[!ht]    
\caption{Impact of the receptive field on the GFCN results, on the IAM dataset. CER is computed over the valid set.}
    \centering
    \resizebox{0.8\linewidth}{!}{
    \begin{tabular}{c c c c c c}
    \hline
    Number of ending Gates & \gls{cer} (\%) & \gls{cer} (\%)  & \multirow{2}{*}{Parameters} & Receptive Field\\ 
    (\gls{dsc}+G+D) & 100 epochs & 200 epochs & & (h, w)\\
    \hline
    \hline
    6 (baseline) & 6.82 & \textbf{5.80} & 1,375,792 & (196, 240)\\
    5 & \textbf{6.69} & 5.97 & 1,241,904 & (196, 212)\\
    4 & 8.14 & 7.48 & 1,108,016 & (196, 184)\\
    3 & 6.93 & 6.23 & 974,128 & (196, 156)\\
    2 & 7.35 & 6.63 & 840,240 & (196, 128)\\
    1 & 8.30 & 7.83 & 706,352 & (196, 100)\\
    \hline
    \end{tabular}
    }
    \label{table:gfcn-rf}
\end{table}

One can notice that only the horizontal axis is affected by this number since the \gls{dsc} has a kernel $1 \times 8$. Moreover, we double the horizontal receptive field from 1 to 6 (\gls{dsc}+G+D) which switch from 100 to 240. In the same way, the number of parameters is almost doubled (from 0.7 to 1.4 million).

Considering that characters have an average width of 31 pixels on the IAM test set, the network can roughly model dependencies through 3 successive characters for prediction when using only one (\gls{dsc}+G+D). For the baseline, the receptive field corresponds to 8 characters. We may assume that this very large receptive field enables to compensate the memory capability of \gls{lstm}.

We can easily see a tendency whereby the \gls{cer} is improved when increasing the number of (\gls{dsc}+G+D) and thus the receptive field. We may assume that the high \gls{cer} of example 4 is due to a bad initialization that would be corrected with more epochs. It would be interesting to cross validate these results by repeating this experiment multiple times to consolidate this assumption. Here, doubling the receptive field this way enables to reduce the \gls{cer} by 2 points at 200 epochs going from 7.83\% down to 5.80\%.

\subsection{Discussion}

We have presented one of the first \acrlong{gfcn}s applied to handwriting line recognition. The proposed model reaches competitive results on the RIMES and IAM datasets, compared to the best recurrent models in the same conditions. This demonstrates that recurrence-free models should be considered for the task of text recognition. The use of \gls{dsc} enables to have very few trainable parameters at stake, and the selective mechanism provided by the gates, combined with the regularization techniques enabled to get a stable deep neural network. One can notice that the vertical receptive field is larger than the height of the text line images (196 compared to 64). The idea behind this \gls{gfcn} network is to have a generic encoder that could then be used to process whole paragraph or page images, as we will see in the next chapters.

However, although we only used convolutional components, which enables highly parallelizable computations, the high stacking of them, combined with the normalization layers, leads to higher prediction time when compared with the standard \acrshort{cnn}+\acrshort{blstm} architecture. 

With some hindsight, the effectiveness of the gates has yet to be proven. Indeed, we propose a new \gls{fcn} architecture in the next chapter for which the use of this gating mechanism was not beneficial anymore. It has to be noted that this latter architecture was trained with data augmentation, contrary to the \gls{gfcn}. In addition, the \gls{fcn} relies on residual connections, which may compensate for the contribution of the gates. For this reason, we do not use any gating mechanism in the next architectures.

\section{Conclusion}

Since this contribution, one can notice a recent trend towards the use of attention-based architectures for \gls{htr} at line level, and more specifically based on transformers \cite{Vaswani2017}. In \cite{Wick2021} and \cite{Kang2020}, the authors proposed some architectures made up of a \gls{cnn} + transformer for encoder and transformer for decoder. They are trained with the cross entropy loss using specific <start-of-line> and <end-of-line> tokens. The authors of \cite{Kang2020} showed that the addition of a \gls{lm} is useless for their model \textit{i. e.} the transformer architecture enabled to model dependencies between the predicted characters. In \cite{Wick2021}, the authors proposed a bidirectional approach for the transformer architecture, combined with a voting algorithm, to reduce the error rate. The authors of \cite{Diaz2022} conducted an extensive study to compare several architectures for text line recognition, including \gls{cnn}+transformer encoder and \gls{gcrl}, both with \gls{ctc} decoding or transformer decoding. They found out that, in their configuration, \textit{i. e.} when using an external dataset and an external \gls{lm}, the combination of \gls{cnn}+transformer encoder trained with the \gls{ctc} loss is better.

The proposed models are more and more efficient. In \cite{Diaz2022}, their best model reached 2.75\% of \gls{cer} on IAM and 1.99\% of \gls{cer} on RIMES at line level, when combined with external data and language models. This is extremely encouraging even if there is still room for improvement. However, one should keep in mind that all these results are given for images of isolated text lines which have been manually annotated or at least manually verified. In real world, handwritten documents are nearly never limited to a single text line: the metrics are biased and does not meet the main use case. 

Although in the majority in the literature, the three-step approach (segmentation then ordering then recognize) has three main drawbacks:
\begin{itemize}
    \item The errors accumulate \textit{i. e.} if a text region is poorly segmented, it will inevitably lead to recognition errors. And even with a perfect segmentation stage, errors can occur during the recognition step. This way, metrics should take into account the prior segmentation and ordering steps which themselves may include some errors.
    \item It requires additional annotations. Since the aim is to recognize the textual content of a document, the minimum requirement is the transcription annotation. Reading order can be deduced from this transcription. However, the segmentation stage requires its own additional annotations, which are difficult to define (see Section \ref{section-segmentation-line}) and costly to produce.
    \item Each stage of the three-step approach being independent, they cannot benefit each other to improve their results. Yet, it is well known that recognition and segmentation should be performed altogether so as to explore and evaluate all the hypotheses as a whole \cite{Sayre1973}. Obviously, the same goes for reading order, since recognition could help a lot to search for the best hypothesis.
\end{itemize}

One way to alleviate these issues is to recognize whole handwritten paragraphs, thus reducing the number of segmentation entities per document required to train the system. We study these paragraph-level approaches in the next chapter.

\glsresetall
\chapter{Handwritten paragraph recognition}
\label{chap:paragraph}
Historically, the methods used for the \gls{htr} task have gradually improved, reducing the need for the segmentation step. We have moved from character-level segmentation to word-level and then line-level segmentation, leading to better and better results. The intuitive next step is the paragraph-level segmentation coupled with an end-to-end paragraph recognition approach. 

\section{Problem statement}
\begin{figure}[ht]
    \centering
    \includegraphics[width=\textwidth]{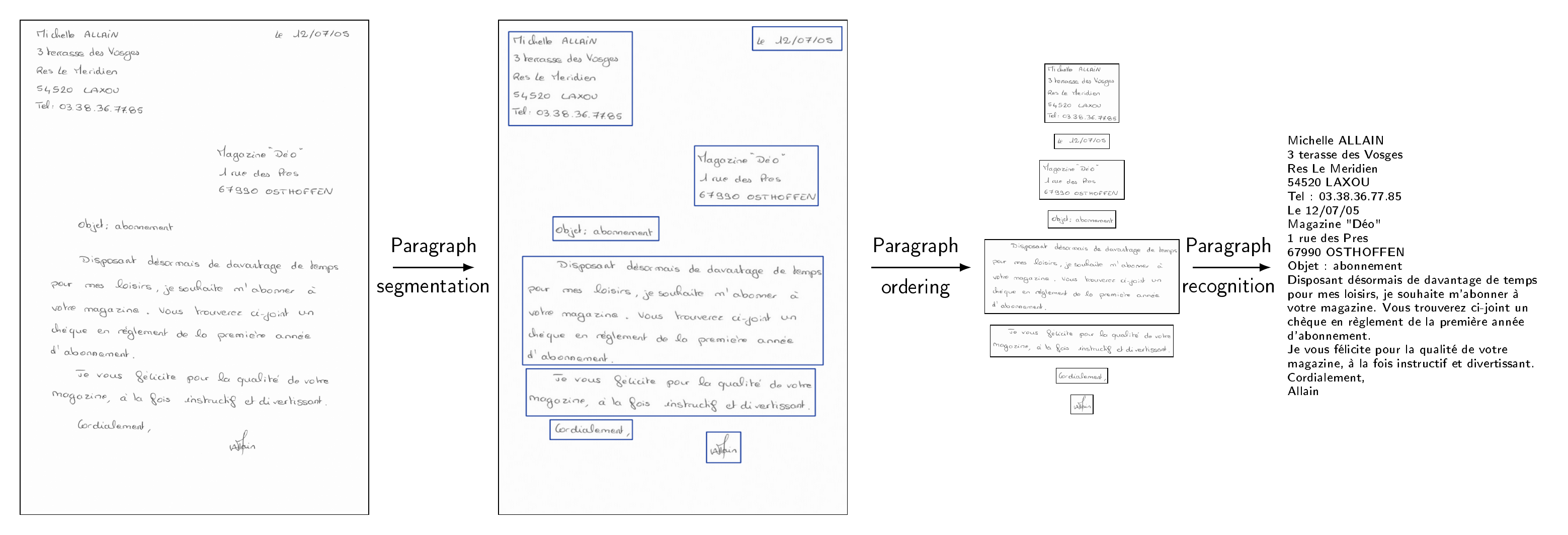}
    \caption{Paragraph-level approach for Handwritten Text Recognition.}
    \label{fig:htr_pg}
\end{figure}

First of all, let's define what we mean by paragraph. In this thesis, paragraph refers to a subpart of the input document image made up of a set of text lines organized as a single column of text. 
Paragraph-level recognition follows the same steps as line-level text recognition: segmentation, ordering and recognition, as depicted in Figure \ref{fig:htr_pg}. It means that paragraph-level recognition keeps the drawback of the errors which accumulate between these various stages. However, the main advantage of processing the input document at paragraph level is the reduced need for segmentation annotations. For the example presented in Figure \ref{fig:htr_pg}, the number of bounding boxes for the segmentation stage decreased from 19 for the line-level approach down to 8 for the paragraph-level approach. In addition, the model can benefit from a larger context and one can expect to get improved predictions. 

However, when switching from line images to paragraph images, one faces new challenges: 
\begin{itemize}
    \item The number of lines varies from one paragraph to another, knowing that each line contains an unknown and variable number of characters.
    \item The layout diversity can be important due to multiple factors such as interline spacing, horizontal alignment or slant.
    \item The recognition of whole paragraphs adds a new level of complexity regarding the reading order. While line-level recognition approaches only rely on a horizontal reading order for Latin languages, paragraph-level recognition approaches also rely on a vertical reading order: from one line to the next. 
\end{itemize}

\section{Related works}
In the literature, only very few works have been devoted to paragraph-level recognition, and most studies have concentrated on line-level recognition, as detailed in Chapter \ref{chap:line}. In addition, to our knowledge, there is no work proposing the full three-step pipeline: the literature only focus on the segmentation stage or on the recognition stage, separately.

\subsection{Paragraph segmentation}
\label{section:related-work-dla}
The paragraph segmentation stage has not been studied as single goal. It is mostly part of a more generic task, named \gls{dla}.
\gls{dla} aims at identifying and categorizing the regions of interest in a document image. 
In \cite{Barlas2014}, the authors proposed a method based on connected components to recognize 8 types of objects on heterogeneous documents: text, photographic image, hand drawn line area, graph, table, edge line, separator and material damage. The authors of \cite{Quiros2018} proposed a 2-stage method based on an artificial neural network for the semantic segmentation task, applied on historical handwritten documents. It detects multiple zone types such as page numbers, marginal notes and main paragraphs. 

Nowadays, as for line-level segmentation, \gls{dla} is mainly handled through pixel-by-pixel classification using \glspl{fcn} (\cite{Soullard2020,DHSegment,Yang2017}). It consists in an elegant and relatively light end-to-end model that does not require to re-scale the input images. The authors of \cite{Yang2017} applied \gls{dla} on printed textual documents: contemporary magazines and academics papers. They trained their model to recognize multiple classes namely figures, tables, section headings, captions, lists and paragraphs. The model presented in \cite{DHSegment} aims at detecting different items from historical documents: text regions, decorations, comments and background. In \cite{Soullard2020}, the model is applied to historical newspapers. It recognizes many textual elements such as titles, text blocks and advertisements, as well as images.

\subsection{Paragraph recognition}
\label{section:related-work-htr-paragraph}
End-to-end approaches for paragraph recognition can be classified into two categories: the one-shot approaches, which predict the whole sequence of characters in a single step, and the attention-based approaches which are based on a recurrent process.

\subsubsection{One-shot approaches}

As for many line-level approaches, one-shot paragraph-level approaches use the CTC loss and decoding process to handle the length variability between the predictions and the ground truths. However, contrary to isolated text lines, paragraphs can contain multiple lines of text and one cannot collapse the vertical dimension columns by columns because it would lead to only one predicted character per column of pixels.

\begin{figure}[h!]
    \centering
    \includegraphics[width=0.45\textwidth]{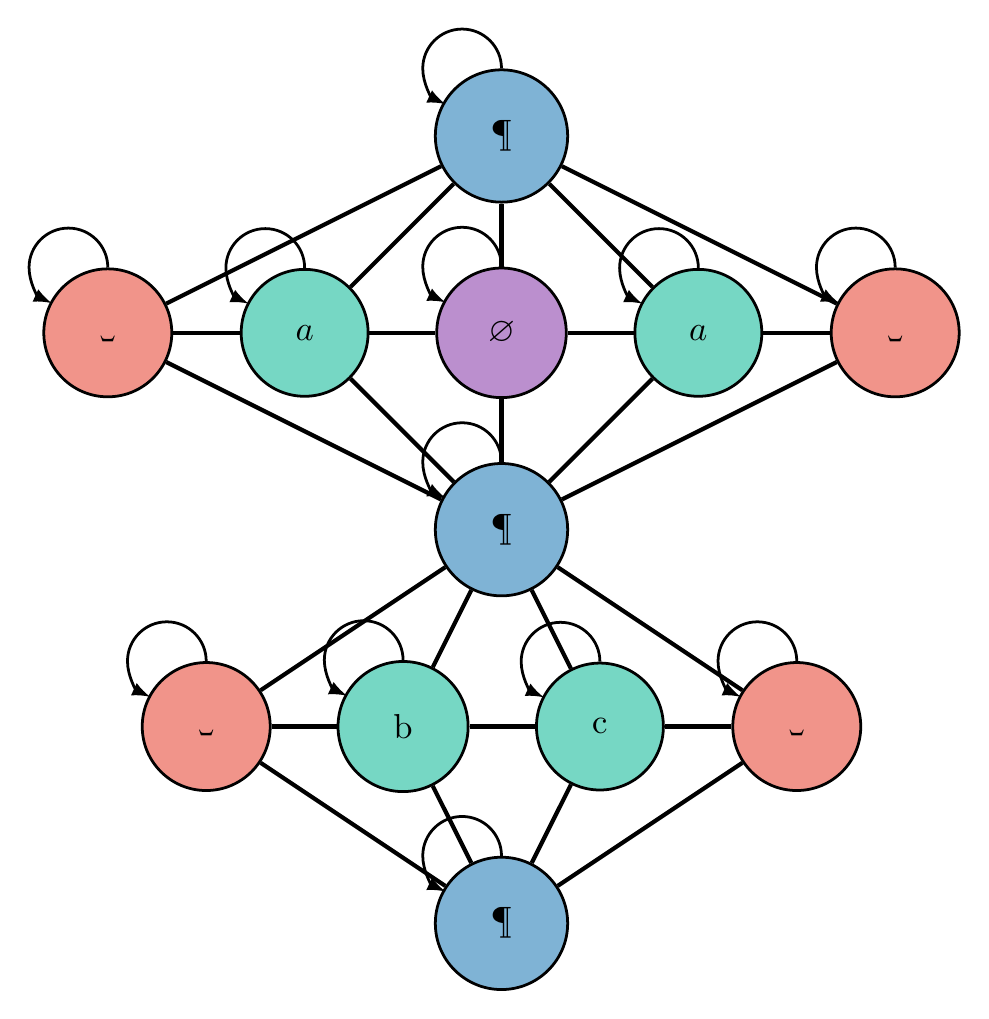}
    \hfill
    \includegraphics[width=0.45\textwidth]{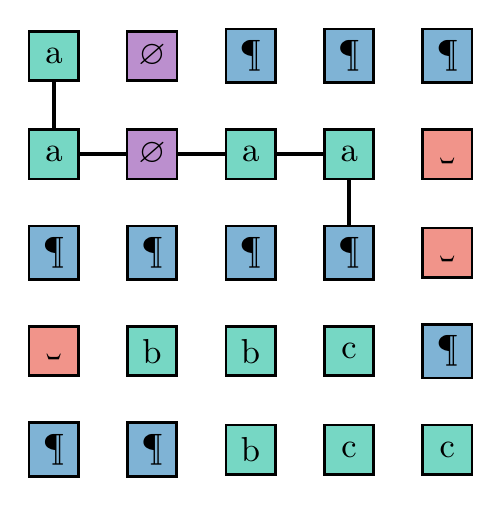}
    \caption{Multi-Dimensional Connectionist Classification. Left: ground truth representation for the text "aa\textbackslash nbc". Right: prediction example.}
    \label{fig:mdcc-crf}
\end{figure}

The authors of \cite{Schall2018} proposed a two-dimensional version of the CTC to tackle this problem: the Multi-Dimensional Connectionist Classification (MDCC). Using a Conditional Random Field (CRF), ground truth transcription sequences are converted into a two-dimensional model (a 2D CRF) able to represent multiple lines, as shown in Figure \ref{fig:mdcc-crf}. A line separator label (¶) is introduced in addition to the CTC null symbol ($\ctcblank{}$). A single white space label (\textvisiblespace) is added at the beginning and at the end of each text line. The CRF graph enables to jump from one line to the following one, whatever the position in the current line. As for the CTC, repeating labels only account for one. An \acrshort{mdlstm}-based network is used to generate probabilities in two dimensions, preserving the spatial nature of the input image. The decoding process is carried out line by line. An example of prediction and correct path (solid line) for the first text line is depicted in Figure \ref{fig:mdcc-crf}.

\begin{figure}[h!]
    \centering
    \includegraphics[width=\textwidth]{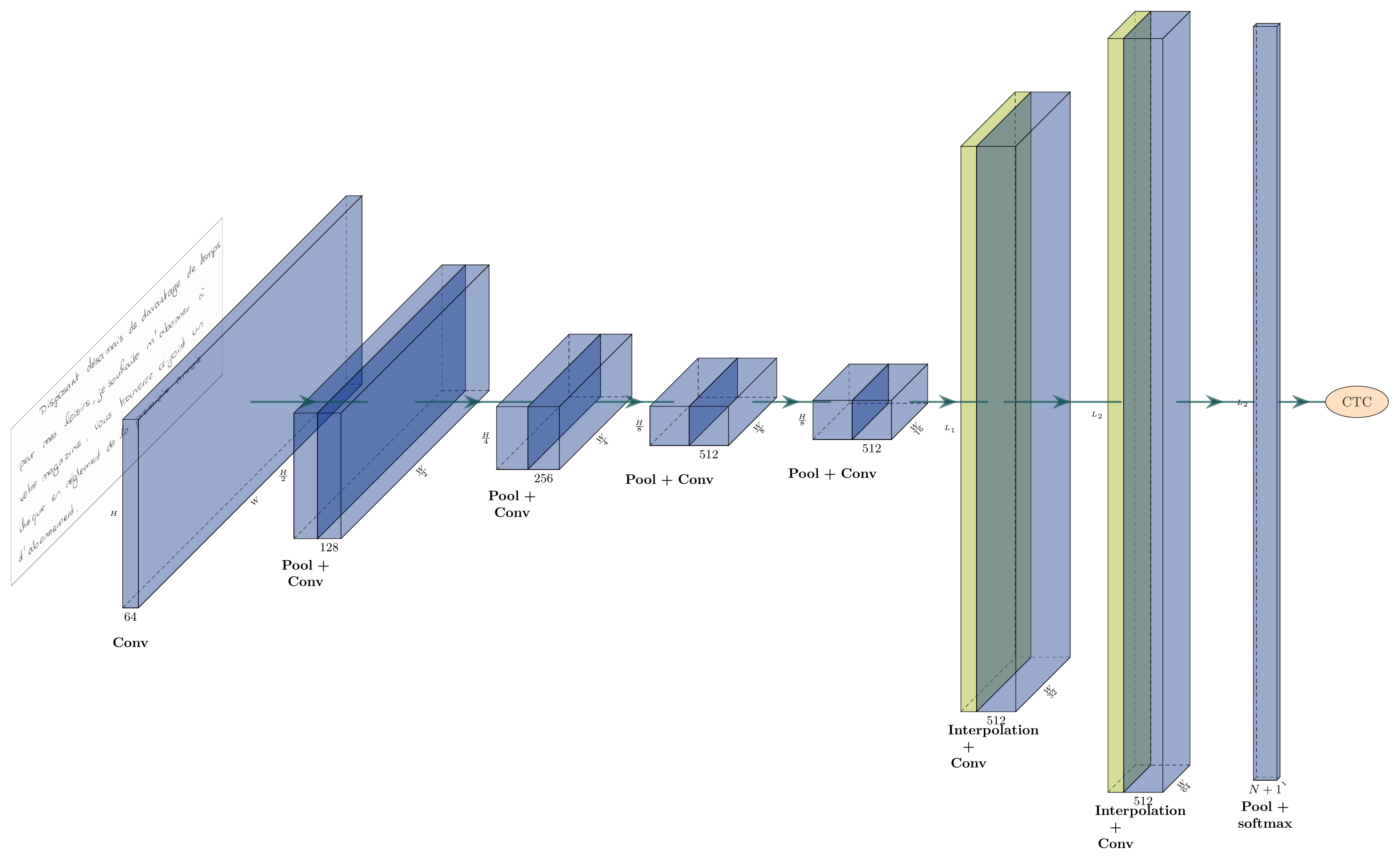}
    \caption{OrigamiNet overview. This FCN integrates two bi-linear interpolation layers in order to concatenate the representation of the different text lines into a single large line.}
    \label{fig:origami}
\end{figure}

Contrary to this approach, which aims at dealing with 2D predictions by proposing a new loss, in \cite{Yousef2020}, the aim is to reformulate the two-dimensional problem as a one-dimensional problem in order to use the standard CTC loss.
Indeed, the authors of \cite{Yousef2020} focused on learning a representation transformation to unfold the input paragraph image into a single text line. The system is trained to concatenate text line representations to obtain a single large text line, before character recognition takes place. This is mainly carried out with bi-linear interpolation layers combined with an FCN encoder, as shown in Figure \ref{fig:origami}. This transformation network enables to use the standard CTC loss and to process the image in a single step. 

\subsubsection{Attention-based approaches}
Attention-based approaches have also been proposed for end-to-end paragraph recognition. It consists in a recurrent process during which the model focuses on specific parts of the input at each time step. To our knowledge, only two works focused on attention-based approaches.

In 2016, the authors of \cite{Bluche2016} proposed a \acrshort{cnn}+\acrshort{mdlstm} model based on line-level attention \textit{i. e.} that the model achieves a kind of implicit line segmentation. The encoder produces feature maps $\mb{f}$ from the input image while the attention module recurrently generates line representations $\mb{l}\mt{t}{}$ applying a weighted sum between the attention weights and the features. Finally, the decoder outputs character probabilities from this line representation. The model iterates a fixed number of times and all the probability lattices of each iteration are concatenated and aligned at paragraph-level with the standard CTC loss. The number of iterations is chosen high enough to cover the maximum number of lines per paragraph in a given dataset.

\begin{figure}[h!]
    \centering
    \includegraphics[width=\textwidth]{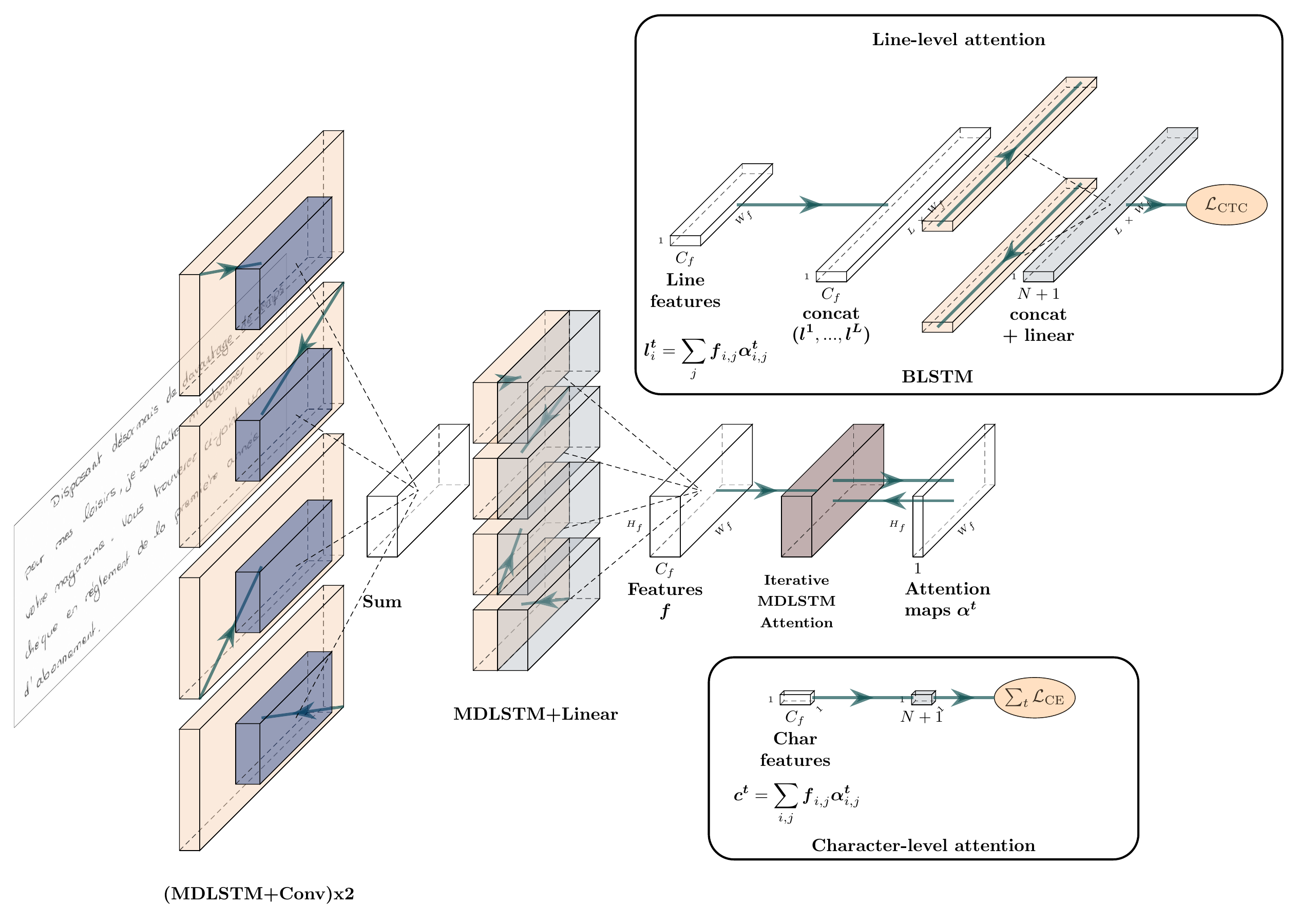}
    \caption{MDLSTM architecture with line-level and character-level attentions. Both approaches are based on the same CNN+MDLSTM encoder to extract 2D features from the input paragraph image. At line-level, the architecture is trained with the CTC loss using the full paragraph transcription, whereas at character-level, the model is trained using the cross-entropy loss, character by character.}
    \label{fig:bluche-attention}
\end{figure}

The next year, T. Bluche \textit{et. al.} proposed a second attention-based \acrshort{cnn}+\acrshort{mdlstm} model in \cite{Bluche2017b}, based on the same encoder. The main difference relates to the attention mechanism which is at character level. It means that the model performs an implicit character segmentation: each iteration is dedicated to the prediction of one character. The model uses specific <start-of-paragraph> and <end-of-paragraph> tokens to initialize and stop the recurrent process, and is trained with the cross entropy loss. Both line-level and character-level models are represented in Figure \ref{fig:bluche-attention}.

While the one-shot approaches rely on a rule-based decoding process to generate the final prediction, the attention-based models are closer to the human way of reading: they learn the reading order. However, contrary to one-shot approaches, the recurrence is inherent to the attention-based models, theoretically leading to higher prediction times, especially with character-level attention. It led us to think that both approaches should be investigated.
This is why we proposed two new architectures for end-to-end paragraph recognition:
\begin{itemize}
    \item The \gls{span}. It is a one-shot approach based on an \gls{fcn} model. Instead of unfolding the input image as in \cite{Yousef2020}, we propose to train the proposed model to both predict and align characters so as to get vertical separation between lines, preserving the two-dimensional nature of the task. In addition, while the approach proposed in \cite{Yousef2020} relies on dataset-specific hyperparameters for the dimension of the bi-linear interpolations, the \gls{span} is able to handle input images of variable sizes, making it flexible enough to be used on multiple datasets without modifying any hyperparameter.
    \item The \gls{van}. It consists in an attention-based model which follows the same idea of implicit line segmentation as in \cite{Bluche2016}. However, we use an \gls{fcn} encoder and an attention module without recurrent layers to reduce the computation time while implying few parameters at the same time. We also proposed a new vertical hybrid attention mechanism as well as a module dedicated to detect the end of the paragraph so as to stop the recurrent process.
\end{itemize}

\section{SPAN: a Simple Predict \& Align Network}
We propose the \gls{span} \cite{Coquenet2021} as an end-to-end one-shot approach for the recognition of handwritten paragraphs. The \gls{span} follows a novel and easy approach for this task. Indeed, it is as simple as line-level \gls{htr} approaches as it relies on a recurrent-free process (we used an \gls{fcn} architecture) and on the standard \gls{ctc} loss. 
In the following, we present the \gls{span} architecture as well as the training strategy we use. We provide an experimental study and a discussion of the model. We provide all source code and pre-trained model weights at \url{https://github.com/FactoDeepLearning/SPAN}.

\subsection{Architecture}
\label{span-architecture}

\begin{figure}[h!]
\centering
    \centering
    \includegraphics[width=\textwidth]{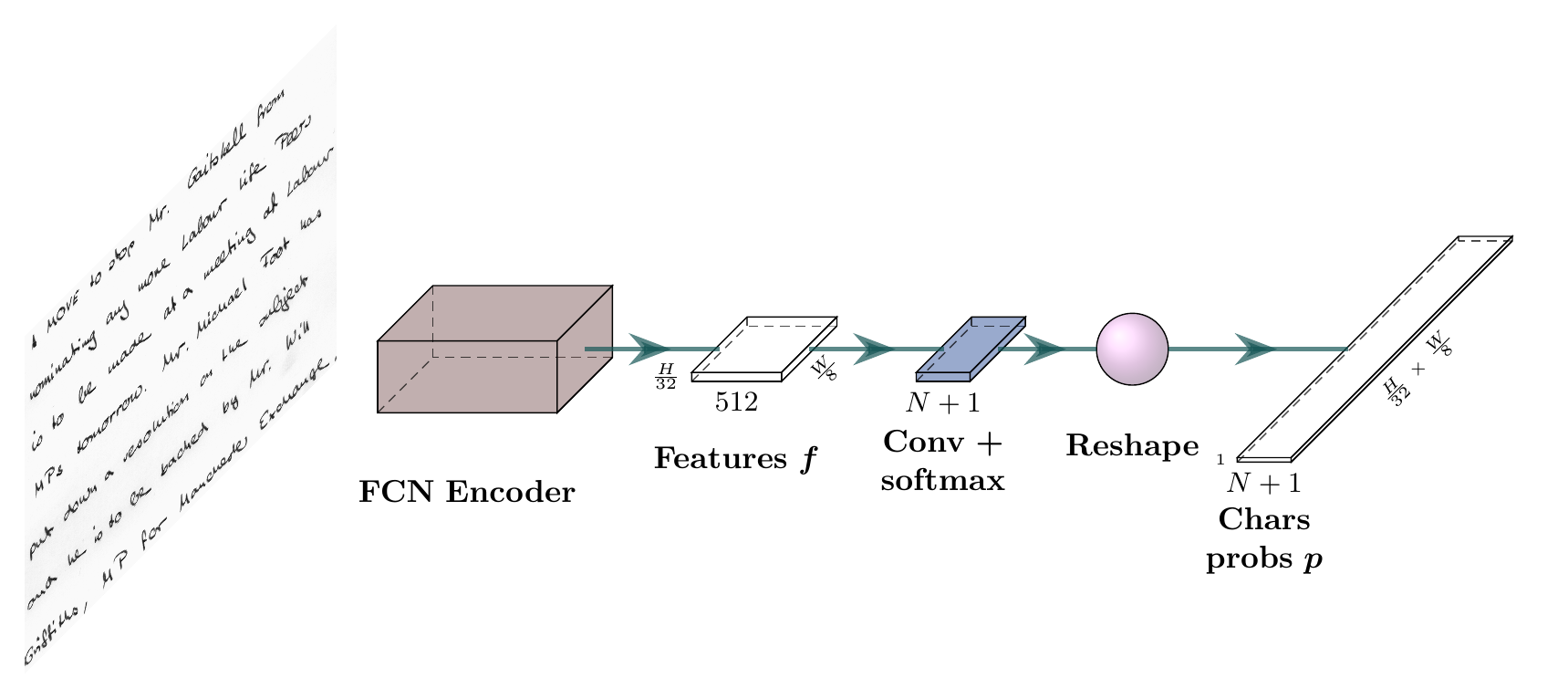}
    \caption{SPAN architecture overview. An FCN Encoder extracts 2D features $\mb{f}$ from the input paragraph image. Probabilities for each character and for the CTC blank label are computed, preserving the 2D representation. The rows of probabilities are then concatenated to get a single row of vectors of probabilities.}
    \label{fig:archi-overview}
\end{figure}

\begin{figure}[h!]
\centering
    \includegraphics[width=\textwidth]{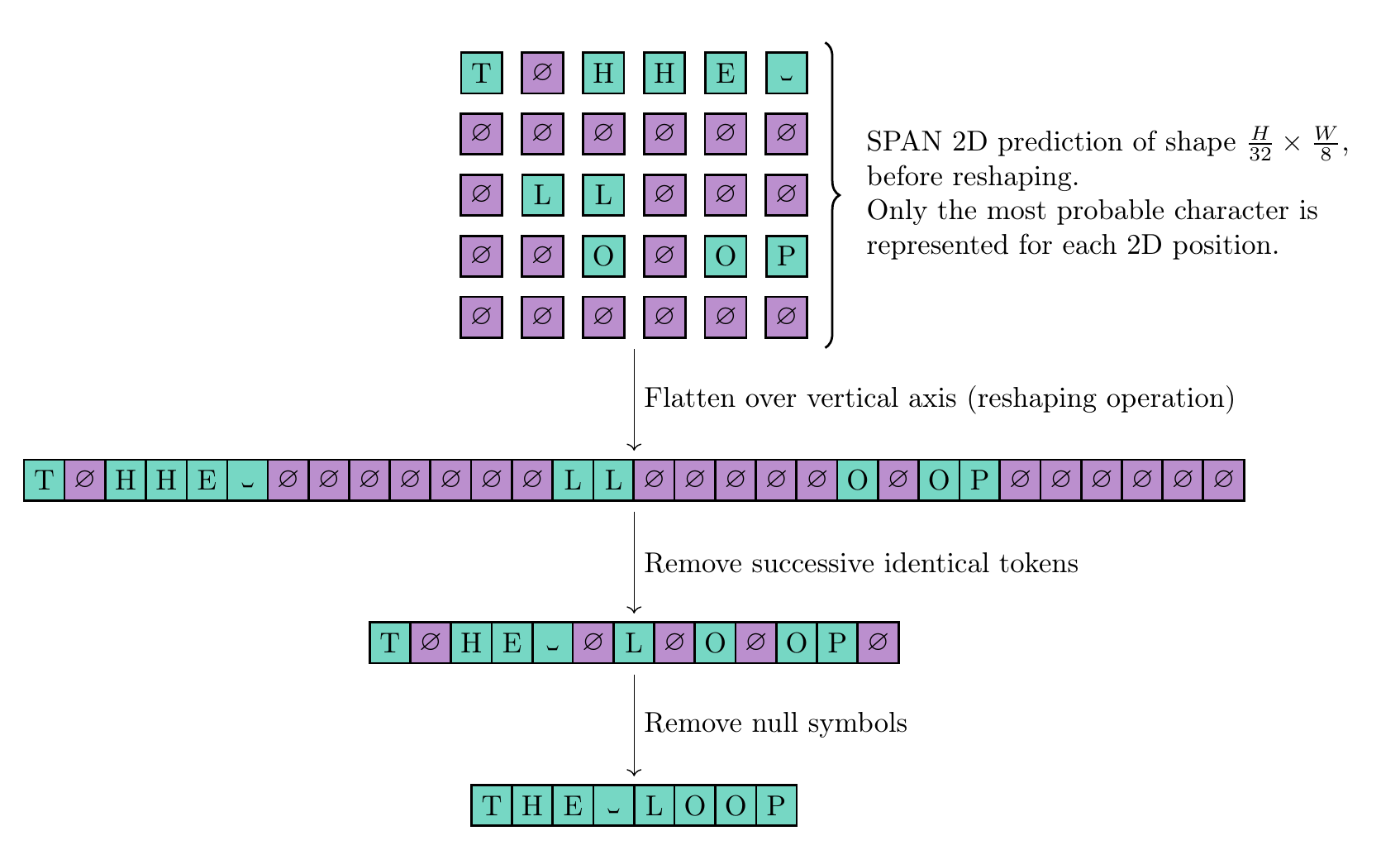}
    \caption{Example of SPAN prediction and CTC decoding process for an input image containing two text lines: "the" and "loop".}
    \label{fig:span-pred}
\end{figure}

The architecture of the proposed \gls{span} is depicted in Figure \ref{fig:archi-overview}. We wanted to keep the original sizes of the input images in order to preserve both their ratio and their details as well as to be flexible enough to adapt to a large variety of datasets. To this end, we use an \gls{fcn} as the encoder to extract features from the 2D paragraph images. 
The decoder consists in a single convolutional layer to predict the characters probabilities. The idea is to predict the character probabilities from the extracted features while keeping the 2D nature of the task. It enables to have a global 2D context which is used to align the characters of a same text lines on the vertical dimension so as to implicitly segment the different text lines.
The convolutional layer is followed by a row concatenation operation, to collapse the vertical dimension, reshaping the 2D latent space into a 1D latent space. As shown on Figure \ref{fig:span-pred}, the vertical prediction alignment enables to preserve the order of the text after the reshaping operation.
This brings us back to a one-dimensional sequence alignment problem which is handled with the standard \gls{ctc} loss.

\subsubsection{Encoder}
\begin{figure}[h!]
    \centering
    \includegraphics[width=\linewidth]{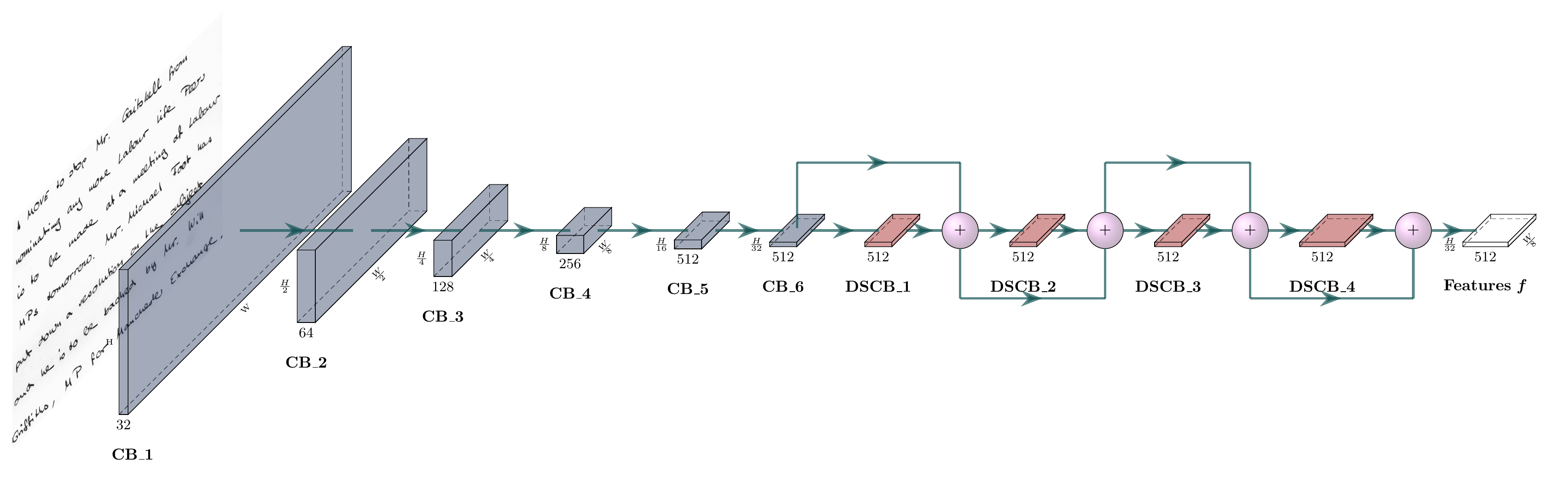}
    \caption{FCN Encoder overview. CB: Convolution Block, DSCB: Depthwise Separable Convolution Block. The height and the width of the feature maps are gradually decreased, while increasing their number. Residual connections are used when the size of the feature maps is preserved from one block to another.}
    \label{fig:encoder-overview}
\end{figure}

The purpose of the encoder is to extract features from the input images. It is made up of some convolutions, with stride greater than $1 \times 1$, in order to reduce the memory consumption: it takes an input image $\mb{X} \in \mathbb{R}^{H \times W \times C}$ and outputs some feature maps $\mb{f} \in \mathbb{R}^{\frac{H}{32} \times \frac{W}{8} \times 512}$ where $H$, $W$ and $C$ are respectively the height, the width and the number of channels ($C=1$ for a grayscale image, $C=3$ for a RGB image). The encoder architecture is depicted in Figure \ref{fig:encoder-overview}. 

It corresponds to an improved version of the \gls{gfcn} we proposed for line-level recognition. The main modifications are as follows. We introduce residual connections in order to have a better feature extraction by aggregating multi-scale information. We made the network simpler by using only two kind of convolution blocks, namely Convolution Blocks (CB) and Depthwise Convolution Blocks (DSCB). We introduce a new dropout strategy named Diffused Mix Dropout (DMD). We remove the gating component as we noticed that it was not beneficial for this new architecture. Stacking these convolutional blocks leads to a receptive field of 961 pixels high and 337 pixels wide.

\paragraph*{Convolutional Block (CB)}
CB is defined as two convolutional layers followed by instance normalization and a third convolutional layer. Each convolutional layer uses $3 \times 3$ kernels and is followed by a ReLU activation function; zero padding is introduced to remove the kernel edge effect. While the first two convolutional layers have a $1 \times 1$ stride, the third one has a $1 \times 1$ stride for CB\_1, $2 \times 2$ stride for CB\_2 to CB\_4 and $2 \times 1$ stride for CB\_5 and CB\_6. The strides are chosen to progressively decrease the dimensions. It is a trade-off between memory consumption and performance. This enables to divide the height by 32 and the width by 8. DMD is applied at three possible locations which are just after the activation layers of the three convolutional layers. 

\paragraph*{Depthwise Separable Convolution Block (DSCB)}
DSCB differs from CB in two respects. On the one hand, the standard convolutional layers are superseded by Depthwise Separable Convolutions (DSC)  \cite{DSC}. The aim is to reduce the number of parameters while keeping the same level of performance. On the other hand, the third convolutional layer has a fixed stride of $1 \times 1$. In this way, the shape is preserved until the last layer of the encoder. This enables to introduce residual connections with element-wise sum operator between the DSCB.

\paragraph*{Diffused Mix Dropout}
\label{section-archi-dropout}
Dropout is commonly used as regularization during training to avoid overfitting. We introduce Diffused Mix Dropout, a new dropout strategy that takes advantage of the two main modes proposed in the literature, namely standard (std.) dropout \cite{Dropout} and spatial (or 2d) dropout \cite{SpatialDropout}. We define the Mix Dropout (MD) layer as the layer which applies one of the two possible dropout modes, randomly. It enables to take advantage of the two implementations in a single layer. Diffused Mixed Dropout (DMD) consists in randomly applying MD at different locations among a set of pre-selected locations. In the model, we use DMD with dropout probability of 0.5 and 0.25 respectively for the standard and the 2d modes. Both modes have equivalent probabilities to be chosen at each execution. The benefit of using this dropout strategy is discussed in Section \ref{section-exp-dropout} through experiments on the \gls{van} architecture.

\subsubsection{Decoder}
\begin{figure}[h!]
\centering
    \includegraphics[width=0.9\textwidth]{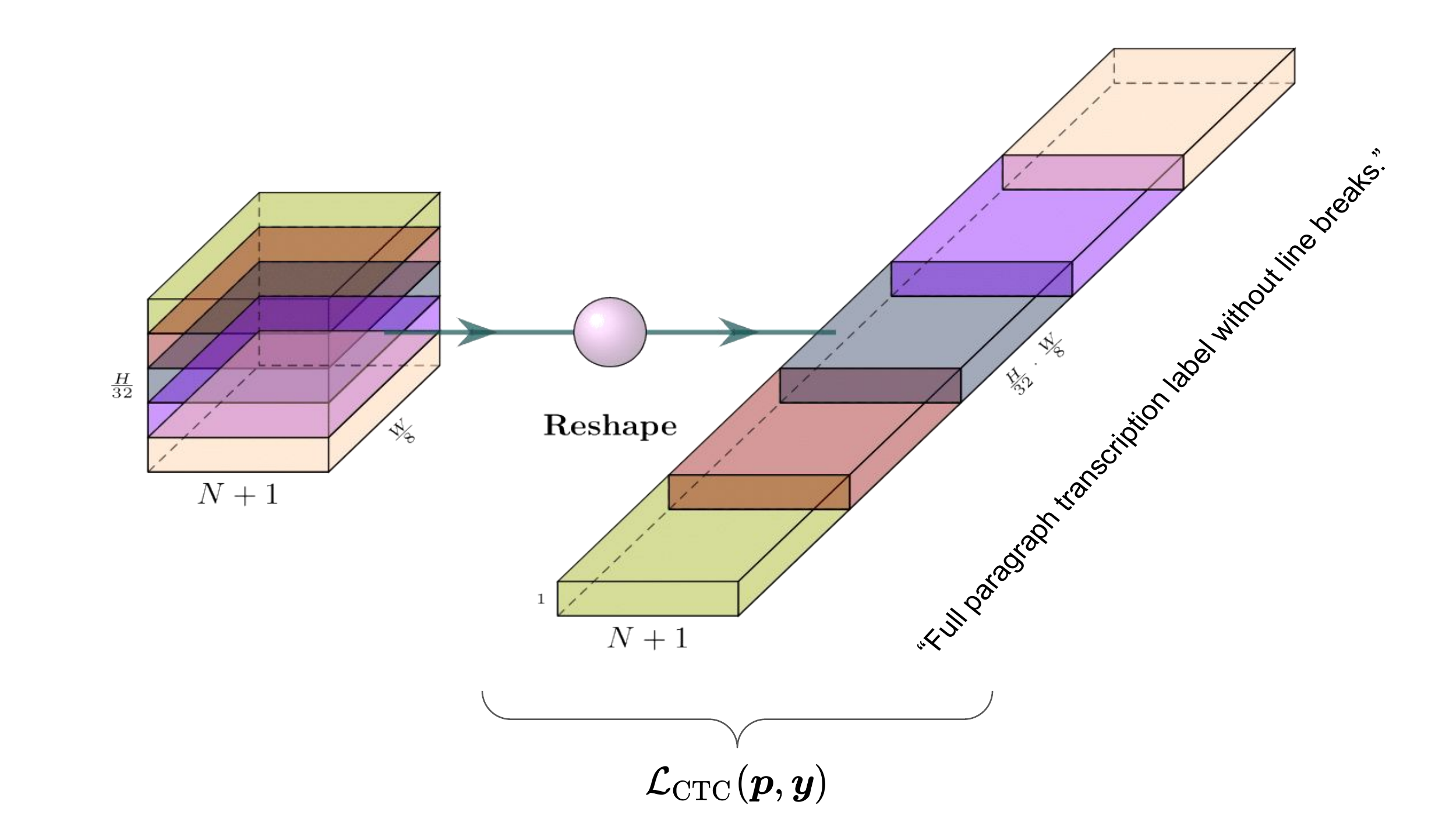}
    \caption{Reshaping operation and loss visualization. No computations are performed in the reshaping operation, both left and right tensors represent characters and CTC blank label probabilities. The CTC loss is computed between the one-dimensional probabilities sequence and the paragraph transcription.}
    \label{fig:reshape}
\end{figure}

The decoder aims at predicting and aligning the probabilities of the characters and of the \gls{ctc} blank label for each 2D position of the features $\mb{f}$.
The decoder part is very simple as it only consists in a single convolutional layer followed by a flatten operation. 
The convolutional layer is performed with a $5 \times 5$ kernel, $1 \times 1$ stride and $2 \times 2$ padding. It outputs $N+1$ channels, $N$ being the size of the charset. Then, the $\frac{H}{32}$ rows are concatenated to obtain the one-dimensional prediction sequence $\mb{p}\in \mathbb{R}^{(\frac{H}{32} \cdot \frac{W}{8}) \times (N+1)}$ as depicted in Figure \ref{fig:reshape}. The \gls{ctc} loss is then computed between this one-dimensional prediction sequence and the paragraph transcription ground truth, without line breaks.

We can highlight some important aspects about the decoder, that are illustrated in Figure \ref{fig:span-pred} with an example of prediction for the target sequence "THE\textbackslash nLOOP":
\begin{itemize}
    \item In this work, the \gls{ctc} blank label (or null symbol $\varnothing$) has a new function. In the standard \gls{ctc} decoding process, the \gls{ctc} blank label is used to recognize two identical successive characters and to predict "nothing", acting like a joker. Here, it is also used to separate lines in a two-dimensional context: predicting blank labels enables to fill the space between two successive text lines, as shown with the second row of prediction in Figure \ref{fig:span-pred}.
    \item One should notice that the prediction occurs before reshaping to 1D, which allows to take advantage of the two-dimensional context in the decision layer. This enables to localize the previous and next lines, and to align the predictions of the same text line on the same row \textit{i.e.} to separate them from the predictions of the other text lines.
    \item Since the prediction rows are concatenated, they are processed sequentially; nothing prevents the model from predicting the beginning of the text line on one row and the end on the next one. This is only feasible if there is enough physical space between this text line and the following one so as not to overlap with the prediction of the next text line. In Section \ref{section:span_results}, we show that this enables the \gls{span} to process inclined lines.
\end{itemize}

\subsection{Experimental conditions}
\label{section:span-exp-conditions}
We now describe the experimental conditions:

\begin{itemize}
\item Datasets: we evaluate the \gls{span} on three public datasets at paragraph level: RIMES 2011, IAM and READ 2016 (as described in Section \ref{section:dataset}).
\item Pre-processings: paragraph images are downscaled by a factor of 2 through a bilinear interpolation leading to a resolution of 150 \acrshort{dpi}. Gray-scaled images are converted into RGB images concatenating the same values three times, for transfer learning purposes. They are then normalized (zero mean and unit variance) considering the channels independently.
\item Data Augmentation: a data augmentation strategy is applied at training time to reduce overfitting. The augmentation techniques are used in this order: resolution modification, perspective transformation, elastic distortion and random projective transformation (from \cite{Yousef2020}), dilation and erosion, brightness and contrast adjustment, and sign flipping. Each transformation has a probability of 0.2 to be applied. Except for perspective transformation, elastic distortion and random projective transformation which are mutually exclusive, each augmentation technique can be combined with the others.
\item Pre-training: the weights of the \gls{span} encoder are initialized with those of a line-level \gls{htr} model trained on isolated text lines of the same dataset. This line-level \gls{htr} model is depicted in Figure \ref{fig:span-line-ocr}. It is made up of the \gls{span} \gls{fcn} encoder, followed by an Adaptive Max Pooling layer which pushes the vertical dimension to collapse. A final convolutional layer predicts the probability of the characters and of the CTC null symbol.

\begin{figure}[ht!]
\centering
\includegraphics[width=\textwidth]{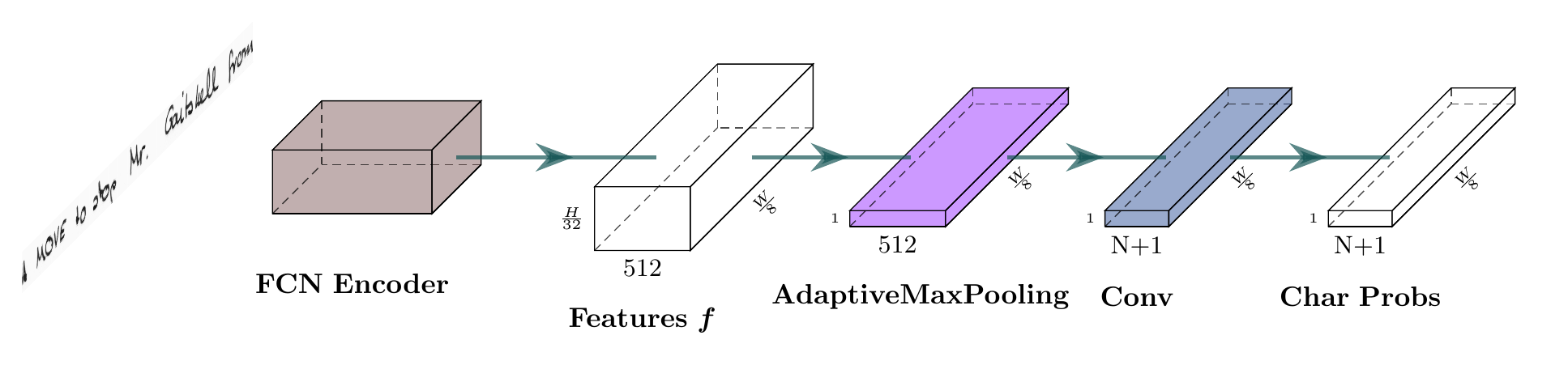}
\caption{Line-level HTR model used for pre-training. The input text line image can be of variable size, the adaptive max pooling operation enables to push the vertical dimension to collapse.}
\label{fig:span-line-ocr}
\end{figure}

\item Training details: we used the Pytorch framework to train and evaluate the models. In all experiments, the networks are trained with the Adam optimizer, with an initial learning rate of $10^{-4}$. Trainings are performed on a single \acrshort{gpu} Tesla V100 (32Gb), during 2 days, with a mini-batch size of 4 for paragraph images and 16 for text lines images. We used the original paragraph ground truth replacing line breaks by space characters.
\item Metrics: we used the \gls{cer} and the \gls{wer} to evaluate the quality of the text recognition. They are both computed with the Levenshtein distance between the ground truth text and the predicted text at paragraph level, without line breaks. Those edit distances are then normalized by the length of the ground truth. For the \gls{wer}, punctuation marks are not counted as words: they are considered as belonging to the previous word. Other metrics are provided in the following experiments such as the number of parameters implied by the models. 
\end{itemize}

We do not use any post-processing \textit{i.e.} we do not use any external language model nor lexicon constraint. Moreover, we only use best path decoding to get the final predictions from the character probabilities lattice. We use exactly the same training configuration from one dataset to another, without model modification, except for the last layer which depends on the charset size.

\subsection{Experimental study}
\label{section:span_results}

In the following, we compare the \gls{span} with the state-of-the-art approaches. We provide a visualization of the vertical alignment learned by the model and we study the impact of pre-training on the \gls{span} results.

\subsubsection{Comparison with the state of the art}

In this section, we compare our approach to state-of-the-art models on the RIMES 2011, IAM and READ 2016 datasets, at paragraph level and in the same conditions \textit{i.e.} without external language model, external data nor lexicon constraint.

\begin{table}[!h]
    \caption{Requirements comparison of the SPAN with the state-of-the-art approaches.}
    \centering
    \resizebox{\linewidth}{!}{
    \begin{threeparttable}
    \begin{tabular}{ l c c c c c c c c}
    \hline
    Architecture & \# Param. & Max memory & Transcription label & Seg. label & PT & CL & HA\\
    \hline
    \hline
    \cite{Carbonell2019} \acrshort{rpn}+\acrshort{cnn}+\acrshort{blstm} & & & Word & Word & \\ 
    \cite{Chung2020} \acrshort{rpn}+\acrshort{cnn}+\acrshort{blstm} & & & Word & Word & \\
    \cite{Wigington2018} \acrshort{rpn}+\acrshort{cnn}+\acrshort{blstm} & & & Line & Line \\
    \cite{Bluche2017b} \acrshort{cnn}+\acrshort{mdlstm}\tnote{**} &   & & Paragraph  & Paragraph & Line & \checkmark & \xmark \\
    \cite{Bluche2016} \acrshort{cnn}+\acrshort{mdlstm}\tnote{*}  & & & Paragraph & Paragraph & Line & \checkmark & \xmark\\
    \cite{Yousef2020} \acrshort{gfcn} & 16.4 M & 8.8 Gb  & Paragraph & Paragraph & \xmark & \xmark & \checkmark\\
    \cite{Coquenet2021} \acrshort{span} (\acrshort{fcn}) &  19.2 M & 5.1 Gb &  Paragraph & Paragraph & Line & \xmark & \xmark\\
    \hline
    \end{tabular}
    \begin{tablenotes}
    \item[*] With line-level attention.
    \item[**] with character-level attention.
    \end{tablenotes}
    \end{threeparttable}
    }
    \label{table:comparison}
\end{table}

Prior to compare the obtained results to the state of the art, it is important to understand the experimental conditions of each of the methods. Table \ref{table:comparison} shows model details that should be taken into account to fairly compare the following tables of results. Quantitative metrics are computed for the IAM dataset, without automatic mixed precision (for a fair comparison with respect to the memory usage). From left to right, the columns respectively denote the architecture, the number of trainable parameters, the maximum \acrshort{gpu} memory usage during training (for a mini-batch size of 1, data augmentation included), the minimum transcription level required, the minimum segmentation level required, the use of Pre-Training (PT) on sub-images, the use of specific Curriculum Learning (CL) and finally the Hyperparameter Adaptation (HA) requirements from one dataset to another.

As one can see, models from \cite{Carbonell2019,Chung2020,Wigington2018} require transcription and segmentation labels at word or line levels to be trained, which implies more costly annotations. The models from \cite{Bluche2016,Bluche2017b} and the \gls{span} are pre-trained on text line images to speed up convergence and to reach better results, thus also using line segmentation and transcription labels even if it is not strictly necessary. \cite{Bluche2016,Bluche2017b} used a specific curriculum learning method for training. In \cite{Yousef2020}, some hyperparameters must be modified from one dataset to another in order to reach optimal performance, namely the fixed input dimension and two intermediate upsampling sizes which are crucial. We do not have such problem since we are working with input images of variable size and we focus on the resolution to be robust to the variety of datasets. Moreover, despite a larger number of parameters (+ 17\% compared to \cite{Yousef2020}), the \gls{span} requires less \acrshort{gpu} memory which is a critical point when training deep neural networks.

\begin{table}[!h]
    \caption{Comparison of the SPAN results with the state-of-the-art approaches on the RIMES 2011, IAM and READ 2016 datasets.}
    \centering
    \resizebox{0.7\linewidth}{!}{
    \begin{threeparttable}[b]
    \begin{tabular}{ l c c c c }
    \hline
    \multirow{2}{*}{Architecture} & \gls{cer} (\%) & \gls{wer} (\%) & \gls{cer} (\%) & \gls{wer} (\%) \\ 
    & validation & validation & test & test\\
    \hline
    \hline
    \textbf{RIMES 2011}\\
    \cite{Bluche2016} \acrshort{cnn}+\acrshort{mdlstm}\tnote{*} & 2.5 & 12.0 & 2.9 & 12.6 \\
    \cite{Wigington2018} \acrshort{rpn}+\acrshort{cnn}+\acrshort{blstm} & & & \textbf{2.1} & 9.3 \\ 
    \cite{Coquenet2021} \acrshort{span} (\acrshort{fcn}) & 3.56 & 14.29 & 4.17 & 15.61 \\
    \\
    \textbf{IAM}\\
    \cite{Carbonell2019} \acrshort{rpn}+\acrshort{cnn}+\acrshort{blstm}\tnote{*}& 13.8 & & 15.6 & \\ 
    \cite{Chung2020} \acrshort{rpn}+\acrshort{cnn}+\acrshort{blstm} & & & 8.5 & \\
    \cite{Wigington2018} \acrshort{rpn}+\acrshort{cnn}+\acrshort{blstm}& & & 6.4 & 23.2  \\
    \cite{Bluche2017b} \acrshort{cnn}+\acrshort{mdlstm}\tnote{**}& &  & 16.2 &  \\
    \cite{Bluche2016} \acrshort{cnn}+\acrshort{mdlstm}\tnote{*}& 4.9 & 17.1 & 7.9 & 24.6 \\
    \cite{Yousef2020} \acrshort{gfcn}& & & \textbf{4.7} & \\
    \cite{Coquenet2021} \acrshort{span} (\acrshort{fcn})  & 3.57 & 15.31 & 5.45 & 19.83 \\
    \\
    \textbf{READ 2016}\\
    \cite{Coquenet2021} \acrshort{span} (\acrshort{fcn}) & 5.09 & 23.69 & \textbf{6.20} & 25.69\\
    \hline
    \end{tabular}
    \begin{tablenotes}
    \item[*] with line-level attention.
    \item[**] with character-level attention.
    \end{tablenotes}
    \end{threeparttable}
    }
    \label{table:sota-span}
\end{table}

Table \ref{table:sota-span} shows the results of the \gls{span} compared to the state-of-the-art approaches, for the RIMES 2011, IAM and READ 2016 datasets respectively. To our knowledge, there are no results reported in the literature on the READ 2016 dataset at paragraph level. One can notice that we reach competitive results on those three datasets, each having its own complexities, with an architecture as simple as an \gls{fcn}. In addition, we used exactly the same hyperparameters from one dataset to another, showing the robustness of this approach.

Results include model pre-training on text line images but the model can be trained without pre-training \textit{i.e.} without using any line-level annotation, while keeping competitive results, as shown in Section \ref{section:pretrain}.

\subsubsection{SPAN prediction visualization}

\begin{figure}[ht!]
{\centering
\includegraphics[width=0.9\textwidth]{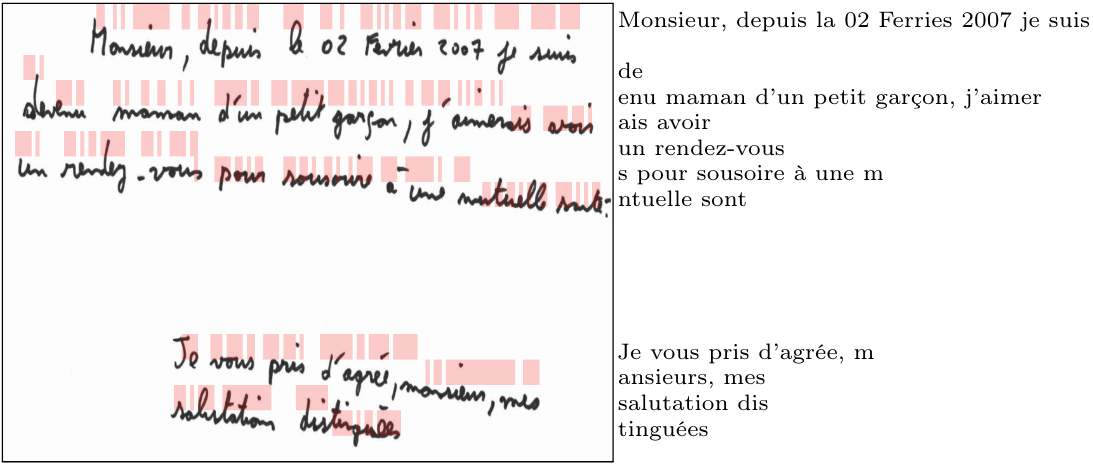}
}
\par\medskip
\footnotesize{Monsieur, depuis l\textbf{a} 02 Fe\textbf{r}rie\textbf{s} 2007 je suis de\textbf{\textit{v}}enu maman d'un petit garçon, j'aimerais avoir un rendez-vous\textbf{s} pour sous\textbf{o}ire à une m\textbf{n}tuelle s\textbf{o}nt\textbf{\textit{é. }}Je vous pris d'agrée, m\textbf{a}nsieurs, mes salutation\textbf{\textit{s}} distinguées}

    \caption{SPAN predictions visualization for a RIMES 2011 test example. Left: 2D characters predictions are projected on the input image. Red color indicates a character prediction while transparency means blank label prediction. Right: row by row text prediction. Bottom: full text prediction where errors are shown in bold and missing letters are shown in italic.}
    \label{fig:viz}
\end{figure}

Figure \ref{fig:viz} presents a visualization of the \gls{span} prediction for an example of the RIMES 2011 test set. Character predictions are shown in red; they seem like rectangle since they are resized to fit the input image size (the features size is $\frac{H}{32} \times \frac{W}{8}$). Combined with the receptive field effect, this explains the shift that can occur between the prediction and the text. As one can notice, text line predictions are totally aligned, or aligned by blocks; the lines are well separated by blank labels, which act as line spacing labels. As one can see, this block alignment enables to handle downward inclined lines, especially for lines 3 and 4. Moreover, the model does not degrade in the presence of large line spacing.

\subsubsection{Impact of pre-training}
\label{section:pretrain}

In this experiment, we try to highlight the impact of pre-training on the \gls{span} results. To this end, we compare two pre-training methods at line level: one focusing only on the optical Recognition task (R) and the second one focusing on both Recognition and prediction Alignment (R\&A). Let's define the following pre-training (PT) approaches:
\begin{itemize}
    \item \gls{span}-Line-R\&A: the \gls{span} is trained with line-level images. Here, the network has to learn both the recognition and the alignment tasks. Indeed, there can be multiple rows of features for a single text line. This way, the model must learn the alignment aspect, in addition to the character recognition, to achieve correct predictions.
    \item Pool-Line-R: it corresponds to the pre-training strategy presented in Section \ref{section:span-exp-conditions}. The line-level \gls{htr} model depicted in Figure \ref{fig:span-line-ocr} is trained on isolated text line images to only focus on the recognition task. Since the vertical dimension is collapsed (through adaptive max pooling), the network does not need to care about the vertical alignment of the predictions.
\end{itemize}

It leads to three different training approaches:
\begin{itemize}
    \item \gls{span}-Scratch: the \gls{span} is trained directly on paragraph images without pre-training.
    \item \gls{span}-PT-R: \gls{span} weights are initialized with Pool-Line-R ones. It is then trained with paragraph images.
    \item \gls{span}-PT-R\&A: \gls{span} weights are initialized with \gls{span}-Line-R\&A ones. It is then trained with paragraph images.
\end{itemize}

One can note that the vertical receptive field is larger than the height of the images of isolated text lines. Thus, when the model switches from line images to paragraph images, the decision benefits from more context, which replaces part of the previously used padding.

Results are given in Table \ref{table:pretrain}. Focusing on the line-level section (pre-training step), one can notice that, as expected, we reached better results on text lines when the task is reduced to optical recognition compared to the task of recognition and alignment, whatever the dataset. This leads to a \gls{cer} improvement of 0.94 point for IAM, 0.79 point for RIMES 2011 and 0.38 point for READ 2016. This can be explained by the difficulty of the task: it is easier to just recognize characters than to recognize characters and align their predictions on the vertical axis. In addition, we assume that aligning the predictions does not help the character recognition.

\begin{table}[!h]
    \caption{Impact of pre-training the SPAN on line images for the IAM, RIMES 2011 and READ 2016 datasets. Results are given on the test sets.}
    \centering
    \resizebox{\linewidth}{!}{
    \begin{tabular}{ l c c c c c c}
    \hline
    \multirow{2}{*}{Approach}& \multicolumn{2}{c}{IAM} & \multicolumn{2}{c}{RIMES 2011} & \multicolumn{2}{c}{READ 2016} \\
     & \gls{cer} (\%) & \gls{wer} (\%) & \gls{cer} (\%) & \gls{wer} (\%) & \gls{cer} (\%) & \gls{wer} (\%) \\ 
    \hline
    \hline
    \textbf{Line-level (pre-training)} \\
    Pool-Line-R & \textbf{4.82} & \textbf{18.17} & \textbf{3.02} & \textbf{10.73} & \textbf{4.56} & \textbf{21.07}\\
    \gls{span}-Line-R\&A & 5.76 & 21.33 & 3.81 & 13.80 & 4.94 & 22.19\\   
    \\
    \textbf{Paragraph-level (training)} \\
    \gls{span}-Scratch & 6.46 & 23.75 & \textbf{4.15} & 16.31 & 9.13 & 36.63\\
    \gls{span}-PT-R & \textbf{5.45} & \textbf{19.83} & 4.74 & \textbf{15.55} & \textbf{6.20} & \textbf{25.69}\\
    \gls{span}-PT-R\&A & 5.78 & 21.16 & 4.17 & 15.71 & 6.62 & 27.38\\
    \hline
    \end{tabular}
    }
    \label{table:pretrain}
\end{table}

Now, comparing the paragraph-level approaches, one can notice that, except for the RIMES 2011 \gls{cer}, pre-training leads to better results, and sometimes by far (-2.93 points of \gls{cer} for READ 2016). Moreover, pre-training on an easier task, \textit{i.e.} only on the optical recognition, is even more efficient. We assume that focusing only on the recognition part enables to improve the feature extraction part compared to both recognizing and aligning the predictions.



    

\subsection{Discussion}

We propose the \acrfull{span} as an end-to-end recurrence-free \gls{fcn} model performing \gls{htr} at paragraph level. It reaches competitive results on the RIMES 2011, IAM and READ 2016 datasets without any architecture or training adaptation from one dataset to another. It shows that one can achieve \gls{htr} at paragraph level with a standard \gls{fcn} network and using the standard \gls{ctc} loss provided that the architecture includes a way to serialize the representation, through a simple flattening operation for instance.

The proposed architecture has other advantages. It only needs transcription label at paragraph level without line breaks, leveraging the need for handmade annotation, which is a critical point for a deep learning system.  It can handle input images of variable sizes, making it robust enough to adapt to multiple datasets; it is also able, through the reshaping operation, to deal with downward inclined text lines. However, this same reshaping operation prevent the model from handling upward ones due to the fixed reshaping order. The use of a second \gls{span} model whose reshaping operation would have a reversed ordering method could enable to deal with both upward and downward inclined lines.

Moreover, since we are using the standard \gls{ctc} loss, one can easily add standard character or word language model to further improve the results. The \gls{span} could also adapt to full page documents since it is based on the image resolution regardless of its size. However, it has to be noticed that this model is limited to single-column multi-line text images due to the row concatenation operation. 

We now present the second paragraph-level architecture we propose: the \gls{van}.

\section{VAN: a Vertical Attention Network}

We propose the \acrlong{van} \cite{Coquenet2022} as an end-to-end attention-based approach for the recognition of handwritten paragraphs. It is based on the idea of a recurrent process which successively focuses on the different text lines, one after the other, and which learns to detect the end of the paragraph in order to stop the prediction. This line attention enables to remain in a one-dimensional sequence alignment problem between the recognized text lines and their ground truth transcriptions, which is handled by the \gls{ctc}.

In the following, we present the \gls{van} architecture and the training strategy. We provide an experimental study, a visualization of the attention process, as well as a discussion of the model. We provide all source code and pre-trained model weights at \url{https://github.com/FactoDeepLearning/VerticalAttentionOCR}.

\subsection{Architecture}
\label{section-archi}

The \gls{van} follows the encoder-decoder principles with an attention module. It takes as input an image $\mb{X}$ and recurrently recognizes the $L$ text lines $[\mb{y}\mt{1}{}, ..., \mb{y}\mt{L}{}]$ present in it. Each text line $\mb{y}\mt{t}{}$ is a sequence of tokens from an alphabet $\mset{A}$, with $|\mset{A}|=N$.

\begin{figure}[htbp!]
\centering
\includegraphics[width=\textwidth]{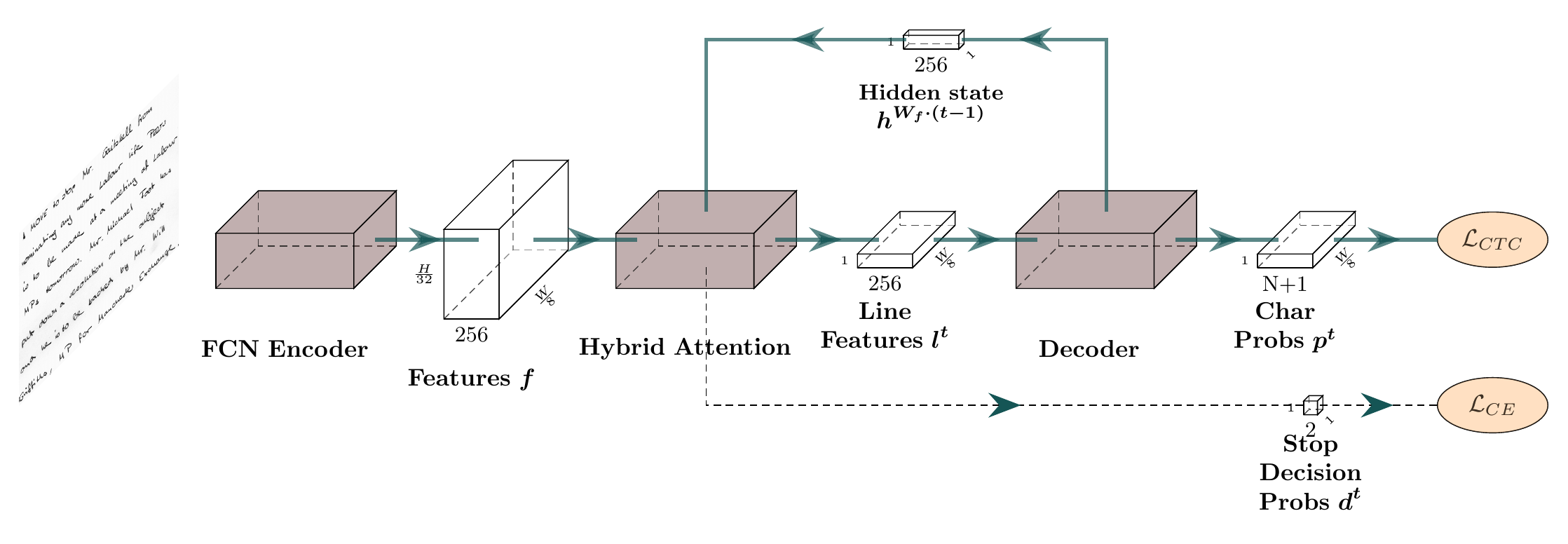}
        \caption{VAN architecture overview. An FCN encoder extracts 2D features $\mb{f}$ from the input paragraph image. Text lines are recognized one after the other. A hybrid attention module iteratively generates the representation of the current text line $\mb{l}\mt{t}{}$ to recognize. The LSTM-based decoder produces a probability lattice for the current text line, as for standard line-level HTR. At each iteration, the probability to detect the end of the paragraph $\mb{d}\mt{t}{}$ is computed to decide when to stop the recognition process.}
        \label{fig:van-overview}
\end{figure}

The overall model is presented in Figure \ref{fig:van-overview}. It first extracts features $\mb{f}$ from $\mb{X}$. We wanted the encoder stage to be modular enough in order to be plugged in different architectures dedicated to text lines, paragraphs or documents recognition without needing any adaptation. To this end, we chose an \gls{fcn} encoder that can deal with input images of variable heights and widths, possibly containing multiple lines of text. The attention module is the main control block: it recurrently produces vertical attention weights that focus where to select the next features for the recognition of the next text line. This way, it recurrently generates as many text line features/representations $\mb{l}\mt{t}{}$ as there are text lines in the input image. This is handled by the detection of the end of the paragraph, which is carried out through the computation of probabilities $\mb{d}\mt{t}{}$ to stop the whole process. Finally, the \acrshort{lstm}-based decoder produces character probabilities $\mb{p}\mt{t}{}$ for each of the $W_f=\frac{W}{8}$ frames of each generated line features $\mb{l}\mt{t}{}$. Here, the different frames correspond to the horizontal axis of the line features. The best path decoding strategy is used to get the final prediction. The model is trained using a combination of two losses: the \gls{ctc} loss for the recognition task, through the line-by-line alignment between recognized text lines (through its probabilities lattice $\mb{p}\mt{t}{}$) and ground truth line transcriptions $\mb{y}\mt{t}{}$, and the \gls{ce} loss for the end-of-paragraph detection. We now describe each module in detail.

\subsubsection{Encoder}
An \gls{fcn} encoder is used to extract features from the input paragraph images. It takes as input an image $\mb{X} \in \mathbb{R}^{H \times W \times C}$, with $H$, $W$ and $C$ being respectively the height, the width and the number of channels. Then, it outputs some feature maps for the whole paragraph image: $\mb{f} \in \mathbb{R}^{H_f \times W_f \times C_f}$ with $H_f=\frac{H}{32}$, $W_f = \frac{W}{8}$ and $C_f = 256$. Those features must contain enough information to recognize the characters afterwards, while preserving the two-dimensional nature of the task.

We used the \gls{span} encoder as it shows interesting results for the same task. The only difference is the maximum number of channels which is set to 256 instead of 512. It considerably reduces the number of trainable parameters at stake, from 18.1M to 1.7M, without impacting the performance for the \gls{van} architecture.

\subsubsection{Attention}
\begin{figure}[h!]
\centering
    \includegraphics[width=0.9\textheight, angle=90]{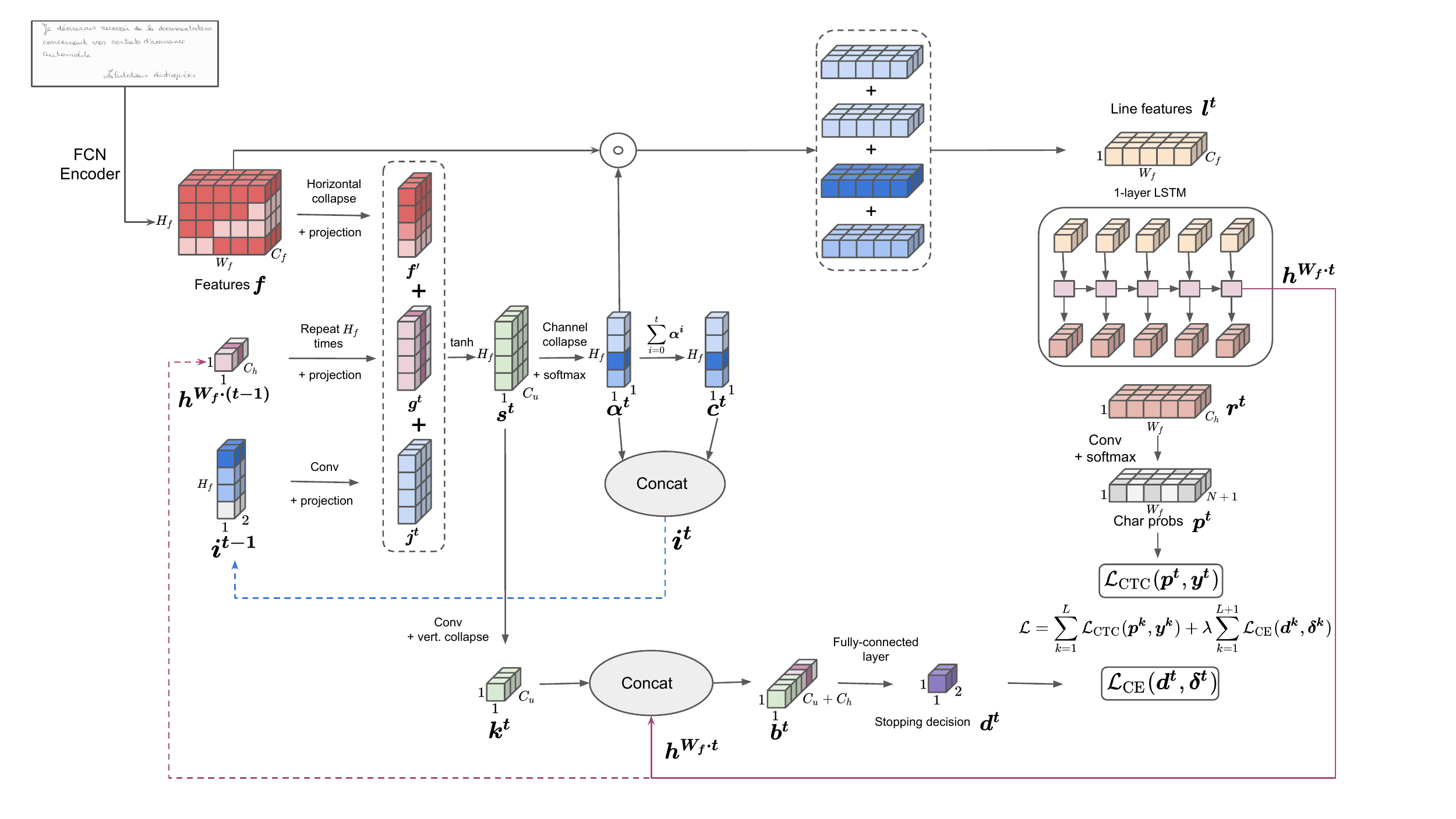}
        \caption{Focus on the VAN hybrid attention mechanism.}
    \label{fig:van-attention}
\end{figure}

At this step, we have computed the features $\mb{f}$. This is a static representation that preserves the two-dimensional aspect of the task: it never changes through the attention process.
The purpose of the attention module is to recursively produce text line representations, in the desired reading order, from top to bottom for example. In addition, it has to decide when to stop generating a new line representation, \textit{i.e.} to detect the end of the paragraph. The whole process is depicted in Figure \ref{fig:van-attention}.

\subsubsection*{Line features generation}
Given an input image with $L$ text lines, the attention module successively produces $L$ line features. In the following, the process of the $t^\mathrm{th}$ line will be referred as attention step $t$.
To produce the successive line features, we chose to use a soft attention mechanism, as proposed by Bahdanau in \cite{Bahdanau}, and more specifically a hybrid attention. The soft attention is a way to focus on specific frames among a one-dimensional sequence of frames. It involves the computation of attention weights $\mb{\alpha}\mt{t}{i}$ for each frame $i$ at attention step $t$. These attention weights sum to 1 and quantify the importance of each frame for the attention step $t$. We apply the attention on the vertical axis only, in order to focus on specific features row at each attention step $t$. This way, we need one weight per features row: we roughly expect one weight to be near 1, selecting the correct feature row and the others to be near 0. Each attention step $t$ is dedicated to process the text line number $t$. The corresponding line features $\mb{l}\mt{t}{}$ can be computed as a weighted sum between the feature rows and the attention weights: 
\begin{equation}
    \mb{l}\mt{t}{} = \sum_{i=1}^{H_f}\mb{\alpha}\mt{t}{i} \cdot \mb{f}_i,
\end{equation}
with $\mb{f}_i$ denoting the $i^\mathrm{th}$ row of $\mb{f}$ ($\mb{f}_i \in \mathbb{R}^{W_f \times C_f}$).
As one can note, this weighted sum enables to provide line features $\mb{l}\mt{t}{}$ to be a one-dimensional sequence and thus to use the standard \gls{ctc} loss for each recognized text line.

The hybrid aspect means that attention weights are computed from both the content and the location output layers. In our case, the attention weights computation combines three elements: 
\begin{itemize}
    \item $\mb{f} \in \mathbb{R}^{H_f \times W_f \times C_f}$: the features in which we want to select specific line features.
    \item $\mb{\alpha}\mt{t-1}{} \in \mathbb{R}^{H_f}$: the previous attention weights. To know which line to select, we must know which ones have already been processed. This is the location-based part.
    \item $\mb{h}\mt{W_f \cdot (t-1)}{} \in \mathbb{R}^{C_h}$: the hidden state of the \acrshort{lstm} from the decoder module (described below), after processing the previous text line (each line features is made up of $W_f$ frames). It contains information about what has been recognized so far. This is the content-based part.
\end{itemize}

$\mb{f}$, $\mb{\alpha}\mt{t-1}{}$ and $\mb{h}\mt{W_f \cdot (t-1)}{}$ cannot be combined directly due to dimension mismatching. Since we want to normalize the attention weights over the vertical axis, we must collapse the horizontal axis of the features. The idea is that we only need to know if there is at least one character in a features row to decide to process it as a text line. Thus, we can collapse this horizontal dimension without information loss. This collapse is carried out in two steps: we first get back to a fixed width of 100 (since inputs are of variable sizes) through Adaptive Max Pooling. Then, a densely connected layer pushes the horizontal dimension to collapse. The remaining vertical representation is called $\mb{f'} \in \mathbb{R}^{H_f \times C_f}$.

Instead of using directly $\mb{\alpha}\mt{t-1}{}$, we found that combining it with a coverage vector $\mb{c}\mt{t}{} \in \mathbb{R}^{H_f}$ is more beneficial. $\mb{c}\mt{t}{}$ is defined as the sum of all previous attention weights:
    \begin{equation}
            \mb{c}\mt{t}{} = \sum_{k=0}^{t}\mb{\alpha}\mt{k}{}.
    \end{equation}
    $\mb{c}\mt{t}{}$ enables to keep track of all previous line positions processed. $\mb{c}\mt{t}{}$ is clamped between 0 and 1 for stability; we only need to know whether a row has already been processed or not. Bigger values would not have sense. We combined $\mb{c}\mt{t}{}$ and $\mb{\alpha}\mt{t}{}$ through concatenation over the channel axis, leading to $\mb{i}\mt{t}{} \in \mathbb{R}^{H_f \times 2}$.
Finally, we extract some context information from $\mb{i}\mt{t}{}$ through a one-dimensional convolutional layer with $C_j=16$ filters of kernel size 15 and stride 1 with zero padding, followed by instance normalization. This outputs $\mb{j}\mt{t}{} \in \mathbb{R}^{H_f \times C_j}$ which contains all the contextualized spatial information from past attention steps.

We can now compute the attention weights $\mb{\alpha}\mt{t}{}$ as follows:
\begin{itemize}
    \item For each row $i$, we compute the associated multi-scale information $\mb{s}\mt{t}{i}$, gathering all the elements we have seen previously:
        \begin{equation}
        \mb{s}\mt{t}{i} = \tanh(
        \mb{W}_f \cdot \mb{f'}_i + \mb{W}_j \cdot \mb{j}\mt{t}{i} + \mb{W}_h \cdot \mb{h}\mt{W_f \cdot (t-1)}{}
        ).
    \end{equation}
    Indeed, $\mb{s}\mt{t}{i}$ contains both local and global information: information from features and previous attention weights can be considered as local since they are position-dependent; and information from the decoder hidden state can be seen as global since it is related to all characters recognized so far. 
    $\mb{W}_f$, $\mb{W}_j$ and $\mb{W}_h$ are weights of densely connected layers unifying the number of channels of the elements to be summed to $C_u=256$ ($\mb{W}_f \in \mathbb{R}^{C_f \times C_u}$, $\mb{W}_j \in \mathbb{R}^{C_j \times C_u}$ and $\mb{W}_h \in \mathbb{R}^{C_h \times C_u}$).
    
    \item Score $\mb{e}\mt{t}{i}$ are then computed for each row $i$:
    \begin{equation}
        \mb{e}\mt{t}{i} = \mb{W}_a \cdot \mb{s}\mt{t}{i},
    \end{equation}
    where $\mb{W}_a$ are weights of a densely connected layer reducing the channel axis to get a single value ($\mb{W}_a \in \mathbb{R}^{C_u \times 1}$).
    
    \item Attention weights are finally computed through softmax activation applied over these scores:
        \begin{equation}
        \mb{\alpha}\mt{t}{i} = \frac
        {\exp(\mb{e}\mt{t}{i})}
        {\displaystyle \sum_{k=1}^{H_f}\exp(\mb{e}\mt{t}{k})}.
    \end{equation}
\end{itemize}

We emphasize that, although the attention module produces a vertical focus $\mb{l}\mt{t}{}$, the line recognizer has a broad view of the input signal due to the size of the receptive field. Therefore, it makes the method robust to inclined or non-straight lines.

\subsubsection*{End-of-paragraph detection}
\label{section-stop-process}

One of the main intrinsic problems of paragraph recognition is the unknown number of text lines to be recognized. As the text recognition is processed sequentially, an end-of-paragraph detection is needed. To solve this problem, we compare three different approaches we named \emph{fixed-stop approach}, \emph{early-stop approach} and \emph{learned-stop approach}.

For each of these approaches, we used the \gls{ctc} loss to align the line prediction (probabilities lattice $\mb{p}\mt{t}{}$ of length $W_f$), with the corresponding line transcription $\mb{y}\mt{t}{}$. It involves the addition of the null symbol (also known as blank), leading to a new alphabet $\mset{A}' = \mset{A} \cup \{blank\}$. 

We also defined a stopping criterion at evaluation time to avoid infinite loops. It consists in a constant $l_\mathrm{max}$ large enough to cover the biggest paragraph of the dataset. This number can easily be set to match the datasets involved; in our case $l_\mathrm{max}=30$.

The \emph{fixed-stop approach} is the simplest way to handle the end-of-paragraph detection issue. The model iterates $l_\mathrm{max}$ times and stops, as proposed in \cite{Bluche2016}.  Additional fictive lines $[\mb{y}\mt{L+1}{}, ..., \mb{y}\mt{l_\mathrm{max}}{}]$ are added to the ground truth as empty strings. The idea is that the extra iterations will focus their attention on interlines, only predicting null symbols, to avoid recognizing the same line multiple times.
During training the loss is defined as follows:
\begin{equation}
    \mathcal{L}_{\mathrm{fs}} = \displaystyle \sum_{k=1}^{l_\mathrm{max}} \mathcal{L}_{\mathrm{CTC}}(\mb{p}\mt{k}{}, \mb{y}\mt{k}{}).
\end{equation}
This approach leads to an additional processing cost during training and evaluation due to the extra iterations needed. 

To alleviate this issue, we propose the \emph{early-stop approach}. The idea is that we can consider that the lines have all been predicted as soon as the current prediction is an empty line (only null symbols predicted). This way, we only have to add one additional line to the ground truth: $y_{L+1}$. The loss is then defined this way:
\begin{equation}
    \mathcal{L}_{\mathrm{es}} = \displaystyle \sum_{k=1}^{L+1} \mathcal{L}_{\mathrm{CTC}}(\mb{p}\mt{k}{}, \mb{y}\mt{k}{}).
\end{equation}

Finally, we propose the \emph{learned-stop approach} as a more elegant way to solve this problem. It consists in learning when to stop recognizing a new text line \textit{i.e.} to detect when the whole paragraph has been processed. This end-of-paragraph detection is performed at each iteration, computing the probability $\mb{d}\mt{t}{}$ to stop or to continue the recognition. More specifically, $\mb{d}\mt{t}{}$ determines whether $\mb{p}\mt{t}{}$ should be considered or not. This way, the model iterates $L+1$ times to learn to predict the end of the process. 

We decided to compute this probability from two elements: the multi-scale information $\mb{s}\mt{t}{}$ which brings some visual information about what has already been processed and what remains to be decoded, and the decoder hidden state $\mb{h}\mt{W_f \cdot t}{}$ which can contain information about what has already been recognized. Indeed, it is more likely to be the end of the paragraph if the last recognized character is a dot for example. 

To this end, we first have to collapse the vertical dimension of $\mb{s}\mt{t}{}$ for dimension matching purposes. This is carried out in the following way:
\begin{itemize}
    \item A one-dimensional convolutional layer with kernel size 5, stride 1 and zero padding is applied on $\mb{s}\mt{t}{}$.
    \item Adaptive Max Pooling is used to reduce the height to a fixed value of 15 (since input are of variable sizes).
    \item A densely connected layer pushes the vertical dimension to collapse, leading to $\mb{k}\mt{t}{} \in \mathbb{R}^{C_u}$.
\end{itemize}
    
Then, the produced tensor $\mb{k}\mt{t}{}$ is combined with $\mb{h}\mt{W_f \cdot t}{}$ through concatenation over the channel axis, leading to $\mb{b}\mt{t}{}$.

Finally, the dimension is reduced to 2 in order to compute the probabilities $\mb{d}\mt{t}{} \in \mathbb{R}^{2}$ through a densely connected layer of weights $\mb{W}_d \in \mathbb{R}^{(C_u+C_h) \times 2}$:
\begin{equation}
    \mb{d}\mt{t}{} = \mb{W}_d \cdot \mb{b}\mt{t}{}.
\end{equation}

This approach leads to the addition of a cross-entropy loss to the \gls{ctc} loss, applied to the decision probabilities $\mb{d}\mt{t}{}$. The corresponding ground truth $\mb{\delta}\mt{t}{}$ is one-hot encoded, deduced from the line breaks. The cross-entropy loss is defined as follows:
\begin{equation}
    \mathcal{L}_{\mathrm{CE}}(\mb{d}\mt{t}{}, \mb{\delta}\mt{t}{}) = -\sum_{i=1}^{2}\mb{\delta}\mt{t}{i}\log{\mb{d}\mt{t}{i}}.
\end{equation}
The final loss is then:
\begin{equation}
\mathcal{L}_\mathrm{ls} = \displaystyle \sum_{k=1}^{L} \mathcal{L}_{\mathrm{CTC}}(\mb{p}\mt{k}{}, \mb{y}\mt{k}{}) + \lambda \displaystyle \sum_{k=1}^{L+1} \mathcal{L}_{\mathrm{CE}}(\mb{d}\mt{k}{}, \mb{\delta}\mt{k}{}),
\end{equation}
where $\lambda$ is set to 1.

These three approaches are compared through experiments presented in Section \ref{section-exp-stop}.

\subsubsection{Decoder}
The decoder aims at recognizing a sequence of characters from the current line features $\mb{l}\mt{t}{}$, \textit{i.e.} a whole text line. To this end, we can process $\mb{l}\mt{t}{}$ as we do in standard line-level \gls{htr} approaches, since the vertical axis is already collapsed: $\mb{l}\mt{t}{}$ is a one-dimensional sequence of features. First, we apply a single \gls{lstm} layer with $C_h=256$ cells that outputs another representation of same dimension $\mb{r}\mt{t}{}$ which includes some context due to the recurrence over the horizontal axis. The \gls{lstm} hidden states $\mb{h}\mt{0}{}$ are initialized with zeros for the first line; they are maintained from one line to the next one to take advantage of the context at paragraph level. Then, a one-dimensional convolutional layer with kernel $1$ going from $256$ to $N+1$ channels is applied in order to produce $\mb{p}\mt{t}{}$, the \textit{a posteriori} probabilities of each character and the \gls{ctc} null symbol, for each of the $W_f$ frames. $N$ is the size of the character set. Best path decoding is used to get the final characters sequence. Successive identical characters and \gls{ctc} null symbols are removed through the \gls{ctc} decoding algorithm, leading to the final text. 
All the recognized text lines are concatenated with a whitespace character as separator to get the whole paragraph transcription $p_\mathrm{pg}$.

\subsection{Experimental conditions}
\label{section-experiment-cond}

The experimental conditions are nearly identical to those used to train the \gls{span}: same datasets (RIMES 2011, IAM, READ 2016), same data augmentation strategy, same post-processing (best path decoding, no external data or language model) and same line-level pre-training strategy if not stated otherwise.
The differences are as follows:
\begin{itemize}
\item We used an additional pre-processing step for the \gls{van}, we zero pad the input images to reach a minimum height of 480 px and a minimum width of 800 px when necessary. This assures that the minimum features width will be 100 and the minimum features height will be 15, which is required by the model as described previously.
\item We considered punctuation characters as words for the \gls{wer} computation, as in the READ 2016 competition \cite{READ2016}.
\end{itemize}

In addition, we propose a comparison with a three-step approach in the following. This way, in addition to the \gls{cer} and to the \gls{wer}, we used two other metrics to evaluate the segmentation stage: the \gls{iou} and the \gls{map} (detailed in Section \ref{section-metric-segmentation}). The segmentation is applied at pixel level with two classes, namely text and background. 
We compute the global \gls{iou} over a set of images by weighting the image \gls{iou} by its number of pixels.
We compute the \gls{map} for an image as the average of AP computed for  \gls{iou} thresholds between 50\% and 95\% with a step of 5\%. Image \gls{map}s are weighted by the number of pixels of the images to give the global mAP of a set.

Other metrics such as the number of parameters, the training time or the prediction time are useful to compare models. The training time is computed as the time to reach 90\% of the convergence. This is a more relevant value since tiny fluctuations can occur after numerous epochs.

\subsubsection*{Training details}
\label{section-training-details}
We used the Pytorch framework with the apex package to enable automatic mixed precision training thus reducing the memory consumption. We used the Adam optimizer for all experiments with an initial learning rate of $10^{-4}$. Trainings are performed during two days on a single \acrshort{gpu} Tesla V100 (32Gb). Models have been trained with mini-batch size of 16 for line-level \gls{htr} model and mini-batch size of 8 for the segmentation model and for the \gls{van}.

We use exactly the same hyperparameters from one dataset to another. Moreover, the model architecture is the same for each dataset: the last layer is the only difference since the datasets do not have the same character set size.

We used the line break annotations to split the ground truth into a sequence of text line transcriptions. This enables us to use the \gls{ctc} loss to train the \gls{van} through a line-by-line alignment between the recognized text lines and the ground truth line transcriptions.

Learned-stop training and prediction processes are respectively detailed in Algorithm \ref{alg:training} and \ref{alg:prediction}. The algorithms for the fixed-stop and early-stop approaches are similar. All the instructions related to the variables $\sigma_\mathrm{CE}$, $\mb{d}\mt{t}{}$ and $\mb{\delta}\mt{t}{}$ must be removed. Also, the for loop iterates $l_\mathrm{max}$ times instead of $L+1$ times for the fixed-stop approach.

\begin{algorithm}
\caption{VAN training process.}
\label{alg:training}
\SetKwInOut{Input}{input}
\Input{paragraph image $\mb{X}$, ground truth transcription $\mb{y}$ composed of $L$ lines $[\mb{y}\mt{1}{}, ..., \mb{y}\mt{L}{}]$}
$\mb{\alpha}\mt{0}{} = \mb{0}$, $\mb{h}\mt{0}{} = \mb{0}$, $\sigma_\mathrm{CTC}=0$, $\sigma_\mathrm{CE}=0$\tcp*[l]{Initialization}

$\mb{f} = \mathrm{Encoder}(\mb{X})$\tcp*[l]{Extract 2d features $\mb{f}$}
\tcp{For each line of the paragraph}
\For{t=1 \KwTo L+1}{
    \tcc{Compute attention weights $\mb{\alpha}\mt{t}{}$ for each features row to generate 1d line features $\mb{l}\mt{t}{}$. Also compute end-of-paragraph probabilities $\mb{d}\mt{t}{}$}
    $\mb{l}\mt{t}{}$,  $\mb{\alpha}\mt{t}{}$ $, \mb{d}\mt{t}{}$ $= \mathrm{Attention}(\mb{f}, \mb{\alpha}\mt{t-1}{}, \mb{h}\mt{W_f \cdot (t-1)}{})$;
    
    \tcc{Determine ground truth for end-of-paragraph detection task}
    \If{$t < L+1$}
    {
        \tcc{Compute character and CTC null symbol probabilities $\mb{p}\mt{t}{}$ for each frame of $\mb{l}\mt{t}{}$}
        $\mb{p}\mt{t}{}, \mb{h}\mt{W_f \cdot t}{} = \mathrm{Decoder}(\mb{l}\mt{t}{}, \mb{h}\mt{W_f \cdot (t-1)}{})$\;
        
        $\sigma_\mathrm{CTC} \mathrel{{+}{=}} \mathcal{L}_\mathrm{CTC}(\mb{p}\mt{t}{}, \mb{y}\mt{t}{})$\tcp*[l]{Compute CTC loss}
        
        $\mb{\delta}\mt{t}{} = \begin{pmatrix}
           0 \\
           1
        \end{pmatrix}$\;
    }
    \Else{
        $\mb{\delta}\mt{t}{} = \begin{pmatrix}
           1 \\
           0
        \end{pmatrix}$\;
    }
    
    $\sigma_\mathrm{CE} \mathrel{{+}{=}} \mathcal{L}_\mathrm{CE}(\mb{d}\mt{t}{}, \mb{\delta}\mt{t}{})$\tcp*[l]{Compute cross-entropy loss}
}
backward($\sigma_\mathrm{CTC}$ $+\sigma_\mathrm{CE}$ $)$\tcp*[l]{Backpropagation}
\end{algorithm}

\begin{algorithm}
\caption{VAN prediction process.}
\label{alg:prediction}
\SetKwInOut{Input}{input}
\Input{paragraph image $\mb{X}$}
$\mb{\alpha}\mt{0}{} = \mb{0}$, $\mb{h}\mt{0}{} = \mb{0}$, $\mb{d}\mt{1}{} = \begin{pmatrix}
   0 \\
   1
\end{pmatrix}$, $l_\mathrm{max}=30$, $t=1$, $p_\mathrm{pg}=$\textquote{ } \;

$\mb{f} = \mathrm{Encoder}(\mb{X})$\;
\While{$t \leq l_\mathrm{max} \And \argmax(\mb{d}\mt{t}{}) == 1$}
{
    $\mb{l}\mt{t}{}, \mb{\alpha}\mt{t}{}, \mb{d}\mt{t}{} = \mathrm{Attention}(\mb{f}, \mb{\alpha}\mt{t-1}{}, \mb{h}\mt{W_f \cdot (t-1)}{})$\;
    \If{$\argmax(\mb{d}\mt{t}{}) == 1$}
    {
        $\mb{p}\mt{t}{}, \mb{h}\mt{W_f \cdot t}{} = \mathrm{Decoder}(\mb{l}\mt{t}{}, \mb{h}\mt{W_f \cdot (t-1)}{})$\;
        $p_\mathrm{pg} = \concatenate(p_\mathrm{pg}, $\textquote{ }$, \mathrm{ctc\_decoding}(\mb{p}\mt{t}{}))$\;
    }
    $t = t + 1$\;
}

\end{algorithm}

We can summarize those processes as follows. First, the input images are pre-processed and augmented (at training time only) as described previously. Then, the encoder extracts features $\mb{f}$ from them. 
The attention module recurrently generates line features $\mb{l}\mt{t}{}$ until $l_\mathrm{max}$ is reached or, for the learned-stop approach, until $\mb{d}\mt{t}{}$ probabilities are in favor of stopping the process.
The decoder outputs character probabilities from each line features $\mb{l}\mt{t}{}$, which are then decoded through the \gls{ctc} algorithm, and merged together, separated by a whitespace character.
We finally get the whole paragraph transcription $p_\mathrm{pg}$.

\subsection{Experiments}
\label{section-experiments}
\renewcommand{\thefootnote}{\alph{footnote}}

This section is dedicated to the evaluation of the \gls{van} for paragraph recognition. We show that the \gls{van} reaches state-of-the-art results on each dataset. We study the need for pre-training on isolated text lines of the target dataset. We show that this can be avoided by using a pre-trained \gls{van} on another dataset. It enables the model to be trained on the target dataset with the paragraph-level ground truth only, without the need for line segmentation ground truth, which is a considerable practical advantage. We study the different stopping strategies on the IAM dataset. We also provide a visualization of the attention process.

\subsubsection{Comparison with state-of-the-art paragraph-level approaches}
\label{section-exp-pg-sota}
The results presented in this section are given for the \gls{van}, with pre-training on line images and using the learned-stop strategy. The following comparisons are made with approaches under similar conditions, \textit{i.e.} without the use of external data (to model the language for example) and at paragraph level.

Comparative results with state-of-the-art approaches on the RIMES 2011 dataset are given in Table \ref{table:van-rimes-pg}. The \gls{van} achieves better results on the test set compared to the other approaches with a \gls{cer} of 1.91\% and a \gls{wer} of 6.72\%. 

\begin{table}[!h]
    \caption{Recognition results  of the VAN and comparison with paragraph-level state-of-the-art approaches on the RIMES 2011 dataset.}
    \centering
    \resizebox{0.8\linewidth}{!}{
    \begin{threeparttable}[b]
        \begin{tabular}{ l c c c c c}
        \hline
        \multirow{2}{*}{Architecture} & \gls{cer} (\%) & \gls{wer} (\%) & \gls{cer} (\%) & \gls{wer} (\%) & \multirow{2}{*}{\# Param.}\\ 
        & valid & valid & test & test\\
        \hline
        \hline
        \cite{Bluche2016} \acrshort{cnn}+\acrshort{mdlstm}\tnote{a} & 2.5 & 12.0 & 2.9 & 12.6 & \\
        \cite{Wigington2018} \acrshort{rpn}+\acrshort{cnn}+\acrshort{blstm}+\acrshort{lm} & & & 2.1 & 9.3 & \\ 
        \cite{Coquenet2021} \acrshort{span} (\acrshort{fcn}) & 3.56 & 14.29 & 4.17 & 15.61 & 19.2 M\\
        \cite{Coquenet2022} \acrshort{van} (\acrshort{fcn}+\acrshort{lstm})\tnote{a} & 1.83 & 6.26 & \textbf{1.91} & \textbf{6.72} & 2.7 M \\
        \hline
        \end{tabular}
        \begin{tablenotes}
            \item[a] With line-level attention.
        \end{tablenotes}
        \end{threeparttable}
    }
    \label{table:van-rimes-pg}
\end{table}

Table \ref{table:van-iam-pg} shows the results compared to the state of the art on the IAM dataset. As one can see, once again the \gls{van} also achieves new state-of-the-art results with a \gls{cer} of 4.45\% and a \gls{wer} of 14.55\% on the test set. One can notice that we use more than 6 times fewer parameters than in \cite{Yousef2020} and \cite{Coquenet2021} with 2.7 M compared to 16.4 M and 19.2 M.

\begin{table}[!h]
    \caption{Comparison of the VAN with the state-of-the-art approaches at paragraph level on the IAM dataset.}
    \centering
    \resizebox{0.9\linewidth}{!}{
    
    \begin{threeparttable}[b]
        \begin{tabular}{ l c c c c c}
        \hline
        \multirow{2}{*}{Architecture} & \gls{cer} (\%) & \gls{wer} (\%) & \gls{cer} (\%) & \gls{wer} (\%) & \multirow{2}{*}{\# Param.}\\ 
        & valid & valid & test & test\\
        \hline
        \hline
        \cite{Bluche2017b} \acrshort{cnn}+\acrshort{mdlstm}\tnote{a} & &  & 16.2 &  & \\
        \cite{Bluche2016} \acrshort{cnn}+\acrshort{mdlstm}\tnote{b} & 4.9 & 17.1 & 7.9 & 24.6 & \\
        \cite{Carbonell2019} \acrshort{rpn}+\acrshort{cnn}+\acrshort{blstm}\tnote{c} & 13.8 & & 15.6 & & \\ 
        \cite{Chung2020} \acrshort{rpn}+\acrshort{cnn}+\acrshort{blstm} & & & 8.5 & &  \\
        \cite{Wigington2018} \acrshort{rpn}+\acrshort{cnn}+\acrshort{blstm}+\acrshort{lm} & & & 6.4 & 23.2 & \\
        \cite{Yousef2020}  \acrshort{gfcn} & & & 4.7 & & 16.4 M \\
        \cite{Coquenet2021} \acrshort{span} (\acrshort{fcn})  & 3.57 & 15.31 & 5.45 & 19.83 & 19.2 M\\
        \cite{Coquenet2022} \acrshort{van} (\acrshort{fcn}+\acrshort{lstm})\tnote{a} & 3.02 & 10.34 & \textbf{4.45} & \textbf{14.55} & 2.7 M\\
        \hline
        \end{tabular}
        
        \begin{tablenotes}
            \item[a] With character-level attention.
            \item [b] With line-level attention.
            \item [c] Results are given for page level.
        \end{tablenotes}
        \end{threeparttable}
    }
    \label{table:van-iam-pg}
\end{table}

To our knowledge, except for the \gls{span}, there are no results reported in the literature on the READ 2016 dataset at paragraph or page level. The recognition results are presented in Table \ref{table:van-read-pg}. The \gls{van} reaches better results than the \gls{span} with a \gls{cer} of 3.59\% and a \gls{wer} of 13.94\%.
\begin{table}[!h]
    \caption{Comparison of the VAN with the state-of-the-art approach for the READ 2016 dataset at paragraph level.}
    \centering
    \resizebox{0.8\linewidth}{!}{
    \begin{tabular}{ l c c c c c}
    \hline
    \multirow{2}{*}{Architecture} & \gls{cer} (\%) & \gls{wer} (\%) & \gls{cer} (\%) & \gls{wer} (\%) & \multirow{2}{*}{\# Param}\\ 
    & validation & validation & test & test\\
    \hline
    \hline
    \cite{Coquenet2021} \acrshort{span} (\acrshort{fcn}) & 5.09 & 23.69 & 6.20 & 25.69 & 19.2 M\\
    \cite{Coquenet2022} \acrshort{van} (\acrshort{fcn}+\acrshort{lstm})\tnote{a} & 3.71 & 15.47 & \textbf{3.59} & \textbf{13.94} & 2.7 M\\
    \hline
    \end{tabular}
    }
    \label{table:van-read-pg}
\end{table}

The proposed \acrlong{van} achieves new state-of-the-art results on these three different datasets. 
For fair comparison with the other competitive approaches of the literature we highlight some comparative features in Table \ref{table:sota-details}. In this table, the approaches are analyzed with regards to the following features: 1- use of an explicit text region segmentation process 2- minimum segmentation level used (whether it is for pre-training, data augmentation or training itself) 3- number of hyperparameters adapted from one dataset to another (except for the last decision layer which is dependent of the alphabet size) 4- use of curriculum learning 5- use of data augmentation. 
Curriculum learning, which consists in progressively increasing the number of lines in the input images, can be considered as data augmentation (crop technique). It is important to note that the use of line segmentation ground truth during training is costly due to the human effort involved to create them; this is even more costly for word segmentation. Data augmentation, for its part, does not require any human effort.

\begin{table}[!h]
    \caption{Training details of state-of-the-art approaches.}
    \centering
    \resizebox{0.8\linewidth}{!}{
    \begin{tabular}{ l c c c c c }
    \hline
    \multirow{2}{*}{Approach} & Explicit & Min. segment. & Hyperparam. & Curriculum  & Data \\ 
    & segmentation & label used & adaptation & learning & augmentation \\
    \hline
    \cite{Carbonell2019} \acrshort{rpn}+\acrshort{cnn}+\acrshort{blstm}& \cmark & Word & \xmark & \xmark & \xmark\\
    \cite{Chung2020} \acrshort{rpn}+\acrshort{cnn}+\acrshort{blstm} & \cmark & Word & \xmark & \xmark & \cmark\\
    \cite{Wigington2018} \acrshort{rpn}+\acrshort{cnn}+\acrshort{blstm} & \cmark & Line & \xmark & \xmark & \cmark\\
    \cite{Yousef2020}  \acrshort{gfcn} & \xmark& Paragraph & 6 &\xmark  & ? \\
    \cite{Bluche2016} \acrshort{cnn}+\acrshort{mdlstm} & \xmark & Line & \xmark & \cmark & \cmark \\
    \cite{Bluche2017b} \acrshort{cnn}+\acrshort{mdlstm} & \xmark& Line & \xmark & \cmark & \cmark \\
    \cite{Coquenet2021} \acrshort{span} (\acrshort{fcn}) & \xmark & Line  & \xmark & \xmark & \cmark\\
    \cite{Coquenet2022} \acrshort{van} (\acrshort{fcn}+\acrshort{lstm}) & \xmark & Line  & \xmark & \xmark & \cmark\\
    \hline
    \hline
    \end{tabular}
    }
    \label{table:sota-details}
\end{table}

The approach proposed in \cite{Yousef2020} is the only one that does not use any segmentation label neither at line nor at word level. However, it is also the only one that requires the architecture to be adapted to each dataset. More specifically, the height and width of the input and of two intermediate upsampling layers are tuned for each dataset. It means that those 6 hyperparameters must be tuned manually to reach the performance reported, which makes the architecture not generic at all. On the contrary, the \gls{van} architecture remains the same for every dataset in all the experiments. Another important point is that the use of line-level segmentation ground truth is not inherent to our approach, as it was only introduced as a pre-training step. As we will see in the following section, pre-training the model at line level is not mandatory. Indeed, similar performance can be obtained with paragraph-level cross-dataset pre-training, without the need for line segmentation ground truth from the target dataset.

\subsubsection{Impact of pre-training}
\label{section-pretraining}

In this section, we study the impact of pre-training on the \gls{van}, using the learned-stop approach. The dataset used for pre-training is named source dataset, and the dataset on which we want to evaluate the model is named target dataset. We conducted three kinds of experiments. They are intended to evaluate the requirement of line-level segmentation labels for the target dataset to reach the best performance. 

A first training strategy consists in training the architecture at paragraph level directly on the target dataset, from scratch. 
A second strategy consists in training the architecture at paragraph level on the target dataset, with pre-trained weights for the encoder and the convolutional layer of the decoder, as detailed in Section \ref{section-training-details}. In this case, pre-training is carried out on the isolated text line images of the target dataset, prior to train the \gls{van} at paragraph level. Here, the source dataset and the target dataset are the same. This pre-training approach is referred to as line-level pre-training.
A third training strategy consists in training the architecture at paragraph level on the target dataset with weights that are initialized with those of another \gls{van}. This other \gls{van} is trained on a source dataset, with the second strategy. The idea is not to use any segmentation label from the target dataset. In this case the training strategy is referred to as cross-dataset pre-training.

\begin{figure}[htbp!]
\centering
\includegraphics[width=0.7\linewidth]{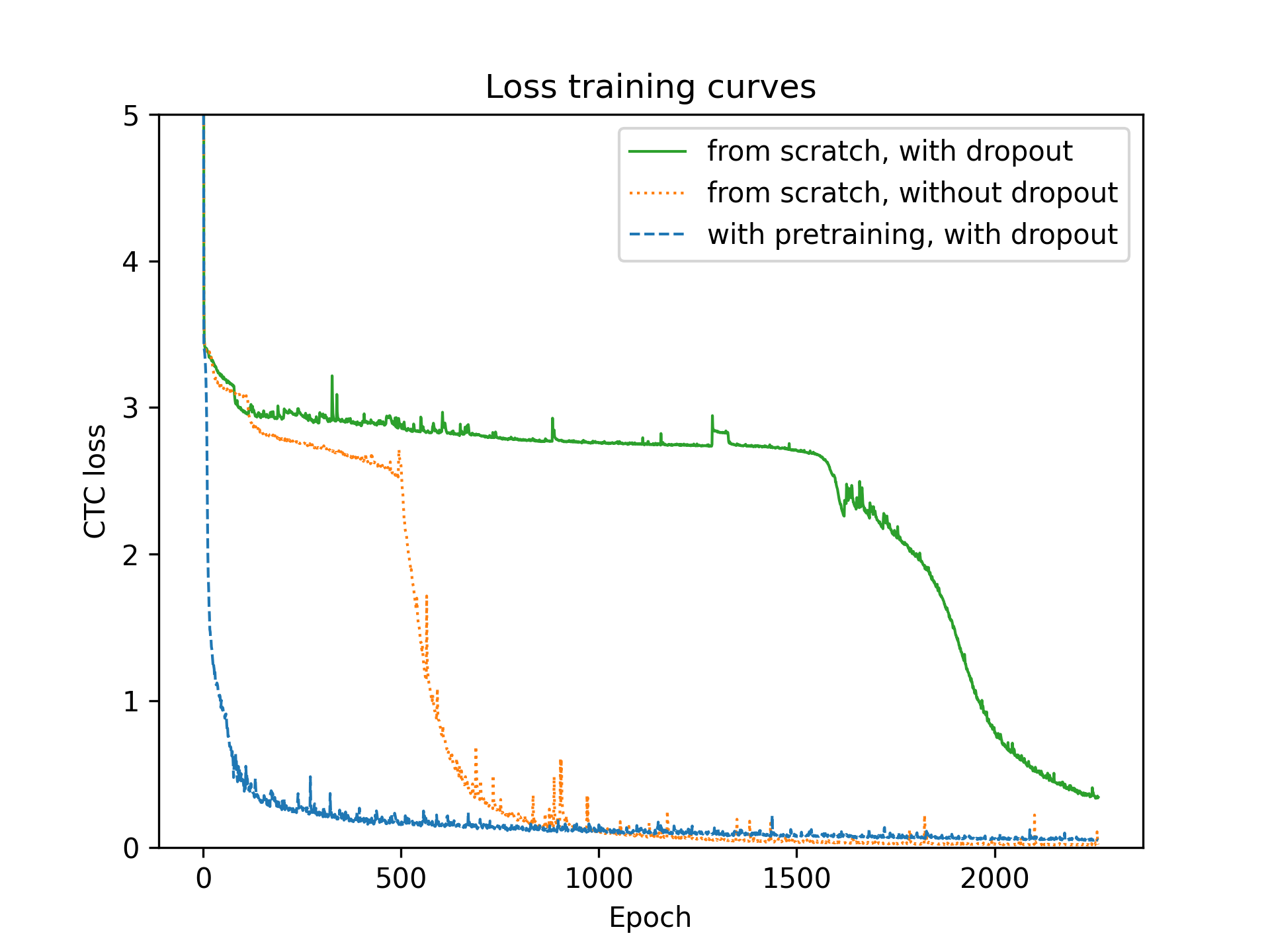}
        \caption{CTC training loss curves comparison for the VAN, with and without pre-training, on the IAM dataset.}
        \label{fig:curves-pretraining}

\end{figure}

Figure \ref{fig:curves-pretraining} shows the evolution of the \gls{ctc} loss on the IAM dataset during training from scratch and training with line-level pre-training. We also highlight the impact of dropout when training from scratch. The recognition performance is given on the test set in Table \ref{table:va-pretrain}. As was expected, we can clearly notice that training the model from scratch is feasible, but it takes much time to converge and it comes at the cost of an important increase of the \gls{cer}, from 4.45\% up to 7.06\%. When training from scratch, the use of dropout highly slows down the convergence but it leads to a lower \gls{cer} of 7.06\% compared to 8.06\%. This experiment shows us that state-of-the-art results are reached with line-level pre-training.

\begin{table}[!h]
    \caption{Impact of the pre-training on lines for the VAN. Results are given on the test set of the IAM dataset.}
    \centering
    \resizebox{0.7\linewidth}{!}{
    \begin{tabular}{ l c c c c }
    \hline
    pre-training  & Dropout & \gls{cer} (\%) & \gls{wer} (\%) & Training time\\ 
    \hline
    \hline
    \xmark & \xmark & 8.06 & 25.38 & 1.39 d \\
    \xmark & \cmark & 7.06 & 22.65 & 1.77 d \\
    \cmark & \cmark & \textbf{4.45} & \textbf{14.55} & 0.63 d \\
    \hline
    \end{tabular}
    }
    \label{table:va-pretrain}
\end{table}

Table \ref{table:cross} reports the recognition performance of the third training strategy, using cross-dataset pre-training. In this table we report the performance of the \gls{van} on the three datasets when pre-trained on one of the two other datasets. One can notice that this pre-training strategy performs almost similarly as line-level pre-training for every dataset, but without using any segmentation label of the target dataset. This is especially true with the two similar datasets RIMES 2011 and IAM, for which the two pre-training strategies reach very similar \gls{cer}: 1.97\% compared to 1.91\% for RIMES 2011 and 4.55\% compared to 4.45\% for IAM.  
Although cross-dataset pre-training is a bit less efficient on READ 2016, it still leads to competitive results. This may be explained by the differences between this dataset (RGB color encoding, historical manuscripts and language) and the other datasets (RIMES 2011 or IAM) on which pre-training is performed.

\begin{table}[!h]
    \caption{Comparison between cross-dataset pre-training and line-level pre-training for the VAN. Results are given on the test sets.}
    \centering
    \resizebox{0.7\linewidth}{!}{
    \begin{tabular}{ l c c c }
    \hline
     \multirow{2}{*}{Source dataset}& RIMES 2011 & IAM & READ 2016\\
     & \gls{cer} (\%) & \gls{cer} (\%) & \gls{cer} (\%) \\
    \hline
    \hline
    \textbf{Cross-dataset pre-training}\\
    RIMES 2011 & \xmark & 4.55 & 4.08 \\
    IAM & 1.97 & \xmark & 4.14\\
    READ 2016 & 2.36 & 5.20 & \xmark \\
    \\
    \textbf{Line-level pre-training}\\
    Target dataset & \textbf{1.91} & \textbf{4.45} & \textbf{3.59}\\
    
    \hline
    \end{tabular}
    }
    \label{table:cross}
\end{table}

These experiments demonstrate the importance of pre-training for the \gls{van}. However, we show that we can alleviate the need for line-level segmentation label of the target dataset through cross-dataset pre-training. We assume that the important pre-training effect is mainly due to the attention mechanism. Indeed, the authors of \cite{Bluche2016,Bluche2017b}, who also proposed attention-based models, used curriculum learning to tackle convergence issues. We assume this is due to the direct relation between recognition and implicit segmentation (through soft attention). When the encoder is pre-trained, the \gls{van} only has to learn the attention module and the decoder, as the pre-trained features may contain all the information needed for the recognition of the characters. Considering random distribution of the attention weights over the vertical axis at first, training will progressively increases values of weights where the features correspond to the correct characters. But when the encoder is randomly initialized, the features extraction must also be learned, which slows the whole training. The use of dropout makes this phenomenon even worse, but it is essential in this architecture to avoid overfitting. It seems that cross-dataset pre-training skips this issue since the attention mechanism is already learned and the encoder only needs to be fine tuned.

Finally, we can notice that without introducing any pre-training, the \gls{van} architecture is able to converge using the paragraph annotations only, but it is not competitive anymore compared to the state of the art.
As a preliminary conclusion, we can highlight the capacity of the \gls{van} to achieve state-of-the-art performance without no need to adapt the architecture to each dataset considered. Moreover, cross-dataset pre-training allows to reach similar results compared to line-level pre-training, without the need to use any line-level segmentation ground truth from the target dataset.

\subsubsection{Learning when to stop}
\label{section-exp-stop}
We now compare the three stopping strategies mentioned in Section \ref{section-stop-process}, namely fixed-stop, early-stop and learned-stop approaches.

Performance evaluation of these three methods is given in Table \ref{table:va-stop} for comparison purpose. Line-level pre-training is used for each approach, as detailed in Section \ref{section-training-details}. 
We define $d_\mathrm{mean}$, as the average of the absolute values of the differences between the actual number of lines in the image $n_\mathrm{i}$ and the number of recognized lines $n_\mathrm{r}$. This metric is used to evaluate the efficiency of the early-stop and learned-stop approaches. For $K$ images in the dataset:

\begin{equation}
    d_\mathrm{mean} = \frac{1}{K}\displaystyle \sum_{k=1}^{K} |n_{\mathrm{i}_k}-n_{\mathrm{r}_k}|.
\end{equation}

As one can see, equivalent \gls{cer} and \gls{wer} are obtained for each stopping strategy. Moreover, the prediction time is not significantly impacted since data formatting, tensor initialization and encoder-related computations take much longer than the recurrent process, which is made up of only a few layers at each iteration.

\begin{table}[!h]
    \caption{Comparison between fixed-stop, early-stop and learned-stop approaches with the VAN on the test set of the IAM dataset.}
    \centering
    \resizebox{0.8\linewidth}{!}{
    \begin{tabular}{ l c c c c c}
    \hline
    Stop method  & \gls{cer} (\%) & \gls{wer} (\%) & Train. time & Pred. time &$d_\mathrm{mean}$\\ 
    \hline
    \hline
    Fixed & \textbf{4.41} & 14.69 & 1.20 d & 33 ms & 20.32 \\
    Early & \textbf{4.41} & \textbf{14.39} & 0.41 d & 32 ms & 0.02 \\
    Learned & 4.45 & 14.55 & 0.63 d & 32 ms & 0.03 \\
    \hline
    \end{tabular}
    }
    \label{table:va-stop}
\end{table}

\begin{figure}[htbp!]
\centering
\includegraphics[width=0.7\linewidth]{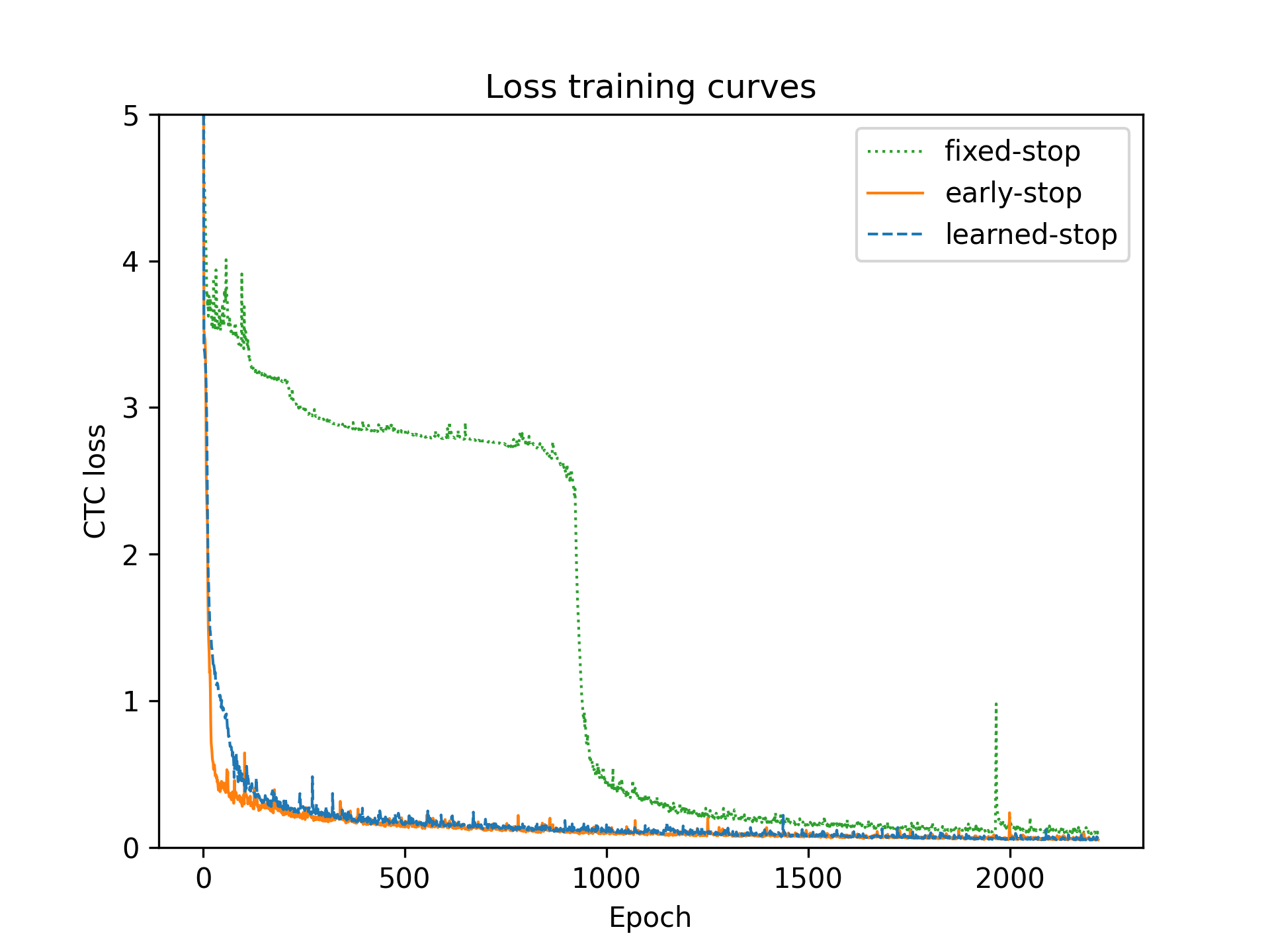}
        \caption{CTC training loss curves comparison for the VAN for each stopping approach on the IAM dataset.}
        \label{fig:loss-stop}

\end{figure}

Figure \ref{fig:loss-stop} illustrates the evolution of the \gls{ctc} loss for the three approaches. One should keep in mind that the comparison is biased since the approaches do not iterate the same number of times for a same example. However, one can clearly notice that the early-stop and learned-stop approaches  converge similarly, in contrast to the fixed-stop approach, which requires far more epochs to converge. But in the end, they reach almost identical \gls{cer}.

\begin{figure}[htbp!]
\centering
\includegraphics[width=0.7\linewidth]{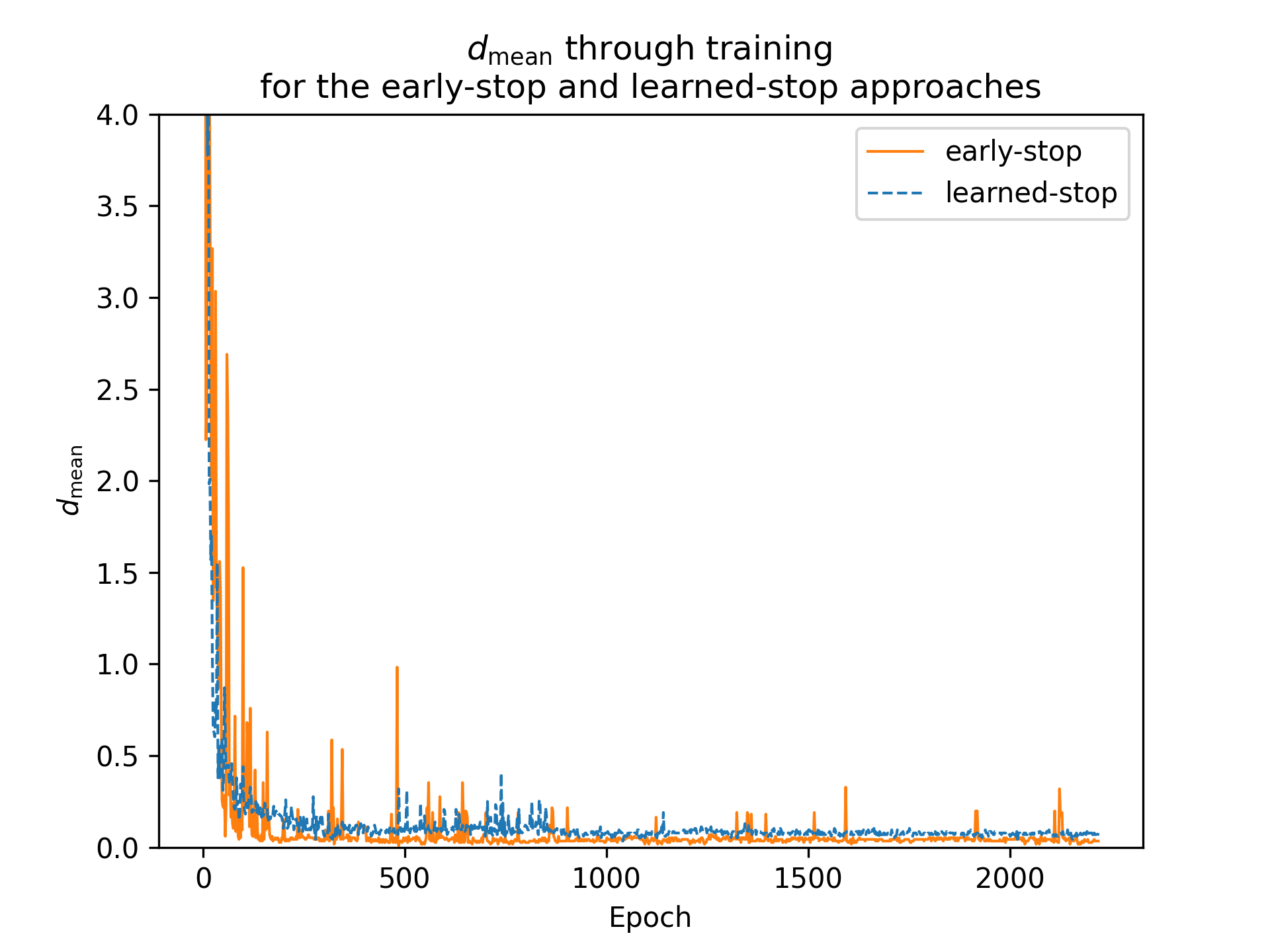}
        \caption{Comparison of the evolution of $d_\mathrm{mean}$ (the mean difference between the true and estimated number of lines in the image ) on the validation set of IAM dataset for the early-stop and learned-stop approaches.}
        \label{fig:diff-stop}
\end{figure}

Figure \ref{fig:diff-stop} compares the $d_\mathrm{mean}$ for the early-stop and learned-stop approaches on the IAM validation dataset. For visibility, the fixed-stop approach curve, which is a plateau at $d_\mathrm{mean}=20.32$, is not depicted in this figure. The learned-stop approach leads to a faster convergence of $d_\mathrm{mean}$ compared to the early-stop approach, whose curve is less stable. It results in a $d_\mathrm{mean}$ of 0.03 for the learned-stop approach and 0.02 for the early-stop approach for the test set of IAM.

With the learned-stop approach, the model successfully learns both tasks: it recognizes text lines with a state-of-the-art \gls{cer} of 4.45\% and it determines when to stop with a high precision since the $d_\mathrm{mean}$ on the test set is only 0.03. It means that, on average, one line is missed or processed twice every 33 paragraph images.
We choose to keep this stopping strategy since it slightly improves the stability of the convergence through the epochs while achieving nearly identical results.

\subsection{Visualization of the vertical attention }
\begin{figure}[htbp!]
\centering
    \begin{subfigure}[t]{0.32\textwidth}
    \includegraphics[width=\textwidth, height=3cm, frame]{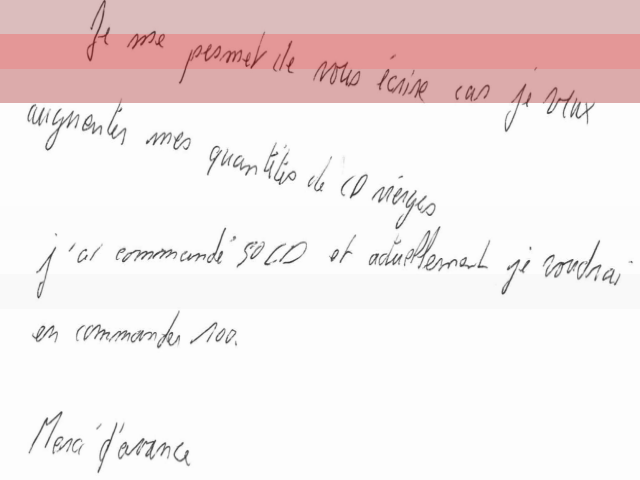}    
    \caption*{1: Je me permet  de vous écrire ca\textbf{s} je v\textbf{M}ux}
    \end{subfigure}
    \hfill
    \begin{subfigure}[t]{0.32\textwidth}
    \includegraphics[width=\textwidth, height=3cm, frame]{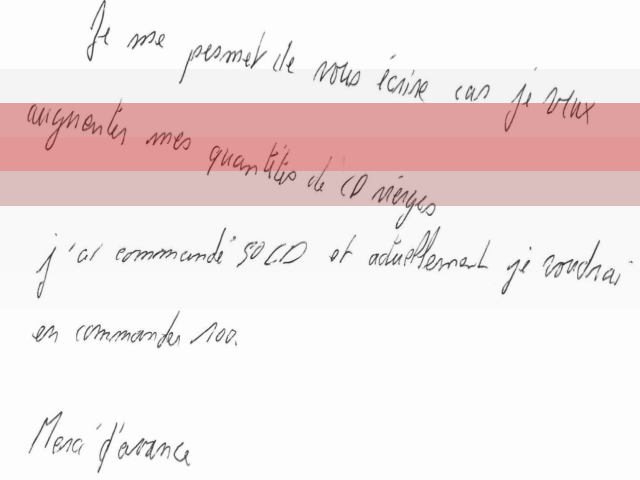}    
     \caption*{2: augmente\textbf{s} mes quantités de CD vierges}
    \end{subfigure}
    \hfill
    \begin{subfigure}[t]{0.32\textwidth}
    \includegraphics[width=\textwidth, height=3cm, frame]{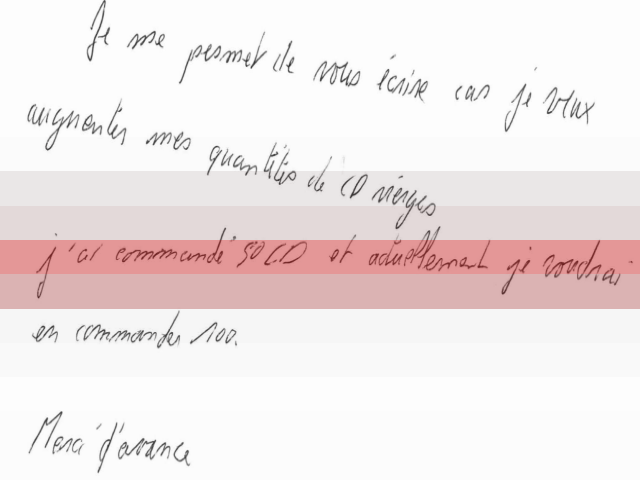}   
     \caption*{3: j'ai commandé 50 CD et a\textbf{d}uellement je voudrai}
    \end{subfigure}
    
    \par\bigskip

    \begin{subfigure}[t]{0.49\textwidth}
    \centering
    \includegraphics[width=0.64\textwidth, height=3cm, frame]{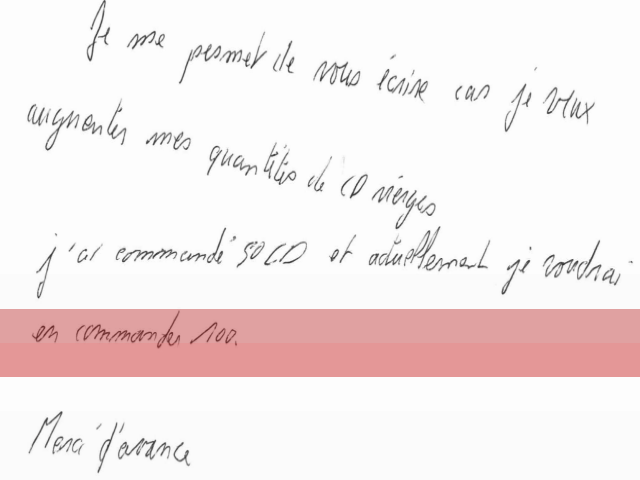}   
     \caption*{4: en commander 100.}
    \end{subfigure}
    \hfill
    \begin{subfigure}[t]{0.49\textwidth}
    \centering
    \includegraphics[width=0.64\textwidth, height=3cm, frame]{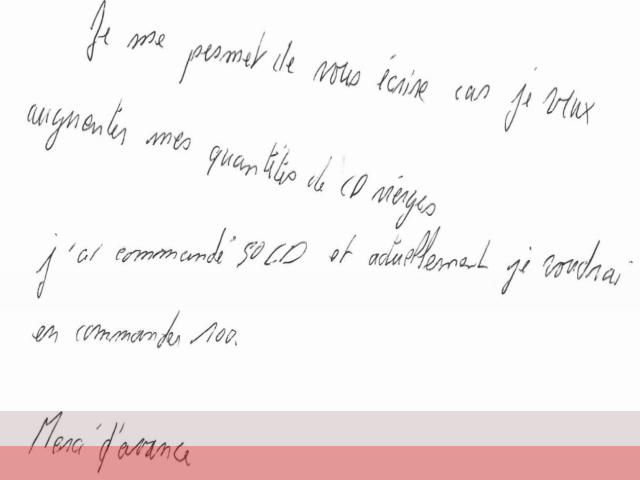}   
    \caption*{5: Merci d'avance}
    \end{subfigure}
    
    \caption{Attention weights visualization on a sample of the RIMES 2011 validation set. Recognized text is given for each line and errors are shown in bold.}
    \label{fig:viz-rimes}
\end{figure}

Figure \ref{fig:viz-rimes} shows the processing steps of an image from the RIMES 2011 validation set with a complex layout. Images represent the attention weights of the 5 attention iterations, each one recognizing a line. The intensity of the weights is encoded with the transparency of the red color. Given that attention weights are only computed along the vertical axis, the intensity is the same for every pixel at the same vertical position. Attention weights are rescaled to fit the original image height; indeed, attention weights are originally computed for the features height, which is 32 times smaller. The recognized text lines are given for each iteration, below the image. 

As one can see, the \gls{van} has learned the reading order, from top to bottom. The attention weights clearly focus on text lines following this reading order. Attention weights focus mainly on one features line, with smaller weights for the adjacent vertical positions. 

One can notice that, sometimes, the focus is not perfectly centered on the text line. This may be due to rescaling, but this can also be a normal behavior due to the large size of the receptive field, which enables to manage slightly inclined lines. The second iteration shows this phenomenon very well with only one misrecognized character on an inclined line. 
Furthermore, one can notice that the attention is less sharp when the layout is more complex between two successive lines, as in the third image, but it does not disturb the recognition process however.

\begin{figure}[htbp!]
\centering
    \begin{subfigure}[t]{0.32\textwidth}
    \includegraphics[width=\textwidth, height=3cm, frame]{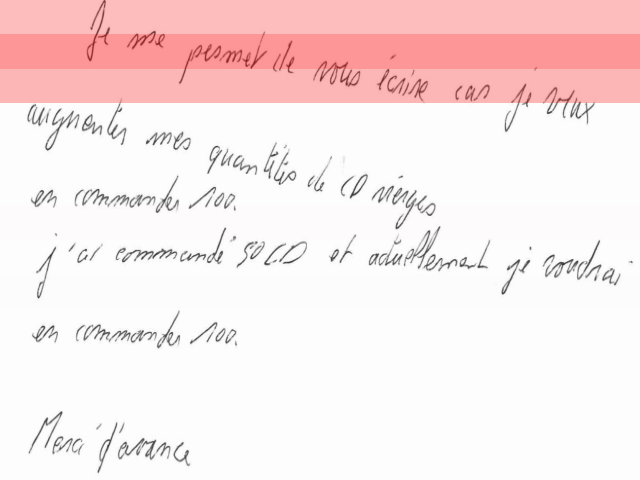}    
    \caption*{1: Je me permet  de vous écrire car je v\textbf{M}ux}
    \end{subfigure}
    \hfill
    \begin{subfigure}[t]{0.32\textwidth}
    \includegraphics[width=\textwidth, height=3cm, frame]{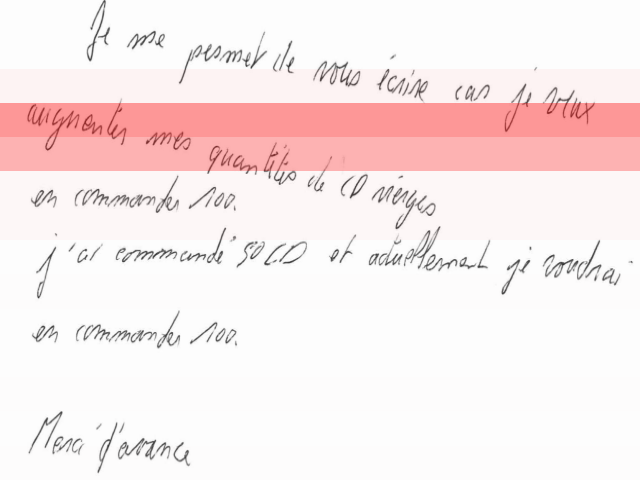}    
     \caption*{2: augmente\textbf{s} mes quantités \textbf{d'is vièrire cas je MuX}}
    \end{subfigure}
    \hfill
    \begin{subfigure}[t]{0.32\textwidth}
    \includegraphics[width=\textwidth, height=3cm, frame]{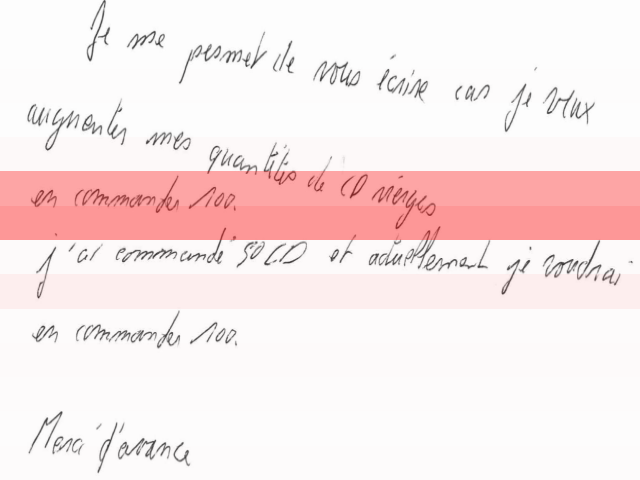}   
     \caption*{3: en commander 100.}
    \end{subfigure}
    
    \par\bigskip

    \begin{subfigure}[t]{0.32\textwidth}
    \centering
    \includegraphics[width=\textwidth, height=3cm, frame]{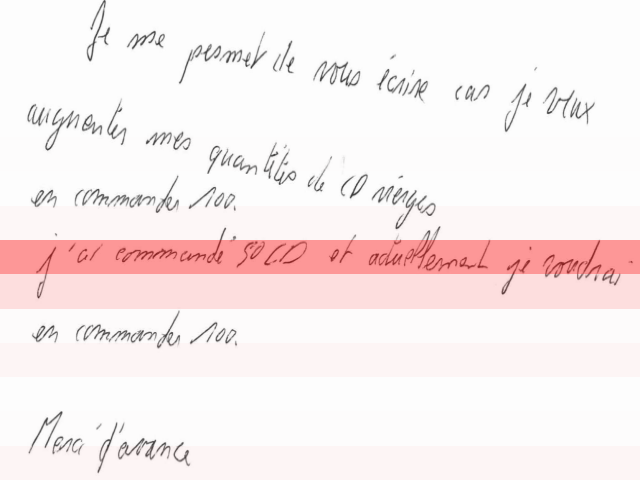}   
     \caption*{4: J'ai commandé 50 CD et a\textbf{d}uellement je voudrai}
    \end{subfigure}
    \hfill
    \begin{subfigure}[t]{0.32\textwidth}
    \centering
    \includegraphics[width=\textwidth, height=3cm, frame]{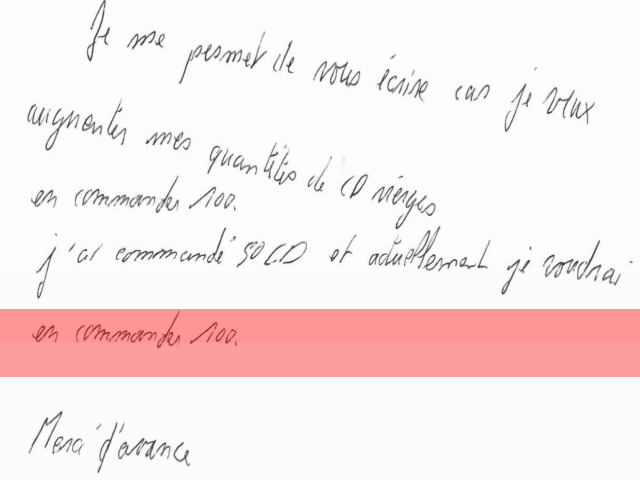}   
    \caption*{5: en commander 100.}
    \end{subfigure}
    \hfill
    \begin{subfigure}[t]{0.32\textwidth}
    \includegraphics[width=\textwidth, height=3cm, frame]{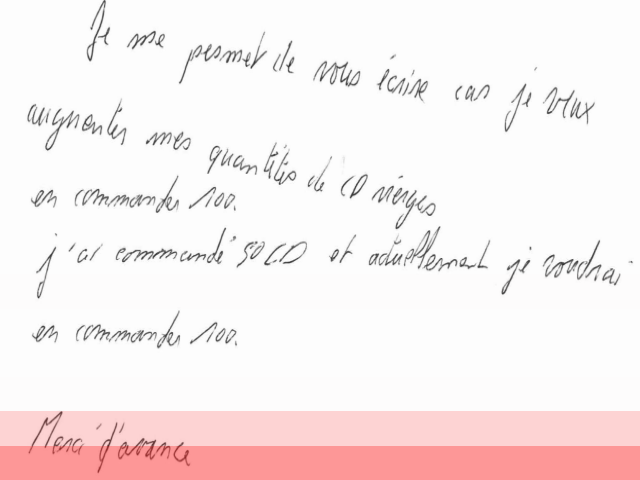}   
     \caption*{6: Merci d'avance}
    \end{subfigure}
    
    \caption{Attention weights visualization on a synthetic sample of the RIMES 2011 validation set. Recognized text is given for each line and errors are shown in bold.}
    \label{fig:viz-rimes2}
\end{figure}

One should note that processing inclined text lines is only possible when the lines, although inclined, do not share the same vertical position. The \gls{van} cannot handle the case where lines would overlap vertically, because the attention weights would mix the two lines in this case. This is illustrated on Figure \ref{fig:viz-rimes2} with a fake input image, which consists in the previous image in which we duplicated a text line.

As one can see, some text is predicted twice (“écrire car je veux”) and other is ignored (“de CD vierges”). But one can note that only the text part which includes overlaps in horizontal projection is altered: it does not affect the quality of the recognition for the remaining text.

So far, we have provided strong results in favor of the \acrlong{van} by achieving state-of-the-art performance on three datasets. We also evaluated the need for pre-training and the efficiency of the stopping strategies we propose.

\subsection{Additional experimental studies}
\label{section-additional-exp}

We now provide additional results which show the superiority of the \gls{van} on many criteria compared with a standard line-level three-step approach (line segmentation followed by ordering and recognition).
We also highlight the positive contribution of the \gls{van} applied to paragraph images compared with the single-line recognition approaches.
Finally, we highlight the positive effect of the proposed dropout strategy (Diffused Mix Dropout) compared to standard ones.

\subsubsection{Comparison with the standard line-level three-step approach}
In this section, we compare the \gls{van} with the standard line-level three-step approach on the IAM dataset. In this respect, two different models are used for the line-level segmentation and recognition stages; the ordering stage is performed through a rule-based algorithm. Both models are trained separately and they do not use any pre-training strategy since they are already working at line level.

\begin{figure}[htbp!]
\centering
\includegraphics[width=\textwidth]{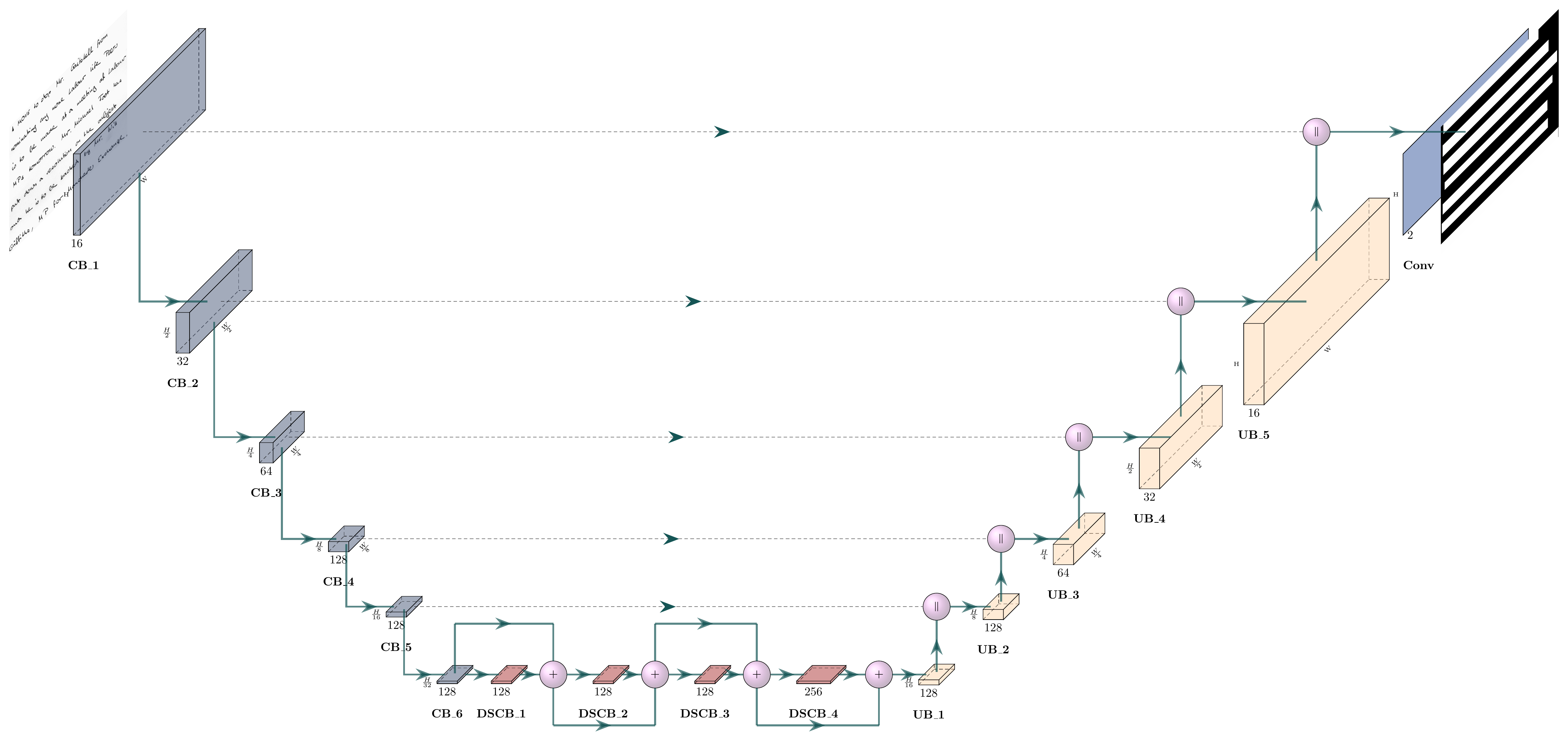}
\caption{U-net architecture for text line segmentation. }
\label{fig:van-unet}
\end{figure}

The line segmentation model, depicted in Figure \ref{fig:van-unet} follows a U-net shape architecture and is based on the \gls{van} \gls{fcn} encoder. Indeed, the features $\mb{f}$ are successively upsampled to match the the shape of the feature maps of CB\_5, CB\_4, CB\_3, CB\_2 and CB\_1 (Figure \ref{fig:encoder-overview}). This upsampling process is handled by Upsampling Blocks (UB). UB consists in \gls{dsc} layer followed by \gls{dsc} Transpose layer and instance normalization. This block also includes DMD layers. Each UB output is concatenated with the feature maps from its corresponding CB. A final convolutional layer outputs only two feature maps, to classify each pixel of the original image between text and background. 
The  line-level \gls{htr} model corresponds to the one used for the \gls{van} pre-training (Figure \ref{fig:span-line-ocr}).

The line segmentation model is trained on the paragraph images at pixel level with ground truth of line bounding boxes. It is trained with the cross-entropy loss. The line-level \gls{htr} model is trained on the isolated text line images with the \gls{ctc} loss. We used a mini-batch size of 8 for the segmentation task and of 16 for the \gls{htr}.

We now detail the three steps of this approach. In a first step, paragraphs are segmented into lines: 
\begin{itemize}
    \item Ground truth bounding boxes are modified in order to avoid overlaps: we divide their height by 2.
    \item A paragraph image is given as input of the network.
    \item A 2-class pixel segmentation (text or background) is output from the network.
    \item Adjacent text-labeled pixels are grouped to form connected components.
    \item Bounding boxes are created as the smallest rectangles containing each connected component; their height is multiplied by 2.
    \item The input image is cropped using those bounding boxes to generate line images.
\end{itemize}

In a second step, the line images are ordered given their vertical position, from top to bottom.
The last stage consists in recognizing the line images with the line-level \gls{htr} model, trained on the IAM dataset. The recognized text lines are concatenated to compute the \gls{cer} and the \gls{wer} at paragraph level.

The performances of both tasks (segmentation and recognition) taken separately and together are shown in Table \ref{table:seg-reco}. As one can see, the results are good for both tasks separately: we get 81.51\% for the \gls{iou} and 85.09\% for the \gls{map} concerning the  text line segmentation task; and a \gls{cer} of 5.01\% and a \gls{wer} of 16.49\% for the line-level \gls{htr}. 
However, when we take the output of the segmentation as input for the \gls{htr} model, it leads to a \gls{cer} increase of 1.54 points. Indeed, the errors of the segmentation stage induce recognition errors.

\begin{table}[!h]
    \caption{Results of the three-step approach on the test set of IAM.}
    \centering
    \resizebox{0.8\linewidth}{!}{
    \begin{tabular}{ l c c c c c}
    \hline
    Architecture &  \gls{iou} (\%) & mAP (\%)& \gls{cer} (\%) & \gls{wer} (\%) & \# Param.\\ 
    \hline
    \hline
    Line seg. model & 81.51 & 85.09 & \xmark &  \xmark & 1.8 M\\
     \gls{ocr} on lines & \xmark & \xmark & 5.01 & 16.49 & 1.7 M \\
    Three-step approach & 81.51 & 85.09 & 6.55 & 18.54 & 1.8+1.7M\\
    \hline
    \end{tabular}
    }
    \label{table:seg-reco}
\end{table}

We can now compare the \acrshort{van} to the three-step approach. Comparison on the IAM test set is summarized in Table \ref{table:summary}. First, one can notice that the \gls{van} reaches a better \gls{cer} of 4.45\% compared to 6.55\%.
Prediction time is computed as the average prediction time, on the test set, to process a paragraph image. As one can see, even though the segmentation step is without recurrence, it requires much more time for prediction due to the formatting of the input required for the \gls{htr}, including the bounding boxes extraction from the original images. Moreover, it cumulates prediction times of the two models involved. 
Despite its recurrent process, the total prediction time for the \gls{van} is shorter than that of the three-step approach since one iteration is very fast. In addition, it implies fewer parameters, this is notably due to the two models required by the three-step approach. Except for the training time, which is a bit higher for the \gls{van}, it only provides advantages compared to the three-step approach.

\begin{table}[!h]
    \caption{Comparison of the three-step approach with the VAN, results are given for the test set of the IAM dataset.}
    \centering
    \resizebox{0.8\linewidth}{!}{
    \begin{tabular}{ l c c c c c }
    \hline
    \multirow{2}{*}{Architecture} & \multirow{2}{*}{\gls{cer} (\%)} & \multirow{2}{*}{\gls{wer} (\%)} & \multirow{2}{*}{\# Param.} & Training & Prediction \\ 
    & &  & & time & time \\
    \hline
    \hline
    Three-step & 6.55 & 18.54 & 1.8+1.7 M & 0.03+0.59 d & 749+28 ms\\
   \gls{van} & \textbf{4.45} & \textbf{14.55} & 2.7 M & 0.59+0.63 d & 32 ms \\
    
    \hline
    \end{tabular}
    }
    \label{table:summary}
\end{table}

\subsubsection{Line-level analysis}
\label{section-exp-line}

In this section, we compare the results of the \gls{van} with the state-of-the-art approaches evaluated in similar conditions \textit{i.e.} at line level and without using external data. We also provide results for the line-level \gls{htr} model (Figure \ref{fig:span-line-ocr}), which has the same encoder, and for the \gls{van} applied at paragraph level. The aim is to highlight the contribution of the \gls{van} processing full paragraph images instead of isolated lines.

To have a fair comparison between the results obtained at paragraph and line levels, one should consider the difference in the ground truth: the paragraph transcriptions contain line breaks (or space characters in our case) between the different line transcriptions. Table \ref{table:linebreak} shows the \gls{van} results difference at paragraph level when removing the interline characters from the metrics. As one can see, the removal of the interline character leads to an increase of the \gls{cer} of 0.24, 0.09 and 0.24 on the test set of RIMES 2011, IAM and READ 2016 respectively. The \gls{wer} is not impacted by this modification. In the following tables of this section, we will use the results without considering interline characters for fair comparison with line-level approaches.

\begin{table}[!h]
    \caption{VAN results at paragraph level with and without interline characters in ground truth for RIMES 2011, IAM and READ 2016 dataset.}
    \centering
    \resizebox{0.6\linewidth}{!}{
    \begin{threeparttable}[b]
        \begin{tabular}{ l c c c c c }
        \hline
        \multirow{2}{*}{Dataset} & Line  & \gls{cer} (\%) & \gls{wer} (\%) & \gls{cer} (\%) & \gls{wer} (\%) \\ 
        & break & valid & valid & test & test\\
        \hline
        \hline
        \multirow{2}{*}{RIMES 2011} & \cmark & \textbf{1.83} & \multirow{2}{*}{6.26} & \textbf{1.91} & \multirow{2}{*}{6.72} \\
        & \xmark & 2.10 &  & 2.15 & \\
        \hline
        \multirow{2}{*}{IAM} & \cmark  & \textbf{3.02} & \multirow{2}{*}{10.34} & \textbf{4.45} & \multirow{2}{*}{14.55} \\
        & \xmark & 3.07 & & 4.54 & \\
        \hline
        \multirow{2}{*}{READ 2016} & \cmark  & \textbf{3.71} & \multirow{2}{*}{15.47} & \textbf{3.59} & \multirow{2}{*}{13.94} \\
        & \xmark & 4.01 & & 3.83 & \\
        \hline
        \end{tabular}
    \end{threeparttable}
    }
    \label{table:linebreak}
\end{table}

Table \ref{table:rimes-line} shows state-of-the-art results on the RIMES 2011 dataset at line level. We report competitive results with a \gls{cer} of 3.04\% for the line-level \gls{htr} model and 3.08\% for the \gls{van} on the test set compared to the model of \cite{Puigcerver2017} which reached 2.3\%. On should notice that Puigcerver \textit{et al.} \cite{Puigcerver2017} does not use exactly the same dataset split for training and validation. In conclusion we can highlight the performance of the \gls{van} obtained at paragraph level which achieves a \gls{cer} of 2.15\% on the test set, which corresponds to decreasing the \gls{cer} by 0.93 compared to processing isolated lines.
\begin{table}[!h]
    \caption{Comparison of the VAN with the state of the art on the line-level RIMES 2011 dataset.}
    \centering
    \resizebox{0.9\linewidth}{!}{
    \begin{threeparttable}[b]
        \begin{tabular}{ l c c c c r}
        \hline
        \multirow{2}{*}{Architecture} & \gls{cer} (\%) & \gls{wer} (\%) & \gls{cer} (\%) & \gls{wer} (\%) & \multirow{2}{*}{\# Param.}\\ 
        & valid & valid & test & test\\
        \hline
        \hline
        \cite{Voigtlaender2016} \acrshort{mdlstm}+\acrshort{lm} &  & & 2.8 & \textbf{9.6} & \\
        \cite{Puigcerver2017} \acrshort{cnn}+\acrshort{blstm}\tnote{a} & 2.2 & 9.6 & \textbf{2.3} & \textbf{9.6} & 9.6 M\\
        \cite{Coquenet2022} Line-level \acrshort{htr} model (FCN) & 2.20 & 6.26 & 3.04 & 8.32 & 1.7 M\\
        \cite{Coquenet2022} \acrshort{van} on lines (\acrshort{fcn}+\acrshort{lstm})  & 1.97 & 6.09 & 3.08 & 8.14 & 2.7 M \\
        \hline
        \cite{Coquenet2022} \acrshort{van} on paragraphs (\acrshort{fcn}+\acrshort{lstm}) & 2.10 & 6.26 & \textbf{2.15} & \textbf{6.72} & 2.7 M \\
        \hline
        \end{tabular}
        
        \begin{tablenotes}
        \item [a] This work uses a slightly different split (10,203 for training, 1,130 for validation and 778 for test).
        \end{tablenotes}
    \end{threeparttable}
    
    }
    \label{table:rimes-line}
\end{table}

Comparison with state-of-the-art results on the IAM dataset is presented in Table \ref{table:iam-line}. We reach competitive results with a \gls{cer} of 4.97\% on the test set. Models proposed in \cite{Yousef_line} and \cite{Michael2019} reach similar results but the former implies a large number of parameters compared to the \gls{van} and the latter is more complex including a recurrent process with attention at character level. It should be noticed however that, in \cite{Puigcerver2017,Michael2019,Yousef_line}, the authors use a slightly different split from ours. As a matter of fact, since the \gls{van} training implies pre-training at line level, it was not possible to use the same split since some lines for training and validation are extracted from the same paragraph image for example. On the IAM dataset, the \gls{van} also reaches a better \gls{cer} at paragraph level than at line level with 4.54\% compared to 4.97\%.
\begin{table}[!h]
    \caption{Comparison of the VAN with the state of the art at the line level on the IAM dataset.}
    \centering
    \resizebox{0.9\linewidth}{!}{
    \begin{threeparttable}[b]
        \begin{tabular}{ l c c c c r}
        \hline
        \multirow{2}{*}{Architecture} & \gls{cer} (\%) & \gls{wer} (\%) & \gls{cer} (\%) & \gls{wer} (\%) & \multirow{2}{*}{\# Param.}\\ 
        & validation & validation & test & test\\
        \hline
        \hline 
        \cite{Voigtlaender2016} \acrshort{cnn}+\acrshort{mdlstm}+\acrshort{lm} & \textbf{2.4} & \textbf{7.1} & \textbf{3.5} & \textbf{9.3} & 2.6 M \\
        \cite{Puigcerver2017} \acrshort{cnn}+\acrshort{blstm}\tnote{a}  & 3.8 & 13.5 & 5.8 & 18.4 & 9.3 M\\
        \cite{Yousef_line}  \acrshort{gfcn}\tnote{a} & 3.3 & & 4.9 & & > 10 M\\
        \cite{Michael2019} \acrshort{cnn}+\acrshort{blstm}\tnote{a} & & & 4.87 & & \\
        \cite{Coquenet2022} Line-level \acrshort{htr} model (FCN) & 3.37 & 11.52 & 5.01 & 16.49 & 1.7 M\\
        \cite{Coquenet2022} \acrshort{van} on lines (\acrshort{fcn}+\acrshort{lstm}) & 3.15 & 10.77 & 4.97 & 16.31 & 2.7 M\\
        \hline
        \cite{Coquenet2022} \acrshort{van} on paragraphs (\acrshort{fcn}+\acrshort{lstm}) & 3.07 & 10.34 & 4.54 & 14.55 & 2.7 M\\
        \hline
        \end{tabular}
     
        \begin{tablenotes}
        \item [a] These works use a slightly different split (6,161 for training, 966 for validation and 2,915 for test).
        \end{tablenotes}
        
    \end{threeparttable}
    }       
    \label{table:iam-line}
\end{table}

The results on the READ 2016 dataset are gathered in Table \ref{table:read2016-line}. We reached state-of-the-art \gls{cer} on the test set with 4.10\% compared to 4.66\% for \cite{Michael2019}. Again, the paragraph-level \gls{van} reaches better results than the \gls{van} applied at line level with a \gls{cer} of 3.83\%.

\begin{table}[!h]
    \caption{Comparison of the VAN with the state-of-the-art line-level recognizers on READ 2016 dataset.}
    \centering
    \resizebox{0.9\linewidth}{!}{
        \begin{threeparttable}[b]
            \begin{tabular}{ l c c c c r}
            \hline
            \multirow{2}{*}{Architecture} & \gls{cer} (\%) & \gls{wer} (\%) & \gls{cer} (\%) & \gls{wer} (\%) & \multirow{2}{*}{\# Param.}\\ 
            & validation & validation & test & test\\
            \hline
            \hline
            \cite{Michael2019} \acrshort{cnn}+\acrshort{blstm} & & & 4.66 & & \\
            \cite{READ2016}\tnote{a}\: \acrshort{cnn}+\acrshort{mdlstm}+\acrshort{lm} & & & 4.8 & 20.9\\
            \cite{READ2016}\tnote{b}\: \acrshort{cnn}+\acrshort{rnn} & & & 5.1 & 21.1\\
            \cite{Coquenet2022} Line-level \acrshort{htr} model (FCN) & 4.49 & 18.22 & 4.25 & 17.14 & 1.7 M\\
            \cite{Coquenet2022} \acrshort{van} on lines (\acrshort{fcn}+\acrshort{lstm}) & 4.42 & 18.17 & \textbf{4.10} & \textbf{16.29} & 2.7 M\\
            \hline
            \cite{Coquenet2022} \acrshort{van} on paragraphs (\acrshort{fcn}+\acrshort{lstm}) & 4.01 & 15.47 & \textbf{3.83} & \textbf{13.94} & 2.7 M\\
            \hline
            \end{tabular}
            
            \begin{tablenotes}
                \item [a] results from RWTH.
                \item [b] results from BYU.
            \end{tablenotes}
        \end{threeparttable}
    }
    \label{table:read2016-line}
\end{table}

In conclusion, one can notice that the \gls{van}, applied on isolated lines, performs at least similarly as the line-level model, for each dataset (except for RIMES with a small \gls{cer} increase of 0.04 points). It also achieves state-of-the art results at line level on the READ dataset. 

The results also highlight the superiority of the \acrlong{van} on the RIMES 2011, IAM and READ 2016 datasets, applied to whole paragraph images, compared to isolated lines. Multiple factors can explain this result: segmentation ground truth annotations of text lines are prone to variations from one annotator to another, bringing variability that is not present when dealing with paragraph images directly. Indeed, the model implicitly learns to segment the lines so it does not have to adapt to pre-formatted lines; it uses more context (with a large receptive field) and uses it to focus on the useful information for the recognition purpose. 
Moreover, the \gls{van} decoder contains a \gls{lstm} layer that may have a positive impact acting as a language model without any loss of context when moving from one line to the next, when producing the output character sequence.

A key element to reach such results with a deep network is to use efficient regularization strategies. We discuss the new dropout strategy we propose in the following paragraph.

\subsubsection{Dropout strategy}
We use dropout to regulate the network training and thus avoid overfitting. We defined Diffused Mix Dropout (DMD) in Section \ref{section-archi-dropout} to improve the results of the model. We carried out some experiments to highlight the contribution of DMD over commonly used standard and 2d dropout layers. Experiments are performed with the line-level \gls{htr} model and the \gls{van} on the IAM dataset; \gls{van} is pre-trained with weights from the corresponding line-level model. Results for the test set are shown in Table \ref{table:dropout}. The columns from left to right correspond respectively to the number of dropout layers per block (CB and DSCB), the type of dropout layer used (mix, standard or spatial), the associated dropout probabilities, the use of the diffuse option (using only one or all dropout layers per block) and the \gls{cer} and \gls{wer} for both models.

\label{section-exp-dropout}
\begin{table}[!h]
    \caption{Dropout strategy analysis. Results are given for the IAM test set.}
    \centering
    \resizebox{\linewidth}{!}{
    \begin{tabular}{ l c c c c c c c r}
    \hline
    & \multirow{2}{*}{\#} & \multirow{2}{*}{type} & \multirow{2}{*}{p} &\multirow{2}{*}{diffused} &  \multicolumn{2}{c}{line-level model} & \multicolumn{2}{c}{\acrshort{van}} \\ 
    & & & & & \gls{cer} (\%) & \gls{wer} (\%) & \gls{cer} (\%) & \gls{wer} (\%)\\
    \hline
    \hline
    Baseline & 3 & mix & 0.5/0.25 &\cmark & \textbf{5.01} & \textbf{16.49} & 4.45 & \textbf{14.55}\\
    (1) & 3 & std. & 0.5 &\cmark & 5.24 & 17.23 & 4.46 & 14.88 \\
    (2) & 3 & 2d & 0.25 & \cmark & 5.38 & 17.64 & 4.72 & 15.63 \\
    (3) & 1 & mix & 0.5/0.25 & \xmark & 5.33 & 17.70 & \textbf{4.43} & 15.25\\
    (4) & 1 & std. & 0.5 & \xmark & 5.56 & 18.40 & 4.78 & 15.91\\
    (5) & 1 & 2d & 0.25 & \xmark & 5.70 & 18.92 & 4.93 & 16.80\\
    (6) & 3 & mix & 0.5/0.25 & \xmark & 6.76 & 21.13 & 6.32 & 19.73\\
    (7) & 3 & mix & 0.16/0.08 & \xmark & 6.71 & 20.91 & 4.64 & 15.36 \\
    (8) & 3 & mix & 0.16/0.08 & \cmark & 7.51 & 23.60 & 5.57 & 18.50\\
    
    \hline
    \end{tabular}
    }
    \label{table:dropout}
\end{table}

In (1) and (2), Mix Dropout layers are respectively replaced by standard and 2d dropout layers, preserving their corresponding dropout probability. Using Mix Dropout leads to an improvement of 0.23 points of \gls{cer} compared to standard dropout and of 0.37 compared to 2d dropout for the line-level model. These improvements are lower for the \gls{van} with 0.01 and 0.27 points.

In (3), only one Mix Dropout is used, after the first convolution of the blocks, leading to a higher \gls{cer} than the baseline, with a difference of 0.34 points for the line-level model. The \gls{cer} is decreased by 0.02 points for the \gls{van} but the \gls{cer} is increased by 0.70 points.
In (4) and (5), we are in the same configuration as (3) \textit{i.e.} with only one dropout layer per block. Mix Dropout is superseded by standard dropout in (4) and by 2d dropout in (5) resulting in an increase of the \gls{cer} compared to (3). This shows the positive impact of Mix Dropout layers in another configuration.

In (6) and (7), Mix Dropout layers are set at each of the three positions \textit{i.e.} they are all used at each execution, contrary to the baseline, which uses only one dropout layer per execution. While (6) keeps the same dropout probabilities, (7) divides them by 3. In both cases, the associated \gls{cer} are higher than the baseline.

Finally, in (8), we are in the same context than the baseline, but dropout probabilities are divided by 3, leading to higher \gls{cer}.

We can conclude that our dropout strategy leads to a \gls{cer} improvement of 0.55 points for the line-level model and of 0.33 for the \gls{van}, when compared to (4) and (5) that do not use Mix Dropout or diffuse option. 

\subsection{Discussion}
\label{section-discussion}
We proposed the \acrlong{van} as a novel end-to-end encoder-decoder segmentation-free architecture using hybrid attention for the task of handwritten paragraph recognition.
As we have seen, the \gls{van} achieves state-of-the-art results on multiple datasets at paragraph level. However, there is one point that should be notice. Modern deep neural systems involve many training strategies (hyperparameters, optimizer, regularization strategies, pre-processings, data augmentation techniques, transfer learning, curriculum learning, and many others). This makes the comparison between architectures very difficult as some training tricks are more suited for some architectures than some others. This is why one should be convinced that the state-of-the-art results obtained by the \gls{van} are due to the whole proposition, including training strategies, and not only to the \gls{van} architecture. However, we have provided experimental results that show the interest of the proposed training strategies of the generic \gls{van} architecture.

We compared different stopping strategies and showed that the \gls{van} can learn to detect the end of paragraph. This additional task has no significant impact on the performance and slightly improves the stability through training. We also compared favorably the \gls{van} to a standard two-step approach and showed the positive impact of processing paragraph-level images compared with line-level ones, for this architecture. The new dropout strategy we propose enabled to reach even better results.

Moreover, the \gls{van} has multiple advantages. The \gls{van} is robust: whether it is at paragraph or line level, and no matter the dataset used, we did not adjust any hyperparameter for each dataset. The \gls{van} takes input of variable sizes, so it could handle whole page images without any modification. As mentioned previously, the \gls{van} can handle slightly inclined lines. However, it is limited to layouts in which there is no overlap between lines on their horizontal projection. Indeed, this case remains to be solved. A standard n-gram language model could process the outputs of the \gls{van} architecture but its impact on the performance remains to be determined through experiments.

However, there is still room for improvement.  Notably, we showed that the \gls{van} needs pre-training on isolated text lines of the target dataset to reach state-of-the-art results. But the need for line-level annotations is not inherent to the \gls{van}, this is only related to this pre-training step. As a matter of fact, we demonstrate that pre-training on another dataset (cross-dataset pre-training) can alleviate this issue, even if the datasets are really different. Indeed, for the three datasets, cross-dataset pre-training leads to results similar to those from line-level pre-training, but without using any line-level annotation from the target dataset.

The \gls{van} should be considered to process single-column text document only. As a matter of fact, as it is the case for \cite{Bluche2016} and supposedly \cite{Yousef2020}, the models are designed and limited to process single-column multi-line text documents with relatively horizontal text lines.

\section{Conclusion}
We have presented two different architectures, the \gls{span} and the \gls{van}, which follow two totally different approaches to deal with \gls{htr} at paragraph-level. Although they reach competitive and state-of-the-art results on 3 public datasets for the \gls{span} and the \gls{van} respectively, both approaches present the same limitations:
\begin{itemize}
    \item They both rely on a pre-training stage based on the use of line-level segmentation annotations: this reduces the interest of dealing with paragraph-level images. The only end-to-end model of the literature which does not use any line-level segmentation annotation during training (\cite{Yousef2020}) compensate with the tuning of dataset-specific hyperparameters, making it not generic at all.
    \item They are, by design, limited to single-column text. It means that, although it has not been tested, they could theoretically handle whole documents with simple layout such as single-column pages. However, they cannot handle multi-column pages. 
    \item These models are quite limited to tackle slanted text lines.
\end{itemize} 

These issues could be partly solved with the use of a character-level attention mechanism. Indeed, with attention at character level, we do not use the \textit{a priori} assumption that there can be only one text line per vertical position. In addition, the attention weight is not associated to a whole row of pixels or features: the attention mechanism can follow the slant of the text lines. 

The authors of \cite{Singh,Rouhou2021} recently proposed a \acrshort{cnn} combined with a transformer decoder module for \gls{htr} through a recurrent character-level attention process. However, the proposed approaches in the literature based on character-level attention are hardly competitive with a \gls{cer} of 16.2\% in \cite{Bluche2017b} and 6.7\% in \cite{Singh} on the test set of the IAM dataset at paragraph-level. In addition, they rely on synthetic samples based on word segmentation annotations. 

In the next chapter, we propose to handle the task of \gls{htr} at page and double-page level with a character-level attention model, also recognizing the layout, without using any physical layout annotations for training.

\glsresetall
\chapter{Handwritten document recognition}
\label{chap:document}

A handwritten document is a complex structure composed of handwritten text blocks structured according to a specific layout. It can also contain non textual items such as images and tables. In this thesis, we define \gls{hdr} as the joint recognition of both text and layout. 
In Chapter \ref{chap:line} and Chapter \ref{chap:paragraph}, we have studied the task of \gls{htr} through the use of a three-step approach at line and paragraph levels. Although \gls{dla} is not necessary, it can be carried out during the segmentation step, leading to a semantic segmentation task. In this case, \gls{dla} and \gls{htr} are two tasks that are generally processed in an independent way in the literature. We assume that this does not seem relevant since these two tasks are closely related. In addition, this three-step approach suffers from multiple drawbacks.
It relies on segmented entities which do not have a clear definition from the image point of view. For instance, text lines can be defined by X-heights, baselines, bounding boxes or polygons. This approach also requires segmentation annotations, which are very costly to produce.
The resulting predictions accumulate errors between each stage, leading to high error rates.
In this chapter, we propose to tackle the task of \gls{hdr} by using a single end-to-end model, free from the aforementioned issues, and able to recognize both text and layout from a document image.

\section{Problem statement}
\begin{figure}[t!]
    \centering
    \includegraphics[width=\linewidth]{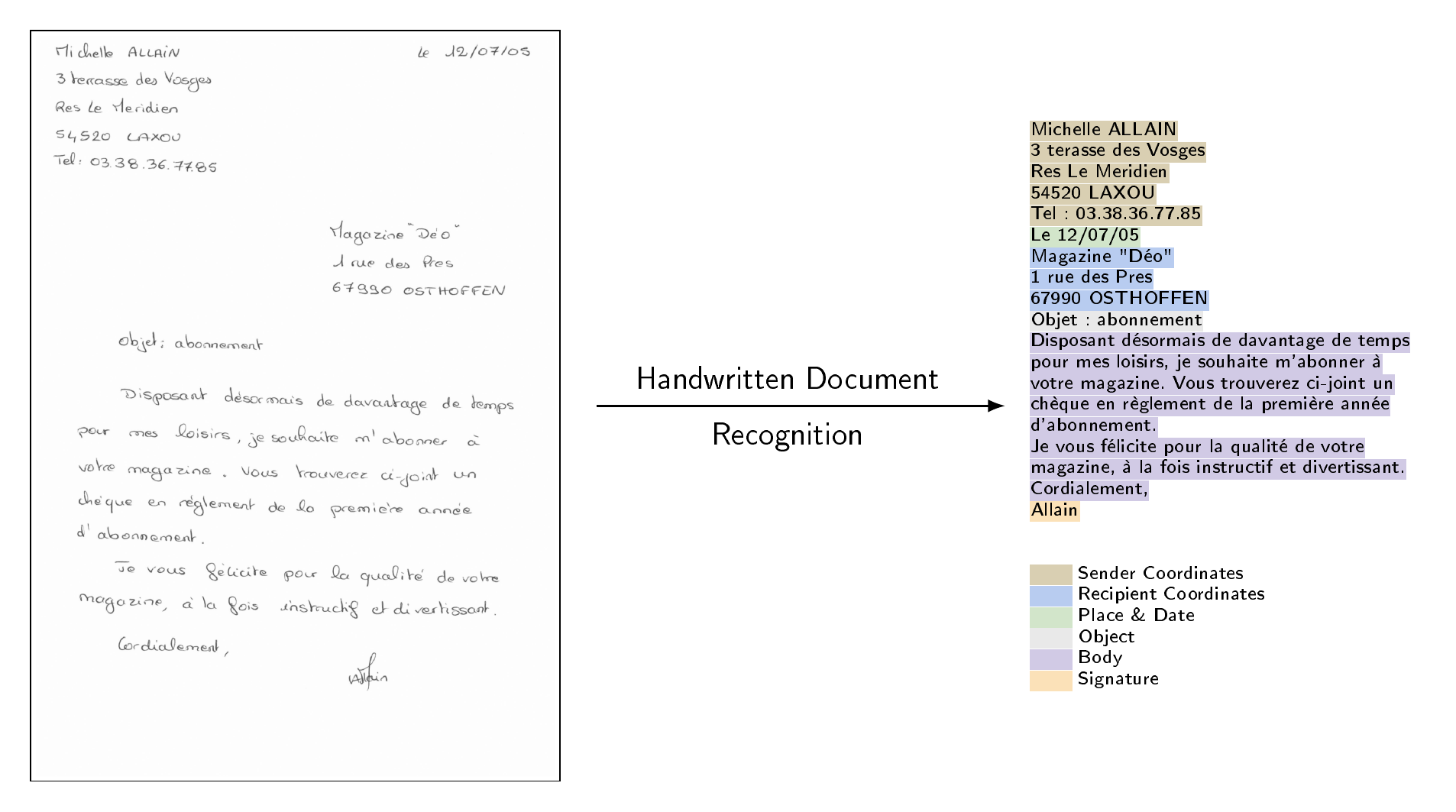}
    \caption{Overview of Handwritten Document Recognition. Not only the text is recognized in a correct order, but it is also labeled given some layout classes.}
    \label{fig:hdr}
\end{figure}

We defined \gls{hdr} as the joint recognition of both text and layout. By this, we mean that the goal is not only to recognize all the text from the input document in a humanly correct order, but also to tag sub-sequences of the predicted sequence of characters given some layout classes. In other words, contrary to the three-step approach, whose segmentation (or \gls{dla}) stage consists in associating a layout class to some pixels of the input image, we propose to associate the layout class directly to a sequence of characters. This approach is depicted in Figure \ref{fig:hdr} with a letter example. As one can note, text parts, in addition to be recognized, are labeled with one of the following layout classes: sender coordinates, recipient coordinates, object, body and signature.
In this example, layout classes correspond to logical layout entities (including some semantic) but it can also be physical layout entities such as header, footer, different title levels, section or paragraph. In this chapter, we only focus on textual entities.

Processing whole documents instead of isolated paragraphs raises new challenges:
\begin{itemize}
    \item Paragraph-level models usually take benefits from the fact that all horizontal elements at the same vertical position belong to the same text line. This assumption is no longer valid at document-level.
    \item The input images are larger and the associated target sequences are longer for documents than for paragraphs, leading to more complex training procedures, especially in the case of attention-based models, which involves a growing need for \acrshort{gpu} memory. 
    \item While paragraphs are read in a monotonous order (characters are read from left to right and lines from top to bottom in case of common occidental languages for example), documents reading order is layout dependent \textit{i. e.} paragraphs of a single-column document are read from top to bottom only, whereas a multi-column document is commonly read column by column, adding a horizontal constraint on the reading order.
\end{itemize}

\section{Related works}

Deep learning models are becoming more and more powerful and can now process entire documents. As we aim at recognizing both the text and the logical layout information of a handwritten document image, the task falls into both the \gls{htr} field and the \gls{dla} field, thus sharing links with document understanding. To our knowledge, we propose the first end-to-end approach for \gls{hdr}. 

Therefore, this section is dedicated to document understanding in a general way, and then focuses on layout analysis and handwriting recognition. 

\subsection*{Document understanding}
Document understanding includes a set of tasks whose purpose is to extract, classify, interpret, contextualize, and search information from documents. It implies, among others, \gls{dla} and \gls{ocr}. But it also consists in understanding complex structures such as tables, schemes or images. This is a developing field and here are some related works.
The authors of \cite{Chargrid,Chargrid_OCR} proposed Chargrid, a way to represent textual documents as a 2D representation of one-hot encoded characters, which are produced by an \gls{ocr}. The idea is to keep the spatial information between the textual entities. Chargrid is then used as the input for a Key Information Extraction (KIE) task implying three sub-tasks: bounding box regression, semantic segmentation and box masking. 
In \cite{Bertgrid}, the Chargrid paradigm is also used but the character's encoding are superseded by word embeddings through the use of the BERT model \cite{BERT}.
In \cite{LayoutLMv2}, the authors tackle the task of Visually-rich Document Understanding (VDU). They used a transformer architecture applied on multiple modalities: \gls{ocr} text and bounding boxes, and visual embedding. The authors of \cite{Tilt} proposed a transformer-based model able to tackle multiple tasks such as document classification, KIE and Visual Question-Answering (VQA).
All these works imply the use of large datasets, mainly synthetic: they need a lot of information either as input or as ground truth annotation (notably for bounding boxes).
One can notice that document understanding area is still in its early stages. It mainly focuses on 2D document representation and information extraction. Our work differs from document understanding in which we propose to extract all the text from a document; we do not aim at retrieving only specific information. 

\subsection*{Document layout analysis}
Related works for \gls{dla} were described in Section \ref{section:related-work-dla}. \gls{dla} focuses on identifying physical regions of interest whether they are textual or not. It is driven by physical ground truth annotations accounting for semantic labels that are associated to document logical elements. \gls{hdr} solely focuses on textual components for which we target their recognition and semantic labeling.

\subsection*{Handwritten text recognition}
We presented related works for \gls{htr}  in Section \ref{section:related-work-htr-line} (line-level) and Section \ref{section:related-work-htr-paragraph} (paragraph-level). Among the paragraph-level \gls{htr} approaches, a single work has been devoted to the recognition of some layout items, in addition to the text. As a matter of fact, the model from \cite{Singh} is trained to recognize, in addition to the text, the presence of non-textual areas: tables, drawings, math equations and deleted text. As one can note, this model recognizes non-textual items but it does not tag the predicted text with a layout class.

The main drawbacks of end-to-end \gls{htr} systems are as follows:
\begin{itemize}
    \item The paragraph limitation. Except for \cite{Singh}, which uses a page-level private dataset, there is no work evaluating their model on a page-level dataset.
    \item The need for segmentation annotations. Even if the models are trained on paragraph-level images, line-level segmentation annotations are used to reach competitive results, either for pre-training (\cite{Coquenet2021,Coquenet2022}) or for synthetic data generation (\cite{Singh,Rouhou2021,Bluche2017b}). The only work which does not use any line-level segmentation annotation \cite{Yousef2020} is limited to single-column text images and uses some dataset-specific hyperparameters to reach its competitive results.
\end{itemize}

We propose the \gls{dan}, an end-to-end transformer-based model for whole \acrlong{hdr}, including the textual components and the logical layout information. This model is trained without using any segmentation label and we evaluate it on two public datasets at page and double-page levels: RIMES 2009 and READ 2016. To our knowledge, this is the first attempt that provides experimental results on such a task.

\section{DAN: a Document Attention Network}
We propose an end-to-end segmentation-free architecture for the task of \gls{hdr}: the \gls{dan} \cite{Coquenet2022b}. In addition to the text recognition, the model is trained to label text parts using begin and end tags in an \acrshort{xml}-like fashion. This model is made up of a \gls{fcn} encoder for feature extraction and a stack of transformer decoder layers for an autoregressive token-by-token prediction process. It takes whole text documents as input and sequentially outputs characters, as well as logical layout tokens.

\begin{figure}[h!]
    \centering
    \includegraphics[width=0.8\linewidth]{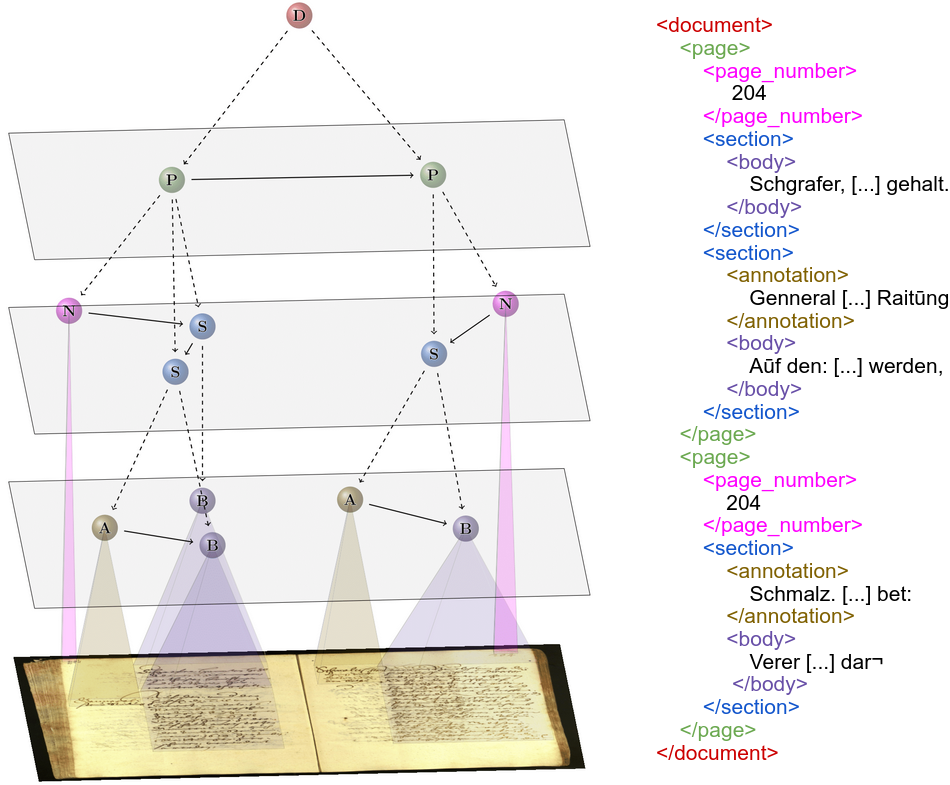}
    \caption{Left: document image with associated layout graph. Right: ground truth transcription including text and layout tokens.}
    \label{fig:xml}
\end{figure}

It is designed to handle whole documents with complex layouts such as double-column pages, and double-page documents, as  depicted in Figure \ref{fig:xml}.

While the textual content can be represented as a sequence of characters, the layout is represented as an oriented graph, in order to model the hierarchy and the reading order of the different layout entities. In figure \ref{fig:xml}, this layout graph is projected on the document image: nodes are layout entities and edges model the relations between them. A membership relation is represented by a dashed arrow, while solid arrows represent the reading order. On this example, the document is made up of two pages, each page containing a page number and a sequence of sections, each section being made up of zero, one or many marginal annotations and a single body. 

Text and layout are intrinsically linked: text recognition may help to label a layout entity, and vice versa. Therefore, we have turned toward the joint recognition of text and layout in a unique model. As shown on the right side, we chose the \acrshort{xml} paradigm to generate a serialized representation of the document that is further used as the ground truth of a handwritten document. Notice that no physical information is encoded in the ground truth. Contrary to the standard \gls{dla} task, the proposed approach does not rely on segmentation labels of text regions such as bounding boxes for instance. Instead, the proposed model is able to provide transcriptions enriched with logical layout information, leading to structured transcriptions. It notably enables to reduce the need for costly segmentation annotations.

The problem can be formalized as follows: the input is a raw document image $\mb{X}$ and the expected output is a sequence $\mseq{y}$ of tokens, of length $L_y$. Tokens are grouped in the same dictionary $\mset{D} = \mset{A} \cup \mset{S} \cup \{ \mathrm{<eot>} \}$, where $\mset{A}$ are tokens of characters from a given alphabet, $\mset{S}$ are specific layout tokens and <eot> is a special end-of-transcription token. Layout tokens are pairs of tokens (begin and end) that tag sequences of character tokens as shown in Figure \ref{fig:xml}. As for characters, they vary according to the dataset used. 

In the following, we present the \gls{dan} architecture and training strategy. We describe two new metrics we propose for the task of \gls{hdr}. We provide an experimental study, including a visualization of the process, and a discussion of the model. We provide all source code and pre-trained model weights at \url{https://github.com/FactoDeepLearning/DAN}.

\subsection{Architecture}
\label{section-dan-architecture}

We propose the \gls{dan}, an end-to-end encoder-decoder architecture that jointly recognizes both text and layout, from whole documents. We opted for an \gls{fcn} as encoder since they are known to be efficient for feature extraction from images and can deal with input of variable sizes. For the decoder, we  chose the transformer \cite{Vaswani2017} because it is currently the state-of-the-art approach for many tasks involving the prediction of sequences of variable lengths.

\begin{figure}[ht!]
    \centering
    \includegraphics[width=\linewidth, max height=0.85\textheight]{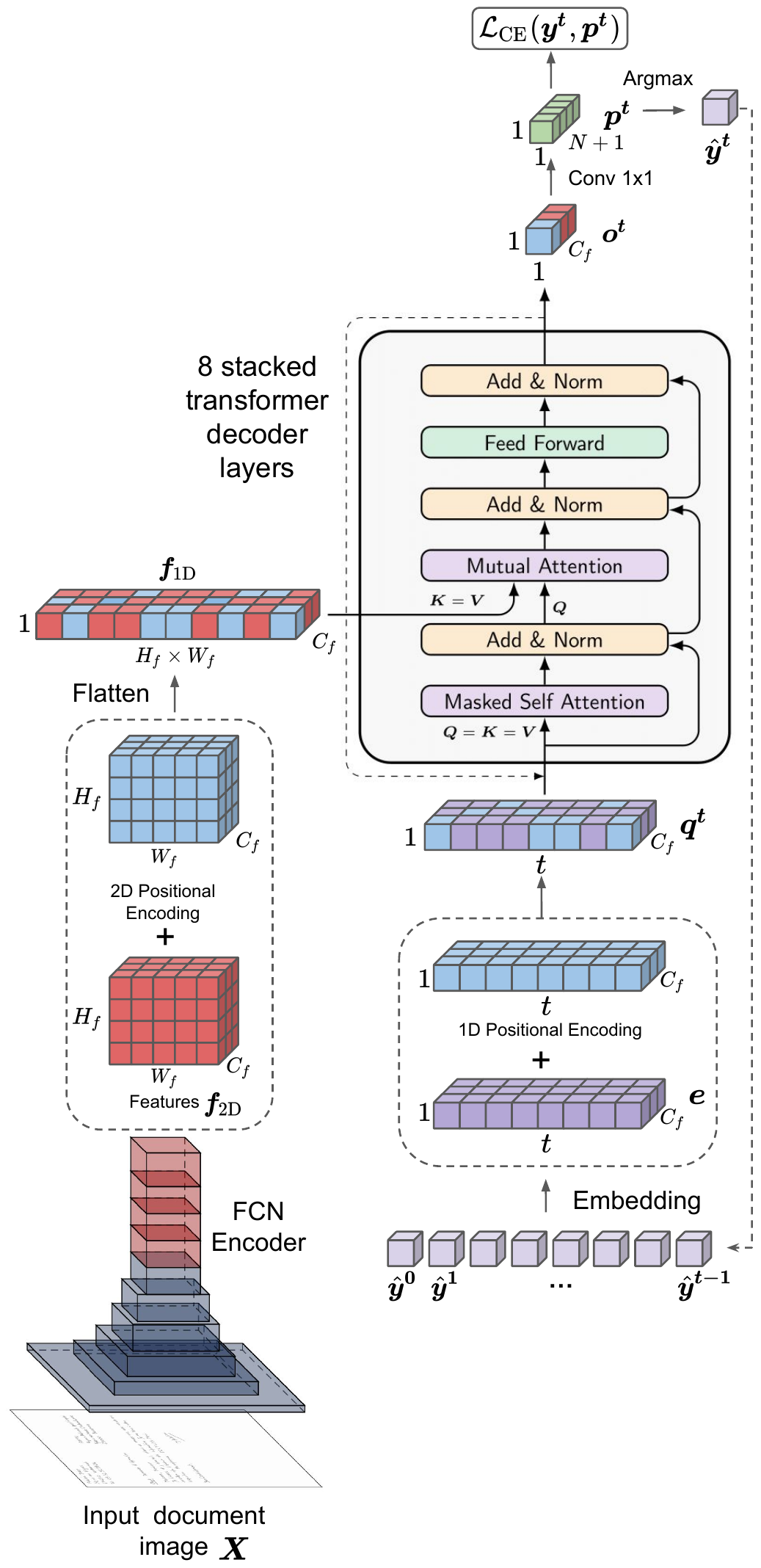}
    \caption{Overview of the DAN architecture (FCN+transformer decoder). Characters $\mb{\hat{y}}\mt{t}{}$ are predicted one after the other, based on the extracted features $\mb{f}_\mathrm{1D}$ and on the previous predictions.}
    \label{fig:model_overview}
\end{figure}

The \gls{dan} architecture is depicted in Figure \ref{fig:model_overview}. It is made up of an \gls{fcn} encoder, to extracts 2D feature maps $\mb{f}_\mathrm{2D}$ from an input document image $\mb{X}$. 2D positional encoding is added to these features in order to keep the spatial information, before being flattened into a 1D sequence of features $\mb{f}_\mathrm{1D}$. This representation is computed only once and serves as input to the transformer decoder. The decoder follows an autoregressive prediction process at character-level: given the previously predicted tokens ($\hat{\mseq{y}}\mtseq{0}{}$, $\hat{\mseq{y}}\mtseq{1}{}$, \ldots, $\hat{\mseq{y}}\mtseq{t-1}{}$) and based on the computed features $\mb{f}_\mathrm{1D}$, it outputs the next token probabilities $\mb{p}\mt{t}{}$ for each token of $\mset{D}$. The final predicted token $\hat{\mseq{y}}\mtseq{t}{}$ is the one with the highest probability. The decoding process starts with an initial <sot> (start-of-transcription) token ($\hat{\mseq{y}}\mtseq{0}{} = \mathrm{<sot>}$), and ends with the prediction of a special <eot> (end-of-transcription) token ($\hat{\mseq{y}}\mtseq{L_y+1}{} = \mathrm{<eot>}$).  We used the cross-entropy loss ($\mset{L}_\mathrm{CE}$) for training.

We now describe the encoder and the decoder with more details.

\subsubsection{Encoder}
As in \cite{Singh}, we opted for an \gls{fcn} encoder in order to better model the local dependencies since inputs are images.
We used the \gls{fcn} encoder of the \gls{van} \cite{Coquenet2022}, for many reasons. It achieves state-of-the art results for \gls{htr} at paragraph level on many public datasets: RIMES 2011, IAM and READ 2016. It can handle input of variable sizes. And it implies few parameters (1.7M) compared to other approaches. 

The encoder takes as input a document image $\mb{X} \in \mathbb{R}^{H \times W \times C}$, with $H$, $W$ and $C$ being respectively the height, the width and the number of channels ($C=3$ for an RGB image). It extracts some features maps for the whole document image: $\mb{f}_\mathrm{2D} \in \mathbb{R}^{H_f \times W_f \times C_f}$ with $H_f=\frac{H}{32}$, $W_f = \frac{W}{8}$ and $C_f = 256$.

The original transformer architecture \cite{Vaswani2017} is defined for 1D sequences. Since the inputs are 2D images, we replaced the 1D positional encoding by 2D positional encoding, as proposed in \cite{Singh}. 2D positional encoding is defined as a fixed encoding based on sine and cosine functions with different frequencies (as in \cite{Vaswani2017}); but instead of encoding a 1D position using all the channels, half is dedicated to vertical positional encoding and the other half to the horizontal positional encoding:

\begin{equation}
\begin{split}
    \mathrm{PE}_\mathrm{2D}(x,y,2k) = \sin(w_k \cdot y), \\
    \mathrm{PE}_\mathrm{2D}(x,y,2k+1) = \cos(w_k \cdot y), \\
    \mathrm{PE}_\mathrm{2D}(x,y,d_{\mathrm{model}}/2 + 2k) = \sin(w_k \cdot x), \\
    \mathrm{PE}_\mathrm{2D}(x,y, d_{\mathrm{model}}/2 +2k+1) = \cos(w_k \cdot x), \\
    \forall k \in \left[0, d_\mathrm{model}/4\right],
\end{split}
\end{equation}
with 
\begin{equation}
    w_k = 1/10000^{2k/d_\mathrm{model}}.
\end{equation}
We set $d_\mathrm{model}=C_f=256$.

Features $\mb{f}_\mathrm{2D}$ are summed with 2D positional encoding before being flattened for transformer decoder requirements:
\begin{equation}
    \mb{f}_{\mathrm{1D}_j} = \mathrm{flatten}(\mb{f}_{\mathrm{2D}_{x,y}} + \mathrm{PE}_\mathrm{2D}(x,y)),
\end{equation}
with 
\begin{equation}
j = y W_f + x.
\end{equation}

\subsubsection{Decoder}
The decoder follows an autoregressive process. At each iteration, it takes as input the flattened visual features $\mb{f}_\mathrm{1D}$ and the previously predicted tokens ($\hat{\mseq{y}}\mtseq{0}{}$, ..., $\hat{\mseq{y}}\mtseq{t-1}{}$), and outputs the probabilities $\mb{p}\mt{t}{}$ for each token in the dictionary $\mset{D}$. We used learned embeddings to convert the tokens to vectors $\mb{e}_{\hat{y}_i}$ of dimension $d_\mathrm{model}$. Embeddings are summed with 1D positional encoding corresponding to the position of the predicted tokens in the predicted sequence, as in \cite{Vaswani2017}:

\begin{equation}
    \mb{q}\mt{t}{i} = \mathrm{PE}_\mathrm{1D}(i) + \mb{e}_{\hat{y}_i},
\end{equation}
with:
\begin{equation}
    \begin{split}
    \mathrm{PE}_\mathrm{1D}(x, 2k) = \sin(w_k \cdot x) \\
    \mathrm{PE}_\mathrm{1D}(x, 2k+1) = \cos(w_k \cdot x) \\
    \forall k \in \left[0, d_\mathrm{model}/2\right].
\end{split}
\end{equation}

The decoder is made up of a stack of 8 transformer decoder layers (as shown in Figure \ref{fig:model_overview}) followed by a convolutional layer with kernel $1 \times 1$ that computes the next token probabilities $\mb{p}\mt{t}{}$ (N character and layout tokens + 1 for the special $<eos>$ token). The transformer decoder layers are based on multi-head attention \cite{Vaswani2017} mechanisms we denote as self attention and mutual attention. Self attention aims at modeling dependencies among the predicted sequence: it corresponds to multi-head attention where queries $\mb{Q}$, keys $\mb{K}$ and values $\mb{V}$ are from the same input. Mutual attention is used to extract visual information from the encoder ($\mb{K}$ and $\mb{V}$ are from $\mb{f}_\mathrm{1D}$), based on $\mb{Q}$ which comes from the previous predictions. In other words, given the previous predictions, it indicates where should the model look at to predict the next token.

We used 8 decoder layers with dimension $d_\mathrm{model}$, feed forward dimension $d_\mathrm{model}$, 4 attention heads, ReLU activation and 10\% of dropout. Self attention is causal since it is based on the previous predictions.  As in \cite{Singh}, we used an attention window of length 100 for the self attention in order to reduce the computation time. It means that given an input sequence $\mseq{s}$ of length $L_\mathrm{s}$, the $t^\mathrm{th}$ output frame $\mb{o}\mt{t}{}$ is computed over the range $[\mseq{s}_a, \mseq{s}_{t-1}]$ with $a = \max(0, t-100)$.
The process starts with a <sos> token and ends when a <eos> token is predicted or after $L_{max}$ iterations. We set $L_{max}=3000$ to match the datasets needs.

The whole model is made up of 7.6 M trainable parameters and is trained in an end-to-end fashion using the cross-entropy loss over the sequence of tokens:
\begin{equation}
    \mset{L} = \sum_{t=1}^{L_y+1} \mset{L}_\mathrm{CE}(\mseq{y}\mtseq{t}{}, \mb{p}\mt{t}{})
\end{equation}

\subsection{Training strategy}
Training a deep attention-based neural network is difficult, especially when dealing with large inputs such as whole documents. The proposed training strategy is designed to improve the convergence while not relying on any segmentation label and dealing with few training data. It is performed in two steps:
\begin{itemize}
    \item Pre-training: the aim is to learn the feature extraction part of the \gls{dan}. We trained a line-level \gls{ocr} model on synthetic printed lines, and used it for transfer learning purposes for the \gls{dan}. We only used synthetic printed lines to avoid using segmentation labels which are costly annotations. Pre-training is carried out during 2 days with a mini-batch size of 16. Pre-processings, data augmentation and curriculum dropout are used during pre-training, as detailed afterwards.
    \item Training: the \gls{dan} is trained using teacher forcing (See section \ref{teacherforcing}) to reduce the training time per epoch. It is trained on both real and synthetic documents. The idea is to learn the attention mechanism, \textit{i. e.} the reading order, through the synthetic images. Indeed, printed text is easier to recognize than handwritten text and the \gls{dan} is pre-trained on printed text lines. Once the reading order is learned, it becomes easier to adapt to real-world images. This is motivated by the nature of the reading order: it is the same between printed and handwritten documents sharing the same layout.
    
    Following this idea, the model is first trained with 90\% of synthetic documents during a curriculum phase to learn the reading order while using few real training samples. Then, this percentage is slowly decreased to reach 20\% through the epochs, in order to fine-tune on the real samples while keeping some synthetic samples acting as unseen training data.
    
    Training is carried out during 4 days. We did not used mini-batch: training is carried out image per image. Pre-processings, data augmentation, curriculum strategies and post-processings are used during training, as described in the following.
\end{itemize}

\subsubsection{Pre-training}
Training deep attention-based models is difficult. It is beneficial to train part of the model beforehand whether it is the attention part or the feature extraction part, as shown in \cite{Coquenet2022}.
As in \cite{Coquenet2022}, we used a line-level pre-training strategy \textit{i. e.} we first train a line-level \gls{ocr} model on isolated line images using the \gls{ctc} loss. This line-level \gls{ocr} model is the same as the one used for the \gls{van} and the \gls{span} (Figure \ref{fig:span-line-ocr}). However, contrary to \cite{Coquenet2022,Coquenet2021}, we do not use real isolated lines (extracted with the bounding boxes annotations) but synthetic printed text lines, generated from the text line transcriptions only. The \gls{dan} is then trained using transfer learning using this model to initialize the weights of the encoder and of the decision layer (last convolutional layer of the decoder) with those of this line-level \gls{ocr} model.

\subsubsection{Curriculum strategies}
We used two curriculum strategies to improve the convergence by slightly increasing the difficulty of the task during training.

A curriculum strategy was used for the generation of synthetic data during the training of the \gls{dan}. Instead of directly generating whole documents, we progressively increase the number of lines per page contained in the generated documents. We set the minimum number of lines to 1 and the maximum number of lines to $l_\mathrm{max}$, to fit the datasets properties. In addition, we also crop the synthetic document image under the lowest text line during this curriculum stage. This way, we slightly increase both the length of the target sequence, through the number of text lines, and the input image size. 

We used a second curriculum strategy regarding the dropout, as defined in \cite{CurriculumDropout}. It means that the dropout rate $\tau$ evolves during training:
\begin{equation}
    \tau_t = (1 - \bar{\tau}) \exp(- \gamma t) + \bar{\tau}, \gamma > 0,
\end{equation}
where $\bar{\tau}$ is the final dropout rate, $t$ is the number of iterations (weight update) and $\gamma = \frac{1}{T}$, $T$ being the total estimated number of weight updates during training. We set $T=5 \times 10^4$.

\subsubsection{Data Augmentation}
We used a data augmentation strategy with a probability of 90\%. This data augmentation strategy consists in applying some transformations, in random order, with a probability of 10\% for each one. These data augmentation techniques are: resolution modification, perspective transformation, elastic distortion, dilation, erosion, color jittering, gaussian blur, gaussian noise and sharpening. Data augmentation is applied on both synthetic and real images.

\subsubsection{Synthetic data}
We generated synthetic printed lines for pre-training and synthetic printed documents for training. 
To this end, we arbitrarily chose a set of fonts $\mathcal{F}$ to introduce diversity in writing styles. We used these different fonts with various font sizes to bring more variability, making the model more robust. The original dataset $\mathcal{D}_\mathrm{doc}$ is used to extract isolated text line transcriptions $y_i$ associated to a layout class $c_i$, leading to a new dataset  $\mathcal{D}_\mathrm{line}$.
Synthetic lines are generated on the fly during pre-training, by randomly selecting a text line transcription from $\mathcal{D}_\mathrm{line}$.

While generating synthetic documents through learning have been studied in \cite{DocSyn} for instance, here we focused on a rule-based approach for simplicity. 
Algorithm \ref{alg:syn} details this generation process of synthetic documents. It is based on a style sheet $s$ which defines the different layout entities (classes) in the document and a set of constraints on them, such as relative and absolute positioning rules. It also defines properties for each layout entity: maximum height or width in pixels, characters per line, line width or number of lines. 

\begin{algorithm}[h]
\caption{Synthetic document generation.}
\label{alg:syn}
\SetKwInOut{Input}{input}
\SetKwInOut{Output}{output}
\Input{
original document image $\mb{X}$, \\
number of lines $l_\mathrm{doc}$,\\
style sheet $s$,\\
line-level dataset $\mathcal{D}_\mathrm{line} = (\mathcal{Y},\mathcal{C})$,\\
set of fonts $\mathcal{F}$.
}
\Output{
synthetic document image $\mb{D}$,\\
ground truth $y$.
}
$y=$\textquote{ }\;
$l_\mathrm{current} = 0$\;
$H, W = \mathrm{size}(\mb{X})$\;
$\mb{D} = \mathrm{zeros}(H, W)$\;
\While{$l_\mathrm{current} < l_\mathrm{doc}$}{
    $c = \mathrm{get\_next\_layout\_class}(\mb{D}, s)$\;
    $l_\mathrm{entity} = \mathrm{get\_num\_lines}(c, s, l_\mathrm{current}, l_\mathrm{doc})$ \;
    \For{$k=1$ \KwTo $l_\mathrm{entity}$}{
        $y_k = \mathrm{get\_random\_text}(\mathcal{D}_\mathrm{line}, c)$ \;
        $f_k = \mathrm{get\_random\_font}(\mathcal{F})$ \;
        $i_k = \mathrm{generate\_text\_line\_image}(y_k, f_k)$ \;
    }
    $i_\mathrm{entity} = \mathrm{merge\_text\_line\_images}(i_1, ..., i_{l_\mathrm{entity}})$ \;    
    $y_\mathrm{entity} = \mathrm{merge\_text\_line\_gt}(y_1, ..., y_{l_\mathrm{entity}})$ \;
    $\mb{D} = \mathrm{add\_layout\_entity}(\mb{D}, i_\mathrm{entity}, s, c)$ \;
    $y = \mathrm{add\_text}(y, y_\mathrm{entity}, c)$ \;
    $l_\mathrm{current} += l_\mathrm{entity}$ \;
}
$\mb{D} = \mathrm{crop\_under\_lowest\_entity}(\mb{D})$ \;
\end{algorithm}

Synthetic documents are produced on the fly. A document image $\mb{X}$ is randomly chosen from the training dataset $\mathcal{D}_\mathrm{doc}$ to get a realistic document shape, which is used as a template for a synthetic document $\mb{D}$ to be generated. Given the current curriculum number of lines per page $l$ (between 1 and $l_\mathrm{max}$), the actual number of lines for the current synthetic document $l_\mathrm{doc}$ is randomly chosen between 1 and $l$. Layout entities are generated one after the other until reaching $l_\mathrm{doc}$: the layout class is chosen through "get\_next\_layout\_class" based on the style sheet definition and the current state of $\mb{D}$. Given the remaining number of lines ($l_\mathrm{doc}$ - $l_\mathrm{current}$), a random number of lines for the given entity $l_\mathrm{entity}$ is chosen in compliance with $s$. $l_\mathrm{entity}$ synthetic text line images are generated using a random font from $\mathcal{F}$ and a random text line from $\mathcal{D}_\mathrm{line}$. These images are concatenated on the vertical axis, introducing some random indent spacing. The associated ground truth transcriptions are also concatenated in the same order. The generated layout entity is then placed into $\mb{D}$ in compliance with $s$. Ground truth transcription of each layout entity are concatenated by adding the corresponding layout tokens, leading to the ground truth of the whole synthetic document $y$. The document image is cropped under the lowest layout entity, as part of the curriculum strategy.

To sum up, we introduced variability in many points to generate different synthetic examples:
\begin{itemize}
    \item different fonts and font sizes for the writing style.
    \item randomness and flexibility in the positioning constraints for the layout.
    \item random number of lines and mixed sample text lines for the content.
    \item cropping under the lowest layout entity for the image size.
\end{itemize}

\subsubsection{Teacher forcing}
\label{teacherforcing}
We used teacher forcing at training time to parallelize the computations by predicting the whole sequence at once: the ground truth is used in place of the previously predicted tokens. To make the \gls{dan} robust to errors occurring at prediction time, we introduced some errors in this sequence of pseudo previously predicted tokens. Some tokens are replaced by a random character or layout token. We found 20\% to be a good error rate through experiments.

\subsubsection{Pre-processings}
To reduce the memory consumption, images are downscaled through a bi-linear interpolation to a 150 dpi resolution. Images are normalized (zero mean, unit variance) based on mean and variance computed on the images of the training set.

\subsubsection{Post-processings}
We used a rule-based post-processing to correct unpaired predicted layout tokens. It is essential to compute the metrics. Let us denote <X> a layout begin token and </X> its associated layout end token.
The post-processing consists in a forward pass on the whole predicted sequence $\hat{\mseq{y}}$ during which only tokens of layout are modified in order to have a coherent global structure. The main rules are: 
\begin{itemize}
    \item a missing end token is added when there are two successive begin tokens.
    \item isolated end tokens are removed. 
\end{itemize}
For instance, omitting text prediction for simplicity, the prediction "<X> <Y> </Y> </Z>" becomes "<X> </X> <Y> </Y>".

In addition, the post processing ensures that the prediction is in accordance with the layout token grammar, \textit{i. e.} the hierarchical relation between tokens are correct. It means that if a layout entity of class A can only be in an entity of class B, by definition, missing tokens are added. For example, the prediction "<A> </Y>" becomes "<B> <A> </A> </B>"
 
We used a second post-processing, for the text prediction: duplicated space characters are removed from the predicted sequence.

\subsection{Metrics}
\label{section-metrics}
The proposed approach aims at jointly recognizing both text and layout. While there exist well established metrics to measure the performance of both tasks independently, we are not aware of an adequate metric to evaluate both tasks when performed altogether.
To our knowledge, there is no prior work handling such task. Therefore, we propose the evaluation of our approach using three different angles: the text recognition only, the layout recognition only and the joint recognition of both text and layout, using two new metrics named \gls{loer} and $\mathrm{mAP}_\mathrm{CER}$.

In the following we will take as example the following predicted sequence $\hat{\mseq{y}}$, after post-processing:\\ 
"<X>text1</X><B><A>text2</A><A>text3</A></B>"

\subsubsection{Evaluation of the text recognition}
To evaluate the text recognition, all layout tokens are removed from the ground truth $\mseq{y}$ and from the prediction $\hat{\mseq{y}}$, leading to $\mseq{y}^\mathrm{text}$ and $\hat{\mseq{y}}^\mathrm{text}$, respectively. \\
Here, $\hat{\mseq{y}}^\mathrm{text}=$   "text1text2text3".

We used the standard \gls{cer} and the \gls{wer} to evaluate the performance of the text recognition. 
They are both computed as the sum of the Levenshtein distances (noted $\mathrm{d}_\mathrm{lev}$) between the ground truths and the predictions, at document level, normalized by the total length of the ground truths $\mseq{y}_{\mathrm{len}_i}^\mathrm{text}$. For K examples in the dataset:
\begin{equation}
    \mathrm{CER} = \frac{\displaystyle \sum_{i=1}^K \mathrm{d_\mathrm{lev}}(\hat{\mseq{y}}_i^\mathrm{text}, \mseq{y}_i^\mathrm{text})}
                {\displaystyle\sum_{i=1}^K{\mseq{y}_{\mathrm{len}_i}^\mathrm{text}}}.
\end{equation}
\gls{wer} is computed in the same way but at word level. Punctuation characters are considered as words. 

One should note that, contrary to text line or paragraph recognition, the reading order is far more complicated to learn. An inversion in the reading order between two text blocks can severely impact the \gls{cer} and \gls{wer} values, even with correctly recognized text blocks.

\subsubsection{Evaluation of the layout recognition}
We cannot use existing \gls{dla} metrics, such as \gls{iou}, \gls{map} \cite{mAP_PascalVOC} or ZoneMap \cite{ZoneMap}, to evaluate the layout recognition, because they are based on physical layout (segmentation) annotations.

We decided to model the layout as an oriented graph to take into account both the reading order and the hierarchical relations between layout entities. To evaluate the layout recognition, we introduce a new metric: the \gls{loer}. 
To this end, we associate to each ground truth and prediction a graph representation: $\mseq{y}^\mathrm{graph}$ and $\hat{\mseq{y}}^\mathrm{graph}$, as shown in Figure \ref{fig:dan-dataset}. 
We propose to generate this graph representation in two steps. First, we compute $\mseq{y}^\mathrm{layout}$ and $\hat{\mseq{y}}^\mathrm{layout}$, the ground truth and the prediction from which all but layout tokens are removed. \\
Here, $\hat{\mseq{y}}^\mathrm{layout}=$ "<X></X><B><A></A><A></A></B>".\\
Second, we map this sequence of layout tokens into a graph following the hierarchical rules of the datasets. 

We designed \gls{loer} following the same paradigm as \gls{cer}, adapting it to graph. It is computed as a \gls{ged}, normalized by the number of nodes and edges in the ground truth. As shown in Figure \ref{fig:dan-dataset}, this graph can be represented by ordering the nodes with respect to a root D which represents the document. This way, the graph can be represented as a multi-level graph where the nodes are the different layout entities, the oriented edges between successive levels (dashed arrows) are their hierarchy and the oriented edges inside a same level (solid arrows) represent their reading order. 

For K samples in the dataset, \gls{loer} is computed as the sum of the graph edit distances,  normalized by the sum of the number of edges $n_\mathrm{e}$ and nodes $n_\mathrm{n}$ in the ground truths:
\begin{equation}
    \mathrm{LOER} = \frac{\displaystyle \sum_{i=1}^K\mathrm{GED}(\mb{y}^\mathrm{graph}_i, \hat{\mb{y}}^\mathrm{graph}_i)}{\displaystyle \sum_{i=1}^K n_{\mathrm{e}_i} + n_{\mathrm{n}_i}}.
\end{equation}
The graph edit distance is computed using a unit cost of edition whether it is for addition, removal or substitution and whether it is for edges or nodes. This computation becomes intractable in a reasonable running time for multiple pages. We circumvented this issue by assuming that the prediction of the page tokens was done in the right order. In this way, the \gls{ged} of a document with several pages corresponds to the sum of the \gls{ged} computed on the sub-graphs representing the isolated pages. Missing ground truth or prediction sub-graphs are compared to null graph.

Combining \gls{cer} and \gls{loer} is not sufficient to evaluate the correct recognition of the document. As a matter of fact, one could reach 0\% for both metrics by predicting first all the character tokens, and then all the layout tokens, each in the correct order. One misses the evaluation of the association between the layout tokens and their corresponding text parts.

\subsubsection{Evaluation of joint text and layout recognition}
We propose a second new metric, $\mathrm{mAP}_\mathrm{CER}$, to evaluate the joint recognition of both text and layout. It is based on the standard \gls{map} score used for object detection approaches \cite{mAP_COCO,mAP_PascalVOC}; but instead of using the \gls{iou} to consider a prediction as true or false, we use the \gls{cer}. It is computed as follows:
\begin{itemize}
    \item The predicted sequence $\hat{\mseq{y}}$ and the ground truth sequence $\mseq{y}$ are split into sub-sequences. These sub-sequences are extracted using the begin and end tokens of a same class. Then, they are grouped by classes into lists of sub-sequences. Results for our prediction example is given in Table \ref{tab:mAP}.
    
    For each layout class, sub-sequences are ordered by their confidence score, computed as the mean between prediction probabilities associated to the begin and end tokens of this class (probabilities from $\mseq{p}\mt{t}{}$). A predicted sub-sequence is considered as true positive if the \gls{cer} between this sub-sequence and a sub-sequence of the ground truth from the same class is under a given threshold. Otherwise, it is considered as false positive. When associated, the used sub-sequences are removed until there is no more sub-sequences in the ground truth or in the prediction.
    
    \begin{table}[h]
     \caption{Sub-sequences are extracted, grouped and ordered by layout classes for $\mathrm{mAP}_\mathrm{CER}$ computation. Left: tokens of the predicted sequence and associated confidence score. Consecutive text tokens are grouped for simplicity, their associated confidence score has been averaged. Right: sub-sequences are extracted by layout tokens and ordered given confidence score.}
     \label{tab:mAP}
        \begin{subfigure}[c]{0.5\textwidth}
             \centering
            \resizebox{0.4\textwidth}{!}{
            \begin{tabular}{c c}
            Token & Confidence\\
            \hline
            <X> & 90\% \\
            text1 & 95\% \\
            </X> & 70\% \\
            <B> & 95\% \\
            <A> & 82\% \\
            text2 & 73\% \\
            </A> & 86\% \\
            <A> & 80\% \\
            text3 & 89\% \\
            </A> & 80\% \\
            </B> & 75\% \\
           \end{tabular}
           }
        \end{subfigure}
        \hfill
        \begin{subfigure}[c]{0.5\textwidth}
            \centering
            \resizebox{0.4\textwidth}{!}{
            \begin{tabular}{c c}
            Score & Text \\
            \hline
            \multicolumn{2}{l}{\textbf{X}} \\
            80\% & text1\\
            \multicolumn{2}{l}{\textbf{A}} \\
            84\% & text2\\
            80\% & text3\\
            \multicolumn{2}{l}{\textbf{B}} \\
            85\% & text2text3\\

            \end{tabular}
            }
        \end{subfigure}
\end{table}
    
    \item The average precision $\mathrm{AP}_c$ for a layout class $c$ corresponds to the Area Under the precision-recall Curve (AUC). As in \cite{mAP_PascalVOC}, it is computed as an approximation by summing the rectangular areas under the curve, formed by each modification of the precision $p_n$ and recall $r_n$:
    \begin{equation}
        \mathrm{AP}_{\mathrm{CER}_c} = \sum (r_{n+1} - r_n) \cdot p_\mathrm{interp}(r_{n+1}),
    \end{equation}
    with 
    \begin{equation}
        p_\mathrm{interp}(r_{n+1}) = \max_{\tilde{r}>r_{n+1}} p(\tilde{r}).
    \end{equation}
    
    \item The average precision is itself averaged for different \gls{cer} thresholds, between $\theta_\mathrm{min} = 5\%$ and $\theta_\mathrm{max} = 50\%$ with a step $\Delta\theta = 5\%$, leading to 10 thresholds:
    \begin{equation}
        \mathrm{AP}_{\mathrm{CER}_c}^{5:50:5} = \frac{1}{10} \displaystyle \sum_{k=1}^{10}\mathrm{AP}_{\mathrm{CER}_c}^{5k}.
    \end{equation}
    
    \item The mean average precision for a document is then computed as a weighted sum over the different layout classes, weighted by the number of characters $\mathrm{len}_c$ in each class $c$:
    \begin{equation}
        \mathrm{mAP}_\mathrm{CER} = \frac{\displaystyle \sum_{c \in S} \mathrm{AP}_{\mathrm{CER}_c}^{5:50:5} \cdot \mathrm{len}_c }{\displaystyle \sum_{c \in S}  \mathrm{len}_c}
    \end{equation}
    
    \item Finally, the global mAP for a set of documents is computed by averaging the mAP of the different documents, weighted by the number of characters in each document.
\end{itemize}

This way, the $\mathrm{mAP}_\mathrm{CER}$ gives an idea of how well the text regions have been classified, based on the recognized text. It could not be used with the \gls{cer} only because it does not evaluate the order of the classified text regions.  Combining \gls{cer}, \gls{loer} and $\mathrm{mAP}_\mathrm{CER}$ enables to have a real estimation of the quality of the prediction for joint text and layout recognition.

However, one could argue that the layout recognition performance is biased by the post-processing we used. To evaluate the impact of the post-processing in the final results, we defined a dedicated metrics: the \gls{pper}. It is used to understand how much of the layout recognition is due to the raw network prediction and how much is due to the post-processing. It is defined by the number of post-processing edition operations (addition or removal of layout tokens) $n_\mathrm{ppe}$, normalized by the number of layout tokens in the ground truth $\mb{y}^{\mathrm{layout}}_\mathrm{len}$. For $K$ examples:
\begin{equation}
    \mathrm{PPER} = \frac{ \sum_{i=1}^{K} n_{\mathrm{ppe}_i}}{\sum_{i=1}^{K} \mb{y}^{\mathrm{layout}}_{\mathrm{len}_i}}.
\end{equation}
One should keep in mind that this metric only quantifies how much of the final layout prediction is due to post-processing through edition operations: these modifications can be either beneficial or unfavorable.

\subsection{Experiments}
\label{section-dan-experiments}
This section is dedicated to the evaluation of the \gls{dan} for document recognition. We evaluate the \gls{dan} on the RIMES 2009 and READ 2016 datasets. We provide a visualization of the attention process and of the predictions. We also provide an ablation study to highlight the key components that made it possible to achieve these results.

\subsubsection{Datasets}
We evaluated the \gls{dan} on two public handwritten document datasets: RIMES \cite{RIMES_page,RIMES_paragraph} and READ 2016 \cite{READ2016}.

\subsubsection*{RIMES}
We used the RIMES 2009 dataset, as detailed in Section \ref{section:dataset}. We used two kinds of annotation from this page-level dataset: transcription ground truth and layout analysis annotations. Text regions are classified among 7 classes: sender (S), recipient (R), date \& location (W), subject (Y), opening (O), body (B) and PS \& attachment (P). We used these classes and associated them to the corresponding text part: we do not use any positional ground truth to train the proposed model. We corrected 3 annotations where at least one text line was missing. Sometimes, the annotators proposed two options for a given word or expression, we systematically selected the first one in every cases. 

Since there is no other work evaluating their model on this dataset, we also evaluate the performance of the \gls{dan} on the RIMES 2009 dataset at paragraph-level, as well as on the RIMES 2011 dataset \cite{RIMES_paragraph} at paragraph and line levels. The idea is to compare the recognition performance between the different segmentation levels and with the state-of-the-art approaches which rely on pre-segmented input at line or paragraph level. One should notice that the paragraphs from RIMES 2011 are not extracted from RIMES 2009 so we cannot directly compare the results on both datasets, but it is the nearest comparison we can have.

\subsubsection*{READ 2016}
We used the READ 2016 dataset at page-level. We also generated a double-page version of this same dataset by concatenating the images of successive pages. Only few pages of the original dataset are not paired and have been removed. Based on their positions, we automatically added a class to each text regions among the 5 following classes: page (P), page number (N), body (B), annotation (A) and section (S) (group of linked annotation(s) and body). For comparison purposes, we also evaluate the \gls{dan} on the READ 2016 dataset at line and paragraph levels.

\begin{figure*}[h!]
\centering
    \begin{subfigure}{0.49\textwidth}
    \centering
    \includegraphics[width=\textwidth]{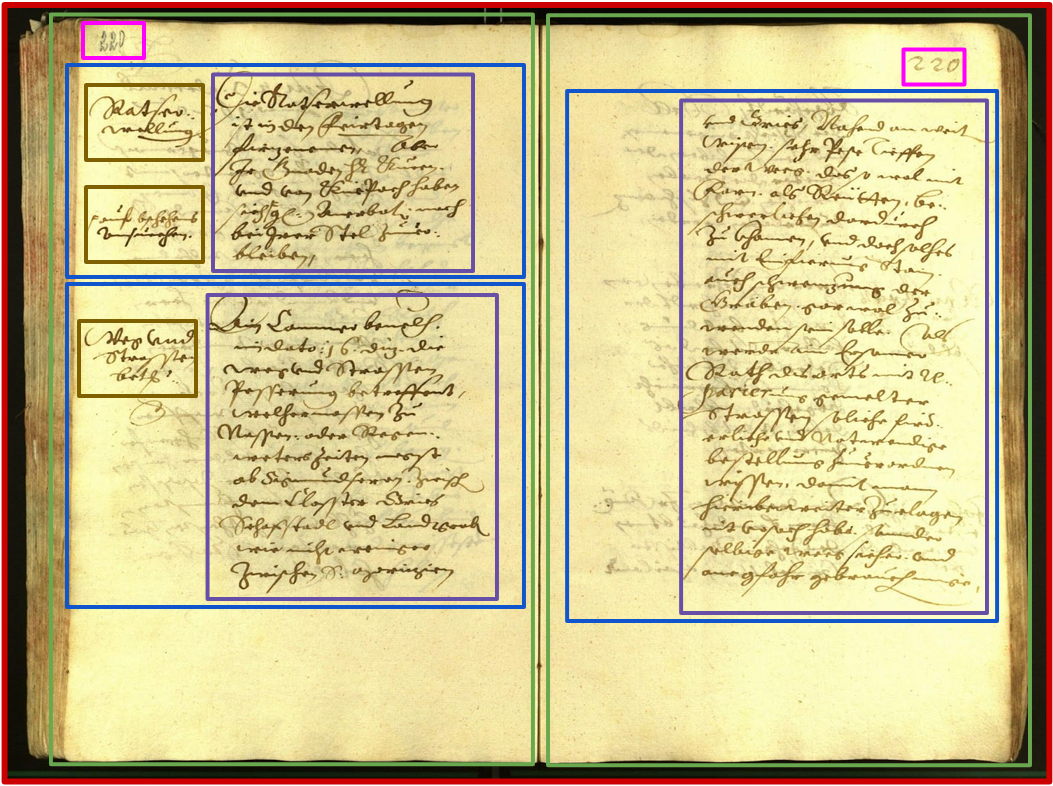}   
    \par\medskip
    \includegraphics[width=0.7\textwidth]{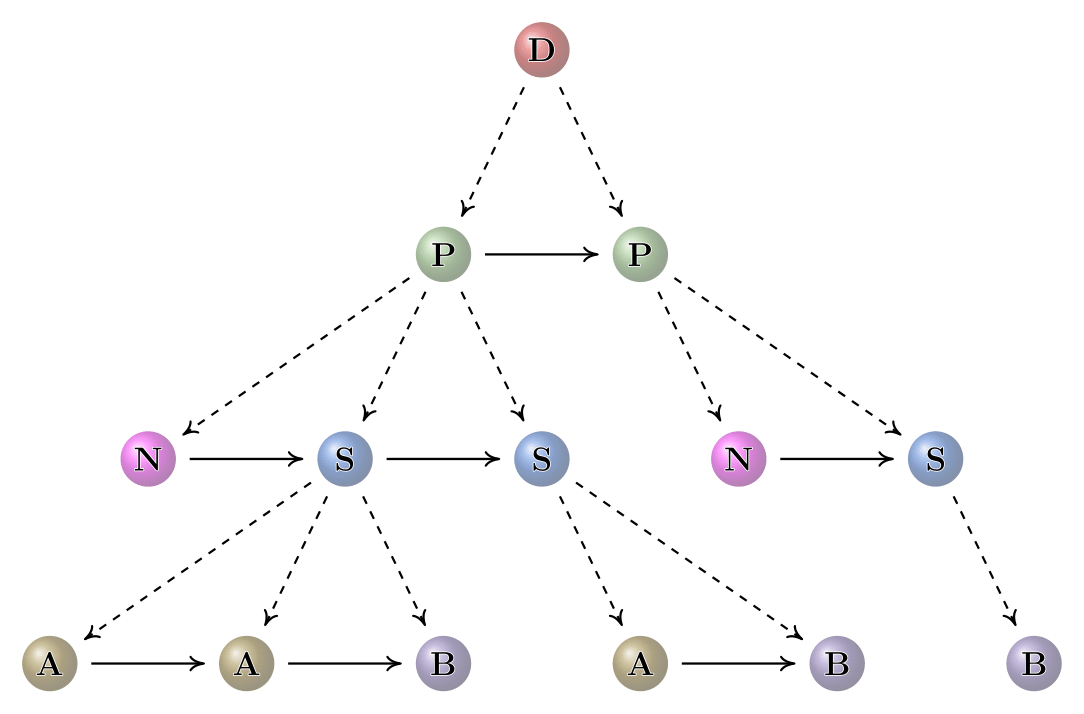}   
    \end{subfigure}
    \hfill
    \begin{subfigure}{0.49\textwidth}
    \centering
    \includegraphics[width=0.7\textwidth]{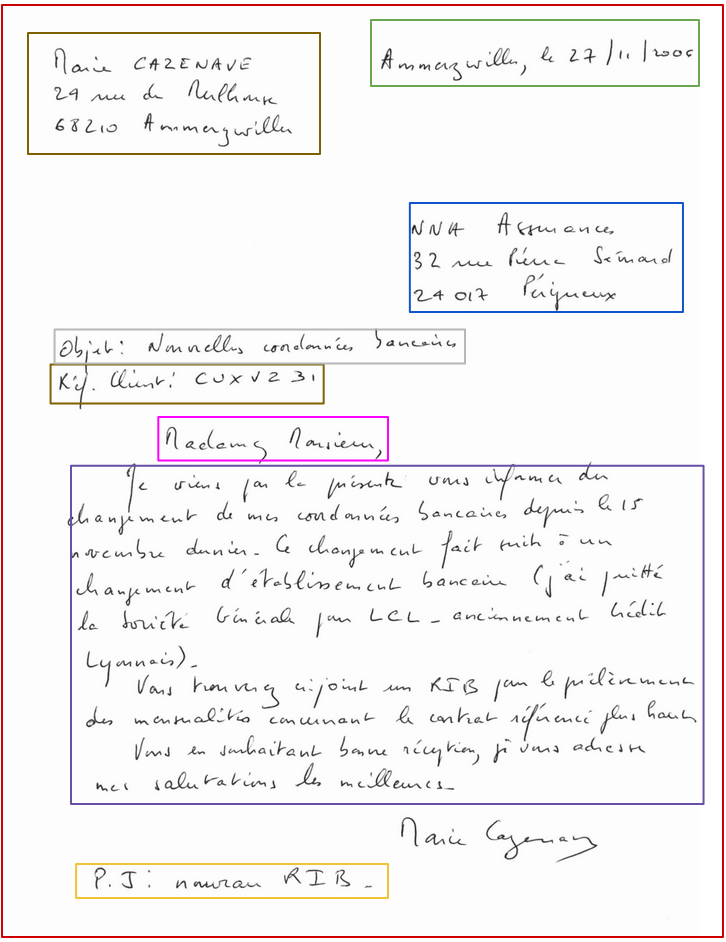}   
    \par\medskip 
    \includegraphics[width=0.9\textwidth]{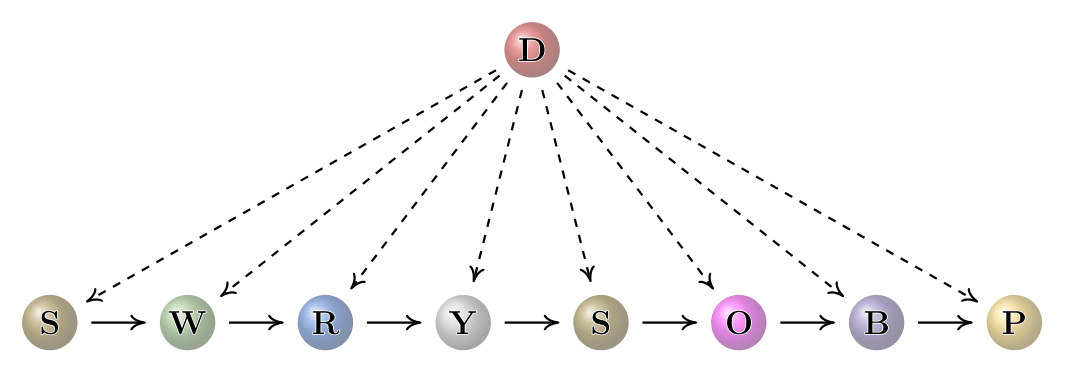}  
    \end{subfigure}

    \caption{Images from READ 2016 and RIMES 2009 and associated layout graph annotation.}
\label{fig:dan-dataset}
\end{figure*}

For both RIMES 2009 and READ 2016 datasets at page or double-page level, the reading order is deduced automatically from the paragraph positions as follows. 
For READ 2016, the reading order is from page to page: first, the page number, then section by section. Among a section, all annotations are read before the body.
For RIMES 2009, text regions are read from top to bottom. If two text regions share a same vertical space, text regions are read from left to right.
An example of each dataset is depicted in Figure \ref{fig:dan-dataset}. Text regions are represented by bounding boxes, colored given their class. We also represented the expected reading order as an oriented graph linking the different text regions. Both datasets have their own specificities. READ 2016 has a more regular layout compared to RIMES 2009 but it includes hierarchical layout tokens: bodies are included into sections, itself included into pages. In addition, it can contain multiple pages. RIMES 2009 provides more variability regarding the layout, but the layout tokens are sequential.

Datasets are split into training, validation and test sets, as detailed in Table \ref{table:split}. It corresponds to official or mainly used splits by the community. We also provide the number of characters for each dataset, as well as the number of layout tokens (two by class: begin and end). We also provide this information at line and paragraph levels for comparison purposes.

\begin{table}[h!]
    \caption{Datasets split in training, validation and test sets and associated number of tokens in their alphabet.}
    \centering
    \resizebox{0.75\linewidth}{!}{
    \begin{tabular}{ c c c c c c c}
    \hline
    \multirow{2}{*}{Dataset} & \multirow{2}{*}{Level} & \multirow{2}{*}{Training} & \multirow{2}{*}{Validation} & \multirow{2}{*}{Test} & \# char  & \# layout\\ 
     & & & & & tokens & tokens\\
    \hline
    \hline     
    \multirow{2}{*}{RIMES 2011} & Line & 10,530 & 801 &  778 & 97 & \xmark\\
    & Paragraph & 1,400 & 100 & 100 & 98 & \xmark\\ 
    \hline  
    \multirow{2}{*}{RIMES 2009} & Paragraph & 5,875 & 540 &  559 & 108 & \xmark\\
    & Page & 1,050 & 100 & 100 & 108 & 14\\
    \hline  
    \multirow{4}{*}{READ 2016} & Line & 8,367 & 1,043 & 1,140 & 88 & \xmark\\
    & Paragraph & 1,602 & 182 & 199 & 89 & \xmark\\
    & Page & 350 & 50 & 50 & 89 & 10\\
    & Double page & 169 & 24 & 24 & 89 & 10\\

    \hline
    \end{tabular}
    }
    \label{table:split}
\end{table}

\subsubsection{Training details}
Pre-training and training are carried out with the same following configuration:
\begin{itemize}
    \item Pytorch framework with automatic mixed precision.
    \item Training with a single \acrshort{gpu} Tesla V100 (32Gb).
    \item Adam optimizer with an initial learning rate of $10^{-4}$.
    \item We use exactly the same hyperparameters for both datasets.
    \item We do not use any external data, external language model nor lexicon constraints.
\end{itemize}

\begin{figure}[ht]
    \centering
    \begin{subfigure}[b]{\linewidth}
        \centering
    {
    \setlength{\fboxsep}{0pt}%
    \setlength{\fboxrule}{1pt}%
    \fbox{\includegraphics[width=0.6\linewidth]{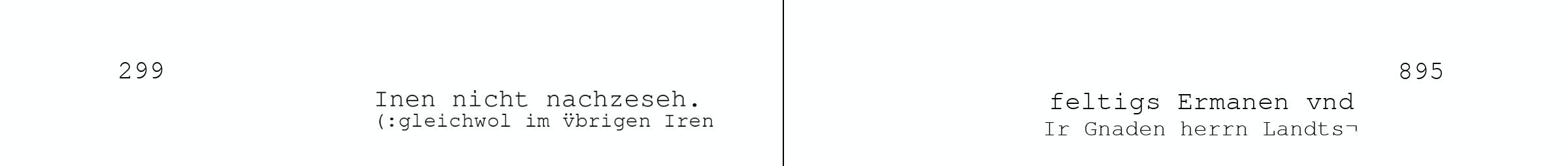}}%
    }
    \caption{$l=3$.}
    \end{subfigure}
    \par\medskip
    \begin{subfigure}[b]{\linewidth}
        \centering
        {
    \setlength{\fboxsep}{0pt}%
    \setlength{\fboxrule}{1pt}%
    \fbox{\includegraphics[width=0.6\linewidth]{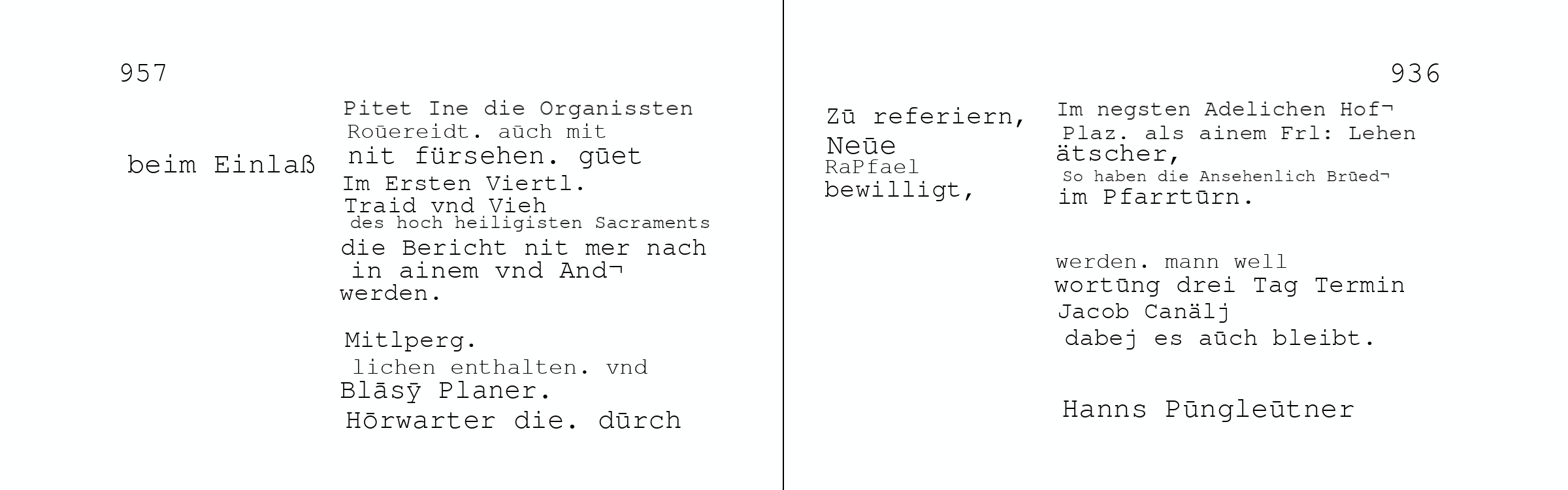}}%
    }
    \caption{$l=15$.}
    \end{subfigure}
    \par\medskip
    \begin{subfigure}[b]{\linewidth}
        \centering
        {
    \setlength{\fboxsep}{0pt}%
    \setlength{\fboxrule}{1pt}%
    \fbox{\includegraphics[width=0.6\linewidth]{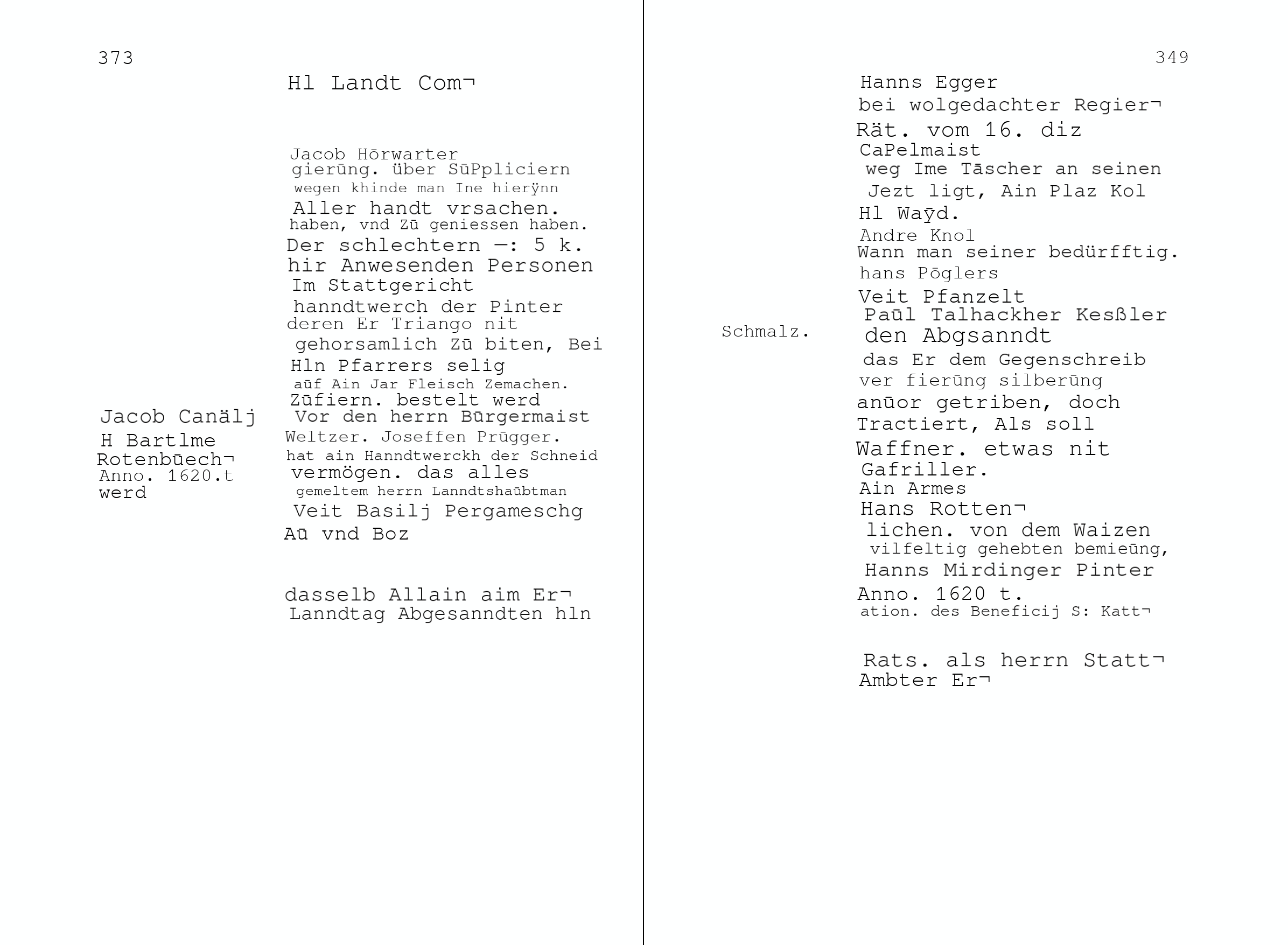}}%
    }
    \caption{$l=l_\mathrm{max}=30$ (end of curriculum stage, no crop).}
    \end{subfigure}
    
    \caption{Illustration of the curriculum learning strategy through the synthetic document image generation process for the READ 2016 dataset at double-page level.}
    \label{fig:syn}
\end{figure}

For the generation of synthetic documents, we set the maximum number of lines per page $l_\mathrm{max}$ to 30 for READ 2016 and to 40 for RIMES 2009 to match the dataset properties. Given a set of arbitrarily-chosen fonts, we only kept those for which all characters were supported, leading to 41 fonts for READ 2016 and 95 for RIMES 2009. The font lists are provided with the code for reproducibility purposes. Figure \ref{fig:syn} illustrates the curriculum process for the generation of synthetic documents for the READ 2016 dataset at double-page level. As one can note, these synthetic documents are far from being visually realistic compared to the original dataset. This does not matter since the objective here is only to learn the reading order.

We also evaluate the \gls{dan} at paragraph and line levels for comparison purposes. For each training of a same dataset, at paragraph and page levels of RIMES 2009 for instance, the same pre-trained weights are used to initialize the model. Each evaluation corresponds to a specific training. Evaluation at paragraph or line levels corresponds to training only on synthetic and real paragraphs or lines, respectively. 

\subsubsection{Evaluation}

To our knowledge, there is no work evaluating their system on the RIMES 2009 and READ 2016 datasets at page level. Comparison at paragraph and line levels are carried out with approaches under similar conditions \textit{i. e.} without external data nor external language model. In Table \ref{table:rimes}, we present the evaluation of the \gls{dan} on the RIMES 2009 dataset at paragraph and page levels. One can notice that we reach very satisfying results at page level for both text and layout recognition with a \gls{cer} of 4.54\%, a \gls{wer} of 11.85\%, a \gls{loer} of 3.82\% and a $\mathrm{mAP}_\mathrm{CER}$ of 93.74\%.
The closest dataset with which we can compare is the RIMES 2011 dataset at paragraph level \cite{RIMES_paragraph}. The \gls{dan} achieves new state-of-the-art results at paragraph level and competitive results at line level on this dataset. One should keep in mind that, as said previously, the comparison between RIMES 2009 and RIMES 2011 at paragraph level is not fair because RIMES 2011 only contains body images whose content seems easier to recognize than that of RIMES 2009. Unique character sequences representing dates, postal codes, product and client references, or even proper nouns like names and places, are mainly in the other text regions. 
In addition, the body images from the RIMES 2011 dataset are not taken from the page images of RIMES 2009. This explains the \gls{cer} difference between RIMES 2009 (5.46\%) and RIMES 2011 (1.82\%) at paragraph level. One can notice a \gls{cer} improvement from line to paragraph level (for RIMES 2011) and from paragraph to page level (for RIMES 2009). This highlights the negative impact of using a prior segmentation step which is prone to annotation variations or errors.

\begin{table}[ht]
    \caption{Evaluation of the DAN on the test set of the RIMES datasets and comparison with the state-of-the-art approaches.}
    \centering
    \resizebox{\linewidth}{!}{
    \begin{threeparttable}[b]
        \begin{tabular}{ l l c c c c c}
        \hline
        Dataset & Approach & \gls{cer} (\%) $\downarrow$ & \gls{wer} (\%) $\downarrow$ & \gls{loer} (\%) $\downarrow$ & $\mathrm{mAP}_\mathrm{CER}$ (\%) $\uparrow$ & \gls{pper} (\%) $\downarrow$\\ 
        \hline
        \hline
        \multirow{9}{*}{\shortstack{RIMES\\2011\\\cite{RIMES_paragraph}}}& \textbf{Line level}\\
        & \cite{Coquenet2022} FCN & 3.04 & 8.32 & \xmark & \xmark & \xmark \\
        & \cite{Puigcerver2017} CNN+BLSTM\tnote{a} & \textbf{2.3} & 9.6 & \xmark & \xmark & \xmark\\
        & \cite{Coquenet2022b} \gls{dan} (FCN+transformer)\tnote{c} & 2.63 & \textbf{6.78} & \xmark & \xmark & \xmark\\
        & \textbf{Paragraph level}\\
        & \cite{Coquenet2021} \acrshort{span} (FCN)  & 4.17 & 15.61 & \xmark & \xmark & \xmark \\
        & \cite{Bluche2016} CNN+MDLSTM\tnote{b} & 2.9 & 12.6 & \xmark & \xmark & \xmark \\
        & \cite{Coquenet2022} \acrshort{van} (FCN+LSTM)\tnote{b}  & 1.91 & 6.72 & \xmark & \xmark & \xmark \\
        & \cite{Coquenet2022b} \gls{dan} (FCN+transformer)\tnote{c} & \textbf{1.82} & \textbf{5.03} & \xmark & \xmark & \xmark\\
        \hline
        \multirow{4}{*}{\shortstack{RIMES\\2009\\\cite{RIMES_page}}} & \textbf{Paragraph level}\\
        & \cite{Coquenet2022b} \gls{dan} (FCN+transformer)\tnote{c} & 5.46 & 13.04 & \xmark & \xmark & \xmark \\
        & \textbf{Page level}\\
        & \cite{Coquenet2022b} \gls{dan} (FCN+transformer)\tnote{c} & 4.54 & 11.85 & 3.82 & 93.74 & 1.45\\
        \hline
        \end{tabular}
        \begin{tablenotes}
            \item [a] This work uses a slightly different split (10,203 for training, 1,130 for validation and 778 for test).
            \item [b] with line-level attention.
            \item [c] with character-level attention.
        \end{tablenotes}
    \end{threeparttable}
    }
    \label{table:rimes}
\end{table}

Table \ref{table:read} provides an evaluation of the \gls{dan} on the READ 2016 dataset, at line, paragraph, single-page and double-page levels. As one can note, we achieve state-of-the-art results at line and paragraph levels. We also reach very interesting results whether it is at single-page or double-page level with a \gls{cer} of 3.53\% and 3.69\%, respectively. It corresponds to slightly higher \gls{cer} compared to the paragraph level (3.22\%).
One can note that the \gls{loer} and the $\mathrm{mAP}_\mathrm{CER}$ are also satisfying, highlighting the good recognition of the layout. Moreover, these metrics are slightly better for the double-page level dataset. This could be explained by the higher necessity to understand the layout for complex samples. This is discussed in Section \ref{section:ablation}.

\begin{table}[ht]
    \caption{Evaluation of the DAN on the test set of the READ 2016 dataset and comparison with the state-of-the-art approaches}
    \centering
    \resizebox{\linewidth}{!}{
    \begin{threeparttable}[b]
        \begin{tabular}{ l c c c c c}
        \hline
        Approach & \gls{cer} (\%) $\downarrow$ & \gls{wer} (\%) $\downarrow$ & \gls{loer} (\%) $\downarrow$ & $\mathrm{mAP}_\mathrm{CER}$ (\%) $\uparrow$ & \gls{pper} (\%) $\downarrow$\\ 
        \hline
        \hline
        \textbf{Line level}\\
        \cite{Michael2019} CNN+BLSTM\tnote{a}  & 4.66 & \xmark  & \xmark & \xmark & \xmark \\
        \cite{READ2016} CNN+RNN & 5.1 & 21.1 & \xmark & \xmark & \xmark\\
        \cite{Coquenet2022} \acrshort{van} (FCN+LSTM)\tnote{b}  & \textbf{4.10} & \textbf{16.29} & \xmark & \xmark & \xmark\\
        \cite{Coquenet2022b} \gls{dan} (FCN+transformer)\tnote{a}  & \textbf{4.10} & 17.64 & \xmark & \xmark & \xmark\\
        \textbf{Paragraph level}\\
        \cite{Coquenet2021} \acrshort{span} (FCN) & 6.20 & 25.69 & \xmark & \xmark & \xmark\\
        \cite{Coquenet2022} \acrshort{van} (FCN+LSTM)\tnote{b}  & 3.59 & 13.94 & \xmark & \xmark & \xmark\\
        \cite{Coquenet2022b} \gls{dan} (FCN+transformer)\tnote{a}  & \textbf{3.22} & \textbf{13.63} & \xmark & \xmark & \xmark\\
        \textbf{Single-page level}\\
        \cite{Coquenet2022b} \gls{dan} (FCN+transformer)\tnote{a}  & 3.53 & 13.33 & 5.94 & 92.57 & 0.15\\
        \textbf{Double-page level}\\
        \cite{Coquenet2022b} \gls{dan} (FCN+transformer)\tnote{a}  & 3.69 & 14.20 & 4.60 & 93.92 & 1.37\\
        \hline
        \end{tabular}
        \begin{tablenotes}
            \item [a] with character-level attention.
            \item [b] with line-level attention.
        \end{tablenotes}
    \end{threeparttable}
    }
    \label{table:read}
\end{table}

One should note that these \gls{cer} values, for both RIMES 2009 and READ 2016 datasets, should not be compared directly to paragraph-level or line-level HTR approaches. Indeed, it is necessary to understand the impact of the reading order. \gls{cer} is computed based on the edit distance between two one-dimensional sequences. This way, if the reading order is wrong, it impacts severely the \gls{cer}. For instance, if the sender coordinates are read before the recipient coordinates while it is the opposite in the ground truth, the edit distance will be very important even if the text is well recognized. It means that part of this \gls{cer} is due to a wrong reading order and not to a wrong text recognition.
However, $\mathrm{mAP}_\mathrm{CER}$ would not be impacted: it is invariant to the reading order between the different text regions.

For both datasets, the \gls{pper} metric is very low. It indicates that the good results obtained for the layout recognition are mainly due to the \gls{dan} itself and not to the post-processing stage.

We now focus on the $\mathrm{mAP}_\mathrm{CER}$ metric with the READ 2016 dataset at double-page level. In Table \ref{tab:map_read}, we detailed this metric for each layout class and for each threshold of \gls{cer}. As one can notice, pages, page numbers, sections and bodies are very well recognized with at least 93\% for a \gls{cer} as of 10\%. The \gls{dan} has more difficulty with the annotations, with an average of 81.64\%. We assume that is due to two main points: it is the layout entity with the more variability. There can be zero, one or multiple annotations per body. In addition, they can be placed wherever along the body, from its beginning to its end. Second point is about the length of the annotations: they are very shorter than bodies. This way, only few errors can lead to an important \gls{cer} increase, leading to lower average precision.
\begin{table}[ht]
    \caption{$\mathrm{mAP}_\mathrm{CER}$ detailed for each class and each CER threshold, for the READ 2016 double-page dataset.}
    \centering
    \resizebox{\linewidth}{!}{
    \begin{tabular}{ l c c c c c c c c c c c}
    \hline
    & 
    $\mathrm{AP}_\mathrm{CER}^{5:50:5}$ & $\mathrm{AP}_\mathrm{CER}^{5}$ & $\mathrm{AP}_\mathrm{CER}^{10}$ & $\mathrm{AP}_\mathrm{CER}^{15}$ &
    $\mathrm{AP}_\mathrm{CER}^{20}$ & $\mathrm{AP}_\mathrm{CER}^{25}$ & $\mathrm{AP}_\mathrm{CER}^{30}$ &
    $\mathrm{AP}_\mathrm{CER}^{35}$ & $\mathrm{AP}_\mathrm{CER}^{40}$ & $\mathrm{AP}_\mathrm{CER}^{45}$ & $\mathrm{AP}_\mathrm{CER}^{50}$\\
    \hline
    \hline
    Page (P) & 97.19 & 78.12 & 93.75 & 100.00 & 100.00 & 100.00 & 100.00 & 100.00 & 100.00 & 100.00 & 100.00\\
    Page number (N) & 98.12 & 96.88 & 96.88 & 96.88 & 96.88 & 96.88 &96.88 & 100.00 & 100.00 & 100.00 & 100.00 \\
    Section (S) & 96.52 & 78.04 & 93.33 & 97.78 & 98.89 & 98.89 & 98.89 & 99.84 & 99.84 & 99.84 & 99.84\\
    Annotation (A) & 81.64 & 47.78 & 66.27 & 76.31 & 85.78 & 89.71 & 89.71 & 89.71 & 89.71 & 89.71& 91.67\\
    Body (B) & 94.33 & 84.63 & 93.15 & 95.19 & 95.19 & 95.19 & 95.19& 96.19& 96.19 & 96.19& 96.19\\
    \hline
    \end{tabular}
    }
    \label{tab:map_read}
\end{table}

\subsubsection{Visualization}
\begin{figure*}[ht]
    \centering
    \includegraphics[width=0.95\textwidth]{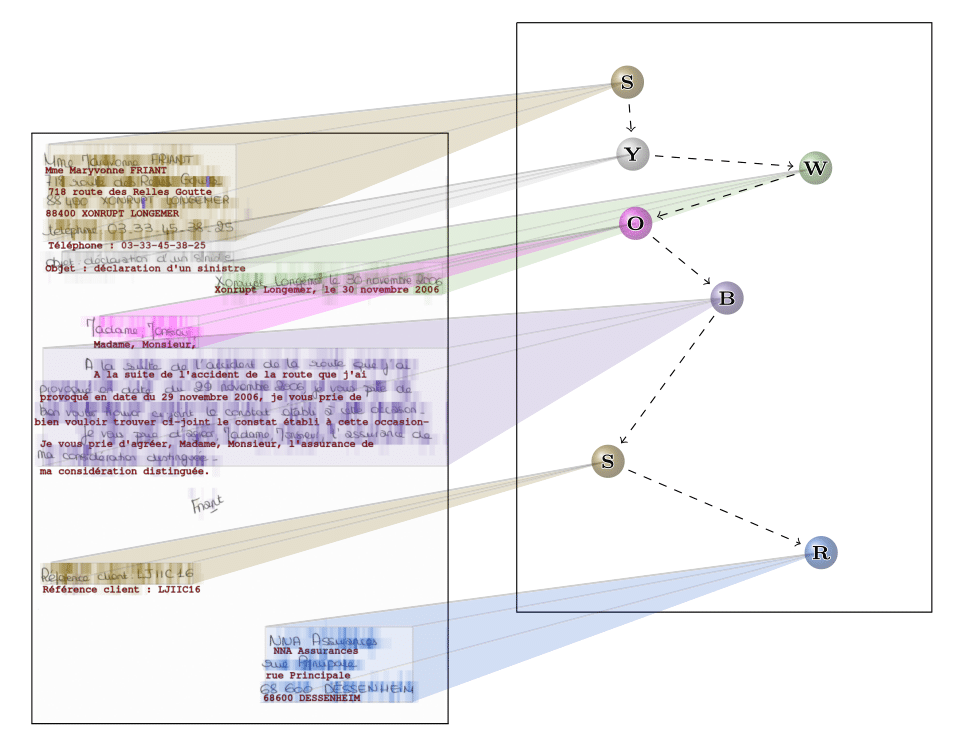}
    \caption{Visualization of the prediction on a RIMES 2009 test sample.}
    \label{fig:dan-viz}
\end{figure*}

A visualization of the prediction for a test sample of the RIMES 2009 dataset is depicted in Figure \ref{fig:dan-viz} \footnote{Full demo at  \url{https://www.youtube.com/watch?v=HrrUsQfW66E}}. On the left, attention weights of the last mutual attention layer are projected on the input image. The colors depend on the last predicted layout token. For visibility, the intensity of the colors is encoded for attention values between 0.02 and 0.25. The text prediction is added in red line by line, under the associated text line. The corresponding layout graph is depicted on the right, with each node corresponding to a text region in the input image. As one can note, even if the \gls{dan} is not trained using any segmentation label, it performs a kind of implicit segmentation in its process, which can be globally retrieve through the attention weights. 
As one can notice, the \gls{dan} performs a document recognition: it recognizes both text and layout.

\begin{figure}
\centering
\includegraphics[width=0.75\linewidth,frame]{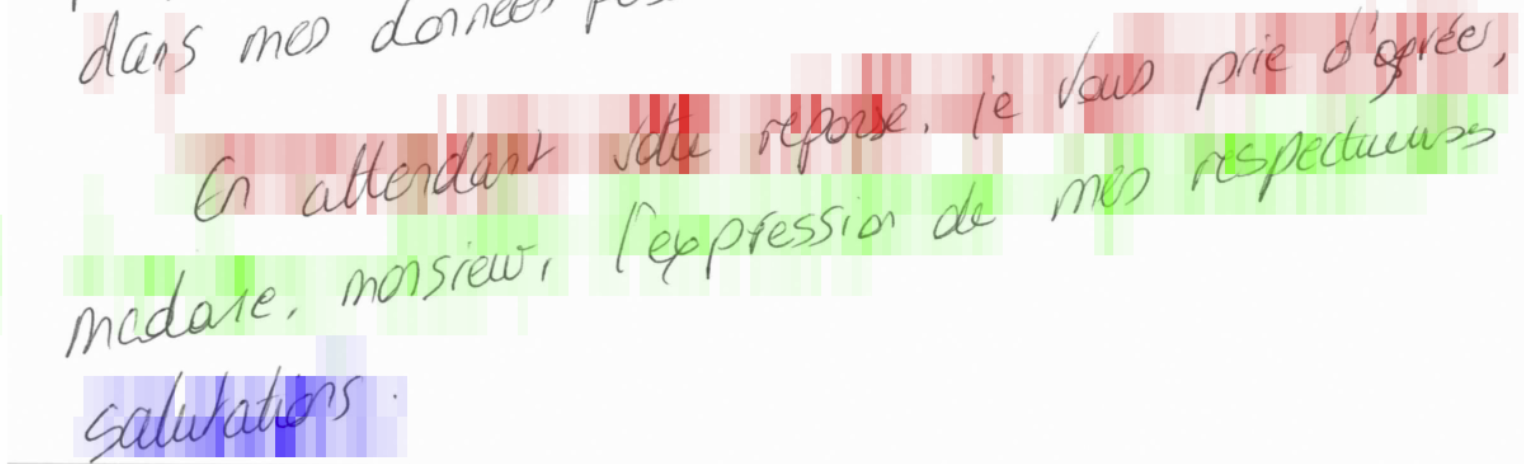}
\par
{\centering
\scriptsize{Prediction: 
\begin{minipage}[t]{0.5\textwidth}
"En attendant votre réponse, je vous prie d'agréer, \\Madame, Monsieur, l'expression de mes respectueuses\\ salutations."
\end{minipage}
}}
\caption{Attention weights visualization. Focus on slanted lines of a validation sample of the RIMES 2009 dataset.}
\label{fig:viz_slanted_lines}
\end{figure}

In addition, the \gls{dan} is able to deal with slanted lines through its character-level attention mechanism. A prediction visualization for such example is depicted in Figure \ref{fig:viz_slanted_lines}. This time, we used one color per line for visibility. As one can note, the attention mechanism correctly follows the slope of the lines, leading to no error in prediction. This is an improvement compared to approaches based on line-level attention, which cannot handle such close and slanted lines.

\subsubsection{Ablation study}
\label{section:ablation}

We provide an extensive ablation study in Table \ref{table:ablation}. The evaluation is carried out on the test set of the RIMES 2009 dataset at page level and the READ 2016 dataset at double-page level, after a 2-day training. We evaluated the proposed training approach through the independent removal of each of the training component we used.

\begin{table}[h!]
    \caption{Ablation study of the DAN on the RIMES 2009 and READ 2016 datasets. Results are given for the test sets for a 2-day training. All metrics are given in percentages.}
    \centering
    \resizebox{1\linewidth}{!}{
    \begin{tabular}{ l | c c c c | c c c c }
    \hline
    & \multicolumn{4}{c|}{RIMES 2009 (single-page)}& \multicolumn{4}{c}{READ 2016 (double-page)}\\
    & \gls{cer} $\downarrow$ & \gls{wer}  $\downarrow$ & \gls{loer} $\downarrow$ & $\mathrm{mAP}_\mathrm{CER}$ $\uparrow$  & \gls{cer}  $\downarrow$ & \gls{wer} $\downarrow$ & \gls{loer} $\downarrow$ & $\mathrm{mAP}_\mathrm{CER}$ $\uparrow$\\ 
    \hline
    \hline
    Baseline & 5.72 & 13.05 & \textbf{4.18} & \textbf{92.86} &  \textbf{4.36} & 16.55 & \textbf{3.80} & \textbf{93.78}\\
    (1) No synthetic data  & 8.26 & 16.45 & 8.18 & 86.34  & 80.75 & 95.65 & 36.77 & 0.13\\
    (2) No curriculum for syn. data  & 7.59 & 16.48 & 6.63 & 88.92  & 81.31 & 92.66 & 21.12 & 0.24\\
    (3) No crop in curr. for syn. data & 5.84 & 13.73 & 4.42 & 91.94 & 99.98 & 100.00 & 85.77 & 0.00 \\
    (4) No data augmentation  & 7.08 & 15.54 & 4.78 & 91.65  & 100.00 & 100.00 & 43.45 & 0.00\\
    (5) No curriculum dropout  & 5.83 & 14.41 & 4.36 & 92.09 & 4.59 & \textbf{16.53} & 4.98 & 92.97\\
    (6) No error in teacher forcing  & 8.09 & 15.12 & 5.91 & 89.24 & 5.44 & 18.67 & 4.98 & 90.99\\ 
    (7) No layout recognition & \textbf{5.30} & \textbf{12.46} & \xmark & \xmark  & 100.00 & 100.00 & \xmark & \xmark\\
    (8) No pre-training  & 71.42 & 87.48 & 18.46 & 12.72 & 5.52 & 19.62 & 6.29 & 89.38\\
    (9) No 1D positional encoding & 8.04 & 16.93 & 5.73 & 90.65 & 5.30 & 18.64 & 6.29 & 90.51\\
    (10) No 2D positional encoding & 12.43 & 20.83 & 8.42 & 89.81 & 70.84 & 93.73 & 23.94 & 20.04\\

    \hline
    \end{tabular}
    }
    \label{table:ablation}
\end{table}

Experiments (1) to (3) are dedicated to the generation of synthetic documents at training time. In (1), we do not generate synthetic documents: the model is only trained on real documents. In (2), the synthetic documents do not follow a curriculum approach: the maximum number of lines per document is directly set to its upper bound ($l=l_\mathrm{max}$). In (3), the synthetic document images are not cropped below the lowest text line during the curriculum phase: they preserve the original image height. As one can note, the removal of either of these three aspects prevents the \gls{dan} from converging, on the READ 2016 dataset. It also leads to poorer results on RIMES 2009, for each metric. One can note an increase of 0.12 points for the \gls{cer} in the best case and of 2.54 points in the worst case. The $\mathrm{mAP}_\mathrm{CER}$ is decreased by 0.92 to 6.52 points.
The removal of the data augmentation strategy during both pre-training and training, experimented in (4), leads to similar results.

In (5), we do not use curriculum dropout strategy neither during pre-training nor during training. This time, it does not prevent the \gls{dan} from converging on the READ 2016 datasets. Except for the \gls{wer} for READ 2016, the results are slightly worse. 
In experiment (6), we do not introduce any error in the teacher forcing strategy: we preserve the ground truth. The introduction of errors improves the results. We assume that training the model with errors helps it to deal with the errors made at prediction time.

The layout tokens are removed from the ground truth in (7) leading to text recognition only. As one can notice, it leads to lower \gls{cer} and \gls{wer} for the RIMES 2009 dataset but prevents the convergence for the READ 2016 dataset. This can be explained by the fact that the reading order is easier for RIMES 2009 than for READ 2016, especially at double-page level. We assume that recognizing the different layout entities enables to understand the spatial relation between them and helps to learn the reading order. Annotations are always at the left of a body: after the prediction of an </annotation> token, the attention must be focused righter to predict a <body> token. 

In (8), the \gls{dan} is trained from scratch, without transfer learning from a prior pre-training step. Results are dramatically worse for the RIMES 2009 dataset. We assume that this is due to its irregular layout. Compared to READ 2016, RIMES 2009 provides simpler reading orders but with more variability in the layout. This way, we assume that for READ 2016, the reading order can be learned jointly with the feature extraction part, directly during the curriculum step. For RIMES 2009, the layout variability slows down the learning of the feature extraction part, leading to slower convergence.

Finally, in (9) and (10), we evaluate the importance of the 1D and 2D positional encoding components. As one can notice, they both enables to improve the results, especially the 2D positional encoding. This is especially true for the READ 2016 dataset. We assume this is due to the double-page nature of this dataset which leads to a more complex layout, resulting in more important jumps from one character prediction to another. For example, from the last character of the left page to the first one of the right page.

\subsection{Discussion}
\label{section-dan-discussion}
We proposed a new end-to-end segmentation-free architecture for the task of \acrlong{hdr}. To our knowledge, it is the first end-to-end approach that can deal with whole documents, recognizing both text and layout.
As we have seen, the \gls{dan} reaches great results on both text and layout recognition whether it is at single-page or double-page level, on the RIMES 2009 and READ 2016 datasets. Indeed, we showed that the \gls{dan} is robust to the  dataset varieties: we used the same hyperparameters with two totally different datasets in terms of layout, language, color encoding as well as number of training samples. 

We proposed an efficient training strategy and we highlighted its impact on an extensive ablation study. This training strategy is based on synthetic line and document generation using digital fonts, to overcome the lack of training data. Pre-training is carried out on synthetic text lines only, avoiding to use any segmentation annotations: this is a great advantage compared to state-of-the-art approaches. We showed that even with complex handwritings such as those of the READ 2016 dataset, for which the fonts aspect is very far from the original writings, this approach still remains effective.

However, it should be noted that the datasets are rather specific: letters for RIMES and historical book pages for READ 2016. The obtained results are very dependent on the quality of the synthetic data \textit{i. e.} they must be close to the original dataset, notably in terms of layout. 
We used a reading order automatically generated from pixel positions of the different text regions. This is effective because we used datasets with rather regular layouts. This would be more complex for datasets of documents with more heterogeneous layouts. However, we assume that this approach could be generalized to heterogeneous documents without any problem by labeling a coherent reading order from one example to another. The main issue is the lack of large-scale public datasets for handwritten document recognition. We hope that this contribution will lead to the production of datasets at lower cost: the \gls{dan} only needs the ordered transcription annotation and layout tags, without the need for any segmentation annotation.

The \gls{dan} learns the reading order through the transcription annotations. We observed an interesting effect linked to this. In the specific case of READ 2016 at double-page level, the page number is identical for both left and right pages. The \gls{dan} focuses on the same area to predict both numbers. Technically, this does not impact the performance, but it shows that the network has not fully learned the concept of reading. We assumed that this phenomenon would disappear by learning on heterogeneous layouts.

The \gls{dan} is based on an autoregressive prediction process. This is not a problem at training time since computation are parallelized through teacher forcing. However, this recurrence issue is significant at prediction time: it grows linearly with the number of tokens to be predicted. This way, the average prediction time for a test sample is 5.8s for RIMES and 4.3s for READ 2016 at single-page level. It becomes 9.7s for READ 2016 at double-page level. This can be an obstacle for an industrial application.  We aim at reducing this prediction time in future works.

\section{Conclusion}
We proposed to handle the task of \gls{hdr} which corresponds to the joint recognition of text and layout. This is a new step toward the global understanding of whole handwritten documents. Moreover, we showed that this additional layout recognition enables to improve the text recognition performance. We assume that this is due to a better understanding of the reading order through the layout entity recognition. We proposed the first model to tackle this task, the \gls{dan}, as well as metrics to evaluate the associated performance. We reached very interesting results on two different but relatively homogeneous datasets. These results are comparable to those obtained at paragraph level, without using any segmentation labels, which is very encouraging. In future works, it would be interesting to go a step further, by recognizing handwritten documents with heterogeneous sizes and layouts. The prediction time, which grows linearly with the length of the character sequence, is another area for improvement.
\glsresetall
\chapter{General conclusion and perspectives}
\label{chap:conclusion}

In this thesis, we focused on the problem of \gls{htr}. As we have seen, this image-to-sequence problem is a difficult task, which implies many challenges. The variety of the writing styles, the variable length of the character sequences from one document to another, which can be relatively large, as well as the involved 2D-to-1D transformation between the input image and the expected sequence of characters are the most important ones. \gls{htr} has long been relying on a prior segmentation step at character, word or line level, and more recently at paragraph level. 


We believe that the line-level approach is quite mature. Numerous architectures have been proposed, studied and successfully applied for this task and we showed that \gls{fcn} architectures can compete with recurrent ones. In addition, they can easily be used for transfer learning purposes for paragraph-level or document-level \gls{htr} models.
We assume that the performances of the current line-level systems are mainly limited by the lack of annotated data and we believe that research should now focus on multi-line approaches. 

To this end, we proposed the \gls{span} and demonstrated that paragraph-level \gls{htr} can be handled in such a simple way. This is the first one-shot approach for paragraph-level \gls{htr} that can deal with inputs of variable sizes. 
We proposed the \gls{van} which reached state-of-the-art results on three public datasets, bridging the gap between paragraph-level and line-level recognition performances. 

However, these paragraph-level approaches still have some drawbacks:
\begin{itemize}
    \item As for the other paragraph-level approaches proposed in the literature, the \gls{span} and the \gls{van} are pre-trained on isolated text lines, using additional line-level segmentation annotations.
    \item They rely on prior segmentation and ordering steps, implying cumulative errors between each stage. However, they are evaluated on manually segmented images: the segmentation stage is almost never taken into account by the metrics.
\end{itemize}

We showed that the first issue can be alleviated by designing a synthetic data generation strategy. Indeed, this task enables us to generate printed documents in a rather easy way and which can be used for line-level pre-training. 
We argue that processing whole documents in an end-to-end way is the solution to consider to solve the second issue. 
Not only does it avoid error accumulation, but it enables to take advantage of layout information for text recognition, and vice versa. 
To this aim, we proposed the \gls{dan} as the first end-to-end model performing the task of \gls{hdr}, recognizing both text and layout.
We showed that it was feasible to process whole documents with rather complex layouts \textit{i. e.} two-column documents, one to two pages long, while reaching competitive results, and without relying on any physical segmentation annotations. 

All the approaches proposed in this thesis represent interesting steps towards the understanding of documents. However, document understanding field is still in its infancy and there are still many areas of research to be explored.

\section*{Perspectives}

\paragraph*{Improving the recognition.}
Whether it is at line, paragraph or document level, there seems to be only little improvement in terms of \gls{cer} these last years.
If we focus on the IAM dataset for example, the approaches of the literature generally reach a \gls{cer} between 4\% and 5\% \cite{Bluche2016,Bluche2017b,Yousef2020,Coquenet2022}. The only works that reach \gls{cer} of less than 4\% \cite{Bluche2017a,Voigtlaender2016}, down to 2.75\% \cite{Diaz2022}, use additional external data and/or an external language model (character and/or word N-gram). On the one hand, designing an external language model implies a multi-step approach; it requires additional training which can be long and whose hyperparameters can be complex to fine-tune. On the other hand, the external data used are mainly private and it is very costly to annotate new training data. In this respect, although the ideal would be to have an large annotated dataset of reference, like ImageNet, it would be interesting to focus on annotation-free approaches to improve the recognition performance.

As we said, this stagnation of the \gls{cer} can be observed at any level (line, paragraph, document). This leads us to think that the feature extraction part (encoder) would be the bottleneck regarding the architectures, rather than the decoder part. In this regard, it would be interesting to study the novel promising image-to-sequence architectures such as the Vision Transformer \cite{VisionTransformer,Chen2021} in the context of \gls{htr}. Moreover, this transformer-based architecture includes a kind of internal language model through its use of self-attention mechanism which could compensate the use of an external language model. However, the drawback of the Vision Transformer is its need for large training datasets so as to reach interesting results. As for all approaches based on neural networks, the size of the dataset is a key issue, especially for this type of architecture.

When proposing the \gls{dan}, we showed that using synthetic printed documents is efficient to train the attention part of the model but we assume that it is quite limited for the learning of the feature extraction part in the context of handwritten text recognition. Self-supervised learning is an emerging field of research which aims at learning from raw inputs, without using any manual annotation. 
This way, it would be interesting to focus on self-supervised learning approaches to take advantage of the astronomical number of handwritten documents that exist on earth and which are not annotated. More specifically, contrastive learning is a category of self-supervised learning methods that seems promising \cite{Roh2021,Caron2020,He2020,Chen2020}. To our knowledge, only one work focused on self-supervised contrastive learning approach for \gls{htr}, at line level \cite{Aberdam2021}. However, the authors of \cite{Caron2021} reaches interesting results by applying a self-supervised approach to the Vision Transformer architecture. 

\paragraph*{Toward the recognition of heterogeneous and complex handwritten documents.}
We proposed the \gls{dan} as the first end-to-end approach for \gls{hdr}. However, we limit ourselves to the evaluation of the \gls{dan} on homogeneous datasets (pages from a same book for READ 2016 and letters for RIMES 2009). Moving from the recognition of homogeneous to heterogeneous documents raises several issues. The layout entities can be very related to the nature of the document: question/answer fields are typical of forms, and the mention of a subject is indicative of a letter. The high variety of existing documents can lead to a relatively high number of classes to be recognized. In addition, if we focus on the \gls{dan} architecture, it requires to serialize the annotation of the document \textit{i. e.} to set a reading order between the different text regions. Given the layout variability from one document to another, it is necessary to annotate the reading order so as to obtain a global coherence, no matter the nature of the document, which can be difficult. 

Moreover, for documents with complex layouts such as schemes or maps, there is no human consensus about a unique reading order. This way, it would be interesting to focus on the training strategy, especially on the loss, to design a way to consider multiple reading orders as valid. The same goes for the metrics in order to have a coherent evaluation of the performance. In this the respect, the standard reference \gls{cer} metric cannot be used in this context. The $\mathrm{mAP}_\mathrm{cer}$ we proposed is a first proposition in that sense. 

Another drawback of the \gls{dan} we mentioned is about its prediction time, which linearly grows with the length of the sequence of characters to be recognized. It results in another point of improvement. It would be interesting to study the use of multiple reading heads that would advance in parallel on several parts of the input document in order to minimize the prediction time. 

Given the interesting results obtained with the Transformer architecture in many tasks, we assume that it would be possible to recognize more than just text and layout. For example, Transformers have been successfully applied for named entity recognition \cite{Rouhou2021} or non-textual item detection \cite{Singh}, in addition to the text recognition. Transformers has also been studied for handwritten mathematical expression recognition for instance \cite{Zhao2021}. This way, we think it would be possible to train a model to recognize all of these items in an end-to-end way: text, layout, mathematical expression, and others such as tables. But as for standard handwritten documents, the main issue is the lack of datasets. We hope that the contributions of this thesis will encourage the community to create datasets for the end-to-end recognition of whole documents.

Handwritten text recognition has been studied for decades and we think it is now time to consider the task of document recognition in a more general way. We assume the coming years will be very interesting from this point of view.

\clearpage

\chapter*{List of Publications}
\addcontentsline{toc}{chapter}{List of Publications}

\paragraph*{}
\fullcite{Coquenet2022b}
\paragraph*{}
\fullcite{Coquenet2022}
\paragraph*{}
\fullcite{Coquenet2021}
\paragraph*{}
\fullcite{Coquenet2020}
\paragraph*{}
\fullcite{Coquenet2019}
\clearpage

\chapter*{List of Communications}
\addcontentsline{toc}{chapter}{List of Communications}
\paragraph*{}
\fullcite{Normastic}
\paragraph*{}
\fullcite{ORASIS}
\paragraph*{}
\fullcite{FeteScience}
\paragraph*{}
\fullcite{SIFED2021}
\paragraph*{}
\fullcite{DS3}
\paragraph*{}
\fullcite{SIFED2020}
\paragraph*{}
\fullcite{SIFED2019}
\clearpage

\begin{appendices}

\chapter{Data augmentation}
\label{section-data-augmentation}

Data augmentation is a way to artificially augment the number of training data. It is based on transformation techniques which alter an original training sample. 

\begin{figure}[ht]
    \centering
    
    \begin{subfigure}[b]{0.45\textwidth}
    \includegraphics[width=\linewidth]{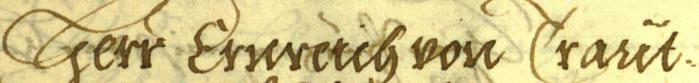}
    \caption{Original.}
    \end{subfigure}
    \hfill
    \begin{subfigure}[b]{0.45\textwidth}
    \includegraphics[width=\linewidth]{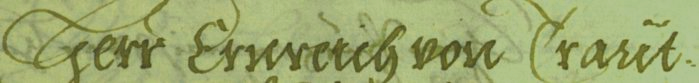}
    \caption{Color jittering.}
    \end{subfigure}
    
    \begin{subfigure}[b]{0.45\textwidth}
    \includegraphics[width=\linewidth]{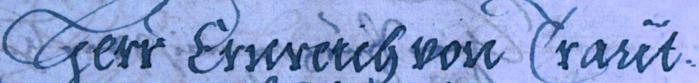}
    \caption{Hue.}
    \end{subfigure}
    \hfill
    \begin{subfigure}[b]{0.45\textwidth}
    \includegraphics[width=\linewidth]{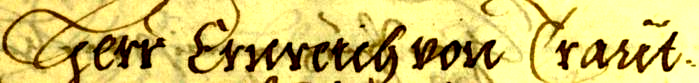}
    \caption{Contrast.}
    \end{subfigure}
    
    \begin{subfigure}[b]{0.45\textwidth}
    \includegraphics[width=\linewidth]{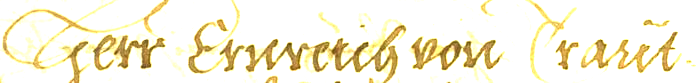}
    \caption{Brightness.}
    \end{subfigure}
    \hfill
    \begin{subfigure}[b]{0.45\textwidth}
    \includegraphics[width=\linewidth]{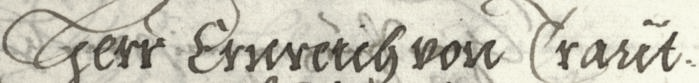}
    \caption{Saturation.}
    \end{subfigure}
    
    \begin{subfigure}[b]{0.45\textwidth}
    \includegraphics[width=\linewidth]{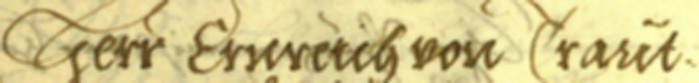}
    \caption{Gaussian Blur.}
    \end{subfigure}
    \hfill
    \begin{subfigure}[b]{0.45\textwidth}
    \includegraphics[width=\linewidth]{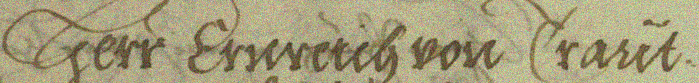}
    \caption{Gaussian Noise.}
    \end{subfigure}
    
    \begin{subfigure}[b]{0.45\textwidth}
    \includegraphics[width=\linewidth]{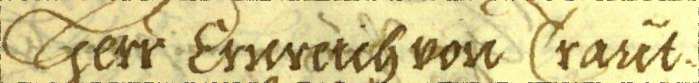}
    \caption{Elastic distortion.}
    \end{subfigure}
    \hfill
    \begin{subfigure}[b]{0.45\textwidth}
    \includegraphics[width=\linewidth]{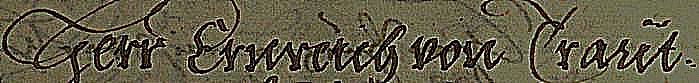}
    \caption{Sharpening.}
    \end{subfigure}
    
    \begin{subfigure}[b]{0.45\textwidth}
    \includegraphics[width=\linewidth]{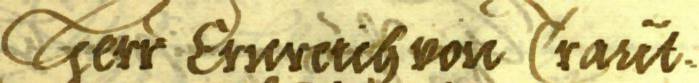}
    \caption{Dilation.}
    \end{subfigure}
    \hfill
    \begin{subfigure}[b]{0.45\textwidth}
    \includegraphics[width=\linewidth]{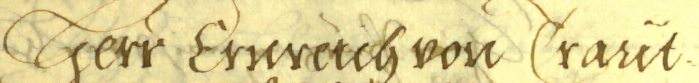}
    \caption{Erosion.}
    \end{subfigure}

    \begin{subfigure}[b]{0.45\textwidth}
    \includegraphics[width=\linewidth]{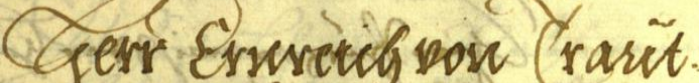}
    \caption{Zoom.}
    \end{subfigure}
    \hfill
    \begin{subfigure}[b]{0.45\textwidth}
    \includegraphics[width=\linewidth]{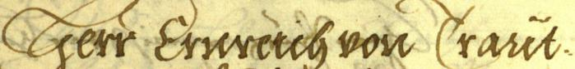}
    \caption{Resolution modification.}
    \end{subfigure}
    
    \begin{subfigure}[b]{0.45\textwidth}
    \includegraphics[width=\linewidth]{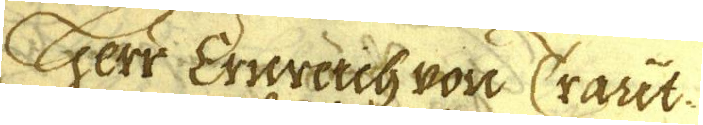}
    \caption{Rotation.}
    \end{subfigure}
    \hfill
    \begin{subfigure}[b]{0.45\textwidth}
    \includegraphics[width=\linewidth]{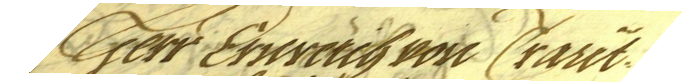}
    \caption{Perspective transformation.}
    \end{subfigure}
    
    \caption{Data augmentation techniques for HTR.}
    \label{fig:data_augmentation}
\end{figure}

The transformations used are subject to only one constraint: the relevant information must be preserved so that the ground truth remains unchanged. In the case of \gls{htr}, the characters must remain in the same order and recognizable. Figure \ref{fig:data_augmentation} illustrates the different data augmentation techniques used in this thesis.

Some of these transformations are based on the modification of color properties such as hue, contrast, saturation and brightness; it is generally referred to as color jittering. 
It can be seen as adding artificial variation in terms of digitization process and conditions. Noise can also be added to make the character recognition more robust through gaussian blurring or gaussian noise addition. One of the main difficult aspects in \gls{htr} is the variation of writing style from one person to another. Many techniques can be used to simulate writing style variety: dilation, erosion and sharpening for the stroke width, resolution modification and zoom for the writing size, perspective transformation for slanted writing and elastic distortion for character shape.
We combined these different data augmentation techniques in order to bring even more variability.

\end{appendices}

\addcontentsline{toc}{chapter}{Bibliography}
\begin{singlespace}
	\setlength\bibitemsep{\baselineskip}
	\printbibliography[title={Bibliography}]
\end{singlespace}

\end{document}